\documentclass[dvipsnames, 12pt]{article}

\pdfoutput=1

\usepackage[preprint]{tmlr}

\usepackage[T1]{fontenc}    
\usepackage[utf8]{inputenc} 
\usepackage{hyperref}       
\hypersetup{
    colorlinks=true,
    citecolor=RoyalBlue, 
    linkcolor=RoyalBlue, 
    filecolor=magenta,      
    urlcolor=RoyalBlue 
}
\usepackage{url}            
\usepackage{booktabs}       
\usepackage{amsfonts}       
\usepackage{nicefrac}       
\usepackage{microtype}      
\usepackage{xcolor}         
\usepackage{amsmath}
\usepackage{graphicx}
\usepackage{enumitem}
\usepackage{multirow}
\usepackage{subcaption}
\usepackage{listings}
\usepackage{tcolorbox}
\usepackage{xspace}
\usepackage{pifont}
\usepackage{bbding}
\usepackage{makecell}
\usepackage{soul}               
\setstcolor{red}            
\usepackage{tikz}

\usepackage{tablefootnote}
\usepackage[edges]{forest}
\definecolor{hidden-draw}{RGB}{20,68,106}
\definecolor{hidden-pink}{RGB}{255,245,247}
\definecolor{red}{RGB}{255,0,0}

\begin{document}
\title{\centering Multilingual Large Language Models: A Systematic Survey}
\author{
\centerline{Shaolin Zhu${^1}$, Supryadi${^1}$, Shaoyang Xu${^1}$, Haoran Sun${^1}$, Leiyu Pan${^1}$, Menglong Cui${^1}$}\\
\centerline{Jiangcun Du${^1}$, Renren Jin${^1}$, António Branco${^2}$${^\dagger}$\hspace{0.5em}, Deyi Xiong${^1}$\thanks{Corresponding author.} 
 \thanks{Advisory role.}}\vspace{0.5em}\\
\centerline{\normalfont{${^1}$TJUNLP Lab, College of Intelligence and Computing, Tianjin University}}\vspace{0.5em}
\centerline{\normalfont{${^2}$NLX, Department 
 of Informatics, University of Lisbon}}\vspace{0.5em}
\centerline{\texttt{\{zhushaolin, dyxiong\}@tju.edu.cn}}
\centerline{\texttt{\{supryadi, syxu, hrsun, lypan, cuimenglongcs, d2000, rrjin\}@tju.edu.cn}}
\centerline{\texttt{antonio.branco@di.fc.ul.pt}}
}
\maketitle

\begin{abstract}
This paper provides a comprehensive survey of the latest research on multilingual large language models (MLLMs). 
MLLMs not only are able to understand and generate language across linguistic boundaries, but also represent an important advancement in artificial intelligence. 

We first discuss the architecture and pre-training objectives of MLLMs, highlighting the key components and methodologies that contribute to their multilingual capabilities. 

We then discuss the construction of multilingual pre-training and alignment datasets, underscoring the importance of data quality and diversity in enhancing MLLM performance.

An important focus of this survey is on the evaluation of MLLMs. 
We present a detailed taxonomy and roadmap covering the assessment of MLLMs' cross-lingual knowledge, reasoning, alignment with human values, safety, interpretability and specialized applications. 
Specifically, we extensively discuss multilingual evaluation benchmarks and datasets, and explore the use of LLMs themselves as multilingual evaluators.

To enhance MLLMs from black to white boxes, we also address the interpretability of multilingual capabilities, cross-lingual transfer and language bias within these models.

Finally, we provide a comprehensive review of real-world applications of MLLMs across diverse domains, including biology, medicine, computer science, mathematics and law. 
We showcase how these models have driven innovation and improvements in these specialized fields while also highlighting the challenges and opportunities in deploying MLLMs within diverse language communities and application scenarios. We listed the paper related in this survey and publicly available at a Github repository.\footnote{\url{https://github.com/tjunlp-lab/Awesome-Multilingual-LLMs-Papers}}

\end{abstract}

\newpage

\tableofcontents

\newpage

\section{Introduction}

Scientific progress and technological innovation are a key driving force of human society, and the recent advancement of language science and technology has been part of them.
While being a crucial mean for people to communicate, language has enhanced knowledge accumulation and cultural inheritance.
And given the multiple languages in the world, 
efforts towards economic, cultural, political and other forms of communication have always require multilingual understanding. With the deepening of globalization, the development of language technology and the search for multilingual comprehension has only accelerated.

In the current information society and Artificial Intelligence (AI) age, 
we are at an unprecedented juncture of knowledge availability and cognitive revolution. 
Being a key factor underlying this transformative shift, generative Large Language Models (LLMs) are fostering a new dimension of intelligence with their extraordinary language processing capabilities — the ``intelligence of language'' \citep{chang2024survey}. 
Though still in its infancy, this form of intelligence is sweeping across every domain of scientific research and technological innovation. 

We are witnessing the emergence of a new 
level of intelligence capabilities that has the potential to transcend human cognitive limits, specially when this encompasses different languages and cultures.
Therefore, endowing LLMs with multilingual capabilities has become a crucial endeavour towards realizing their full potential.

A Multilingual LLM (MLLM) is an LLM that seeks to learn such capabilities across multiple languages by means of a single model.

Efforts to equip AI models with multilingual capabilities can be traced back to machine translation systems based on statistical methods, such as IBM's Candide system \citep{hutchins-1999-retrospect}, and even further back, to rule-based systems, relying on manually designed correspondences between expressions of different languages. 

Over a decade ago, with the rapid advancement of computational power and the arrival of the big data era, neural network-based end-to-end multilingual language models began to emerge. 
In 2014, \citet{DBLP:conf/emnlp/ChoMGBBSB14} proposed an attention-based neural machine translation model, greatly improving translation quality. 
In 2016, Google researchers introduced the GNMT neural machine translation system, built on a large-scale corpus, achieving state-of-the-art performance across multiple language pairs.

These early multilingual models typically required separate training and optimization for each language pair, lacking scalability and generalization capabilities. 
Seeking to address this bottleneck, a new generation of pre-trained multilingual language models emerged.

In 2019, \citet{DBLP:conf/naacl/DevlinCLT19} introduced the BERT model, which achieved stronger cross-lingual transfer abilities through pre-training on large-scale multilingual corpora.

Since then, a series of multilingual pre-trained language models, including XLM \citep{DBLP:conf/nips/ConneauL19}, mBART \citep{DBLP:journals/tacl/LiuGGLEGLZ20}, among many others, have been proposed, broadening the potential of multilingual modeling. Researchers no longer viewed multilingual processing as an extension of machine translation but rather as a holistic ability to understand and handle meaning across languages. 

To improve the training efficiency of models on massive multilingual data, innovative training objectives and architectures have been continuously proposed, such as the denoising autoencoding of XLM-R \citep{DBLP:conf/acl/ConneauKGCWGGOZ20} and the multitask architecture of mT5 \citep{DBLP:conf/naacl/XueCRKASBR21}, just to mention a couple of examples, among countless others. 

The rapid adoption of ChatGPT (OpenAI, 2022), attracting over 100 million users in just two months, has highlighted the transformative potential of LLMs. 
Their capabilities, including natural text generation, code generation and tool usage, coupled with their surprisingly strong multilingual abilities, have both triggered enthusiasm and critical questions \citep{DBLP:journals/corr/abs-2402-12204,DBLP:journals/bjmc/TarsTF22}. 

The path forward for MLLMs is not without challenges.
While many surveys have explored specific aspects of MLLMs \citep{xu2024survey,qin2024multilingual}, such as their training data, architecture or applications, a comprehensive examination of their multilingual capabilities, limitations, and challenges is lacking. 

Additionally, critical issues related to responsible AI, such as fairness and toxicity, have not been adequately addressed in that context \citep{DBLP:conf/acl/PhilippyGH23,DBLP:journals/corr/abs-2402-16438}.

The present survey aims to fill this gap by providing a comprehensive survey of the research on MLLMs. 
We will analyze the specific challenges MLLMs face in handling linguistic diversity, thus including non-English and low-resource languages, and explore data construction strategies, model training and fine-tuning approaches, as well as core responsible AI issues. Furthermore, we will delve into the practical challenges of deploying MLLMs in real-world language communities and application domains. 

This survey seeks thus to provide essential insights for navigating the evolving landscape of MLLMs and their impact on a global scale.

\begin{figure*}[!ht]
\begin{center}
\includegraphics[scale=0.51]{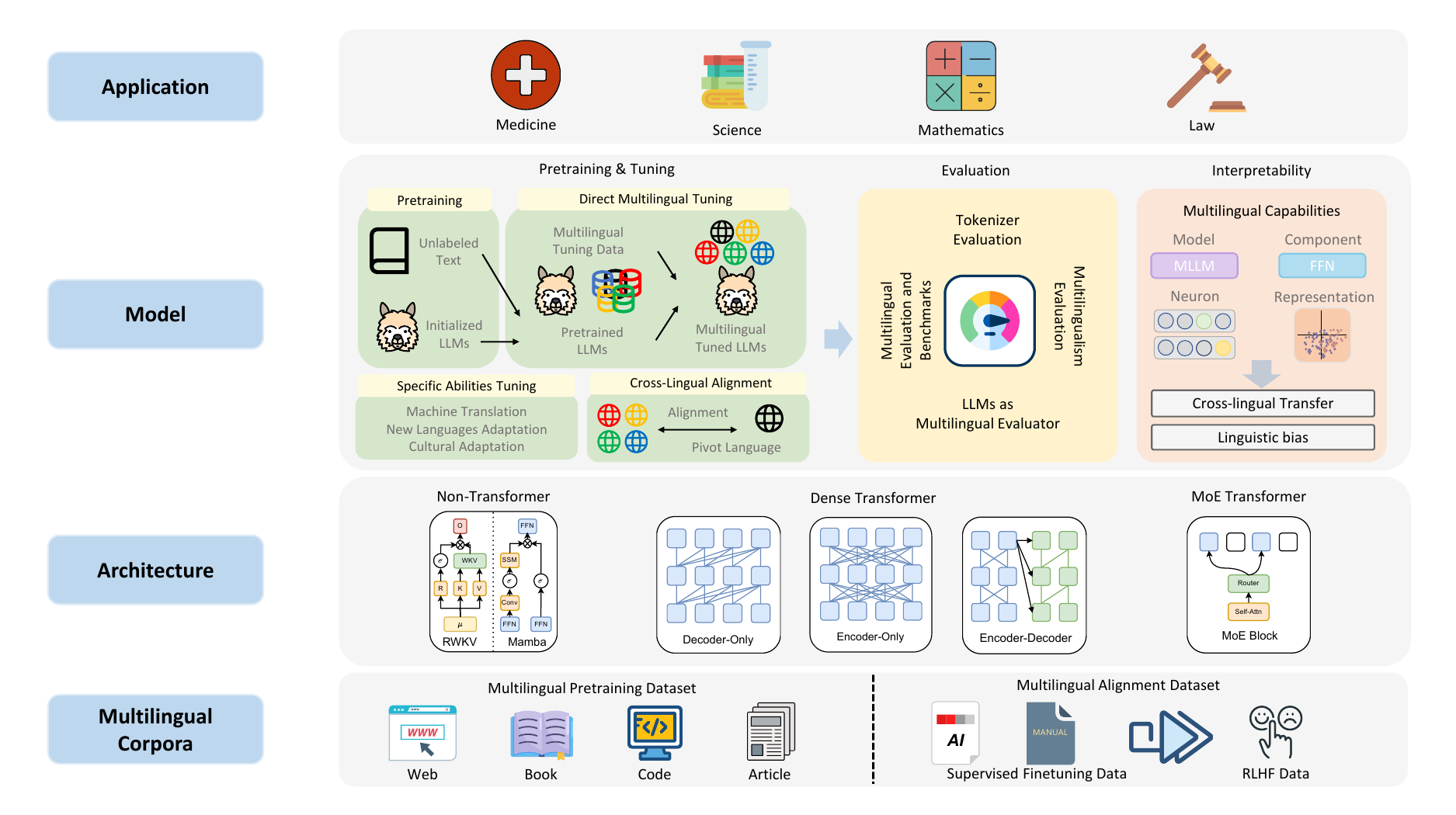}
\caption{Our proposed taxonomy of major categories and sub-categories of MLLMs.}
\label{taxonomy}
\end{center}
\end{figure*}

We will discuss data construction strategies, model training and fine-tuning approaches, and core Responsible AI issues such as fairness, bias and toxicity in the context of MLLMs. 
We will also analyze the specific challenges MLLMs encounter in handling linguistic diversity, along with relevant evaluation methodologies. 
Furthermore, we will delve into the practical challenges of deploying MLLMs in real-world language communities and application domains. 

As the performance of MLLMs continues to improve, leveraging them for developing multilingual language technologies globally has become increasingly viable. 
This survey aims thus to provide researchers in linguistics and artificial intelligence with a cutting-edge perspective on the development of MLLMs. 
We will analyze the opportunities and challenges that MLLMs face in towards realizing the full potential of language intelligence.

Our goal is to facilitate more inclusive and responsible language technology development, ensuring that these powerful tools are accessible and beneficial to diverse language communities worldwide.

By exploring the intricacies of multilingualism, encompassing linguistic diversity, language variations, and multilingual knowledge transfer, we strive to bridge the gap between the current state of LLM research and the urgent need for equitable and effective language understanding across all languages. 
Through this comprehensive examination of MLLMs, we aim to contribute for language intelligence to empower global communication, to foster cultural understanding and to unlock the potential of diverse knowledge systems.

\section{Taxonomy and Roadmap}
\label{Taxonomy and Roadmap of MLLMs} 

To provide a comprehensive overview of the rapidly evolving area of Multilingual Large Language Models, the present survey adopts a structured taxonomy, organizing the research landscape into six fundamental and interconnected domains.

Beginning with the foundational element of multilingual data, this survey considers diverse sources like web crawls, books, and code repositories. 

It then delves into the neural architecture choices for building effective MLLMs, analyzing the strengths and weaknesses of common architectures like decoder-only and encoder-decoder.

Building upon these foundations, this survey addresses the methodologies for pre-training and fine-tuning MLLMs, discussing various objectives like masked language modeling and translation language modeling, as well as fine-tuning techniques like instruction tuning and preference tuning. 

The crucial aspect of evaluating MLLMs performance is addressed through a comprehensive review of benchmarks and datasets, emphasizing the importance of balanced evaluation across diverse languages, including low-resource languages.

This paper then takes into account the ``black box'' dimension of MLLMs and discusses interpretability techniques to understand how these models achieve their multilingual capabilities and how to identify and mitigate potential linguistic biases \citep{gurgurov2024multilingual,blevins2024breaking,nezhad2024drives,kojima2024multilingual,liu2024unraveling}. 

Finally, the survey showcases diverse real-world applications of MLLMs.
Our overarching objective is to address these six fundamental domains, as illustrated in Figure \ref{taxonomy}.

The next Section \ref{Architectures} discusses the key architectural components and pre-training objectives that contribute to the development and training of MLLMs. 
MLLMs typically adopt either the decoder-only or encoder-decoder architecture. 
The acquisition of multilingual knowledge comes from training on multilingual corpora. The main architectural difference between multilingual and monolingual LLMs lies in the vocabulary, which needs to be larger in MLLMs to address the out-of-vocabulary issue. 
In terms of pre-training objectives, MLLMs use similar techniques to monolingual models, such as masked language modeling, but also explore novel objectives like translation language modeling to enhance cross-lingual abilities.

Section \ref{Multilingual Corpora} discusses the training data for MLLMs. 
The pre-training stage requires large amounts of unlabeled text data, while the alignment stage requires supervised fine-tuning with parallel data and reinforcement learning with feedback data. 
In terms of pre-training data, we summarize the characteristics of various web data, book data, code data, and academic paper data. 
Regarding parallel data, we detail the generation methods for supervised fine-tuning data, including human-generated and model-assisted generation, as well as the process of collecting reinforcement learning feedback data. 

The pre-training of MLLMs is discussed in Section \ref{Pre-training}, where the process of data curation is examined, together with the pre-training objectives that should be set and the pre-training strategies that have been explored in the literature.

Section \ref{Multilingual Tuning of LLMs} presents strategies for multilingual tuning, with the goal of adapting the general capabilities of multilingual LLMs to specific objectives. 
It begins by discussing approaches that extend LLMs tuning directly into multilingual contexts, with a focus on multilingual data collection and cross-lingual transfer. Additionally, it introduces strategies to further enhance cross-lingual alignment during multilingual tuning. Finally, it addresses specialized tuning techniques for the enhancement of specific multilingual capabilities.

Next, in Section \ref{Multilingual Evaluation of LLMs}, we address the evaluation of MLLMs by extensively discussing the evaluation, from multilingual tokenization to the datasets and including benchmarks used to evaluate MLLMs' capabilities across diverse tasks in a multilingual context. 
We further discuss the evaluation methods for determining the multilingualism of MLLMs and explore the use of MLLMs themselves in evaluating MLLMs' performance.

In seeking to transform an MLLM from a black box to a white box, Section \ref{Interpretability of MLLMs} addresses the core question of how the model represents multilingual capacities.
It also tackles advanced issues such as cross-lingual transfer and the causes of language bias. 

In section \ref{Application of MLLMs}, we undertake a thorough investigation into the evolutionary pathway and the latest breakthroughs of MLLMs spanning diverse disciplinary realms, with a particular emphasis on their real-world implementations and applications.

Overall, our aim is to address and help to clarify the following fundamental questions:

\begin{itemize}
    \item What are the capabilities of MLLMs?
    
    \item What is the language boundary of MLLMs?
    
    \item What factors must be taken into account when constructing and tuning MLLMs?

    \item How to evaluate the multilingual transfer capabilities of MLLMs?
\end{itemize}

\begin{figure*}[t]
\begin{center}
\includegraphics[width=1.0\linewidth]{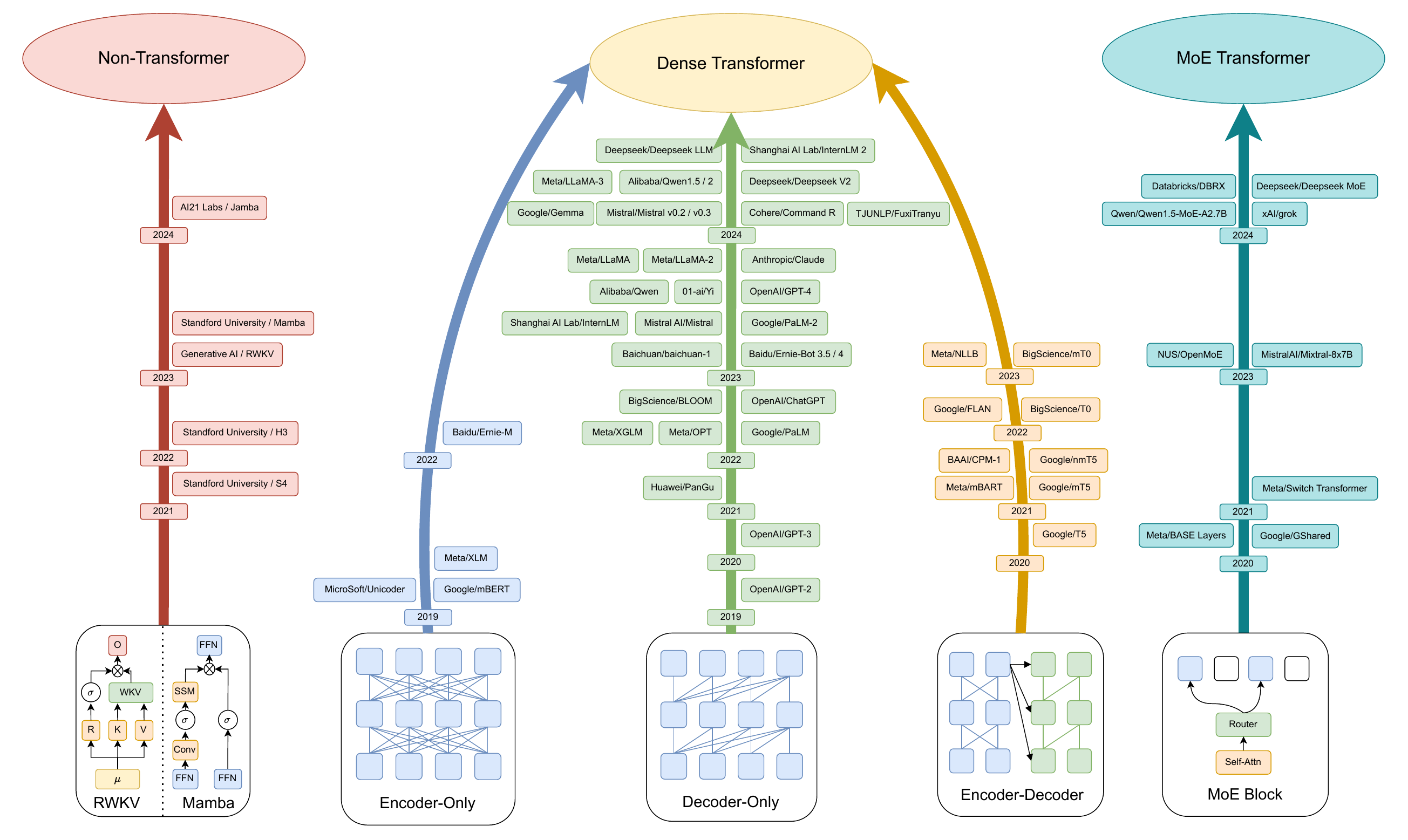}
\caption{The architectures of major MLLMs.}
\label{S3}
\end{center}
\end{figure*}

\section{Architectures}
\label{Architectures}

As large language models can excel in English, their performance varies though across the other languages. 
Accordingly, there is a growing need to advance multilingual large language models in order to ensure a balanced performance across a broader range of languages. Against this context, the goal of the present section is to provide a thorough examination of both LLMs and MLLMs. 
We will discuss the architecture of these models, typically based on the Transformer model \citep{vaswani2017attention}, as well as other architectures that may have the potential to challenge such supremacy of Transformers. Organized according to the major families of architectures, the deployment of major MLLMs along the last few years are summarized in Figure \ref{S3}.

\subsection{Dense Transformer}
MLLMs tend typically to adopt the same architecture as monolingual LLMs, even in the era of pre-trained language models \citep{bendale2024sutra}. 
However, the finer details of the Transformer architecture can vary significantly when building a top performing language model, which inherently affects the performance and stability of pre-training.

In an initial period, the preferred architecture was usually either the encoder-only model \citep{bert, XLM, XLM-R, erine} or the encoder-decoder model \citep{BART, T5, mT5, NLLB}, both of which typically utilize non-causal masked attention to perform denoising tasks.

Following the subsequent success of GPT models \citep{GPT3, InstructGPT}, the preferred approach shifted to the decoder-only model, which excel in generation tasks. 
Useful pre-training of these models involves training with a substantial number of parameters, say over 7 billion, on a massive amount of tokens, over 1 trillion. 
This process requires extensive GPU memory and time, which represents a significant cost. 
To address these challenges and enhance the performance of pre-trained models, various modifications have been made to the vanilla Transformers, often related to the self-attention mechanism, layer normalization, activation functions and position embeddings.

LLaMA \citep{llama-1} is one of the first open foundation models that paved the way for the success of open source MLLMs.
Despite using the vanilla Transformer architecture, LLaMA introduces significant modifications. 
It adopts pre-normalization like GPT-3 \citep{GPT3}, and uses RMSNorm \citep{rmsnorm} instead of the traditional layer normalization, which eliminates the shift operation of layer normalization and improves FLOPs. 
Additionally, LLaMA chooses SwiGLU, following PaLM \citep{chowdhery2022palmscalinglanguagemodeling}, to enhance performance, and Rotary Position Embedding (RoPE) \citep{rope}, following GPTNeo \citep{gpt-neo}, instead of standard absolute position embeddings to support long-context extension. 
LLaMA also removes the linear bias of the feed-forward networks and self-attention mechanism. This design approach has been widely adopted by successors such as Yi \citep{Yi}, Deepseek-LLM \citep{DeepseekLLM}, InternLM \citep{internlm}.

Other MLLMs employ different architectures. 
For example, Qwen1.0 \citep{qwen} retains the bias in self-attention, and Gemma uses GeGLU \citep{Gemma} instead of SwiGLU. 
Falcon \citep{falcon} and BaiChuan-13B \citep{baichuan} uses ALiBi \citep{ALiBi} instead of RoPE for position embeddings. 
To decrease memory usage during inference, Grouped Query Attention (GQA) \citep{GQA} has been implemented in many LLMs, including LLaMA-2 \citep{llama-2}, Yi \citep{Yi}, DeepSeek-LLM \citep{DeepseekLLM}, InternLM 2 \citep{internlm2}, and OpenELM \citep{OpenELM}. Compared to Multi-Head Self-Attention (MHA), GQA groups multiple queries and uses a single key and value head, reducing memory usage.

MLLMs like BLOOM \citep{BLOOM} and PolyLM~\citep{tuning-7} use the GeLU activation function and vanilla layer normalization, as well as ALiBi instead of RoPE for position embeddings. 
These architectural differences are not specific to language, but rather general improvements. 
BLOOM also expanded the vocabulary size to 250K to reduce the risk of under-segmenting for multilingual purposes, whereas other monolingual LLMs from the same period used much smaller vocabulary sizes, typically around 32K. 
Another MLLM FuxiTranyu~\citep{fuxitranyu} makes a balance between performance and training efficiency, using GeLU activation function and RMSNorm. 
Similar to BLOOM, FuxiTranyu expands the vocabulary size to 250K to fit the multilingual corpora. 

Researchers have also proposed new alternatives to MHA or GQA.
For instance, the Mistral-7B model \citep{mistral} adopts Sliding Window Attention (SWA), and DeepSeek-V2 \citep{DeepseekV2} proposes Multi-Head Latent Attention (MLA). These variants aim to reduce the KV cache during inference.

\subsection{MoE Transformer}
The Mixture of Experts (MoE) architecture is an effective approach for scaling model parameters while keeping the FLOPs similar to their dense counterparts. 
Research on MoE models has shown promising results across multiple tasks \citep{Outrageously}.

OpenMoE \citep{OpenMoE} is one of the first attempts to build an open-sourced LLM adopting an MoE approach, with parameters ranging from 650M to 34B, pre-trained on 1T tokens. 
Unlike Switch Transformer or GShard \citep{switchtransformer, GShard}, which replace the Feed-Forward Network (FFN) module with MoE, OpenMoE interleaves the MoE module with the FFN module. 
Mixtral-8x7B \citep{Mixtral}, in turn, is an outstanding LLM that pushes forward the experimentation with MoE. 
It adopts an architecture similar to the dense Mistral-7B model, replacing the FFN module with the MoE module. 
Mixtral-8x7B contains 8 experts per layer and uses a top-2 routing mechanism to select experts for each token. Xai's Grok-1 model\footnote{https://x.ai/blog/grok} follows this setup, scaling the parameters to 314B.

DeepSeek MoE \citep{DeepSeekMoE} employs a similar architecture but with a fine-grained expert design. 
The original $N$ experts are segmented into $mN$ experts and correspondingly routed to $mK$ experts instead of $K$ experts to achieve a more flexible combination of activated experts. 
Additionally, it leaves $K_s$ experts as shared experts, leading to a final number of routed experts of $mK - K_s$ while $mK$ experts are activated. Using this MoE architecture, performance similar to their dense counterparts is achieved, with the same total number of parameters. 
The Qwen1.5-MoE series \citep{qwen_moe} follows DeepSeek MoE's successful approach, using 4 shared experts and 4 of 60 activated experts. 
DBRX \citep{DBRX}, in turn, uses a similar idea to build a fine-grained MoE architecture with 16 experts, activating 4 of them. 
These fine-grained MoE shows promising results compared to Mixtral-8x7B and Grok-1. However, they do not use shared experts.

In summary, the MoE approach provides an effective way to scale the parameters of MLLMs, and fine-grained expert design leads to surprisingly good performance compared to dense counterparts. 
In the context of multilingual models, the main issue with model architecture is often its limited capabilities. 
It remains an open question to be examined whether a fine-grained MoE architecture would significantly boost the performance of MLLMs.

\subsection{Competitors of Transformer}

Despite the supremacy of the Transformer architecture for MLLMs, it has a significant drawback: its time complexity scales quadratically with the sequence length. 
Competitors to the transformer often propose architectures with linear time complexity.

RWKV \citep{rwkv} is a variant of the Recurrent Neural Networks (RNN) architecture that stacks residual blocks with time-mixing and channel-mixing modules. 
RWKV has scaled to 7 billion parameters and matches Transformer performance on multiple benchmarks. 
Mamba \citep{mamba} follows previous state-space models like S4 and H3 \citep{gu2022efficientlymodelinglongsequences, fu2023hungryhungryhipposlanguage}, adopting the state-space architecture while scaling parameters to 1.3 billion. 
It has shown promising results on multiple validation sets using perplexity as the metric, although whether it can scale much larger remains unknown.

Jamba \citep{jamba} can be seen as a successful fusion of Mamba and Transformers, stacking Mamba layers, Mamba with MoE layers, and Transformer layers together. 
Jamba inherits the high-quality output of transformers and the high throughput of Mamba, demonstrating remarkable performance on several benchmarks compared to LLaMA-2 70B and Mixtral 8x7B, with a total of 52 billion parameters and 12 billion activated parameters.

\section{Multilingual Corpora}
\label{Multilingual Corpora}

Training MLLMs typically involves three major stages: pre-training, supervised fine-tuning (SFT), and reinforcement learning from human feedback (RLHF). 
Pre-training requires vast amounts of unlabeled, raw text, while SFT and RLHF occur during the alignment phase. 
The GPT-3 \citep{DBLP:conf/nips/BrownMRSKDNSSAA20}, based on the transformer architecture, demonstrates strong contextual learning abilities with pre-training using only large amounts of unlabeled data. However, it still shows significant gaps in areas such as value alignment. 
The powerful capabilities of ChatGPT arise from quality data used in supervised fine-tuning and intense RLHF \citep{DBLP:conf/nips/Ouyang0JAWMZASR22}.
Subsequently, skillful chat models such as the Qwen-Chat series \citep{DBLP:journals/corr/abs-2309-16609}, LLaMA-Chat series \citep{DBLP:journals/corr/abs-2302-13971}, and Baichuan 2-Chat series \citep{DBLP:journals/corr/abs-2309-10305}  have emerged, all having undergone both SFT and RLHF. 
Consequently, gathering, preparing, curating and handling data deserves and has gained substantial research, as witnessed by the expanding ELRA Language Resources Association's LREC conferences,\footnote{https://www.elra.info/elra-events/lrec/} among others.

In this section, we will introduce the pre-training and alignment data for MLLMs separately.
For the alignment phase, we will discuss SFT data and RLHF data in detail.

\subsection{Multilingual Pre-training Datasets}

Pre-training is the most time-consuming and resource-intensive stage during a typical MLLMs training. 
Compared to alignment data, the amount of pre-training data tends to be much larger. Qwen \citep{DBLP:journals/corr/abs-2309-16609} has undergone pre-training on 2 to 3 trillion tokens, Baichuan 2 \citep{DBLP:journals/corr/abs-2309-10305} on 2.6 trillion tokens, and the LLaMA 3\footnote{\url{https://ai.meta.com/blog/meta-llama-3}} model on 15 trillion tokens. 
To ensure that the model can learn and generalize across input data during the pre-training stage, the pre-training data encompasses a wide range of fields and sources, and data filtering methods should be employed to ensure its quality. 
For MLLMs, a rich collection of multilingual data enables the model to learn connections between languages, and generalize across them, thereby enhancing its multilingual capabilities.

In this section, we will focus on multilingual pre-training data sets.

\newcommand{\gou}{\ding{52}}
\begin{table*}[]
\resizebox{\textwidth}{!}{%
\begin{tabular}{@{}ccccccccc@{}}
\toprule
name & open-source & from/type & size & language & low\% & non-English\% & date & used by \\ \midrule
\multicolumn{1}{c|}{Anna’s Archive\footnote{\url{https://zh.annas-archive.org}}} & \gou & Book & 862.2 TB & Multi & N/A & N/A & 2024-06 & / \\
\multicolumn{1}{c|}{CC100 \citep{DBLP:conf/acl/ConneauKGCWGGOZ20}} & \gou & CommonCrawl 2018 & 2.5TB & 100 & 61\% & 85\% & 2020-07 & XLM-R \\
\multicolumn{1}{c|}{CulturaX \citep{DBLP:conf/coling/NguyenNLMNDRN24}} & \gou & mC4 \& OSCAR & 27TB (6.3T tokens) & 167 & 34\% & 55\% & 2023-09 & / \\
\multicolumn{1}{c|}{mC4 \citep{DBLP:conf/naacl/XueCRKASBR21}} & \gou & CommonCrawl & 251GB & 101 & 46\% & 62\% & 2021-06 & mT5 \\
\multicolumn{1}{c|}{MultiUN \citep{MultiUN}} & \gou & Parallel Corpora & 4353MB & 7 & N/A & N/A & 2010-05 & / \\
\multicolumn{1}{c|}{News-crawl\footnote{\url{https://commoncrawl.org/blog/news-dataset-available}}} & \gou & CommonCrawl & N/A & Multi & N/A & N/A & 2024-03 & / \\
\multicolumn{1}{c|}{OSCAR 23.01\footnote{\url{https://oscar-project.org}} \citep{ortiz-suarez-etal-2020-monolingual,OrtizSuarezSagotRomary2019}} & \gou & CommonCrawl & 9.49TB & 153 & 36\% & 64\% & 2023-01 & / \\
\multicolumn{1}{c|}{ParaCrawl \citep{ParaCrawl}} & \gou & Parallel Corpora & 1527 M sentences & 48 & N/A & N/A & 2021-09 & / \\
\multicolumn{1}{c|}{RedPajama \citep{together2023redpajama}} & \gou & Mixed & 1.2T tokens & Multi & N/A & N/A & 2023-04 & / \\
\multicolumn{1}{c|}{RedPajama v2 \citep{together2023redpajama}} & \gou & CommonCrawl & 30T tokens & 5 & 14\% & 32\% & 2023-10 & / \\
\multicolumn{1}{c|}{ROOTS \citep{ROOTS}} & \gou & OSCAR/Github etc. & 1.6TB & 59 & 15\% & 70\% & 2023-03 & BLOOM \\
\multicolumn{1}{c|}{UNCorpus \citep{UNCorpus}} & \gou & Parallel Corpora & 799,276 docs & 6 & N/A & N/A & 2016-05 & / \\
\multicolumn{1}{c|}{FineWeb} & \gou & CommonCrawl & 29.2 TB & Multi & N/A & N/A & 2024-04 & / \\
\multicolumn{1}{c|}{Gutenberg project\footnote{\url{https://www.gutenberg.org}}} & \gou & Books & 70,000+ Books & Multi & N/A & N/A & 2024-06 & / \\
\multicolumn{1}{c|}{Zyda \citep{tokpanov2024zyda}} & \gou & Extract from existing data & 1.3T tokens & Multi & N/A & N/A & 2024-06 & / \\ \bottomrule
\end{tabular}%
}
\caption{Multilingual data sets used for pre-training. 
We categorize languages with more than 10GB of training data in BLOOM \citep{DBLP:journals/corr/abs-2211-05100} as high-resource languages, while the rest are considered low-resources languages. 
The high-resource languages group includes: Arabic (ar), Bengali (bn), Catalan (ca), English (en), French (fr), Hindi (hi), Indonesian (id), Portuguese (pt), Simplified Chinese (zhs), Spanish (es), and Vietnamese (vi).}
\label{tab:pretrain-data}
\end{table*}

\subsubsection{Multilingual Data from the Web}

Taking into account the sources of data, typically the data can be roughly divided into web crawl, book, code, academic paper, news, and others. 

Considering the distribution of data domains, they can be divided into general and domain-specific data. 

For pre-training datasets, the focus is mainly on general data. 
We list these data sets and summarize its major aspects in Table \ref{tab:pretrain-data}.

The most well-known source of pre-training data for MLLMs is CommonCrawl\footnote{https://commoncrawl.org}, a large-scale collection of web dumps. 
It crawls billions of web pages every month and releases a snapshot, now containing snapshots from 2008 to the present, and it is continuously being updated. 
CommonCrawl contains web data in multiple languages but includes a lot of pornographic, violent, and misaligned information that needs to be filtered before use. 

Nowadays, many web-based corpora are cleaned and obtained from CommonCrawl. 
CC100 \citep{DBLP:conf/acl/ConneauKGCWGGOZ20}, mC4 \citep{DBLP:conf/naacl/XueCRKASBR21}, FineWeb\footnote{https://huggingface.co/spaces/HuggingFaceFW/blogpost-fineweb-v1}, RedPajama v2 \citep{together2023redpajama}, RedPajama (part) \citep{together2023redpajama}, and SlimPajama \citep{cerebras2023slimpajama} are extracted and filtered from CommonCrawl.
They contain multiple languages and represent the largest data volumes. 

\textbf{CC100} \citep{DBLP:conf/acl/ConneauKGCWGGOZ20} is a dataset gathered from the web with over 100 languages, generated from 12 snapshots of CommonCrawl from the year 2018. 
Only one snapshot was used for English, while the other languages were extracted using all 12 snapshots. 
This dataset was processed using CCNET \citep{DBLP:conf/lrec/WenzekLCCGJG20}, an open-source CommonCrawl processing pipeline. 
The CCNET processing pipeline includes language identification, deduplication, and filtering based on language model perplexity.

\textbf{mC4} \citep{DBLP:conf/naacl/XueCRKASBR21} is the multilingual version of the C4 dataset, which is primarily aimed at English while mC4 covers 108 languages. 
It is extracted and deduplicated from CommonCrawl, followed by filtering based on various rules, including line length etc.

\textbf{RedPajama} \citep{together2023redpajama} was initiated to reproduce the training data for LLama. Most of its data comes from CommonCrawl and C4, with a small portion from code, books, papers, and wikipedias.

\textbf{RedPajama v2} \citep{together2023redpajama} is a large-scale dataset processed using the CCNET pipeline on CommonCrawl, containing data in 5 languages.

\textbf{SlimPajama} \citep{cerebras2023slimpajama} is a high-quality dataset derived from RedPajama after extensive filtering and deduplication.

\textbf{FineWeb} is primarily English with a small amount of multilingual data. 
It has undergone more meticulous cleaning, including common URL-based filtering and custom filters for removing list-like documents, documents with repeated lines, and documents with potentially incorrect line formats. 
After thorough processing, FineWeb's data quality surpasses other high-quality datasets such as SlimPajama and C4.

\textbf{Zyda} \citep{tokpanov2024zyda} dataset, encompassing 1.3 trillion tokens, is a high-quality pre-training dataset created by integrating multiple premium datasets such as RefinedWeb, SlimPajama, C4, and arXiv. 
This integration process involved meticulous de-duplication and filtering at a fine-grained level, both within individual datasets and across different datasets.

\textbf{Oscar}\footnote{\url{https://oscar-project.org}} \citep{ortiz-suarez-etal-2020-monolingual,OrtizSuarezSagotRomary2019} is extracted from CommonCrawl through filtering and deduplication, and it also serves as one of the sources for the ROOTS datasets. 

\textbf{CulturaX} \citep{DBLP:conf/coling/NguyenNLMNDRN24} is derived from both Oscar and mC4. 
Consequently, there is a relatively high overlap among these datasets.

\subsubsection{Multilingual Book Datasets}

Books represent a high-quality segment of pre-training data once all books should have undergone a careful manual review, and the complexity and quality are much higher than those of web crawls. 
Two massive e-book projects, Anna's Archive\footnote{\url{https://zh.annas-archive.org}} and the Gutenberg Project\footnote{\url{https://www.gutenberg.org}}, exemplify this type of data. Anna's Archive currently holds over 862.2 TB of data, while the exact volume of data in the Gutenberg Project remains unknown.

\subsubsection{Datasets with Code and Articles}

Programming code and academic papers represent the highest quality data for pre-training. Studies have shown that code data can significantly enhance the performance of models on various tasks. 
Most of the code data is sourced from GitHub\footnote{\url{https://github.com}}, which hosts a vast amount of open-source code, following a series of filtering and cleaning processes.

BigCode includes several code datasets, such as the Stack \citep{DBLP:journals/corr/abs-2211-15533} and Stack v2 \citep{DBLP:journals/corr/abs-2402-19173}, featuring over 600 programming languages including C++, Java, Python etc., which are used by the large code model StarCoder2\citep{DBLP:journals/corr/abs-2402-19173}.

Academic materials primarily include journal and conference papers, mainly sourced from open-access repositories such as arXiv\footnote{\url{https://arxiv.org}}, PubMed\footnote{\url{https://pubmed.ncbi.nlm.nih.gov}}, and PhilPapers.\footnote{\url{https://philpapers.org}} 
These materials provide high-quality knowledge for the model and are typically included as part of the dataset.

\begin{table*}[]
\resizebox{\textwidth}{!}{%
\begin{tabular}{@{}cccccccc@{}}
\toprule
name                                     & open-source & entries       & language & low\%    & non-English\% & date    & used by \\ \midrule
\multicolumn{1}{c|}{Aya Collection \citep{DBLP:journals/corr/abs-2402-06619}}      & \gou        & 513M          & 114      & Yes      & balance       & 2024-02 & N/A     \\
\multicolumn{1}{c|}{Aya Dataset \citep{DBLP:journals/corr/abs-2402-06619}}         & \gou        & 204K          & 65       & Yes      & balance       & 2024-02 & N/A     \\
\multicolumn{1}{c|}{Bactrain-X \citep{DBLP:journals/corr/abs-2305-15011}}          & \gou        & 3M            & 52       & Yes      & balance       & 2023-05 & N/A     \\
\multicolumn{1}{c|}{CAMEL \citep{DBLP:conf/nips/LiHIKG23}}               & \gou        & 1.6M          & multi    & No       & balance       & 2023-03 & N/A     \\
\multicolumn{1}{c|}{Flan 2021 \citep{flan2021}}                  & \gou & 62 datasets   & multi & N/A    & N/A & 2021-09 & Flan LAMDA \\
\multicolumn{1}{c|}{Flan 2022 \citep{flan2022}}           & \gou        & 1836 datasets & multi    & N/A      & N/A           & 2023-01 & Flan T5 \\
\multicolumn{1}{c|}{GuanacoDataset\footnote{\url{https://guanaco-model.github.io}}}      & \gou        & 534K          & 5        & No       & N/A           & 2023-03 & Guanaco \\
\multicolumn{1}{c|}{LMSYS-Chat-1M \citep{DBLP:journals/corr/abs-2309-11998}}       & \gou        & 1M            & multi    & Little   & Little        & 2023-09 &  N/A \\
\multicolumn{1}{c|}{OASST1 \citep{DBLP:conf/nips/KopfKRATSBNSNES23}}              & \gou        & 161K          & 35       & 20\%     & 57\%          & 2023-04 & N/A  \\
\multicolumn{1}{c|}{OpenOrca\footnote{\url{https://huggingface.co/datasets/Open-Orca/OpenOrca}}}            & \gou        & 4.2M          & multi    & Little   & Little        & 2023-06 &  N/A \\
\multicolumn{1}{c|}{Phoenix-sft-data-v1 \citep{phoenix-2023, llm-zoo-2023}} & \gou        & 464K          & multi    & < 20\% & 41\%          & 2023-05 &  Phoenix  \\
\multicolumn{1}{c|}{SUPER-NATURAL INSTRUCTIONS \citep{DBLP:conf/emnlp/WangMAKMNADASPK22}} & \gou & 1616 datasets & 55    & Little & Few & 2022-04 & Tk-Instruct \\
\multicolumn{1}{c|}{xP3 \citep{DBLP:conf/acl/MuennighoffWSRB23}}                 & \gou        & 82 datasets   & 46       & 28\%     & 60\%          & 2022-11 & BLOOMz  \\ \bottomrule
\end{tabular}
}
\caption{Multilingual datasets for Supervised Fine-tuning.}
\label{tab:sft-data}
\end{table*}

\subsection{Multilingual Alignment Datasets}

In this section, we introduce the datasets for so-called alignment. 
Alignment is an important research topic as pre-training enables models to acquire knowledge, while alignment allows models to follow instructions and align to human preferences and ethics. 
The alignment process for MLLMs is usually divided into two stages: Supervised Fine-Tuning (SFT) and Reinforcement Learning from Human Feedback (RLHF). 

\subsubsection{Datasets for Supervised Fine-Tuning}
SFT involves inputting diverse instructions and producing the respective outputs that align with human preferences. 

Two data-related factors influence the effectiveness of instruction fine-tuning. 
First, \textbf{data quality:} high-quality data is more effective than larger quantities of data \citep{llama-2}. 
Second, \textbf{the diversity of instructions:} a richer variety of instruction types tend to better help the alignment of the models.


The development of SFT datasets can be categorized into LLM-generated and human-annotated data. 
According to LLaMA 2 \citep{llama-2}, synthesized SFT data is competitive with human-annotated SFT data. Consequently, a significant portion of SFT data is actually derived from model-assisted generation or entirely model-generated. 
This approach allows for the reallocation of substantial human annotation costs to other alignment data, such as RLHF. 

We classify the data into three categories based on the method of synthesis: AI-Generated, Manually-Created, and mixed, and provide an overview of the SFT datasets. 

AI-generated data is produced by MLLMs like GPT-4\footnote{https://openai.com/index/gpt-4/} or others, sometimes using seed instructions or prompts provided by humans. Manually-created data is constructed entirely by humans. Mixed datasets include portions that are manually-created, AI-generated, and data obtained by applying instruction templates to traditional NLP datasets. A summary of multilingual SFT datasets is shown in Table \ref{tab:sft-data}.


\paragraph{AI-Generated Data}
%
%
\begin{itemize}
    \item The \textbf{alpaca} dataset \citep{alpaca} is a SFT dataset of English instructions generated by GPT-4, containing a total of 52K samples, and serves as a source for some SFT data.
    
    \item The \textbf{Bactrian-X} \citep{DBLP:journals/corr/abs-2305-15011} dataset includes 3.4 million pairs of instructions and responses across 52 languages. These instructions are derived from alpaca dataset and Dolly \citep{DatabricksBlog2023DollyV2} and translated into 52 languages, with model responses generated by GPT-3.5-turbo.\footnote{https://openai.com/index/gpt-3-5-turbo-fine-tuning-and-api-updates/}
    
    \item \textbf{CAMEL} \citep{DBLP:conf/nips/LiHIKG23} provides a role-playing SFT dataset that includes AI society and code instruction fine-tuning data, comprising a total of 584K entries, of which approximately 107K have been translated into multiple languages.
    
    \item The \textbf{GuanacoDataset}\footnote{\url{https://guanaco-model.github.io}} extends the alpaca dataset to include Japanese, German and Chinese, adding 534K entries.
    
    \item \textbf{OpenORCA}\footnote{\url{https://huggingface.co/datasets/Open-Orca/OpenOrca}} is an open-source reproduction based on the Orca paper \citep{DBLP:journals/corr/abs-2306-02707}, primarily in English. It expands entries from FLAN by using ChatGPT 3.5 or ChatGPT 4 to generate additional responses.
\end{itemize}

\paragraph{Manually-Created Data}

\begin{itemize}
    \item The \textbf{Aya} \citep{DBLP:journals/corr/abs-2402-06619} is an extensive multilingual instruction fine-tuning dataset. It includes instructions written by native speakers in 65 different languages. 
    
    \item \textbf{OASST1} \citep{DBLP:conf/nips/KopfKRATSBNSNES23} is a large-scale, multilingual assistant-style conversation corpus entirely generated and annotated by humans. It encompasses 35 languages and includes over 161K messages.
\end{itemize}

\paragraph{Mixed}

\begin{itemize}
    \item The \textbf{Aya collections} \citep{DBLP:journals/corr/abs-2402-06619} comprise data translated from other languages, data converted according to instruction templates, and the original Aya dataset, totaling approximately 513 million samples.
    
    \item \textbf{LMSYS-Chat-1M} \citep{DBLP:journals/corr/abs-2309-11998} includes 1 million real-world conversations with 25 MLLMs. These conversations were collected from 21K unique IP addresses on MLLMs dialogue website.
    
    
    \item \textbf{Phoenix-sft-data-v1} \citep{phoenix-2023, llm-zoo-2023} contains 465K instruction and conversation data entries. The instruction data includes Chinese and English instructions sourced from the Alpaca dataset, as well as instructions in other languages obtained through translation. The outputs for the other language instructions consist of two parts: one part directly translates the outputs from the Alpaca dataset, and the other part translates only the instructions, with the outputs generated by GPT-3.5. Additionally, some instruction data is synthesized manually using a self-instructed approach.

    \item \textbf{Flan2021} \citep{flan2021} consists of 62 instruction datasets built from existing datasets, primarily in English. \textbf{Flan2022} \citep{flan2022} expands on this by adding chain-of-thought (CoT) and dialogue data, encompassing 1,836 tasks.

    \item \textbf{Super-NaturalInstructions} \citep{DBLP:conf/emnlp/WangMAKMNADASPK22} collects over 1,600 NLP tasks and converts them into an instruction dataset format.
\end{itemize}




\subsubsection{RLHF data}

The RLHF datasets, also known as the preference datasets, typically includes dialogue in the form of inputs and outputs, with each output containing preference information from other models or human feedback. 

Compared to SFT data, the cost for RLHF data is much higher. It is estimated that LLama 2 \citep{llama-2} spent around \$8 million to annotate its RLHF dataset.\footnote{\url{https://www.interconnects.ai/p/llama-2-from-meta?sd=pf}} 

Consequently, publicly available preference datasets are relatively scarce, and multilingual preference datasets are even rarer. The main ones include Chatbot Arena Conversations and OASST1 pairwise RLHF reward.


\textbf{OASST1} \citep{DBLP:conf/nips/KopfKRATSBNSNES23} dataset comprises multi-turn conversations between humans and MLLMs, including human feedback on the dialogues. 
This dataset is structured in the form of dialogue trees, with a total of 161K dialogue trees containing 461K quality rates.

\textbf{Chatbot Arena Conversations} \citep{zheng2023judging} includes 33K pairs of human preference data collected from 13K unique IP addresses. 
Each sample features two model-generated results and human preference information in different directions.

\section{Pre-training}
\label{Pre-training}

Thanks to the results obtained by the open-source community, the techniques underlying the pre-training of MLLMs have become increasingly more transparent. 
The primary techniques in the pre-training phase of MLLMs revolve around the collection and pre-processing of multilingual data, which takes a dominant role.
Pre-training objectives and strategies are also crucial to their eventual performance.

\subsection{Data Curation}

Research has demonstrated that a high-quality, de-duplicated pre-training dataset is essential for the enhanced performance of pre-trained LLMs \citep{Yi,deduplicating-lee}. 
Given that a substantial portion of pre-training data is sourced from the web, ensuring the quality and safety of this data is crucial. 
The process of developing such dataset for a well-performing LLM can be organized into five parts: language identification, quality filtering, safety filtering, deduplication, and up-sampling of certain domains.

Language identification is crucial for subsequent filtering methods since heuristic rule filters are typically based on the statistical information of the dataset, which can vary significantly across different languages. 

For quality filtering, combining filters based on heuristic rules with machine learned filters improves data quality as they permit to take into consideration both syntactic and semantic aspects \citep{Yi, DeepseekLLM, internlm2}. 

Heuristic rule filters often involve filtering documents based on pre-defined block lists, such as URLs, and the statistical information of documents, including character/token/digit/symbol ratios and the frequency of repeated words/n-grams/paragraphs \citep{Yi, internlm2, OpenELM}.
These rules are usually designed based on observations of the data. For instance, BLOOM \citep{BLOOM} defines a series of indicators to filter out low-quality data for each language, with indicators selected by fluent speakers of each language. 

Learned filters typically involve task-specific trained models. 
Examples include using KenLM\footnote{https://github.com/kpu/kenlm} to measure document perplexity or trained quality models to score data quality. 
For instance, \citet{internlm2} organize human annotators to label document quality and fine-tune BERT models to serve as filters. 
LLaMA 3 \citep{LLaMA-3} finds that using LLaMA 2 to identify high-quality data yield excellent results, and leverages it to generate high-quality text data.

To enhance the safety of pre-training data, documents containing unsafe content must be filtered out. 
Like quality filters, safety filters are composed of manually designed heuristic rules and machine learned filters, targeting, for instance, personally identifiable information (PII), toxicity or unwanted words or domains \citep{Yi, internlm2, qwen, llama-2, Gemma}. 
As another example, \citep{DBLP:journals/corr/abs-2405-04434} removes also contentious content from the pre-training corpus to mitigate cultural bias in the data.

Turning to text deduplication, it significantly impacts LLM performance. 
Current deduplication methods include exact-match deduplication and fuzzy-match deduplication, using algorithms like MinHash and LSH \citep{Yi, DeepseekLLM, qwen, internlm, LLaMA-3}. 
Effective deduplication ensures that redundant data does not skew the training process.

Some dataset were also subject to up-sampling methods for specific domains or data sources to enhance their diversity \citep{DeepseekLLM, qwen, llama-2}. 
This approach ensures that the pre-training dataset represents a broad range of linguistic and contextual variations.


\subsection{Pre-training Objectives}

Most popular MLLMs have increasingly adopted the decoder-only architecture and causal mask attention mechanism. 
This focus has led to a corresponding focus in terms of pre-training objectives, moving from masked language modeling to next token prediction. 
Under this pre-training objective, MLLMs are trained autoregressively to predict the next token given the current tokens, which can be formulated as:

\begin{equation}
\mathcal{L} = -\sum_t^Tp(x_{t+1} | x_{<t}; \theta)
\end{equation}
where $x$ stands for a sequence of tokens, $t$ represents each time step, and $\theta$ the parameters of the model.

Besides the next token prediction objectives, masked language modeling tasks can also be used to train LLMs. 
UL2 \citep{UL2} proposes using a combination of denoising objectives for pre-training. 
It includes R-Denoiser, S-Denoiser, and X-Denoiser. R-Denoiser is the standard span corruption introduced in T5 \citep{T5}, typically masking 15\% of the input tokens in spans of 2-5 tokens. 
S-Denoiser is akin to prefix language modeling, partitioning the input sequence into two sub-sequences corresponding to source and target. 
X-Denoiser is similar to R-Denoiser but with longer spans and a higher corruption rate. 

By incorporating these denoisers, UL2 pre-training objectives can be applied to any model architecture. 
It has been successfully applied to PaLM-2, showing promising results.
OpenMoE \citep{OpenMoE} also experimented with UL2 as the pre-training objective but found that the model's improvement slowed beyond the early training stage. 
Consequently, its training reverted to next token prediction.

\subsection{Pre-training Strategies}

To fully utilize computing resources during the pre-training stage, packing \citep{T5} is used instead of padding, as padding introduces meaningless computation. 
Packing involves collecting different sequences into a single document and then truncating it into multiple smaller segments of a predefined sequence length. 
However, this method disregards sentence semantics, potentially truncating sentences in the middle or concatenating multiple irrelevant sequences into a single segment.

Going beyond a single pre-training stage, many MLLMs incorporate a multi-stage pre-training strategy aimed at extending the context length \citep{internlm2, DeepseekV2}. 
The quadratic complexity of self-attention computation related to context length results in a significant increase in both computation and memory costs. 
And implementing a multi-stage pre-training with an expanded long context window has proven to be an effective solution, as this stage requires fewer tokens than the initial stage. 

For example, InternLM 2 \citep{internlm2} dedicates 90\% of total training steps to a 4096 context length, while the remaining 9\% of training steps use a 32K context length.

In addition to the stage that extends context length, MiniCPM \citep{MiniCPM} proposes another pre-training strategy to enhance MLLM performance on downstream tasks, called the warmup-stable-decay learning rate scheduler (WSD LRS). 
The warmup and stable stages are the same as in conventional pre-training. 
In the decay stage, the learning rate rapidly decreases from the maximum to the minimum. The loss curve demonstrates that this strategy achieves significantly better performance compared to the original pre-training strategy. 
MiniCPM suggests that 10\% of training steps dedicated to the decay stage is sufficient.

\begin{figure*}[!ht]
\begin{center}
\includegraphics[scale=0.6]{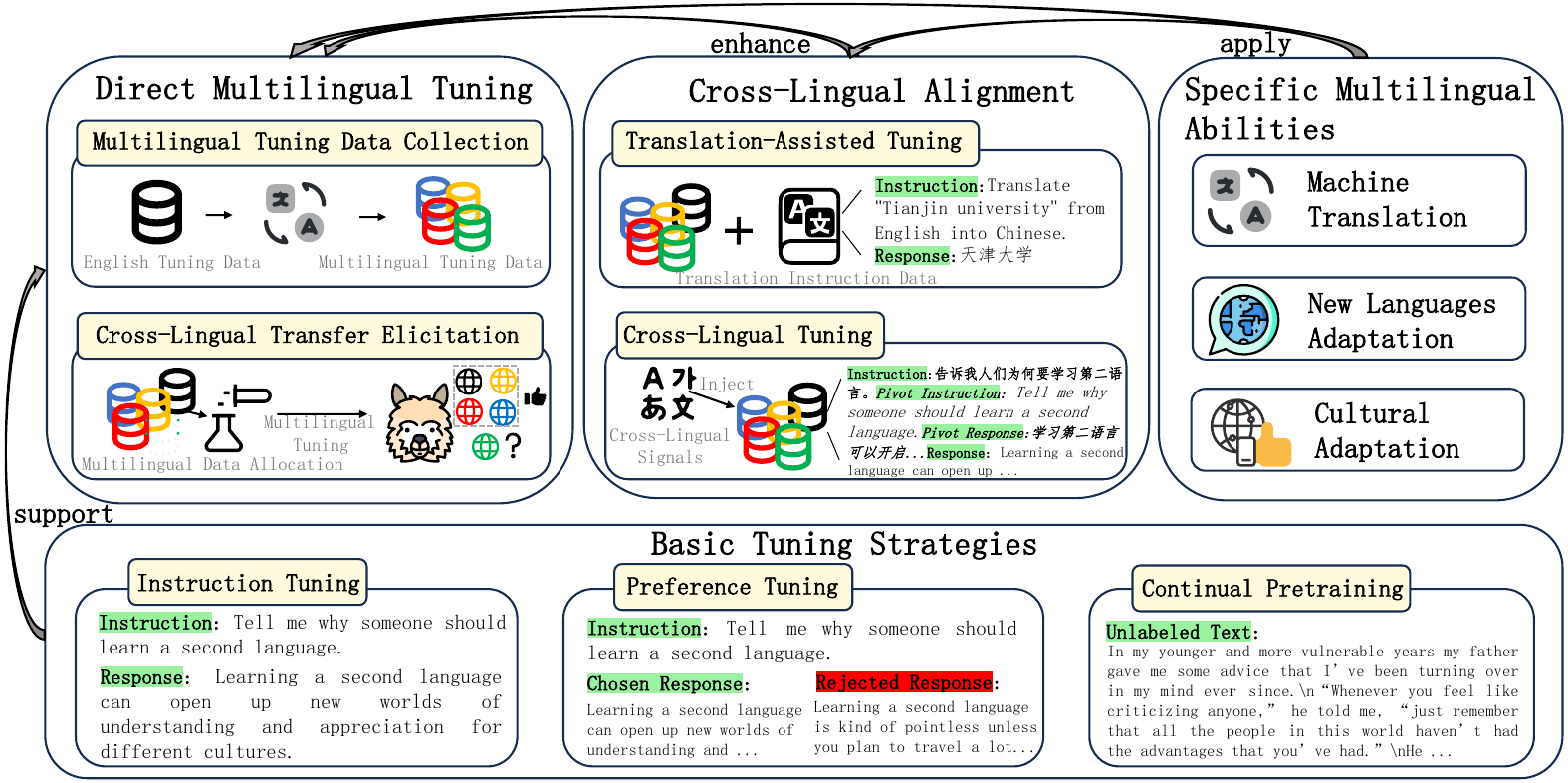} 
\caption{Multilingual tuning of MLLMs.}
\label{fig:multilingual tuning}
\end{center}
\end{figure*}

\section{Multilingual Tuning}
\label{Multilingual Tuning of LLMs}

There are over 7,000 languages spoken on the planet, yet most existing LLMs are primarily English-based or have limited multilingual capabilities. 
This hampers access to cutting-edge Artificial Intelligence for non-English speakers and poses the risk of cultural uniformisation and the loss of linguistic diversity.
To democratize LLMs across languages, a number of efforts have sought to extend tuning of MLLMs to multilingual scenarios. 

As summarized in Figure \ref{fig:multilingual tuning}, in this section, we start by introducing common tuning strategies to adapt the general capabilities of LLMs to specific objectives. 
Then, we present two common multilingual tuning strategies. 
The first is direct multilingual tuning, which simply expands tuning data to multiple languages; the second focuses on cross-lingual alignment during multilingual tuning, transferring capabilities from the pivot language (primarily English) to other languages.
We will also discuss specific tuning methods used to enhance particular multilingual capabilities.

\subsection{Basic Tuning Strategies}

Here we introduce foundational tuning techniques for adapting the general capabilities of MLLMs to specific objectives. 
These techniques also serve as the basis for multilingual tuning. 
They are: instruction tuning, preference tuning, and continual pre-training.

\subsubsection{Instruction Tuning}

In general, MLLMs are trained using a self-supervised learning objective, which involves predicting a token on the basis of the preceding or surrounding tokens \citep{DBLP:conf/nips/BrownMRSKDNSSAA20,DBLP:journals/jmlr/ChowdheryNDBMRBCSGSSTMRBTSPRDHPBAI23,DBLP:journals/corr/abs-2104-12369,DBLP:journals/corr/abs-2205-01068,DBLP:journals/corr/abs-2211-05100,DBLP:journals/corr/abs-2112-11446,DBLP:journals/corr/abs-2302-13971,shaham2024multilingual}. 
When a prompt is entered, the MLLMs generate a completion for that prompt that is contextually appropriate. 
However, despite the appropriateness of these completions, they may not always align with human preferences, thereby limiting the practical applications of MLLMs. 
To address this issue, instruction tuning has been proposed, which aims to enhance the alignment of generated responses with human preferences and has become the most widely method adopted to fine-tune MLLMs.

Instruction tuning is a technique designed to bridge the gap between the next-token prediction objective of MLLMs during pre-training and the goal of providing desired responses for humans in practical scenarios. 
Unlike the pre-training stage, which typically involves training the LLMs on document corpora, instruction tuning involves training the models with instructional prompts and corresponding completions. 

The pioneering work on instruction tuning involves fine-tuning MLLMs on instructions paraphrased from samples of various NLP tasks, which substantially enhances the performance of MLLMs on unseen tasks \citep{DBLP:conf/iclr/WeiBZGYLDDL22,DBLP:conf/iclr/SanhWRBSACSRDBX22}. However, fine-tuning MLLMs solely on data from NLP tasks can result in models that excel at completing these tasks rather than effectively interacting with humans.

Notably, \citet{DBLP:conf/nips/Ouyang0JAWMZASR22,DBLP:journals/corr/abs-2204-05862,DBLP:journals/corr/abs-2112-00861} fine-tuned MLLMs with instruction data that more closely align with everyday human instructions. 
This type of instruction data can be authored by either humans or LLMs themselves \citep{DBLP:conf/acl/WangKMLSKH23,DBLP:journals/corr/abs-2304-12244,DBLP:journals/corr/abs-2306-08568}. A critical difference between pre-training and instruction tuning is that the latter employs a prompt template that specifies the roles and tasks for MLLMs, which can be populated with samples to conduct the tuning process. Additionally, while pre-training typically computes loss over all tokens in a training sample, instruction tuning can ignore the loss of prompt tokens \citep{DBLP:journals/corr/abs-2308-06259}. 
However, recent studies suggest that optimizing the loss over both prompt tokens and response tokens yields superior performance compared to optimizing the loss over response tokens alone \citep{shi2024instruction,huertaenochian2024instruction}.

\subsubsection{Preference Tuning}

In contrast to instruction tuning, which involves training large language models with instructional prompts and their corresponding completions \citep{DBLP:journals/corr/abs-2308-10792}, preference tuning typically employs preference data. This data is generated by annotating the more preferred completion between two options \citep{DBLP:journals/corr/abs-2204-05862} or by ranking completions according to human preference \citep{DBLP:conf/nips/Ouyang0JAWMZASR22}. Currently, two primary techniques are used for preference tuning: Reinforcement Learning from Human Feedback (RLHF) \citep{DBLP:journals/corr/abs-1909-08593,DBLP:conf/nips/StiennonO0ZLVRA20,DBLP:journals/corr/abs-2112-09332,DBLP:conf/nips/Ouyang0JAWMZASR22} and Direct Preference Optimization (DPO) \citep{DBLP:conf/nips/RafailovSMMEF23}.

RLHF comprises three essential steps for training LLMs: (1) training LLMs with instruction tuning, also known as supervised fine-tuning; (2) training a reward model using preference data to predict human preferences; and (3) applying reinforcement learning to further train MLLMs from step 1, thereby enhancing the alignment between MLLMs and human preferences. 

In step 2, the trained reward model produces a scalar reward to quantify the degree of human preference for given completions, where a higher reward denotes greater alignment with human preferences. The parameters $\theta$ of the reward model are optimized based on the assumption that the preference data used for training are sampled from a preference function that adheres to the Bradley-Terry model \citep{bradley1952rank}.

Let $x$ denote the prompt, and let $y_{w}$ and  $y_{l}$ denote the ``win'' and ``lose'' responses, respectively, where ``win'' and ``lose'' represent the preferred and less preferred responses as chosen by humans. Let $r$ denote the preference function that maps a prompt and its corresponding completion to a scalar value, reflecting human preferences. The sigmoid function is denoted by $\sigma \left ( \cdot  \right )$. The notation $y_{w} \succ y_{l}$ indicates that $y_{w}$ is more preferred than $y_{l}$. According to the Bradley-Terry model, the probability that $y_w$ is preferred over $y_l$ can be expressed as follows:

\begin{align}
    P\left ( y_{w} \succ y_{l} | x \right ) &= \frac{\mathrm{exp} \left ( r\left ( x, y_{w} \right ) \right )}{\mathrm{exp}\left ( r\left ( x, y_{w} \right ) + \mathrm{exp}\left ( r\left ( x, y_{l} \right ) \right ) \right )} \notag \\
    &= \sigma \left ( r\left ( x, y_{w} \right ) - r\left ( x, y_{l} \right ) \right )
\end{align}

Consequently, the reward model $r_{\theta}$ can be trained to minimize the following objective:
\begin{equation}
    \mathcal L\left ( r_{\theta} \right ) = - \mathrm{log} \sigma \left ( r_{\theta} \left ( x, y_{w} \right ) - r_{\theta} \left ( x, y_{l} \right ) \right )
\end{equation}

All parameters of the reward model, with the exception of the final embedding layer, can be initialized from either pre-trained MLLMs \citep{DBLP:journals/corr/abs-2310-01377,DBLP:journals/corr/abs-2311-04919,DBLP:journals/corr/abs-2310-06452} or instruction-tuned LLMs \citep{DBLP:journals/corr/abs-2307-09288}. 
For MLLMs, the final embedding layer maps the hidden states of each token to the probability distribution over the vocabulary. However, to adapt the MLLMs into a reward model, the final embedding layer is replaced with a randomly initialized linear layer. This new layer maps the hidden states of each token to a scalar reward.

In step 3, reinforcement learning techniques, such as Proximal Policy Optimization (PPO) \citep{DBLP:journals/corr/SchulmanWDRK17}, are employed to train the LLMs to maximize the following objective:

\begin{equation}
    J\left ( \pi_{\phi } \right ) = r_{\theta }\left ( x, y \right ) - \beta \mathrm{log}\frac{\pi_{\phi} \left ( y | x \right)}{\pi_{\mathrm{ref}} \left ( y | x \right)}
    \label{RLHF_obj}
\end{equation}
where $y$ is the corresponding completion for prompt $x$, $r_{\theta }$ is the reward model trained in step 2, $\pi_{\phi }$ is the trained policy (trained LLM) and $\pi_{\mathrm{ref}}$ is a reference MLLM, which can be the instruction-tuned MLLMs. 
The hyperparameter $\beta$ can be tuned to control the strength of the restriction on the distribution divergence between the trained policy and the reference LLM, thereby ensuring the trained policy remains close to the reference MLLM.

Although RLHF has demonstrated impressive performance in aligning MLLMs with human preferences, achieving stable training in practice remains challenging and computationally expensive. 
To address these challenges, DPO has been proposed. While the objective of DPO is derived from Eq.~\ref{RLHF_obj}, which is also the objective of RLHF, DPO circumvents the need to explicitly train a reward model by expressing human preferences in terms of the optimal policy $\pi^{*}$:


\begin{equation}
    P\left ( y_{w} \succ y_{l} | x \right ) = \frac{1}{1 + \mathrm{exp}\left ( \beta \mathrm{log} \frac{\pi^{*}\left ( y_{l} | x \right )}{\pi_{\mathrm{ref}}\left ( y_{l} | x \right )} - \beta \mathrm{log}\frac{\pi^{*}\left ( y_{w} | x \right )}{\pi_{\mathrm{ref}}\left ( y_{w} | x \right )} \right )}
    \label{DPO_obj}
\end{equation}

Consequently, the training objective of DPO can be derived from Eq.~\ref{DPO_obj} by minimizing the following objective:


\begin{equation}
    \mathcal L\left ( \pi_{\phi } \right ) = - \mathrm{log} \sigma \left ( \beta \mathrm{log} \frac{\pi_{\phi }\left ( y_{w} | x \right )}{\pi_{\mathrm{ref}}\left ( y_{w} | x \right )} - \beta \mathrm{log}\frac{\pi_{\phi}\left ( y_{l} | x \right )}{\pi_{\mathrm{ref}}\left ( y_{l} | x \right )} \right )
\end{equation}

Apart from RLHF and DPO, there has been a surge of preference tuning approaches proposed in recent years, including IPO \citep{DBLP:conf/aistats/AzarGPMRVC24}, PRO \citep{DBLP:conf/aaai/00010LYHLW24}, RRHF \citep{DBLP:journals/corr/abs-2304-05302}, KTO \citep{DBLP:journals/corr/abs-2402-01306}, and SLiC-HF \citep{DBLP:journals/corr/abs-2305-10425}, among others \citep{wu2024betadpodirectpreferenceoptimization, DBLP:journals/corr/abs-2405-14734, xiong2024iterativepreferencelearninghuman}.

\subsubsection{Continual Pre-training}

Pre-trained MLLMs can be further adapted to specific objectives through methods such as instruction tuning and preference tuning. 
Continual pre-training serves as an intermediary step between the initial pre-training and task-specific fine-tuning, adapting MLLMs' general knowledge to specific domains or integrating entirely new knowledge~\citep{cpt_1,mu2024revealing,li2024comparison}. 
This process enhances the adaptability of MLLMs to specific tasks, thereby improving task performance.

The training objective during continual pre-training is next-token prediction, similar to that in the pre-training stage. 
However, a significant challenge in continual pre-training is catastrophic forgetting~\citep{cpt_1}, which occurs when MLLMs are optimized on the target data that are domain-specific or entirely new compared to the pre-training data. 
In essence, this phenomenon arises from the distributional differences between the target data and the original pre-training data. Catastrophic forgetting is a critical issue because it not only erases previously acquired knowledge but also hinders effective knowledge transfer.

To mitigate catastrophic forgetting, common approaches include replay, regularization, and architecture-based methods. 
Specifically, replay-based methods involve incorporating samples from the original training data alongside new target data~\citep{cpt_2,cpt_3,cpt_4,cpt_5,cpt_6,cpt_7}. Regularization-based methods impose constraints to minimize discrepancies between updated and original model parameters~\citep{cpt_8,cpt_9}, while architecture-based methods primarily focus on parameter-efficient fine-tuning techniques~\citep{cpt_10,cpt_11,cpt_12,cpt_13,cpt_14,cpt_15,cpt_16}.

\subsection{Direct Multilingual Tuning}

To democratize MLLMs across languages, many studies directly extend the tuning of MLLMs to multilingual contexts. 
The majority of these efforts involve direct multilingual instruction tuning, enabling MLLMs to follow instructions in multiple languages \citep{tuning-1,tuning-2,tuning-3,tuning-4,tuning-5,tuning-7,tuning-8,tuning-9,tuning-10,tuning-11,tuning-12,tuning-13,tuning-14,tuning-15}, while others focus on multilingual preference tuning, aligning MLLMs with human preferences across multiple languages \citep{tuning-1,tuning-6,tuning-12}. 
These studies adopt significantly different approaches to collecting multilingual tuning data. 
Additionally, the research question of how to effectively stimulate cross-lingual transfer during the multilingual tuning process is widely discussed.

\subsubsection{Multilingual Tuning Data Collection}

Expanding the tuning of MLLMs into multilingual scenarios necessitates the collection of multilingual tuning data. 
Here, we'll outline this process separately for multilingual instruction tuning and multilingual preference tuning.

\paragraph{Multilingual Instruction Tuning} Combining P3 \citep{p3} with 30 other multilingual datasets, \citet{tuning-3} created the xP3 dataset, which encompasses 46 languages. 
They subsequently translated the English prompts in xP3 into non-English languages using the Google Cloud API, resulting in the xP3-mt dataset. 

Additionally, Bactrian-X \citep{tuning-2} collected 67K English instructions from Alpaca \citep{alpaca} and Dolly \citep{dolly}, which were then translated into 51 languages using Google Translate. 
They further employed ChatGPT to generate multilingual responses to mitigate issues such as ``translationese'' and non-native answer styles. 

In addition to the strategy of first translating instructions and then generating language-specific responses, Phoenix \citep{tuning-4} included additional data by directly translating instructions and responses using GPT-4. 
Their multilingual dataset was also derived from English Alpaca and expanded to over 40 languages. 

Similarly, based on Alpaca, Polylm \citep{tuning-7} leveraged 175 English task seeds from it and iteratively collected and filtered samples for 11 languages using a Self-Instruct \citep{DBLP:conf/acl/WangKMLSKH23} method.

SeaLLMs \citep{tuning-8}, in turn, focused on nine Southeast Asian languages. 
To address the scarcity of Southeast Asian data (3.3\%) in their instruction dataset, they adopted a hybrid training strategy, merging the instruction data with multilingual pre-training data to achieve more balanced language ratios. 

In contrast, \citet{tuning-14} considered a broader range of languages, 101 in total. 
Various methods were employed to construct large-scale multilingual datasets, including aggregating and refining multilingual templates, as well as carefully selecting elusive human annotations from fluent speakers of different languages. 
Additionally, data augmentation strategies, such as machine translation and generating synthetic data combined with translation, were also utilized.

\paragraph{Multilingual Preference Tuning} To gather multilingual preference data, Okapi \citep{tuning-1} employed a series of procedures. 
Initially, they expanded the 52K English Alpaca dataset to 158K using Self-Instruct \citep{DBLP:conf/acl/WangKMLSKH23} techniques. 

Subsequently, ChatGPT was used to translate both instructions and responses into 26 languages. 
After translation, the 52K dataset for each language was first used to perform instruction tuning on MLLMs. 
Then, 42K instructions were fed into the instruction-tuned model to sample responses. Finally, ChatGPT was employed to translate the obtained responses back into English and rank preferences, resulting in the final preference data. 

It's worth noting that they conducted this process separately for each language. Afterward, they used the collected preference data to train a reward model for each language, and further trained a preference-tuned model for each language using the remaining 64K instructions per language.

\subsubsection{Cross-Lingual Transfer Elicitation}

In the process of direct multilingual tuning, effectively stimulating cross-lingual transfer under limited resources is an important research question \citep{xu2023language}. 
Here, we introduce how this challenge is addressed in terms of multilingual instruction tuning and multilingual preference tuning, respectively. 
Overall, these studies indicate that in direct multilingual tuning, cross-lingual transfer is influenced by factors such as the selection of source languages, linguistic relationships, the number of languages involved, and the scale of multilingual instruction data \citep{razumovskaia2024analyzing,faisal2024efficient,kim2024efficient}.

\paragraph{Multilingual Instruction Tuning} First, several studies suggest that simply performing monolingual instruction tuning on a single language can lead to certain degrees of zero-shot cross-lingual transfer \citep{tuning-3, tuning-5, tuning-13}. Specifically, \citet{tuning-5} found that various source languages yield varying levels of cross-lingual transfer, with English, Italian, and Spanish demonstrating superior results in their experimental settings, while \citet{tuning-13} noted that adjusting instruction tuning hyperparameters for multilinguality and using sufficient instruction data can improve zero-shot cross-lingual transfer.

Furthermore, assuming multilingual instruction data is available, several studies explore key factors to promote cross-lingual transfer, such as the number of languages and the volume of instruction data. 
Comparing multilingual and monolingual instruction tuning, \citet{tuning-5} found that replacing even a small number of monolingual (English) training samples (as few as 40) with multilingual ones significantly improves performance on these languages. 
This finding aligns with other research \citep{tuning-11,tuning-15}, emphasizing the advantages of multilingual instruction tuning over monolingual instruction tuning, especially in resource-constrained settings. 
Additionally, considering the amount of multilingual instruction data, \citet{tuning-15} suggested that sufficient instruction data is essential for achieving better multilingual performance. It is worth noting, however, that this stands in contrast to the Superficial Alignment Hypothesis \citep{lima}.

Regarding the number of languages, it is suggested that including just a few languages (e.g., 2-4) can enhance cross-lingual transfer, especially for languages not encountered during pre-training \citep{tuning-5, tuning-10}. 
However, \citet{tuning-9} suggested that increasing the number of languages may further enhance multilingual performance. 
They also explored other factors such as language similarity and concluded that the optimal number of languages depends on both language similarity and downstream evaluation.

\paragraph{Multilingual Preference Tuning} While the aforementioned studies explore cross-lingual transfer in direct multilingual instruction tuning, other works address it in multilingual preference tuning \citep{chai2024xcot}. 
Specifically, \citet{tuning-6} found that reward models trained on source languages can be effectively utilized for cross-lingual preference tuning in target languages. 
They even observed cases where using reward models from source languages outperforms those trained on target languages, possibly due to a regularization effect.
Additionally, high-resource languages (e.g., English) tend to be more effective in inducing cross-lingual preference transfer than low-resource languages \citep{tuning-6, tuning-12}.

\subsection{Multilingual Tuning Augmented by Cross-Lingual Alignment}

While direct multilingual tuning can enhance the multilingual capabilities of MLLMs to some extent, many approaches augment multilingual tuning with cross-lingual alignment to further improve cross-lingual transfer \citep{li2024improving,peng2024concept}. 
They typically use English, the best resourced language, as the pivot language, aligning the understanding, reasoning, and generation capabilities of MLLMs in non-English languages with those in English. 

Based on how cross-lingual signals are incorporated to achieve alignment, these works can be classified into two categories: Translation-Assisted Tuning and Cross-Lingual Tuning. 

Specifically, Translation-Assisted Tuning defines auxiliary translation-related tasks and combines it with original multilingual tuning, while Cross-Lingual Tuning transforms multilingual tuning tasks into cross-lingual forms without explicitly relying on translation task.

It's worth noting that multilingual tuning here generally refers to multilingual instruction tuning, as most approaches perform cross-lingual alignment at this stage. The only exception is the work done by \citet{trans-10}, which models cross-lingual alignment as a preference optimization problem.

\subsubsection{Translation-Assisted Tuning}

Assuming the internal translation capability of MLLMs can facilitate cross-lingual alignment, Translation-Assisted Tuning methods define auxiliary translation-related tasks and combine them with original multilingual tuning. 
Such tasks involve machine translation \citep{trans-1, trans-2, trans-6, trans-5} or other variations \citep{trans-3, trans-4}.

The most straightforward approach is to convert parallel data into translation instruction format and incorporate them to the original multilingual instruction data. While \citet{trans-1} focused on bilingual scenarios, \citet{trans-2} initially explored the correlation between the scale of translation instructions and translation performance through bilingual training, which subsequently served as a reference for determining the allocation of translation instructions in multilingual settings. 

Specifically, for multilingual reasoning tasks, \citet{trans-5} proposed translation instruction tuning of parallel reasoning questions (Question Alignment) before English instruction tuning for reasoning tasks, thereby facilitating the transfer of English reasoning capabilities of MLLMs to non-English languages.

Similarly, starting with Question Alignment, \citet{trans-6} further integrated general multilingual translation instructions (from English to non-English) with English instruction tuning for reasoning tasks, enhancing the ability of MLLMs to generate multilingual reasoning outputs.

Other studies explore variations of traditional machine translation tasks and transform them into instruction tuning formats. 

Specifically, \citet{trans-3} extended translation instruction tuning to a multi-turn interactive setting, aiming to simultaneously enhance the cross-lingual alignment and instruction-following capabilities of MLLMs. 

Additionally, \citet{trans-4} introduced a cross-lingual semantic similarity task, where LLMs are trained to determine the semantic relationship between two parallel sentences. They also introduced a bilingual denoising task, training MLLMs to reconstruct the target side of input parallel sentences that have been noised.

\subsubsection{Cross-Lingual Tuning}

Without relying on auxiliary translation-related tasks, the approaches in Cross-Lingual Tuning reconstruct multilingual instruction tuning into a cross-lingual form, enhancing the cross-lingual alignment of MLLMs during instruction learning in a more integrated manner \citep{si2024mpn}. 

In \citep{trans-7}, cross-lingual signals are introduced by specifying instructions in non-English languages and responses in English. Certain tokens in the non-English instructions may be replaced with English to construct code-switched instructions.

In \citep{trans-8} and \citep{trans-9}, MLLMs were trained to respond to non-English instructions by first thinking in English and then responding in a non-English language. When the input is a non-English instruction, MLLMs are trained to sequentially output the English translation of the instruction, an English response, and finally a non-English response. The Random Online CoT introduced by \citet{trans-7} serves a similar purpose.

Other works promote cross-lingual alignment by enhancing the cross-lingual consistency of model responses. 
When responses are required to be in English, the Cross-lingual Distillation method introduced by \citet{trans-7} minimizes the difference between generated English responses when inputs are in different languages. 
For multilingual responses, \citet{trans-10} employed high-performing multilingual translation models (e.g., NLLB-600M-distilled) to compute cross-lingual consistency between non-English and English responses. 
They further framed cross-lingual alignment as a preference optimization problem, utilizing the computed cross-lingual consistency as rewards for preference tuning.

\subsection{Enhancement of Specific Multilingual Abilities}

The direct multilingual tuning and cross-lingual alignment discussed above serve as foundational strategies for enhancing MLLMs' capabilities in multilingual instruction following, preference alignment and complex reasoning \citep{zhao2024adamergex}. 
Beyond this, there is interest in more multilingual abilities of MLLMs, including diverse capabilities of understanding and generation. Those specific multilingual abilities can be enhanced using particular techniques, which we will address here.

We will focus on three of such abilities: adaptation to new languages, machine translation and cultural adaptation, as they inherently involve multiple languages or cultures.
Adaptation to new languages primarily involves adapting English-centric MLLMs to other languages; machine translation focuses on translating between different languages; and cultural adaptation addresses the cultural diversity of our society, with language considered one of the proxies of culture.

\subsubsection{Adaptation to New Languages}

Although the techniques employed for adaptation to new languages are general, primarily focusing on vocabulary expansion and continual pre-training, the target languages for adaptation vary across different works \citep{adapt-1, adapt-2, adapt-3, adapt-4, adapt-5}. 

For instance, \citet{adapt-1} aimed to extend English-based LLMs to Chinese.
To enhance the efficiency of MLLMs in encoding and decoding Chinese text, they trained an additional tokenizer on Chinese corpora before conducting continual pre-training, expanding the original vocabulary of MLLMs. 

Other studies extend the vocabulary and perform continual pre-training on different target languages, such as Korean \citep{adapt-2}, Japanese \citep{adapt-3}, or multiple languages \citep{adapt-4, adapt-5}, and so forth.

Additionally, \citet{adapt-2} proposed a seven-stage continuous pre-training strategy to enable LLMs to better learn new token representations.

Adaptation to new languages presents a significant challenge when data for the target languages are not available. 

In one such case, to enable monolingual instruction-tuned MLLMs to perform multilingual mathematical reasoning tasks, \citet{adapt-6} employed an external multilingual encoder and initialized a trainable linear layer as a bridge between the encoder and the MLLMs. 
They enhanced them with multilingual understanding capabilities by aligning the representation space of this encoder and the MLLMs.

In another case, considering the high computational and data resource costs of the strategy of first expanding the vocabulary and then performing continual pre-training, \citet{adapt-7} advocated for more efficient reuse of the original vocabulary rather than for expanding it. 
Given that English-centric vocabularies primarily consist of Roman letters, they proposed romanizing the text of new languages. 
Factors considered in this romanization scheme include the similarity between the romanized text and the typical writing style, compatibility with the original LLM tokenizer, and the lossiness of the conversion process.

Other studies conduct comparative analyses of key techniques for adapting to new languages \citep{adapt-8, adapt-9, adapt-10}. 

Specifically, \citet{adapt-8} compared three languages adaptation strategies, viz. continual pre-training, MAD-X \citep{MAD-X}, and $(\mathrm{IA})^3$ \citep{IA3}, and experimentally found that adapter-based methods are more effective than continual pre-training for MLLMs. 

On the other hand, \citet{adapt-9} investigated the suitability of vocabulary expansion, concluding that such an extension may not be advantageous when the scale of the training dataset is below billions of tokens. 

Similar to \citep{adapt-9}, \citet{adapt-10} suggested that while expanding the vocabulary for new languages may accelerate encoding efficiency, it may not necessarily improve performance. 
They also gathered empirical evidence that training from a base model rather than an instruction-tuned model could result in better languages adaptation.

\subsubsection{Machine Translation}

Many works use MLLMs to implements machine translation task \citep{zhu2024towards,zhu2024feds,cui2024efficiently,zhu2024landermt,chen2024dual,zhang2024paying}.
To adapt MLLMs to machine translation, \citet{mt-1} utilized extensive parallel corpora (exceeding 300M) for fine-tuning.

However, it has been observed that using excessive parallel data (e.g., 5M or 20M) for translation task training on well pre-trained MLLMs can have adverse effects, potentially due to catastrophic forgetting \citep{mt-2}. 
To address this issue, \citet{mt-2} proposed a two-stage fine-tuning approach. Initially, they conducted continual pre-training on the languages involved in the translation task to enhance MLLMs' proficiency in those languages. 
Then, they performed translation instruction tuning on a small scale (e.g., 58K) but with high-quality parallel data. 

Other studies \citep{mt-3, mt-4} conducted detailed analyses on multilingual translation instruction tuning, taking into account factors such as language similarity, the volume of pre-training and instruction tuning data, and the translation directions.

Although commonly used to enhance MLLMs' translation capabilities, translation instruction tuning is regarded as having certain limitations. 
On the one hand, the quality of translation instruction data can restrict the performance of MLLMs \citep{mt-5}. 
On the other hand, MLLMs fine-tuned with translation instructions may overlook specific contextual knowledge, forget instructions, and consequently encounter issues such as hallucinations, over-translation or translation omissions \citep{mt-6, mt-7, mt-8}.

To address these challenges, some efforts have opted for preference tuning to further enhance the translation capabilities of MLLMs. 

To build preference pairs, \citet{mt-5} employed different MLLMs to sample translation results, which were then scored by reference-free evaluation models. 

To efficiently collect high-quality preference data, \citet{mt-9} aligned multilingual versions of books to acquire human translations, forming preference pairs alongside machine-generated translations. 

Additionally, \citet{mt-6} utilized external word aligners to annotate the degree of word alignment for translation candidates, while \citet{mt-7} employed output comparison and preference comparison strategies to construct preference data.

Instead of performing preference tuning, \citet{mt-8} proposed to strengthen the global representations of instructions, aiming to mitigate MLLMs' tendency to forget instructions.

The lack of cross-lingual alignment may also affect machine translation. To enhance cross-lingual alignment, \citet{mt-10} introduced XConST, which employs Kullback-Leibler (KL) regularization on semantically equivalent parallel sentences. 

Furthermore, \citet{mt-11} initially utilized external tools to extract statistical word alignment signals and then trained MLLMs to discern these signals.

\citet{mt-12} proposed Relay Decoding, a method aligning two LLMs supporting the source and target languages by training a straightforward mapping layer. 

Additionally, \citet{mt-13} proposed m3P, a multimodal neural machine translation framework to bridge language disparities through universal visual features.

\subsubsection{Cultural Adaptation} 

As research continues to advance MLLMs in their capabilities, assessing and enriching their cultural and value diversity becomes increasingly crucial, especially considering their extensive utilization across diverse populations worldwide \citep{c-1}. 

Significant efforts have been dedicated to exploring the multicultural knowledge embedded within LLMs, as well as the extent of their cultural adaptation and alignment \citep{c-3, c-4, c-5, c-6, c-7, c-8, c-9, c-10, c-11, c-12, c-13, c-14, c-15, c-16, c-17, DBLP:journals/corr/abs-2404-16019, c-19, c-20, c-21, c-22, c-23, c-24, c-25, c-26, c-27, c-28, c-29, c-30, c-31, c-32, c-33}. 

However, efforts to fine-tune MLLMs for cultural adaptation are relatively sparse, or limited to English.

Recently, \citet{c-34} introduced CultureLLM, a methodology aimed at instilling MLLMs with cultural differences. 
First, they collected seed data from the World Values Survey \citep{wsv}, spanning 9 cultures, and utilized GPT-4 for data augmentation. Subsequently, they fine-tuned LLMs on the augmented data, creating culture-specific models for each culture alongside a unified model. 

Furthermore, \citet{c-34-1} proposed CultureSPA, a framework that achieves pluralistic cultural alignment in MLLMs by leveraging their internal cultural knowledge. This framework involves generating diverse questions from seed prompts, yielding both culture-unaware and culture-aware MLLM outputs, collecting culture-related QA pairs and conducting culture-joint and specific SFT. They validated the effectiveness of the method across 18 cultures spanning five continents.

With a proposed technique for extracting culture-related instruction data from unstructured data, \citet{c-35} performed instruction tuning on this data to enhance the cultural reasoning abilities of MLLMs.

Despite recognizing that language often acts as a proxy for culture \citep{c-2}, these works employed English for data collection, leaving exploration in multilingual contexts for future research.

Other works analyze the relationship between cultural adaptation and multilingual instruction tuning \citep{c-36, c-37}. 

Specifically, \citet{c-36} conducted multilingual instruction tuning using the translated Alpaca dataset, investigating the influence of language-specific instruction tuning and pre-training data on cultural adaptation, as well as the optimal method to elicit cultural knowledge from MLLMs. 

In turn, \citet{c-37} explored the impact of selected languages and data sources on the shift of cultural values during model fine-tuning.

\section{Multilingual Evaluation}
\label{Multilingual Evaluation of LLMs}

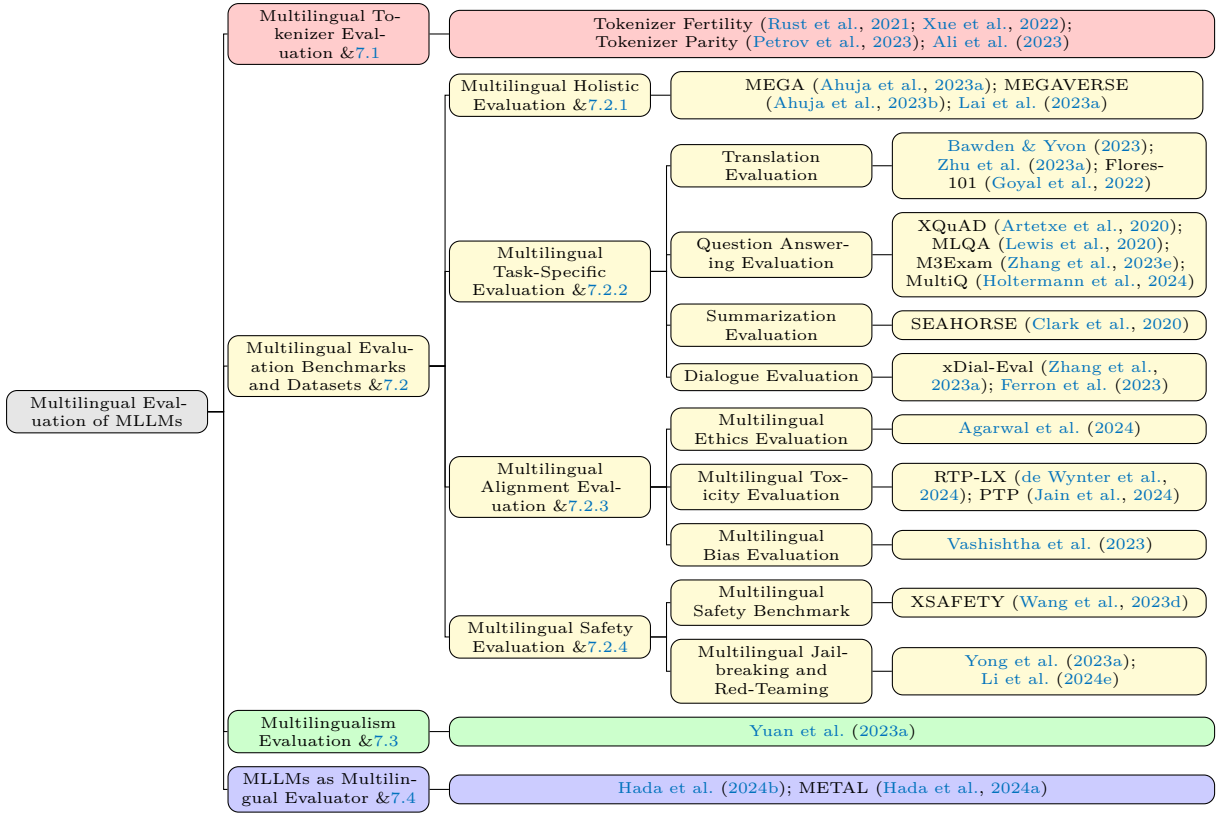
\begin{figure*}[t]
    \tiny
    \begin{forest}
        for tree={
            forked edges,
            grow'=0,
            draw,
            rounded corners,
            node options={align=center},
            calign=edge midpoint,
        },
        [Multilingual Evaluation of MLLMs, text width=2.5cm, fill=black!10
            [Multilingual Tokenizer Evaluation  \&\ref{Multilingual Tokenizer Evaluation}, text width=2.5cm, for tree={fill=red!20}
                    [                    
                    Tokenizer Fertility \citep{DBLP:conf/acl/RustPVRG20,xue-etal-2022-byt5};
                    Tokenizer Parity \citep{DBLP:conf/nips/PetrovMTB23};
                    \citet{DBLP:journals/corr/abs-2310-08754},
                    text width=10.0cm
                    ]
            ]
            [Multilingual Evaluation Benchmarks and Datasets  \&\ref{Multilingual Evaluation Benchmarks and Datasets}, text width=2.5cm, for tree={fill=yellow!20}
                    [Multilingual Holistic Evaluation  \&\ref{Multilingual Holistic Evaluation}, text width=2.5cm, for tree={fill=yellow!20}
                        [
                            MEGA \citep{DBLP:conf/emnlp/AhujaDHORJNGSAB23};
                            MEGAVERSE \citep{DBLP:journals/corr/abs-2311-07463};
                            \citet{DBLP:conf/emnlp/LaiNVMDBN23},
                            text width=6.9cm
                        ]
                    ]
                    [Multilingual Task-Specific Evaluation   \&\ref{Multilingual Task-Specific Evaluation}, text width=2.5cm, for tree={fill=yellow!20}
                        [Translation Evaluation, text width=2.5cm, for tree={fill=yellow!20}
                            [
                            \citet{DBLP:conf/eamt/BawdenY23}; \citet{DBLP:journals/corr/abs-2304-04675};
                            Flores-101 \citep{DBLP:journals/tacl/GoyalGCCWJKRGF22},
                            text width=4.0cm]
                        ]
                        [Question Answering Evaluation, text width=2.5cm, for tree={fill=yellow!20}
                        [
                            XQuAD \citep{DBLP:conf/acl/ArtetxeRY20}; MLQA \citep{DBLP:conf/acl/LewisORRS20};
                            M3Exam \citep{DBLP:conf/nips/ZhangAGCB23};
                            MultiQ \citep{DBLP:journals/corr/abs-2403-03814},
                            text width=4.0cm]
                        ]
                        [Summarization Evaluation, text width=2.5cm, for tree={fill=yellow!20}
                            [
                            SEAHORSE \citep{DBLP:journals/tacl/ClarkPNCGCK20},
                            text width=4.0cm]
                        ]
                        [Dialogue Evaluation, text width=2.5cm, for tree={fill=yellow!20}
                        [
                            xDial-Eval \citep{DBLP:conf/emnlp/ZhangDTST023};
                            \citet{DBLP:conf/emnlp/FerronSMA23},
                            text width=4.0cm]
                        ]
                    ]
                    [Multilingual Alignment Evaluation   \&\ref{Multilingual Alignment Evaluation}, text width=2.5cm, for tree={fill=yellow!20}
                        [Multilingual Ethics Evaluation, text width=2.5cm, for tree={fill=yellow!20}
                            [\citet{DBLP:conf/coling/AgarwalTKC24},
                            text width=4.0cm]
                        ]
                        [Multilingual Toxicity Evaluation, text width=2.5cm, for tree={fill=yellow!20}
                        [
                            RTP-LX \citep{DBLP:journals/corr/abs-2404-14397};
                            PTP \citep{jain2024polyglotoxicityprompts},
                            text width=4.0cm]
                        ]
                        [Multilingual Bias Evaluation, text width=2.5cm, for tree={fill=yellow!20}
                            [
                            \citet{DBLP:conf/acl/VashishthaAS23},
                            text width=4.0cm]
                        ]
                    ]
                    [Multilingual Safety Evaluation   \&\ref{Multilingual Safety Evaluation}, text width=2.5cm, for tree={fill=yellow!20}
                        [Multilingual Safety Benchmark, text width=2.5cm, for tree={fill=yellow!20}
                            [XSAFETY \citep{DBLP:journals/corr/abs-2310-00905},
                            text width=4.0cm]
                        ]
                        [Multilingual Jailbreaking and Red-Teaming, text width=2.5cm, for tree={fill=yellow!20}
                        [
                            \citet{DBLP:journals/corr/abs-2310-02446};
                            \citet{DBLP:journals/corr/abs-2401-16765},
                            text width=4.0cm]
                        ]
                    ]
            ]
            [Multilingualism Evaluation \&\ref{Multilingualism Evaluation}, text width=2.5cm, for tree={fill=green!20}
                [
                \citet{DBLP:journals/corr/abs-2311-09071},
                text width=10.0cm
                ]
            ]
            [MLLMs as Multilingual Evaluator \&\ref{LLMs as Multilingual Evaluator}, text width=2.5cm, for tree={fill=blue!20}
                [
                \citet{DBLP:conf/eacl/HadaGWDACBS24};
                METAL \citep{DBLP:journals/corr/abs-2404-01667},
                text width=10.0cm
                ]
            ]
        ]
    \end{forest}
    \caption{Multilingual evaluation of MLLMs.}
    \label{fig:Multilingual Evaluation of MLLMs}
\end{figure*}

The evaluation of MLLMs are crucial to have a better understanding towards their performance and capabilities, being also important to assess human-values compliance and to address safety concerns \citep{zhang2024enhancing,park2024multiprageval,shen2024language,doddapaneni2024cross}. 
In multilingual contexts, the evaluation should be broaden towards diverse languages, where the performance towards each language should be understandable. 
In this section, we discuss the multilingual evaluation of MLLMs, focusing on tokenizers, benchmarks and datasets, multilingualism of MLLMs, and MLLMs as multilingual evaluators.
The framework in this section is shown in Figure \ref{fig:Multilingual Evaluation of MLLMs}.

\subsection{Multilingual Tokenizer Evaluation}
\label{Multilingual Tokenizer Evaluation}

A tokenizer plays an important role for supporting multilingual processes in LLMs, as tokenizers separate sentences into tokens and map them into numerical ids that are the input for MLLMs.
Tokenizers that supports multilingual tokens are crucial to enhance the performance of MLLMs.

The fertility of a tokenizer is a metric aimed at assessing its quality, which is defined as the average number of sub-words produced per tokenized word \citep{DBLP:conf/acl/RustPVRG20}. A lower fertility score indicates a better quality tokenizer.
Fertility is also known as the tokenizer's compression rate \citep{xue-etal-2022-byt5}. 

Fertility tests have been conducted to evaluate tokenizers on multilingual sentences across various MLLMs, including BLOOM \citep{DBLP:journals/corr/abs-2211-05100} and OpenAI models \citep{DBLP:conf/emnlp/AhujaDHORJNGSAB23}. Based on their findings, the fertility score of OpenAI tokenizers is higher in low-resource languages compared to BLOOM's tokenizer.

Another way to evaluate a tokenizer is using parity, which is introduced in \citep{DBLP:conf/nips/PetrovMTB23} and it is motivated by the unequal treatment of tokenization across different languages. 
For instance, a Japanese kanji character can be tokenized into three tokens in GPT-2. To address this issue, the concept of tokenizer parity is introduced to evaluate how fairly tokenizers treat equivalent sentences in different languages. 
A tokenizer achieves parity when the ratio of the tokenization results for a sentence in language A compared to a sentence in language B is almost equal to 1.

A comprehensive evaluation of tokenizers was undertaken by \citet{DBLP:journals/corr/abs-2310-08754}, dividing the assessment into intrinsic and extrinsic evaluations. 

The intrinsic evaluation focuses solely on metrics like fertility and parity. 

The extrinsic evaluation evaluates the impact of the tokenizer on the performance of LLMs, with respect to downstream tasks performance and computational costs. 

The research found that tokenizers trained with a balanced share across languages achieve better fertility and parity scores. The impact of worst, higher fertility will increase computational costs.

\subsection{Multilingual Evaluation Benchmarks and Datasets}
\label{Multilingual Evaluation Benchmarks and Datasets}

\begin{table*}[t]
\resizebox{\textwidth}{!}{%
\begin{tabular}{l|cll}
\hline
\textbf{Name} &
  \textbf{\#Languages} &
  \textbf{Language Family} &
  \textbf{Type} \\ \hline
XNLI \citep{DBLP:conf/emnlp/ConneauRLWBSS18}&
  15 &
  Indo-European, Turkic, Afro-Asiatic, Austronesian, Austro-Asiatic, &
  Natural Language Inference \\
   & &
  Tai-Kadai, Sino-Tibetan, Atlantic-Congo & \\
Indic-XNLI \citep{DBLP:conf/emnlp/Aggarwal0K22}&
  11 &
  Indo-European, Dravidian &
  Natural Language Inference \\
GlueCoS \citep{DBLP:conf/acl/KhanujaDSSC20}&
  3 &
  Indo-European, Dravidian &
  Language Identification, POS Tagging, \\
   &
   &
   & NER, Sentiment Analysis, QA, NLI \\
XCOPA \citep{DBLP:conf/emnlp/PontiGMLVK20}&
  11 &
  Uralic, Indo-European,  Austronesian, Quechuan, Atlantic-Congo,  &
  Commonsense Reasoning \\ 
   &
   &
  Dravidian, Kra-Dai, Turkic, Sino-Tibetan, Austro-Asiatic &
   \\ 
XStoryCloze \citep{DBLP:conf/emnlp/LinMAWCSOGBDPSK22}&
  10 &
  Indo-European, Sino-Tibetan, Afro-Asiatic, Austronesian,&
  Reasoning \\
   &
   &
  Dravidian, Atlantic-Congo, Basque &
   \\
PAWS-X \citep{DBLP:conf/emnlp/YangZTB19}&
  6 &
  Indo-European, Sino-Tibetan,   Japonic, Koreanic &
  Paraphrase Identification \\
EN-ES-CS \citep{DBLP:conf/lrec/VilaresAG16}&
  2 &
  Indo-European &
  Sentiment Analysis \\
XQuAD \citep{DBLP:conf/acl/ArtetxeRY20}&
  11 &
  Indo-European, Sino-Tibetan, Austro-Asiatic &
  Question Answering \\
MLQA \citep{DBLP:conf/acl/LewisORRS20}&
  7 &
  Indo-European, Afro-Asiatic, Austro-Asiatic, Sino-Tibetan &
  Question Answering \\
TyDiQA-GOLDP \citep{DBLP:journals/tacl/ClarkPNCGCK20}&
  11 &
  Indo-European, Afro-Asiatic, Uralic, Japonic, &
  Question Answering \\
 &
   &
   Atlantic-Congo, Koreanic, Dravidian, Kra-Dai &
   \\
IndicQA \citep{DBLP:journals/corr/abs-2212-05409}&
  11 &
  Indo-European, Dravidian &
  Question Answering \\
PAN-X/WikiANN \citep{DBLP:conf/acl/PanZMNKJ17}&
   &
  Indo-European, Afro-Asiatic, Turkic, Sino-Tibetan, Austronesian,&
  Named Entity Recognition,  \\
 &
   &
   Uralic, Atlantic-Congo, Kartvelian,   Japonic, Dravidian,  &
   \\
 &
   &
  Koreanic, Kra-Dai, Nilo-Saharan, Austro-Asiatic, Mongolic &
   \\
UD v2.2 \citep{UD22}&
  71 &
  Indo-European, Afro-Asiatic,   Basque, Sino-Tibetan, Uralic, Austronesian,  &
  Sequence Labeling \\
 &
   &
  Japonic, Turkic, Koreanic, Atlantic-Congo, Dravidian, Kra-Dai, Austro-Asiatic &
   \\
XL-Sum \citep{DBLP:conf/acl/HasanBIMLKRS21}&
  47 &
  Afro-Asiatic, Turkic, Austro-Asiatic, Sino-Tibetan, Indo-European, Japonic, &
  Summarization \\
 &
   &
   Koreanic, Dravidian, Atlantic-Congo, Austronesian, Austro-Asiatic, Kra-Dai &
   \\
Jigsaw&
  6 &
  Indo-European, Turkic &
  Toxic Classification \\
Wino-MT \citep{DBLP:conf/acl/StanovskySZ19}&
  8 &
  Indo-European, Afro-Asiatic &
  Gender Bias in Translation \\
Belebele \citep{DBLP:journals/corr/abs-2308-16884}&
  122 &
  Afro-Asiatic, Indo-European, Turkic, Mande, Sino-Tibetan,  &
  Question Answering \\
 &
   &
  Austronesian, Uralic, Basque, Atlantic-Congo,   Tupian, Japonic, Dravidian, &
   \\
 &
   &
   Kartvelian, Mongolic, Austroasiatic, Koreanic,   Kra-Dai, Nilo-Saharan &
   \\
AfriQA \citep{DBLP:journals/corr/abs-2305-06897}&
  10 &
  Atlantic-Congo, Afro-Asiatic &
  Question Answering \\
XRiSAWOZ \citep{DBLP:conf/acl/MoradshahiSBCCG23}&
  5 &
  Indo-European, Koreanic,   Sino-Tibetan &
  Dialogue \\
IN22 \citep{DBLP:journals/corr/abs-2305-16307}&
  22 &
  Indo-European, Austronesian,   Dravidian &
  Translation \\
MaRVL \citep{DBLP:conf/emnlp/0001BPRCE21}&
  5 &
  Austronesia,   Sino-Tibetan, Atlantic-Congo, Dravidian, Turkic &
  Image-Text   Reasoning \\
XM-3600 \citep{DBLP:conf/emnlp/ThapliyalPCS22}&
  36 &
  Afro-Asiatic,   Indo-European, Uralic, Austronesian, Japonic, Koreanic, Quechuan &
  Image Captioning \\
MultiCoNER \citep{DBLP:conf/semeval/MalmasiFFKR22}&
  11 &
  Sino-Tibetan,   Indo-European, Koreanic, Turkic &
  NER \\
SMiLER \citep{DBLP:conf/eacl/SegantiFSSA21}&
  14 &
  Koreanic,   Indo-European, Afro-Asiatic &
  Relation   Extraction \\
X-CSQA \citep{DBLP:conf/acl/LinLQ020}&
  15 &
  Indo-European,   Japonic, Sino-Tibetan, Afro-Asiatic, Atlantic-Congo, Austro-Asiatic &
  Commonsense   Reasoning \\
Wikipedia Cloze QA \citep{kakwani-etal-2020-indicnlpsuite}&
  11 &
  Indo-European, Dravidian &
  Question   Answering \\
Flores-101 \citep{DBLP:journals/tacl/GoyalGCCWJKRGF22}&
  101 &
  Indo-European,   Afro-Asiatic, Turkic, Sino-Tibetan, Austronesian, Uralic,   &
  Translation \\
 &
   &
  Atlantic-Congo, Kartvelian, Japonic, Dravidian, Koreanic,  &
   \\
 &
   &
  Kra-Dai, Nilo-Saharan,   Austro-Asiatic, Mongolic &
   \\
M3Exam \citep{DBLP:conf/nips/ZhangAGCB23}&
  9 &
  Indo-European,   Sino-Tibetan, Austro-Asiatic, Kra-Dai, Atlantic-Congo &
  Question Answering \\
MultiQ \citep{DBLP:journals/corr/abs-2403-03814}&
  137 &
  Afro-Asiatic, Altaic,   Austro-Asiatic, Austronesian, Aymaran, Basque,  &
  Question Answering \\
 &
   &
  Dravidian, Indo-European,   Kra-Dai, Japonic, Koreanic, Kartvelian, Mande, &
   \\
 &
   &
   Atlantic-Congo, Quechuan,   Sino-Tibetan, Tupian, Uralic, Turkic &
  \\
SEAHORSE \citep{DBLP:journals/tacl/ClarkPNCGCK20}&
  6 &
  Indo-European, Turkic,   Austro-Asiatic &
  Summarization \\
xDial-Eval \citep{DBLP:conf/emnlp/ZhangDTST023}&
  9 &
  Sino-Tibetan, Indo-European,   Japonic, Koreanic, Afro-Asiatic &
  Dialogue \\
RTP-LX \citep{DBLP:journals/corr/abs-2404-14397}& 27
   & Afro-Asiatic, Indo-European, Austronesian, Kra-Dai, Atlantic-Congo, 
   & Toxicity
   \\
 & 
   & Sino-Tibetan, Japonic, Koreanic, Turkic, Uralic
   & Toxicity
   \\
PolygloToxicityPrompts \citep{jain2024polyglotoxicityprompts}& 17
   & Afro-Asiatic, Sino-Tibetan, Indo-European, Austronesian, Japonic, Koreanic
   & Toxicity
   \\
XSAFETY \citep{DBLP:journals/corr/abs-2310-00905} & 10
   & Indo-European, Sino-Tibetan, Afro-Asiatic, Japonic
   & Safety
   \\
MultiJail \citep{DBLP:journals/corr/abs-2310-06474}& 9
   & Sino-Tibetan, Indo-European, Austro-Asiatic, Afro-Asiatic, 
   & Jailbreaking
   \\
  & & Koreanic, Kra-Dai, Atlantic-Congo, Austronesian
   & 
   \\
\hline
  
\end{tabular}
}
\caption{Multilingual evaluation datasets.}
\label{tab:eval-data}
\end{table*}

The evaluation of MLLMs should cover various dimensions, including accuracy, alignment with human values and safety. 
In the context of multilingual evaluation, these dimensions should be evaluated for each language, from high-resource to low-resource languages.
In this section, we divide the evaluation benchmarks of MLLMs into a holistic evaluation and a task-specific, alignment and safety evaluation.
We summarize the available datasets in Table \ref{tab:eval-data}.

\subsubsection{Multilingual Holistic Evaluation}
\label{Multilingual Holistic Evaluation}

Several studies have conducted comprehensive multilingual evaluations of MLLMs, covering diverse languages and tasks. 
\citet{DBLP:conf/emnlp/AhujaDHORJNGSAB23} proposed MEGA, the first benchmark for evaluating generative AI in a multilingual context. 
This benchmark covers five categories of NLP tasks: classification, question answering, sequence labeling, natural language generation, and responsible AI.
These categories are represented by 16 datasets encompassing 70 languages. 

Concerning the classification task, the datasets used are XNLI \citep{DBLP:conf/emnlp/ConneauRLWBSS18}, Indic-XNLI \citep{DBLP:conf/emnlp/Aggarwal0K22}, GLUECos NLI \citep{DBLP:conf/acl/KhanujaDSSC20}, XCOPA \citep{DBLP:conf/emnlp/PontiGMLVK20}, XStoryCloze \citep{DBLP:conf/emnlp/LinMAWCSOGBDPSK22}, PAWS-X \citep{DBLP:conf/emnlp/YangZTB19}, and EN-ES-CS \citep{DBLP:conf/lrec/VilaresAG16}.

For the question answering (QA) task, the dataset used for this task are XQuAD \citep{DBLP:conf/acl/ArtetxeRY20}, MLQA \citep{DBLP:conf/acl/LewisORRS20},  TyDiQA-GoldP \citep{DBLP:journals/tacl/ClarkPNCGCK20}, and IndicQA \citep{DBLP:journals/corr/abs-2212-05409}. 

For sequence labeling task, PAN-X \citep{DBLP:conf/acl/PanZMNKJ17} and UDPOS \citep{UD22} datasets were used. 

For the natural language generation, XL-Sum \citep{DBLP:conf/acl/HasanBIMLKRS21} dataset was resorted to. 

For responsible AI evaluation, Jigsaw \citep{jigsaw-multilingual-toxic-comment-classification} and WinoMT \citep{DBLP:conf/acl/StanovskySZ19} datasets were used.

In the next version, MEGAVERSE \citep{DBLP:journals/corr/abs-2311-07463} was distributed, an expanded version of MEGA multilingual evaluation benchmark. 
This benchmark was expanded to 22 datasets covering 81 languages.

The additional datasets include Belebele \citep{DBLP:journals/corr/abs-2308-16884}, a multiple choice machine reading comprehension in 122 languages; AfriQA \citep{DBLP:journals/corr/abs-2305-06897}, a question answering dataset for 10 African languages; XRiSAWOZ \citep{DBLP:conf/acl/MoradshahiSBCCG23}, a task oriented dialogue modeling dataset originally in Chinese and translated into English, Hindi, French, Korean and English-Hindi code-mixed setting; IN22 \citep{DBLP:journals/corr/abs-2305-16307}, a translation benchmark for all 22 scheduled Indic languages; MaRVL \citep{DBLP:conf/emnlp/0001BPRCE21}, a multicultural reasoning over vision and language dataset in 5 distinct languages, comprises of image and its caption; XM-3600 \citep{DBLP:conf/emnlp/ThapliyalPCS22}, a multilingual image captioning dataset consisting of 3600 geographically diverse images directly captioned in 36 different languages.

Another work explored the multilingual evaluation of ChatGPT \citep{DBLP:conf/emnlp/LaiNVMDBN23}. 
This evaluation was conducted for 7 diverse NLP tasks in 37 diverse languages. 

In part-of-speech (POS) tagging task, XGLUE-POS \citep{DBLP:conf/emnlp/LiangDGWGQGSJCF20} dataset was used, which covers 18 languages. For named entity recognition (NER) task, MultiCoNER \citep{DBLP:conf/semeval/MalmasiFFKR22} dataset that supports 11 languages was resorted to. For relation extraction (RE) task, SMiLER \citep{DBLP:conf/eacl/SegantiFSSA21} dataset contains 14 languages was used. For natural language inference (NLI) and QA task, the dataset used is the same as MEGA benchmark, the XNLI and XQuAD dataset. For common sense reasoning (CSR) task, two datasets were included: X-CSQA \citep{DBLP:conf/naacl/TalmorHLB19,DBLP:conf/acl/LinLQ020} in English and the translation into 15 languages, and Wikipedia Cloze QA from IndicNLPSuite \citep{kakwani-etal-2020-indicnlpsuite} in 11 low-resource Indian languages.

\subsubsection{Multilingual Task-Specific Evaluation}
\label{Multilingual Task-Specific Evaluation}

LLMs are able to solve diverse kind of tasks, from question answering to dialogue generation and including translation, among many others. 
The trained MLLMs have strengths in supporting tasks in diverse kind of languages as they are trained in multilingual datasets. 
Recent studies have explored the multilingual evaluation of LLMs in specific tasks.

\paragraph{Translation Evaluation} Numerous studies have demonstrated that LLMs possess the ability to perform multilingual translation tasks. 
Research has specifically investigated the multilingual translation capabilities of models like BLOOM \citep{DBLP:conf/eamt/BawdenY23} and diverse kinds of LLMs \citep{DBLP:journals/corr/abs-2304-04675}. 

BLOOM's translation abilities were evaluated in \citep{DBLP:conf/eamt/BawdenY23} by resorting to several datasets and exploring various prompting strategies. 
This evaluation focused on English-French and English-Hindi parallel corpora and utilized the Flores-101 dataset \citep{DBLP:journals/tacl/GoyalGCCWJKRGF22}. 
The empirical results indicate that few-shot prompting strategies significantly improve translation quality. 
However, translating low-resource languages remains challenging, even when the language is included in the training data.

For a more extensive study, \citet{DBLP:journals/corr/abs-2304-04675} evaluated LLMs in 102 languages and 606 translation directions that are English-centric, French-centric and Chinese-centric.
MLLMs were compared with machine translation models with varying model sizes, resorting to the Flores-101 dataset \citep{DBLP:journals/tacl/GoyalGCCWJKRGF22}. 

 
\paragraph{Question Answering Evaluation} MLLMs can provide valuable knowledge and insights on the basis of the questions that are prompted. 
The most prominent datasets for the evaluation of this task are XQuAD \citep{DBLP:conf/acl/ArtetxeRY20} and MLQA \citep{DBLP:conf/acl/LewisORRS20}. 

In MLLMs in multilingual context, several studies have undertaken evaluation of multilingual question answering.
M3Exam was proposed by \citet{DBLP:conf/nips/ZhangAGCB23}, a benchmark to evaluate LLMs when they handle human exam questions in multilingual, multimodal and multilevel contexts. This dataset contains more than 12K questions in 9 languages by gathering the official exams from different countries. 
It has also questions in multimodal context where questions are provided in text and need to be answered based on images. 
It is divided into 3 levels: primary, middle and high school.

For more diverse languages evaluation, MultiQ was proposed in \citet{DBLP:journals/corr/abs-2403-03814}, a benchmark to evaluate basic question answering task in 137 languages. 
This dataset comprises more than 27K prompts. 
It is originally from 200 English prompts, 100 prompts gathered from LMSYS-Chat-1M \citep{DBLP:journals/corr/abs-2309-11998} and 100 prompts curated manually by prompting GPT-4 to provide a question and the answer. 
These questions are then translated automatically into 136 other languages.

\paragraph{Summarization Evaluation} Summarization is a common task in natural language processing (NLP), where the input is a document and the output is another one that is a shorter version of it. 

SEAHORSE \citep{DBLP:journals/tacl/ClarkPNCGCK20} is a dataset for multilingual and multifaceted summarization evaluation. This dataset consists of 96K summaries in 6 languages and 9 different outputs (1 human summaries and 8 language models). 
It includes also ratings from human annotators along 6 dimensions: comprehensibility, repetition, grammar, attribution, main ideas and conciseness. 
The articles for the summarization are based on XSum \citep{DBLP:conf/emnlp/NarayanCL18}, XL-Sum \citep{DBLP:conf/acl/HasanBIMLKRS21}, MLSum \citep{DBLP:conf/naacl/PagnoniBT21}, and WikiLingua \citep{DBLP:conf/emnlp/LadhakDCM20}.

\paragraph{Dialogue Evaluation} With MLLMs, dialogues are supported in more than one languages, being thus important to evaluate dialogues for different languages.

xDial-Eval \citep{DBLP:conf/emnlp/ZhangDTST023} is a multilingual benchmark for evaluating the open-domain dialogue task. 
This dataset is sourced from the 12 turn-level and 6 dialogue-level datasets in English. The size of the collected datasets are 14930 annotated turns and 8691
annotated multi-turn dialogues. 
These datasets are translated into 9 diverse languages by utilizing machine translation, which was validated by human evaluation.

In another study, \citet{DBLP:conf/emnlp/FerronSMA23} evaluate the engagingness in dialogue, especially in multilingual contexts. 
An engaged reply is increasing the attention, interest and participation of the users. 
5 subdimensions of engagingness are proposed: response diversity, interactional quality, interestingness, contextual specificity and othering. 
4 turn- and dialogue-level datasets were used that cover 3 languages: English, Chinese, and Spanish.

\subsubsection{Multilingual Alignment Evaluation}
\label{Multilingual Alignment Evaluation}
The alignment of MLLMs towards human-values are crucial, by means of which LLMs should follow human preferences, not adhering to bias or producing toxic outputs \citep{wang2024probing}. 
It is thus important to evaluate MLLMs not only in their alignment with respect to English only, but also in multilingual terms.

\paragraph{Multilingual Ethics Evaluation}
MLLMs should follow human ethics and preferences also in multilingual contexts. 
An ethical MLLM is a MLLM that is able to 
discern what is morally good or bad in multilingual contexts. \citet{DBLP:conf/coling/AgarwalTKC24} proposed an evaluation benchmark in multilingual ethical reasoning across 7 different languages. 
The ethical categories are divided into virtue, deontology and consequentialism. 
This dataset is originally in English \citep{DBLP:conf/emnlp/RaoKTAC23} and it was translated automatically into 6 languages. 
The experimental results show that high-resource languages has a strong ethical reasoning capability, but the capability is lower in low-resource languages.


\paragraph{Multilingual Toxicity Evaluation}
Toxicity or toxic degeneration is defined as a disrespectful, rude or unreasonable text and make people leave a discussion \citep{jain2024polyglotoxicityprompts}. 
A benchmark for evaluating toxicity in English, named RealToxicityPrompts (RTP), was developed in \citep{DBLP:conf/emnlp/GehmanGSCS20}. 
For expanding the evaluation of toxicity into multilingual contexts, RTP-LX (RTP-Language eXpanded) dataset was curated \citep{DBLP:journals/corr/abs-2404-14397}. 
This dataset used RealToxicityPrompts as the seed corpus and added cultural-specific human-crafted prompts. It cover 27 languages with more than 1K prompts.

In another study, PolygloToxicityPrompts (PTP) was developed as the first large-scale multilingual benchmark to evaluate toxic degeneration in MLLMs \citep{jain2024polyglotoxicityprompts}. 
Differently from RTP-LX, this dataset is scraped from mC4 and THE PILE corpora. 
It consists of 425K prompts with the respective toxicity scores and covers 17 languages, with 25K prompts for each language. 

\paragraph{Multilingual Bias Evaluation}
Harmful biases need to be evaluated and mitigated, where
a most concerning issue is the gender bias. 
In multilingual bias evaluation, \citet{DBLP:conf/acl/VashishthaAS23} proposed two benchmarks to evaluate biases across languages. 
One of them is Discovery of Correlations (DisCo) \citep{DBLP:journals/corr/abs-2010-06032} concerning a measurement of unfairness or bias in MLLMs to predict particular gender. Another one uses multilingual bias evaluation (MBE) score \citet{DBLP:conf/naacl/KanekoIBO22} and it a bias evaluation dataset in 8 high-resource languages.

\subsubsection{Multilingual Safety Evaluation}
\label{Multilingual Safety Evaluation}
The safety of MLLMs is a major concern before deployment for public use. 
It is crucial to evaluate their robustness and risk mitigation against adversarial attacks, malicious inputs and the generation of unsafe outputs. 
In a multilingual context, safety evaluations must be conducted for each language, encompassing both high-resource and low-resource languages, ensuring that MLLMs maintain consistent safety standards across all languages.

\paragraph{Multilingual Safety Benchmark}
There are different kind of safety scenarios that must be addressed before deploying MLLMs, e.g. privacy, illegal activities and harmful contents. \citet{DBLP:journals/corr/abs-2310-00905} created the XSAFETY dataset, a multilingual safety dataset encompassing 14 different safety scenarios. 
This dataset is curated from the translation of SAFETYPROMPTS \citep{DBLP:journals/corr/abs-2304-10436} and SAFETEXT \citep{DBLP:conf/emnlp/LevyASCPMW22} datasets into 9 languages.
They evaluated several MLLMs, including ChatGPT, PaLM2, LLaMA2-Chat-13B and Vicuna-13B. Results indicate that all MLLMs produce more unsafe responses in non-English languages.

\paragraph{Multilingual Jailbreaking and Red-Teaming} 
In real-life scenarios, users may enter malicious inputs, such as harmful instructions or adversarial attacks. 
Jailbreaking or red-teaming are methods recently used to evaluate the robustness of LLMs. 
If MLLMs are not jailbroken and continue to produce safe content after red-teaming attempts, they are considered to be robust.

\citet{DBLP:journals/corr/abs-2310-02446} conducted multilingual jailbreaking towards GPT-4 with AdvBench dataset \citep{DBLP:journals/corr/abs-2307-15043}, which is translated into 12 languages that high-resource, medium-resource and low-resource. 
The results indicate that chances to jailbreak GPT-4 increases with the usage of low-resource languages. 

Jailbreaking in multilingual contexts were also studied by \citet{DBLP:journals/corr/abs-2310-06474}, who developed MultiJail, a dataset for jailbreaking sampled from Antrophic's red teaming dataset \citep{DBLP:journals/corr/abs-2209-07858} and translated into 9 languages that are high-resource, medium-resource and low-resource. 
The evaluation metric consists in counting attack success rates. 
Their findings suggest that low-resource languages can be potential targets for jailbreaking. 
In other words, the robustness of MLLMs is primarily focused on English, while other languages exhibit lower robustness.

Extensive empirical study of multilingual jailbreaking was undertaken in \citep{DBLP:journals/corr/abs-2401-16765}.
The dataset used was originally in English, inspired by previous studies, and it was translated into 8 languages with a semantic-preserving algorithm, with the final dataset containing 365 multilingual questions. 
This study also performed MLLM representation analysis, addressing the attention visualization of malicious questions. 
An empirical result obtained indicate that MLLMs focus on specific keywords in questions without jailbreak templates, leading to non-responses, while questions with templates have a more dispersed attention.

\subsection{Multilingualism Evaluation}
\label{Multilingualism Evaluation}

Recent approaches to evaluating the multilingual capabilities of MLLMs involve testing them with evaluation datasets and assessing the generated text \citep{zhao2024large,aggarwal2024maple}. 
However, a deep understanding of MLLMs' multilingualism remains underexplored. 

Some studies have conducted extensive analyses and used interpretability methods to investigate models' inner workings. 
For instance, \citet{DBLP:journals/corr/abs-2311-09071} proposed a benchmark to evaluate multilingual capabilities of MLLMs that when fine-tuned in one source language can be applied to other languages. 
It uses Embed FT, a method that only trains the embedding layer and keep the parameters frozen. 
Embed FT was trained with 101 language pairs, from and to English. 

To proceed with the analysis, languages were divided into four multilingual quality quadrants based on the results. 
The selfish quadrant contains languages whose model's capabilities only increase for the corresponding language pair. 
The reciprocal quadrant have languages whose training with bilingual data improves corresponding language pair and boost multilingual capabilities. 
The altruistic quadrant encompasses languages that primarily improve the multilingual performance. 
The idle quadrant contain languages that do not improve neither bilingual nor multilingual performance.

\subsection{MLLMs as Multilingual Evaluator}
\label{LLMs as Multilingual Evaluator}

Evaluating MLLMs is a complex and resource-intensive task, particularly when evaluating performance across numerous languages, diverse tasks, and various domains \citep{mu2024large}. 
Curating high-quality evaluation datasets demands significant resources, in terms of financial support and human annotators. 
To mitigate this, several studies have explored the use of MLLMs as evaluators for MLLMs, leveraging their capabilities to evaluate MLLMs in multilingual context \citep{DBLP:conf/eacl/HadaGWDACBS24, DBLP:journals/corr/abs-2404-01667}.

\citet{DBLP:conf/eacl/HadaGWDACBS24} explored the potential of an LLM-based evaluator, having resorted to GPT-4 for multilingual evaluation. 
They evaluate tasks in open prompt, continue writing and summarization in 8 languages: English, French, German, Spanish, Chinese, Japanese, Italian, Portuguese and Czech. 
The evaluation metrics including linguistic acceptability (LA), output content quality (OCQ), task quality (TQ), problematic content (PC), and hallucination (H). 
They performed inner-annotator agreement (IAA) analysis between human annotators and GPT-4.
The results indicate that GPT-4 demonstrates relatively high consistency for non-English languages.
However, there is a bias compared to human judgments in low-resource and non-Latin script languages.

METAL, an end-to-end multilingual meta-evaluation framework, was proposed in \citet{DBLP:journals/corr/abs-2404-01667} to assess MLLMs, as an evaluator in multilingual context. 
Firstly, the meta-evaluation dataset was created by prompting GPT-4 to generate data samples, whose evaluation is performed by native speakers.
With this curated dataset, covering 10 languages, with a total of 1000 summaries, the evaluation was conducted by prompting MLLMs with its test examples and evaluating the outcome with five metrics, following previous work \citep{DBLP:conf/eacl/HadaGWDACBS24}.

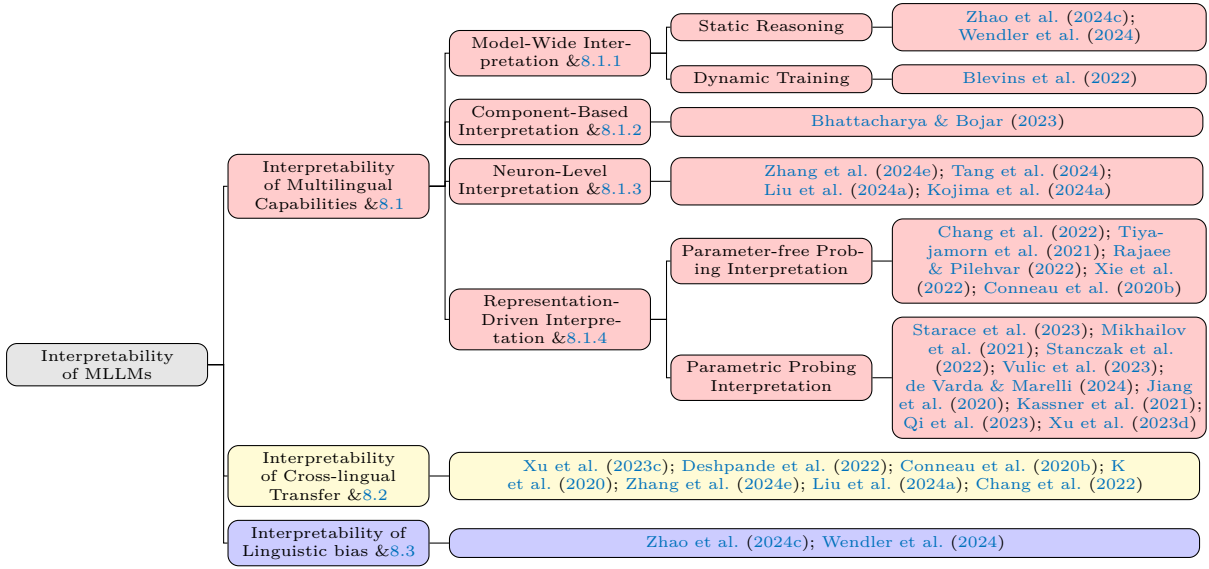
\begin{figure*}
    \tiny
    \begin{forest}
        for tree={
            forked edges,
            grow'=0,
            draw,
            rounded corners,
            node options={align=center},
            calign=edge midpoint,
        },
        [Interpretability of MLLMs, text width=2.5cm, fill=black!10
            [Interpretability of Multilingual Capabilities  \&\ref{Interpretability of Multilingual Capabilities}, text width=2.5cm, for tree={fill=red!20}
                    [Model-Wide Interpretation  \&\ref{Model-Wide Interpretation}, text width=2.5cm, for tree={fill=red!20}
                        [Static Reasoning, text width=2.5cm, for tree={fill=red!20}
                            [\citet{DBLP:journals/corr/abs-2402-18815}; \citet{DBLP:journals/corr/abs-2402-10588},
                            text width=4.0cm]
                        ]
                        [Dynamic Training, text width=2.5cm, for tree={fill=red!20}
                            [\citet{DBLP:conf/emnlp/BlevinsGZ22},
                            text width=4.0cm]
                        ]
                    ]
                    [Component-Based Interpretation   \&\ref{Component-Based Interpretation}, text width=2.5cm, for tree={fill=red!20}
                        [\citet{DBLP:conf/blackboxnlp/BhattacharyaB23},
                        text width=6.9cm]
                    ]
                    [Neuron-Level Interpretation   \&\ref{Neuron-Level Interpretation}, text width=2.5cm, for tree={fill=red!20}
                        [\citet{DBLP:journals/corr/abs-2402-14700};
                         \citet{DBLP:journals/corr/abs-2402-16438};
                         \citet{DBLP:journals/corr/abs-2402-16367};
                         \citet{DBLP:journals/corr/abs-2404-02431},
                         text width=6.9cm]
                    ]
                    [Representation-Driven Interpretation   \&\ref{Representation-Driven Interpretation}, text width=2.5cm, for tree={fill=red!20}
                        [Parameter-free Probing Interpretation, text width=2.5cm, for tree={fill=red!20}
                            [\citet{DBLP:conf/emnlp/ChangTB22};
                            \citet{DBLP:conf/emnlp/TiyajamornKAO21};
                            \citet{DBLP:conf/acl/RajaeeP22};
                            \citet{DBLP:conf/emnlp/XieZ0L22};
                            \citet{DBLP:conf/acl/ConneauWLZS20},
                            text width=4.0cm]
                        ]
                        [Parametric Probing Interpretation, text width=2.5cm, for tree={fill=red!20}
                            [\citet{DBLP:conf/emnlp/StaracePCPRLS23};
                            \citet{DBLP:journals/corr/abs-2104-12847};
                            \citet{DBLP:conf/naacl/StanczakPHCA22};
                            \citet{DBLP:conf/eacl/VulicGLCPK23};
                            \citet{DBLP:conf/coling/VardaM24};
                            \citet{DBLP:conf/emnlp/JiangAADN20};
                            \citet{DBLP:conf/eacl/KassnerDS21};
                            \citet{DBLP:conf/emnlp/QiFB23};
                            \citet{DBLP:conf/emnlp/XuLX23},
                            text width=4.0cm]
                        ]
                    ]
            ]
            [Interpretability of Cross-lingual Transfer  \&\ref{Interpretability of Cross-lingual Transfer}, text width=2.5cm, for tree={fill=yellow!20}
                [\citet{DBLP:conf/emnlp/XuZYZH23};
                 \citet{DBLP:conf/naacl/DeshpandeTN22};
                 \citet{DBLP:conf/acl/ConneauWLZS20};
                 \citet{DBLP:conf/iclr/KWMR20};
                 \citet{DBLP:journals/corr/abs-2402-14700};
                 \citet{DBLP:journals/corr/abs-2402-16367};
                 \citet{DBLP:conf/emnlp/ChangTB22},
                 text width=9.8cm]
            ]
            [Interpretability of Linguistic bias  \&\ref{Interpretability of Linguistic bias}, text width=2.5cm, for tree={fill=blue!20}
                [\citet{DBLP:journals/corr/abs-2402-18815};
                 \citet{DBLP:journals/corr/abs-2402-10588},
                 text width=9.8cm]
            ]
        ]
    \end{forest}
    \caption{Interpretability of MLLMs.}
    \label{fig:Interpretabilities of MLLMs}
\end{figure*}

\section{Interpretability}
\label{Interpretability of MLLMs}


Training an MLLM with multilingual data and subsequently fine-tuning it with multilingual instruction data can result in a highly effective MLLM. 
However, understanding the underlying mechanisms of such a model becomes challenging due to the diverse distribution of the training data and the complexity of the model's structure. 
Exploring the interpretability of MLLMs, or just MLLMs in general, is crucial for several reasons. 

1) It enhances the credibility and reliability of these models in practical use. When processing sensitive information or making important decisions, it is essential to explain the model's behavior and inferential processes to ensure compliance with legal, ethical or social standards. 

2) Interpretability helps identify biases and unfairness in the models. Due to imbalances in the quantity and quality of linguistic data, MLLMs may perform better in some languages than others, leading to linguistic bias. 
By interpreting the models' internal decision-making processes, we can better identify and correct these biases, making the models fairer. 

3) Analyzing the internal decision-making processes in depth allows us to discover patterns and regularities that can further improve model performance. For instance, understanding the mechanisms behind cross-lingual transfer in MLLMs can help enhance the performance of low-resource languages.

While the interpretability of MLLMs is challenging, the interpretability of MLLMs brings specific complexities. 
It centers on how the unique attributes of various languages and their interactions influence these models. 

Our discussion on the interpretability of MLLMs will begin by exploring three main issues: how these models manage multilingual capabilities, how they perform cross-lingual transfer, and the reasons behind language bias. 

The first issue is the basis for investigating the interpretability of MLLMs, while the latter two issues are phenomena that arise after modeling multilingual capabilities in MLLMs. 

\begin{figure*}[t]
\begin{center}
\includegraphics[scale=0.8]{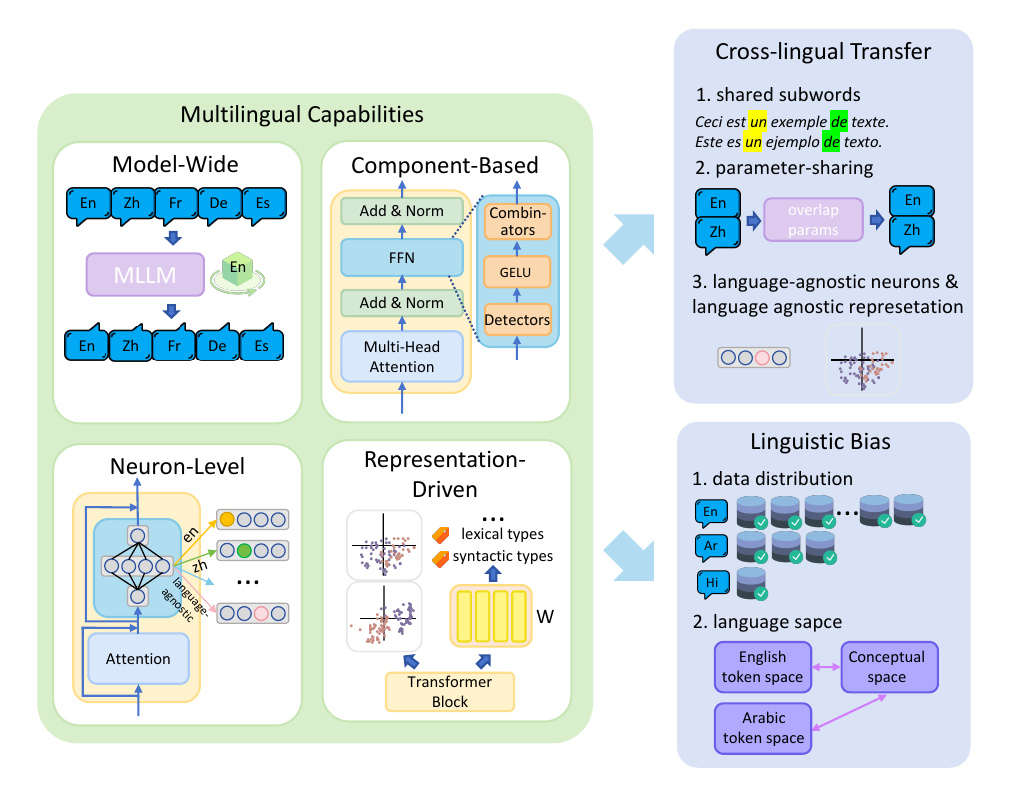} 
\caption{MLLMs interpretability perspectives and approaches.}
\label{fig:Multilingual LLMs Interpretability Perspectives and Approaches}
\end{center}
\end{figure*}

The framework of this section is shown in Figure \ref{fig:Interpretabilities of MLLMs}, and the specific perspectives and approaches are shown in Figure \ref{fig:Multilingual LLMs Interpretability Perspectives and Approaches}.

\subsection{Interpretability of Multilingual Capabilities}
\label{Interpretability of Multilingual Capabilities}


Studying the interpretability of MLLMs first requires understanding how these models specifically handle multilingual capabilities, how inputs from different languages are represented and processed \citep{zhao2024large}. 
This involves examining how the model shares semantic information across languages and manages language-specific features. 
Previous studies have explored the multilingual capabilities of MLLMs through four perspectives: model, component, neuron and representation.

At the model level, the behavior and decision paths of the entire model is analyzed to understand its overall multilingual processing capabilities. At the component level, the focus is on the internal components, such as feed-forward network and attention module, and investigates their roles in a multilingual environment. The neuron level delves into finer granularity, exploring the functions of individual neurons in multilingual tasks. Finally, the representation level examines how the model learns multilingual representations, investigating how inputs from different languages are represented and distinguished in high-dimensional space. Through these levels of analysis, a comprehensive understanding is gained on how MLLMs operate and process multilingual information.


\subsubsection{Model-Wide Interpretation}
\label{Model-Wide Interpretation}

To uncover the internal mechanisms of MLLMs modeling multilingual capabilities, some studies begins by examining the model as a whole. Generally, this aims to divide the model's processing of multilingual data into distinct stages. Specifically, these studies are categorized into two perspectives: static reasoning and dynamic training. Static reasoning focuses on the trained model, while dynamic training examines the model during its training process.

\paragraph{Static Reasoning}


Existing studies suggest that the layers closest to a model's input or output exhibit more language-specific behaviors than the layers in the middle \citep{DBLP:conf/blackboxnlp/BhattacharyaB23}. 
For instance, \citet{DBLP:journals/corr/abs-2402-18815} summarized the multilingual workflow of an MLLM by observing changes in the ratio of languages between layers. 
In this workflow, multilingual inputs are converted to English near the input layer. 
The model uses English for processing and integrates multilingual knowledge through specific structures in the intermediate layers. 
Near the output layer, the information is converted back to the original input language. 

Additionally, some studies interpret the intermediate process as encoding into a conceptual space, noting that the model appears to think in concepts that favor English, as the English space seems to be closer to this conceptual space than other languages spaces \citep{DBLP:journals/corr/abs-2402-10588}.

\paragraph{Dynamic Training}


patterns in the dynamic training process of MLLMs were uncovered by \citet{DBLP:conf/emnlp/BlevinsGZ22}. 
Compared to monolingual LLMs, MLLMs acquire information within each language in essentially the same order as monolingual models. 
However, this internal language learning occurs early in the training process. 
In contrast, cross-lingual information transfer is learned throughout pre-training, with the sequence of transferring linguistic information between specific languages varying considerably.

\subsubsection{Component-Based Interpretation}
\label{Component-Based Interpretation}


The components in MLLMs play a crucial role in modeling multilingual capabilities. 
In \citet{DBLP:conf/blackboxnlp/BhattacharyaB23}, distinct patterns of multilingual processing in the sub-layers of the model's feed-forward network were discovered. 
It divided FFNs into detectors and combiners, finding that the detectors in the early and late layers are multilingual, while the combiners in the middle layer also exhibit multilingual properties.

\subsubsection{Neuron-Level Interpretation}
\label{Neuron-Level Interpretation}


Due to the complexity of MLLM structures, where each component can have multiple functions, numerous studies have investigated how MLLMs model multilingual capabilities at a finer granularity--—the neuron level. 
A core region corresponding to multilingual capabilities in MLLMs was identified in \citet{DBLP:journals/corr/abs-2402-14700}, as perturbing parameters in this region were demonstrated to affect performance across all languages. 
Additionally, it found a region related to specific monolingual capabilities, where altering parameters decreases the performance of the corresponding language. 

Another perspective differentiates between language-agnostic neurons, responsible for processing generalized knowledge, and language-specific neurons, which handle language-specific vocabulary, grammar, and idiomatic expressions \citep{DBLP:journals/corr/abs-2402-16438, DBLP:journals/corr/abs-2402-16367, DBLP:journals/corr/abs-2404-02431}.

\subsubsection{Representation-Driven Interpretation}
\label{Representation-Driven Interpretation}


Considering interpretability from the perspective of model-specific structures, including the overall model, components and neurons, is one of the research approaches. Another approach focuses solely on the intermediate representations produced by the model. 
This involves transforming high-dimensional model representations into low-dimensional, human-understandable features through specific operations, typically probe methods. These probe methods can be divided into two categories: parameter-free probing methods, such as direct feature dimensionality reduction; and parametric probing methods, such as those using diagnostic classifiers to classify features.

\paragraph{Parameter-free Probing Interpretation}


The parameter-free probing approach primarily targets encoder-based MLLMs. 

Studies have shown that after applying mean-centering operations on the representations of MLLMs, different languages occupy similar linear subspaces. 
Additionally, it has been found that these MLLMs encode information along orthogonal axes that are either language-sensitive or language-neutral, addressing both language-agnostic and language-specific information \citep{DBLP:conf/emnlp/ChangTB22}.

In contrast, some other studies aim at maximizing the use of language-agnostic information in representations while ignoring language-specific information. 
This is achieved through methods like clustering-based anisotropy enhancement \citep{DBLP:conf/acl/RajaeeP22} or identifying low-rank subspaces that encode semantically irrelevant information \citep{DBLP:conf/emnlp/XieZ0L22}, which can improve the performance of language-agnostic tasks, such as cross-lingual text similarity \citep{DBLP:conf/emnlp/TiyajamornKAO21}.

Other studies have explored the reason of cross-lingual generic representations, identifying the significant role of shared parameters. 
Surprisingly, these studies also found similarities in the representations of models from different languages even when parameters are not shared \citep{DBLP:conf/acl/ConneauWLZS20}.

\paragraph{Parametric Probing Interpretation}


Some parametric probing methods focus on uncovering the underlying cross-lingual syntactic and morphological representations. 
By applying probing classifiers to syntax and morphology-related tasks, researchers have found that MLLMs learn syntactic and morphological content in different languages in a very similar way \citep{DBLP:journals/corr/abs-2104-12847}. 
These models can even learn this content across languages in a monolingual-like manner \citep{DBLP:conf/emnlp/StaracePCPRLS23}. 

The intrinsic probing approach suggests this may be because some neurons can encode universal syntactic and morphological information. 
However, since subsets of neurons from different languages do not completely overlap, the complexity of syntactic morphological information and the distance between languages may affect the learning of syntactic morphology across languages \citep{DBLP:conf/naacl/StanczakPHCA22}.


In more advanced research, some approaches explore lexical and semantic representations in MLLMs. 
Cross-lingual lexical fine-tuning followed by interpolation with static linguistic word embeddings was used in \citet{DBLP:conf/eacl/VulicGLCPK23} to reveal cross-lingual lexical knowledge from MLLMs. 

Additionally, \citet{DBLP:conf/coling/VardaM24} trained linear probes using extracted representations to identify neurons most relevant to the target lexicon, discovering significant cross-lingual overlap in these neurons.


A special case of lexical representation is factual knowledge representation. 
Existing MLLMs lack cross-lingual consistency in encoding factual knowledge, leading to different answers when the same factual question is asked in different languages \citep{DBLP:conf/emnlp/JiangAADN20, DBLP:conf/eacl/KassnerDS21, DBLP:conf/emnlp/QiFB23}. One potential solution is to explicitly transfer relatively rich factual knowledge from English to non-English languages \citep{DBLP:conf/emnlp/XuLX23}.

\subsection{Interpretability of Cross-lingual Transfer}
\label{Interpretability of Cross-lingual Transfer}


Cross-lingual transfer refers to the transfer of knowledge from one language to another. Despite the challenges posed by differences between languages, similar concepts can facilitate this process. 

Studies have shown that zero-shot and few-shot cross-lingual transfer occurs in MLLMs \citep{DBLP:conf/emnlp/XuZYZH23}. 
While some attribute this phenomenon to shared sub-word tokens \citep{DBLP:conf/naacl/DeshpandeTN22}, others argue that parameter sharing \citep{DBLP:conf/acl/ConneauWLZS20} and network depth \citep{DBLP:conf/iclr/KWMR20} are more crucial. 

Additionally, language-agnostic neurons \citep{DBLP:journals/corr/abs-2402-14700, DBLP:journals/corr/abs-2402-16367} and language-agnostic representations \citep{DBLP:conf/emnlp/ChangTB22} in MLLMs may play a key role, storing or representing cross-lingual semantic knowledge and enabling conceptual alignment between languages.

\subsection{Interpretability of Linguistic Bias}
\label{Interpretability of Linguistic bias}


Language bias refers to the tendency to favor or discriminate against certain groups in language use \citep{zhang2024getting}. 
This issue arises in MLLMs due to the imbalance in the distribution of high- and low-resource languages within their training data. 
Consequently, MLLMs can exhibit language bias, such as not fully understanding the cultural context of certain languages. 

This bias may stem from the distribution imbalance between high- and low-resource datasets, causing the model to shift inputs from various languages to high-resource languages, particularly English, to process information \citep{DBLP:journals/corr/abs-2402-18815}. 

Additionally, it has been suggested that this bias could be due to the close proximity between the space of English tokens and abstract concepts, further contributing to linguistic bias in MLLMs \citep{DBLP:journals/corr/abs-2402-10588}.

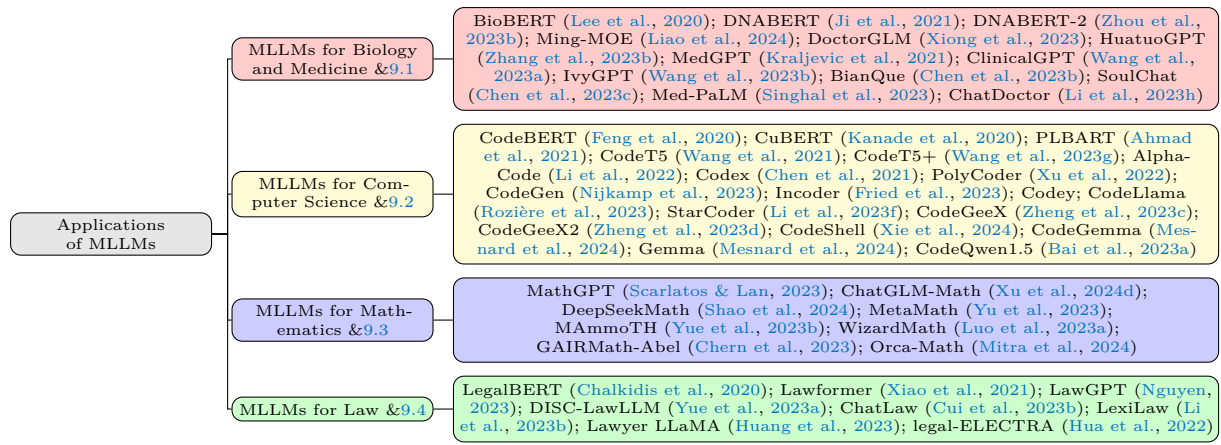
\begin{figure*}
    \tiny
    \begin{forest}
        for tree={
            forked edges,
            grow'=0,
            draw,
            rounded corners,
            node options={align=center},
            calign=edge midpoint,
        },
        [Applications of MLLMs, text width=2.5cm, fill=black!10
            [MLLMs for Biology and Medicine  \&\ref{MLLMs for Biology and Medicine}, text width=2.5cm, for tree={fill=red!20}
                    [                    
                    BioBERT \citep{DBLP:journals/bioinformatics/LeeYKKKSK20};
                    DNABERT \citep{DBLP:journals/bioinformatics/JiZLD21};
                    DNABERT-2 \citep{DBLP:journals/corr/abs-2306-15006};
                    Ming-MOE \citep{DBLP:journals/corr/abs-2404-09027};
                    DoctorGLM \citep{DBLP:journals/corr/abs-2304-01097};
                    HuatuoGPT \citep{DBLP:conf/emnlp/ZhangCJYCCLWZXW23}; 
                    MedGPT \citep{DBLP:journals/corr/abs-2107-03134};
                    ClinicalGPT \citep{DBLP:journals/corr/abs-2306-09968};
                    IvyGPT \citep{DBLP:conf/cicba/WangDLCXCLPT23};
                    BianQue \citep{DBLP:journals/corr/abs-2310-15896};
                    SoulChat \citep{DBLP:conf/emnlp/ChenXLZWLX23};
                    Med-PaLM \citep{DBLP:journals/corr/abs-2305-09617};
                    ChatDoctor \citep{DBLP:journals/corr/abs-2303-14070},
                    text width=10.0cm
                    ]
            ]
            [MLLMs for Computer Science  \&\ref{MLLMs for Computer Science}, text width=2.5cm, for tree={fill=yellow!20}
                [                    
                CodeBERT \citep{DBLP:conf/emnlp/FengGTDFGS0LJZ20};
                CuBERT \citep{DBLP:conf/icml/KanadeMBS20};
                PLBART \citep{DBLP:conf/naacl/AhmadCRC21};
                CodeT5 \citep{DBLP:conf/emnlp/0034WJH21};
                CodeT5+ \citep{DBLP:conf/emnlp/WangLGB0H23};
                AlphaCode \citep{DBLP:journals/corr/abs-2203-07814};
                Codex \citep{DBLP:journals/corr/abs-2107-03374};
                PolyCoder \citep{DBLP:conf/pldi/Xu0NH22};
                CodeGen \citep{DBLP:conf/iclr/NijkampPHTWZSX23};
                Incoder \citep{DBLP:conf/iclr/FriedAL0WSZYZL23};
                Codey; 
                CodeLlama \citep{DBLP:journals/corr/abs-2308-12950};
                StarCoder \citep{DBLP:journals/corr/abs-2305-06161};
                CodeGeeX \citep{DBLP:conf/kdd/ZhengXZDWXSW0LS23};
                CodeGeeX2 \citep{DBLP:journals/corr/abs-2303-17568};
                CodeShell \citep{DBLP:journals/corr/abs-2403-15747};
                CodeGemma \citep{DBLP:journals/corr/abs-2403-08295};
                Gemma \citep{DBLP:journals/corr/abs-2403-08295};
                CodeQwen1.5 \citep{DBLP:journals/corr/abs-2309-16609},
                text width=10.0cm
                ]
            ]
            [MLLMs for Mathematics  \&\ref{MLLMs for Mathematics}, text width=2.5cm, for tree={fill=blue!20}
                [
                MathGPT \citep{scarlatos-lan-2023-tree};
                ChatGLM-Math \citep{DBLP:journals/corr/abs-2404-02893}; DeepSeekMath \citep{DBLP:journals/corr/abs-2402-03300};
                MetaMath \citep{DBLP:journals/corr/abs-2309-12284};
                MAmmoTH \citep{DBLP:journals/corr/abs-2309-05653};
                WizardMath \citep{DBLP:journals/corr/abs-2308-09583};
                GAIRMath-Abel \citep{abel};
                Orca-Math \citep{DBLP:journals/corr/abs-2402-14830},
                text width=10.0cm
                ]
            ]
            [MLLMs for Law  \&\ref{MLLMs for Law}, text width=2.5cm, for tree={fill=green!20}
                [
                LegalBERT \citep{DBLP:journals/corr/abs-2010-02559}; 
                Lawformer \citep{DBLP:journals/aiopen/XiaoHLTS21};
                LawGPT \citep{DBLP:journals/corr/abs-2302-05729};
                DISC-LawLLM \citep{DBLP:journals/corr/abs-2309-11325};
                ChatLaw \citep{DBLP:journals/corr/abs-2306-16092};
                LexiLaw \citep{DBLP:conf/sigir/LiACDW0CT23};
                Lawyer LLaMA \citep{DBLP:journals/corr/abs-2305-15062};
                legal-ELECTRA \citep{DBLP:journals/corr/abs-2212-08204},
                text width=10.0cm
                ]
            ]
        ]
    \end{forest}
    \caption{Applications of MLLMs.}
    \label{fig:Application of MLLMs}
\end{figure*}

\section{Application}
\label{Application of MLLMs}

Due to its excellent multi-language understanding and cross-language generalization capabilities, MLLMs have demonstrated excellent performance in various downstream tasks.
Therefore, MLLM is widely used in various professional domains, including biology and medicine, computer science, mathematics, law, etc. 

These domain-specific MLLMs have demonstrated outstanding capabilities and promising perspectives in related domains, and have even surpassed human levels in some aspects. 
As a consequence, MLLMs provide a new approach for the integration of artificial intelligence and these domains. 
Nevertheless, the application of MLLMs in these domains remains challenging, mainly due to the reliance on specific expertise and data collection requirements. 

At present, MLLMs abound more for English and Chinese, and there are fewer suitable for low-resource languages, which greatly hinders the development of generative AI on a global scale. 
In this section, we address the development trajectory and recent progress of MLLMs across different domains, focusing on their practical applications, as shown in Figure \ref{fig:Application of MLLMs}.

\subsection{MLLMs for Biology and Medicine}
\label{MLLMs for Biology and Medicine}

The integration of MLLMs has shown tremendous potential in the health sector, particularly in applications such as medical Q\&A, intelligent diagnosis, and psychological counseling \citep{qiu2024towards,lifelo2024adapting,garcia2024medical}. 

By bridging language barriers and enhancing data-driven research and clinical practice, MLLMs have significantly advanced the fields of biology and medicine. 
Their ability to understand and generate human language across multiple languages has demonstrated substantial promise in improving medical diagnostics, treatment, and research in diverse linguistic contexts globally.

BioBERT \citep{DBLP:journals/bioinformatics/LeeYKKKSK20} is a domain-specific language representation model pre-trained on a large biomedical corpus. It significantly outperforms BERT in numerous biomedical text mining tasks.

DNABERT \citep{DBLP:journals/bioinformatics/JiZLD21} is a pre-trained bidirectional encoder representation designed to capture a global and transferable understanding of genomic DNA sequences based on upstream and downstream nucleotide contexts. DNABERT-2 \citep{DBLP:journals/corr/abs-2306-15006} is trained on multi-species genomes and is more efficient, powerful, and easy to use than its first generation.

Ming-MOE \citep{DBLP:journals/corr/abs-2404-09027}, a novel Mixture-of-Expert-based medical large language model designed to manage diverse and complex medical tasks without requiring task-specific annotations, thus enhancing its usability across extensive datasets.

DoctorGLM \citep{DBLP:journals/corr/abs-2304-01097} is a Chinese medical inquiry model based on ChatGLM-6B.
It is pre-trained using various techniques on a collected Chinese medical dialogue database.
%
%

HuatuoGPT \citep{DBLP:conf/emnlp/ZhangCJYCCLWZXW23} leverage both distilled data from ChatGPT and real-world data from doctors in the supervised fine-tuned stage.

MedGPT \citep{DBLP:journals/corr/abs-2107-03134} is a large medical language model based on LLaMA-13B, fine-tuned through supervised learning for multiple tasks. 
It excels in various applications, including disease inquiry, differential diagnosis, recommending tests and examinations, summarizing medical records, interpreting examination results, and providing diagnostic outcomes and treatment plans.

ClinicalGPT \citep{DBLP:journals/corr/abs-2306-09968} is a language model specifically designed and optimized for clinical settings. 
Trained on real-world medical records and domain-specific knowledge, it excels in tasks such as medical knowledge question-answering, physical examinations, patient consultations and medical record analysis.

IvyGPT \citep{DBLP:conf/cicba/WangDLCXCLPT23}, a model based on LLaMA, has been trained and fine-tuned using over 300,000 high-quality medical question-answer (QA) instances and reinforced learning from human feedback (RLHF). 
It demonstrates strong multi-turn dialogue capabilities, providing more detailed and human-like diagnostic and treatment responses.

BianQue \citep{DBLP:journals/corr/abs-2310-15896} utilizing ChatYuan as its base model, is a large-scale medical dialogue model fine-tuned through a combination of instruction and multi-turn inquiry dialogues. 
It has been refined on a mixed dataset of over 9 million Chinese medical question-answering instructions and multi-turn dialogues.

SoulChat \citep{DBLP:conf/emnlp/ChenXLZWLX23} employs the ChatGLM-6B model as its initialization framework and has undergone comprehensive fine-tuning on both single-turn and multi-turn mental health counseling dialogue datasets. 
It is capable of demonstrating empathy, encouraging users to express themselves, and providing reasonable advice.

Med-PaLM \citep{DBLP:journals/corr/abs-2305-09617}, a large language model from Google Research, designed to provide high quality answers to medical questions.  
It was the first AI system to overcome the pass mark (>60\%) in the U.S.A. 

ChatDoctor \citep{DBLP:journals/corr/abs-2303-14070} is a medical assistant utilizing the LLaMA model, trained with integrated medical knowledge. 
It has been fine-tuned on over 100,000 real doctor-patient conversations. 
It not only demonstrates fluent conversational abilities but also exhibits a high level of understanding and diagnostic accuracy in the medical domain.

\subsection{MLLMs for Computer Science}
\label{MLLMs for Computer Science}

Recently, there have been significant advancements in the application of large language models within the domain of computer science, particularly in tasks such as code generation and text-to-SQL conversion.
The application of large language models is reshaping the domain of computer science.
The paradigm has shifted from manually writing code to generating it and making human corrections.
In this subsection, we will explore the development history and recent advancements of MLLMs in the domain of computer science.

\textbf{Encoder-Only: } Due to the Encoder-Only architecture's ability to effectively capture the global dependencies and features of input sequences, it excels in code detection and classification tasks.

\citet{DBLP:conf/emnlp/FengGTDFGS0LJZ20} proposed CodeBERT, a pre-trained model specifically designed for programming languages. 
This model comprises 124 million parameters and supports 6 different programming languages.

In the same year, CuBERT \citep{DBLP:conf/icml/KanadeMBS20} was introduced, focusing on training BERT on source code to obtain contextual embeddings. This model was used to identify code block defects and detect duplicate code blocks.

\textbf{Encoder-Decoder: } 
The Encoder-Decoder architecture is a neural network model widely used in natural language processing and other sequence-to-sequence tasks. It has now been extensively applied to tasks such as text generation, question answering, and code generation.

PLBART \citep{DBLP:conf/naacl/AhmadCRC21} adopts the encoder-decoder BART architecture and is pre-trained on an extensive collection of programming languages (PL) and natural languages (NL) via denoising autoencoding. It is primarily used for code summarization, text-to-code generation, code-to-code translation, and code refinement.

CodeT5 \citep{DBLP:conf/emnlp/0034WJH21} supports eight common programming languages and employs the same pre-training approach as T5 \citep{DBLP:journals/jmlr/RaffelSRLNMZLL20}. It focuses on masked span prediction, denoising sequence reconstruction and masked identifier prediction with a bimodal dual generation strategy. 
This strategy encourages better alignment between NL and PL, allowing CodeT5 to significantly outperform PLBART across all generation tasks.

CodeT5+ \citep{DBLP:conf/emnlp/WangLGB0H23} is an enhanced version of CodeT5, employing a larger parameter scale to deliver enhanced performance and more accurate code comprehension. It mainly features three parameter levels: 2B, 6B and 16B. Across various code-related tasks such as code completion, code recommendation and code classification, it demonstrates superior performance compared to CodeT5.

In 2022, DeepMind announced the launch of AlphaCode \citep{DBLP:journals/corr/abs-2203-07814}, a code generation system that can create novel solutions to problems requiring deeper reasoning, using a Transformer-based model. It was trained on over 715 GB of data from GitHub and Codeforces issues and solutions, and it supports twelve of the most common programming languages.

\textbf{Decoder-Only: } 
The Decoder-Only architecture is currently the most commonly used MLLM architecture, typically employed for sequence generation tasks such as text generation and machine translation.

Codex \citep{DBLP:journals/corr/abs-2107-03374}, a model provided exclusively through OpenAI's API, is a descendant of GPT-3. 
It is available in three sizes: 300M, 2.5B and 12B, and it powers GitHub Copilot, a well-known and robust model renowned for its performance. 
While excelling in Python, it also demonstrates proficiency in a variety of other programming languages, for tasks such as code translation, code explanation and code refactoring. 
However, unlike other models, Codex is not publicly downloadable. 

PolyCoder \citep{DBLP:conf/pldi/Xu0NH22} is an MLLM based on the GPT-2 architecture.
It has been trained on 249GB of code across twelve common programming languages and is available in three sizes: 160M, 0.4B and 2.7B. 
In the C programming language, PolyCoder outperforms all previous models, including Codex.

CodeParrot is a GPT-2 model trained exclusively on 180GB of Python code. It is available in two sizes, 110M and 1.5B parameters. 

CodeGen \citep{DBLP:conf/iclr/NijkampPHTWZSX23} is an autoregressive language model designed for program synthesis, sequentially trained on The Pile, BigQuery and BigPython datasets. 
Its goal is to enhance developer productivity by converting metadata into readable and maintainable source code.

Incoder \citep{DBLP:conf/iclr/FriedAL0WSZYZL23} is trained on code using a causal masking objective, enabling code insertion/filling as well as standard left-to-right generation. It is trained on public open-source repositories with permissive, non-Copyleft licenses from GitHub, GitLab, and StackOverflow. 
These repositories predominantly contain Python and JavaScript but also include code in twenty eight other languages sourced from StackOverflow. 
Incoder is available in two variants, 1.3B and 6B parameters.

Codey, a fine-tuned model of PaLM2,\footnote{https://lablab.ai/tech/google/codey\#codey-google-ais-revolutionary-coding-assistant} is capable of performing varying coding tasks. 
It is fine-tuned with extensive high-quality code and coding documentation. Google claims that Codey can code in more than twenty programming languages. 
It is used to enhance Google products like Google Colab, Android Studio, and more.

The CodeLlama \citep{DBLP:journals/corr/abs-2308-12950} release introduces a series of models with 7B, 13B and 34B parameters. 
These base models are initialized from Llama 2 and then trained on 500B tokens of code. Meta subsequently fine-tuned these base models into two distinct flavors: a Python specialist variant (with an additional 100B tokens) and an instruction fine-tuned version capable of comprehending natural language instructions. 
These models exhibit state-of-the-art performance across various programming languages.

StarCoder \citep{DBLP:journals/corr/abs-2305-06161} is trained on permissively licensed data from GitHub, encompassing over eighty programming languages, Git commits, GitHub issues and Jupyter notebooks. 
Similar to LLaMA, it is a 15 billion parameter model trained on 1 trillion tokens. StarCode outperforms existing open code language models on popular programming benchmarks and matches or surpasses closed models such as Codex.

ChatGPT and GPT-4 are advanced language models developed by OpenAI. 
They utilize Reinforcement Learning with Human Feedback (RLHF) to enhance their program synthesis capabilities. 
These models have demonstrated proficiency in code generation tasks, often surpassing human-level performance.

CodeGeeX \citep{DBLP:conf/kdd/ZhengXZDWXSW0LS23} and CodeGeeX2 \citep{DBLP:journals/corr/abs-2303-17568} are multilingual code generation models developed by Tsinghua University. 
Unlike CodeGeeX, CodeGeeX2 is built on the ChatGLM2 architecture with added code pre-training. Leveraging the superior performance of ChatGLM2, CodeGeeX2 achieves performance improvements across multiple benchmarks. 
Remarkably, with only 6B parameters, it surpasses the 15B parameter StarCoder-15B by nearly 10\%.

The original training data for CodeShell \citep{DBLP:journals/corr/abs-2403-15747} is based on self-collected GitHub data, the Stack and StarCoder datasets, and a small amount of high-quality Chinese and English data. 
It underwent cold start training on 500B tokens, with a context window length of 8192. CodeShell's code generation performance surpasses that of CodeLlama-7B and StarCoder-7B.

CodeGemma \citep{DBLP:journals/corr/abs-2403-08295} is a family of code-specialist LLM models by Google, based on the pre-trained 2B and 7B Gemma \citep{DBLP:journals/corr/abs-2403-08295} checkpoints. 
CodeGemma are further trained on an additional 500B tokens of primarily English language data, mathematics and code to improve on logical and mathematical reasoning, and are suitable for code completion and generation.

CodeQwen1.5 \citep{DBLP:journals/corr/abs-2309-16609} is the code-specific version of Qwen1.5, pre-trained on a vast corpus of code data.
With a context length of 64K tokens and support for ninety two programming languages, it exhibits robust code generation capabilities across a range of benchmark tests. Additionally, it demonstrates outstanding performance in tasks such as text-to-SQL conversion and error correction.

\subsection{MLLMs for Mathematics}
\label{MLLMs for Mathematics}

In recent years, there has been a significant surge in the development of large language models aimed at automating the process of solving mathematical problems, which given
the wide-ranging and diverse nature of mathematical problem, presents a challenge to the development of this emerging field.

Multilingual large language models perform exceptionally well in tasks such as text completion and machine translation. However, they exhibit notable limitations in solving, explaining, answering and recommending solutions for mathematical problems.

In this subsection, we will discuss recent advancements of MLLMs in the domain of mathematics.

MathGPT \citep{scarlatos-lan-2023-tree} is a large-scale model targeting global mathematics enthusiasts and research institutions with problem-solving and teaching algorithms at its core. Its mathematical computation capabilities span across primary, middle and high school levels. In evaluations across six publicly available math test sets--—CEval-Math, AGIEval-Math, APE5K, CMMLU-Math, high school math exams, and Math401--—MathGPT has achieved the highest scores in multiple tests, even surpassing those of GPT-4.

The ChatGLM-Math \citep{DBLP:journals/corr/abs-2404-02893} model is an LLM that enhances mathematical problem-solving abilities through a self-critical pipeline. This model not only improves mathematical skills but also preserves and enhances language capabilities, resulting in performance enhancements across various tasks.

The DeepSeekMath \citep{DBLP:journals/corr/abs-2402-03300} model underwent pretraining with a total of 500 billion tokens, which included mathematical-related texts from Common Crawl as well as natural language and code data. Built upon the foundational architecture of DeepSeek-Coder-v1.5 7B, the model received specialized instruction tuning and reinforcement learning training to enhance its mathematical problem-solving and tool utilization abilities. Additionally, DeepSeekMath 7B achieved performance levels similar to Gemini-Ultra and GPT-4 in the highly competitive MATH Challenge.

MetaMath \citep{DBLP:journals/corr/abs-2309-12284}, fine-tuned on the MetaMathQA dataset using the LLaMA-2 architecture, is a large language model specialized in mathematical reasoning (both forward and backward). MetaMath-70B has surpassed ChatGPT in performance on mathematical reasoning datasets, achieving state-of-the-art results.

MAmmoTH \citep{DBLP:journals/corr/abs-2309-05653} underwent instruction fine-tuning on the MathInstruct dataset, which covers various mathematical domains and complexities, blending Concept of Thinking (CoT) with Programming of Thinking (PoT). The MAmmoTH-34B model has surpassed the CoT results of GPT-4 on competition-level datasets.

WizardMath \citep{DBLP:journals/corr/abs-2308-09583} combines reinforcement learning with math-specific instruction data to enhance the abilities of LLMs in mathematical reasoning. It surpasses advanced models such as ChatGPT, Claude Instant-1, PaLM-2, and Minerva on the GSM8K dataset.

GAIRMath-Abel \citep{abel} achieves state-of-the-art performance across open-source LLMs solely through the utilization of Parental Oversight, a Babysitting Strategy for Supervised Fine-tuning, without requiring tools, continued pretraining, reward modeling or reinforcement learning from human feedback (RLHF).

Orca-Math \citep{DBLP:journals/corr/abs-2402-14830}, a small language model constructed with 700M parameters, is fine-tuned on the Mistral-7B architecture. It redefines the traditional approach to teaching mathematical word problems through creative synthetic datasets and iterative learning mechanisms. Notably, Orca-Math has achieved significant advancements on the GSM8K benchmark.

\subsection{MLLMs for Law}
\label{MLLMs for Law}

While general multilingual large language models have the capacity to address a broad spectrum of knowledge domains, the legal field is a highly specialized domain that necessitates both utmost accuracy and timely relevance.
General MLLMs may be deficient in their grasp of legal statutes, judicial interpretations and nuances particular to specific jurisdictions, thereby leading to imprecise responses or the omission of crucial information.

Consequently, the development of specialized legal MLLMs becomes imperative to overcome these limitations and cater to the specific demands of the legal sector.

In this section, we address the most recent advancements of MLLMs within the domain of law.

LegalBERT \citep{DBLP:journals/corr/abs-2010-02559}, based on the BERT model, is the first large-scale model pre-trained on legal texts. It was trained on the entire Harvard Law School case corpus, which comprises 37GB of legal data and includes 3,446,187 legal decisions from both federal and state courts. The model demonstrates exceptional performance on downstream legal tasks.

Lawformer \citep{DBLP:journals/aiopen/XiaoHLTS21}, based on Longformer, is pre-trained on a large-scale corpus of Chinese legal texts. As the first Chinese legal pre-training model, it does not use standard self-attention but instead combines local sliding windows with a global attention mechanism to capture long-range dependencies. It demonstrates exceptional performance on long-text legal tasks.

LawGPT \citep{DBLP:journals/corr/abs-2302-05729} builds on existing general-purpose Chinese language models by expanding the vocabulary with legal-specific terms and conducting pre-training on a large-scale corpus of Chinese legal texts. This enhances the model's foundational semantic understanding in the legal domain. Additionally, LawGPT undergoes instruction fine-tuning on legal domain dialogue question-answering datasets and the Chinese judicial examination dataset, improving its comprehension and execution of legal content.

Fudan University has introduced DISC-LawLLM \citep{DBLP:journals/corr/abs-2309-11325}, a large-scale model designed to provide users with professional, intelligent and comprehensive legal services. It surpasses ChatGPT in performance on the latest legal evaluation benchmark, Lawbench, and is second only to GPT-4.

The ChatLaw \citep{DBLP:journals/corr/abs-2306-16092} model comes in two versions, one based on the Ziya-13B model and the other on the Anima-33B model. It is pre-trained on extensive legal dialogue data, including legal news, forums, statutes, judicial interpretations, legal consultations, judicial exam questions and court judgments. ChatLaw demonstrates excellent performance on various legal tasks.

LexiLaw \citep{DBLP:conf/sigir/LiACDW0CT23}, built on the ChatGLM-6B architecture, has been fine-tuned on legal domain datasets to enhance its performance and professionalism in providing legal consultation and support.

Lawyer LLaMA \citep{DBLP:journals/corr/abs-2305-15062}, based on the LLaMA-13B architecture, underwent continual pretraining on a large-scale legal corpus. Subsequently, the model was instruction fine-tuned using collected legal data, enabling a deep understanding of common areas in Chinese law, including civil law, criminal law, administrative law, and procedural law.

legal-ELECTRA \citep{DBLP:journals/corr/abs-2212-08204} is built upon the ELECTRA model and further pre-trained on a dataset of 20M high-quality judicial documents, enabling the model to better comprehend legal texts.

\section{Challenges and Future Directions}


The rapid advancement of MLLMs has opened exciting new avenues for cross-lingual language understanding and generation. 
However, several critical challenges remain, hindering the full realization of their potential and necessitating further research and development.

\subsection{Democratizing Language Technology}


Democratizing language technology and effectively addressing linguistic diversity represent paramount challenges in MLLM research.
Current models often exhibit a significant performance disparity, favoring high-resource languages like English due to data abundance. 
Bridging this gap requires innovative approaches to cross-lingual transfer learning, enabling efficient knowledge transfer from resource-rich to resource-scarce languages. 
This involves exploring advanced techniques such as adapter-based transfer, meta-learning for cross-lingual adaptation, and developing robust interlingua representations that capture universal linguistic properties. 
Furthermore, addressing the scarcity of data in low-resource languages necessitates sophisticated data augmentation and generation methods. 
Techniques like back-translation, paraphrasing, and leveraging synthetic data generation through controlled text generation can supplement limited datasets, though careful consideration must be given to cultural sensitivity and appropriateness during data generation. 
Ultimately, democratizing language technology hinges on developing MLLMs that are genuinely inclusive, capable of understanding and generating text across a wide spectrum of languages, regardless of their resource availability.

Beyond supporting a multitude of languages, a crucial aspect of linguistic diversity lies in the rich tapestry of language variations and dialects. 
MLLMs frequently struggle with these nuances, as they deviate significantly from standardized language forms commonly used in training data. 
Therefore, future research must focus on incorporating dialectal information during both training and evaluation.
This may involve developing specialized modules within MLLM architectures that are sensitive to dialectal variations or by leveraging techniques like code-switching and dialect adaptation during fine-tuning. 
Creating dedicated evaluation benchmarks that accurately assess MLLM performance on dialectal data is also essential, emphasizing aspects such as comprehension, fluency, and cultural relevance. 
Furthermore, code-switching and language mixing, prevalent phenomena in multilingual communities, pose additional complexities. 
Effectively handling these intricate language patterns necessitates developing models capable of explicitly recognizing and adapting to code-switching dynamics, potentially through dedicated modules and by training on large-scale code-switching datasets.

\subsection{Towards Culturally-Aware and Adaptive MLLMs}


Developing truly multilingual AI necessitates moving beyond purely linguistic considerations and embracing the rich tapestry of cultural nuances embedded within language. 
Towards this goal, building culturally-aware and adaptive MLLMs is a critical research frontier. 
Current MLLMs, often trained predominantly on English data, risk perpetuating cultural biases and failing to capture the diverse perspectives and values reflected in different languages and cultures.
Future research must prioritize integrating cultural knowledge into MLLMs, ensuring they are not only multilingual but also multicultural in their understanding. 
Training on culturally diverse datasets, encompassing various perspectives and worldviews, is essential. 
This involves carefully curating data that represents different cultural contexts, including literature, social media conversations, and news articles from diverse sources. 
Moreover, developing evaluation metrics that capture cultural nuances is crucial. These metrics should move beyond simple accuracy and fluency, assessing aspects such as cultural sensitivity, appropriateness, and the ability to adapt communication styles based on cultural context. 
This shift towards culturally-aware evaluation will encourage the development of MLLMs that are not only proficient in multiple languages but also respectful and understanding of the diverse cultures they represent.

Adaptability is another key dimension of culturally-aware MLLMs. 
Language is not static; it evolves and adapts to changing cultural contexts. 
Therefore, MLLMs should be equipped to learn and adapt to these evolving cultural dynamics. 
This involves developing models that can dynamically adjust their understanding and generation based on cultural cues, potentially by incorporating mechanisms for continuous learning and adaptation. 
For example, an adaptive MLLM could learn to recognize and respond appropriately to culturally-specific humor, idioms, and social norms. 
Furthermore, research should explore methods for personalizing MLLMs to individual cultural preferences, enabling users to tailor the model's behavior to align with their own cultural background and values. 
This personalization can enhance user experience and promote inclusivity, ensuring that MLLMs are accessible and beneficial to diverse cultural communities worldwide. 
Building culturally-aware and adaptive MLLMs requires a multidisciplinary approach, drawing on insights from linguistics, anthropology, sociology, and other fields to develop models that are truly representative of the rich diversity of human culture.

\subsection{Ensuring Safety, Fairness, and Interpretability}




Ensuring safety, fairness, and interpretability in MLLMs is paramount, especially as these powerful models become increasingly integrated into diverse real-world applications. 
Their widespread deployment necessitates a rigorous focus on responsible AI principles to mitigate potential risks and ensure ethical and equitable outcomes. Safety, in the context of MLLMs, encompasses safeguarding against adversarial attacks, malicious inputs, and the generation of unsafe or harmful content. 
This requires developing robust defense mechanisms, including techniques like adversarial training and input sanitization, to enhance MLLM resilience against malicious exploitation. 
Furthermore, establishing clear safety guidelines and protocols for MLLM deployment is crucial, outlining best practices for data handling, model training, and user interaction. 
These guidelines should address potential risks associated with specific application domains, such as healthcare, finance, and legal services, ensuring that MLLMs are used responsibly and ethically.

Fairness is another critical concern, particularly given the potential for MLLMs to perpetuate or amplify existing societal biases. 
Bias can manifest in various forms, including gender bias, racial bias, and cultural bias, leading to discriminatory or unfair outcomes. 
Addressing this challenge requires a multi-pronged approach, encompassing both data and model-level interventions. 
Developing methods for detecting and mitigating biases in training data, such as debiasing techniques and data augmentation strategies, is essential. 
Furthermore, incorporating fairness metrics into evaluation frameworks can help quantify and monitor bias in MLLM outputs, encouraging the development of fairer and more equitable models. 
Promoting fairness also necessitates fostering responsible data collection and annotation practices, ensuring that data used to train MLLMs is representative and inclusive of diverse populations.

Interpretability and explainability are crucial for building trust and ensuring accountability in MLLMs. 
The complex nature of these models, often referred to as ``black boxes" makes it challenging to understand their internal decision-making processes.
Enhancing interpretability involves developing techniques for visualizing cross-lingual representations, analyzing attention patterns, and providing human-understandable explanations for MLLM outputs.
This requires moving beyond simply generating text to providing insights into how the model arrived at a particular conclusion or prediction. 
For example, methods like probing tasks and attention visualization can help reveal the underlying linguistic knowledge and reasoning processes employed by MLLMs. 
Furthermore, research should focus on developing methods for explaining cross-lingual transfer, providing insights into how knowledge is transferred between languages and how this transfer can be optimized for improved performance. Ultimately, achieving interpretability and explainability in MLLMs is essential for promoting transparency, fostering trust, and enabling effective human-model collaboration.

\subsection{Towards Efficient and Sustainable MLLMs}

The computational demands of MLLMs present significant challenges to their widespread accessibility and sustainability. 
Training and deploying these massive models require substantial resources, raising concerns about their environmental impact and limiting their use by researchers and practitioners with constrained budgets. 
Addressing this challenge necessitates a multifaceted approach focusing on efficiency improvements across various dimensions. 
Research should prioritize developing more efficient training algorithms, including techniques like sparse training, pruning, and quantization, as well as exploring lightweight model architectures and model compression methods such as knowledge distillation. 
Furthermore, investigating hardware acceleration strategies, including specialized hardware and distributed training, can significantly reduce computational costs. Enhancing data efficiency through few-shot and zero-shot learning, alongside exploring alternative data sources like synthetic data and unsupervised learning, can further reduce resource requirements. 
Ultimately, building efficient and sustainable MLLMs demands a holistic approach, optimizing not only model architecture and training but also hardware, data usage, and the overall environmental footprint, ensuring these powerful tools are both accessible and environmentally responsible.

Future research should focus on developing more efficient training algorithms, exploring model compression techniques, and investigating hardware acceleration strategies to make MLLMs more accessible and computationally sustainable. 
This includes exploring lightweight architectures, knowledge distillation methods, and distributed training strategies to reduce computational costs while maintaining performance.

\section{Conclusion}

The development pace of MLLMs has been astonishing, showcasing remarkable progress across numerous tasks. However, despite ushering in a new era of artificial intelligence, our understanding of this novel form of intelligence remains relatively limited.

It is crucial to delineate the boundaries of MLLMs' capabilities, understand their performance in various domains, and explore how to harness their potential more effectively. 
This necessitates a comprehensive evaluation framework to guide the direction of MLLM development.

This survey has provided a systematic and thorough elaboration on the core capabilities of MLLMs, encompassing critical aspects like cross-lingual knowledge, reasoning, alignment with human values and safety. Furthermore, it has delved into the interpretability of multilingual capabilities, cross-lingual transfer, and language bias within these models, transforming them from black boxes to white boxes.

Most importantly, the survey has explored the potential applications of MLLMs across diverse domains, including biology, medicine, computer science, mathematics and law. 
It has discussed how these models have driven innovation and improvements in these specialized fields, while also highlighting the challenges and opportunities in deploying MLLMs within diverse language communities and application scenarios.

\newpage


\setcitestyle{authoryear,open={(},close={)}}
\bibliography{main}

\begin{thebibliography}{409}
\providecommand{\natexlab}[1]{#1}
\providecommand{\url}[1]{\texttt{#1}}
\expandafter\ifx\csname urlstyle\endcsname\relax
  \providecommand{\doi}[1]{doi: #1}\else
  \providecommand{\doi}{doi: \begingroup \urlstyle{rm}\Url}\fi

\bibitem[Agarwal et~al.(2024)Agarwal, Tanmay, Khandelwal, and Choudhury]{DBLP:conf/coling/AgarwalTKC24}
Utkarsh Agarwal, Kumar Tanmay, Aditi Khandelwal, and Monojit Choudhury.
\newblock Ethical reasoning and moral value alignment of llms depend on the language we prompt them in.
\newblock In Nicoletta Calzolari, Min{-}Yen Kan, V{\'{e}}ronique Hoste, Alessandro Lenci, Sakriani Sakti, and Nianwen Xue (eds.), \emph{Proceedings of the 2024 Joint International Conference on Computational Linguistics, Language Resources and Evaluation, {LREC/COLING} 2024, 20-25 May, 2024, Torino, Italy}, pp.\  6330--6340. {ELRA} and {ICCL}, 2024.
\newblock URL \url{https://aclanthology.org/2024.lrec-main.560}.

\bibitem[Aggarwal et~al.(2022)Aggarwal, Gupta, and Kunchukuttan]{DBLP:conf/emnlp/Aggarwal0K22}
Divyanshu Aggarwal, Vivek Gupta, and Anoop Kunchukuttan.
\newblock Indicxnli: Evaluating multilingual inference for indian languages.
\newblock In Yoav Goldberg, Zornitsa Kozareva, and Yue Zhang (eds.), \emph{Proceedings of the 2022 Conference on Empirical Methods in Natural Language Processing, {EMNLP} 2022, Abu Dhabi, United Arab Emirates, December 7-11, 2022}, pp.\  10994--11006. Association for Computational Linguistics, 2022.
\newblock \doi{10.18653/V1/2022.EMNLP-MAIN.755}.
\newblock URL \url{https://doi.org/10.18653/v1/2022.emnlp-main.755}.

\bibitem[Aggarwal et~al.(2024)Aggarwal, Sathe, Watts, and Sitaram]{aggarwal2024maple}
Divyanshu Aggarwal, Ashutosh Sathe, Ishaan Watts, and Sunayana Sitaram.
\newblock Maple: Multilingual evaluation of parameter efficient finetuning of large language models.
\newblock \emph{arXiv preprint arXiv:2401.07598}, 2024.

\bibitem[Ahmad et~al.(2021)Ahmad, Chakraborty, Ray, and Chang]{DBLP:conf/naacl/AhmadCRC21}
Wasi~Uddin Ahmad, Saikat Chakraborty, Baishakhi Ray, and Kai{-}Wei Chang.
\newblock Unified pre-training for program understanding and generation.
\newblock In Kristina Toutanova, Anna Rumshisky, Luke Zettlemoyer, Dilek Hakkani{-}T{\"{u}}r, Iz~Beltagy, Steven Bethard, Ryan Cotterell, Tanmoy Chakraborty, and Yichao Zhou (eds.), \emph{Proceedings of the 2021 Conference of the North American Chapter of the Association for Computational Linguistics: Human Language Technologies, {NAACL-HLT} 2021, Online, June 6-11, 2021}, pp.\  2655--2668. Association for Computational Linguistics, 2021.
\newblock \doi{10.18653/V1/2021.NAACL-MAIN.211}.
\newblock URL \url{https://doi.org/10.18653/v1/2021.naacl-main.211}.

\bibitem[Ahuja et~al.(2023{\natexlab{a}})Ahuja, Diddee, Hada, Ochieng, Ramesh, Jain, Nambi, Ganu, Segal, Ahmed, Bali, and Sitaram]{DBLP:conf/emnlp/AhujaDHORJNGSAB23}
Kabir Ahuja, Harshita Diddee, Rishav Hada, Millicent Ochieng, Krithika Ramesh, Prachi Jain, Akshay~Uttama Nambi, Tanuja Ganu, Sameer Segal, Mohamed Ahmed, Kalika Bali, and Sunayana Sitaram.
\newblock {MEGA:} multilingual evaluation of generative {AI}.
\newblock In Houda Bouamor, Juan Pino, and Kalika Bali (eds.), \emph{Proceedings of the 2023 Conference on Empirical Methods in Natural Language Processing, {EMNLP} 2023, Singapore, December 6-10, 2023}, pp.\  4232--4267. Association for Computational Linguistics, 2023{\natexlab{a}}.
\newblock \doi{10.18653/V1/2023.EMNLP-MAIN.258}.
\newblock URL \url{https://doi.org/10.18653/v1/2023.emnlp-main.258}.

\bibitem[Ahuja et~al.(2023{\natexlab{b}})Ahuja, Aggarwal, Gumma, Watts, Sathe, Ochieng, Hada, Jain, Axmed, Bali, and Sitaram]{DBLP:journals/corr/abs-2311-07463}
Sanchit Ahuja, Divyanshu Aggarwal, Varun Gumma, Ishaan Watts, Ashutosh Sathe, Millicent Ochieng, Rishav Hada, Prachi Jain, Maxamed Axmed, Kalika Bali, and Sunayana Sitaram.
\newblock {MEGAVERSE:} benchmarking large language models across languages, modalities, models and tasks.
\newblock \emph{CoRR}, abs/2311.07463, 2023{\natexlab{b}}.
\newblock \doi{10.48550/ARXIV.2311.07463}.
\newblock URL \url{https://doi.org/10.48550/arXiv.2311.07463}.

\bibitem[AI@Meta(2024)]{LLaMA-3}
AI@Meta.
\newblock Llama 3 model card.
\newblock 2024.
\newblock URL \url{https://github.com/meta-llama/llama3/blob/main/MODEL_CARD.md}.

\bibitem[Ainslie et~al.(2023)Ainslie, Lee-Thorp, de~Jong, Zemlyanskiy, Lebr{\'o}n, and Sanghai]{GQA}
Joshua Ainslie, James Lee-Thorp, Michiel de~Jong, Yury Zemlyanskiy, Federico Lebr{\'o}n, and Sumit Sanghai.
\newblock Gqa: Training generalized multi-query transformer models from multi-head checkpoints.
\newblock \emph{arXiv preprint arXiv:2305.13245}, 2023.

\bibitem[Ali et~al.(2023)Ali, Fromm, Thellmann, Rutmann, L{\"{u}}bbering, Leveling, Klug, Ebert, Doll, Buschhoff, Jain, Weber, Jurkschat, Abdelwahab, John, Suarez, Ostendorff, Weinbach, Sifa, Kesselheim, and Flores{-}Herr]{DBLP:journals/corr/abs-2310-08754}
Mehdi Ali, Michael Fromm, Klaudia Thellmann, Richard Rutmann, Max L{\"{u}}bbering, Johannes Leveling, Katrin Klug, Jan Ebert, Niclas Doll, Jasper~Schulze Buschhoff, Charvi Jain, Alexander~Arno Weber, Lena Jurkschat, Hammam Abdelwahab, Chelsea John, Pedro~Ortiz Suarez, Malte Ostendorff, Samuel Weinbach, Rafet Sifa, Stefan Kesselheim, and Nicolas Flores{-}Herr.
\newblock Tokenizer choice for {LLM} training: Negligible or crucial?
\newblock \emph{CoRR}, abs/2310.08754, 2023.
\newblock \doi{10.48550/ARXIV.2310.08754}.
\newblock URL \url{https://doi.org/10.48550/arXiv.2310.08754}.

\bibitem[Aljundi et~al.(2018)Aljundi, Babiloni, Elhoseiny, Rohrbach, and Tuytelaars]{cpt_9}
Rahaf Aljundi, Francesca Babiloni, Mohamed Elhoseiny, Marcus Rohrbach, and Tinne Tuytelaars.
\newblock Memory aware synapses: Learning what (not) to forget.
\newblock In Vittorio Ferrari, Martial Hebert, Cristian Sminchisescu, and Yair Weiss (eds.), \emph{Computer Vision - {ECCV} 2018 - 15th European Conference, Munich, Germany, September 8-14, 2018, Proceedings, Part {III}}, volume 11207 of \emph{Lecture Notes in Computer Science}, pp.\  144--161. Springer, 2018.
\newblock \doi{10.1007/978-3-030-01219-9\_9}.
\newblock URL \url{https://doi.org/10.1007/978-3-030-01219-9\_9}.

\bibitem[AlKhamissi et~al.(2024)AlKhamissi, ElNokrashy, AlKhamissi, and Diab]{c-3}
Badr AlKhamissi, Muhammad~N. ElNokrashy, Mai AlKhamissi, and Mona~T. Diab.
\newblock Investigating cultural alignment of large language models.
\newblock \emph{CoRR}, abs/2402.13231, 2024.
\newblock \doi{10.48550/ARXIV.2402.13231}.
\newblock URL \url{https://doi.org/10.48550/arXiv.2402.13231}.

\bibitem[Almazrouei et~al.(2023)Almazrouei, Alobeidli, Alshamsi, Cappelli, Cojocaru, Debbah, Goffinet, Hesslow, Launay, Malartic, et~al.]{falcon}
Ebtesam Almazrouei, Hamza Alobeidli, Abdulaziz Alshamsi, Alessandro Cappelli, Ruxandra Cojocaru, M{\'e}rouane Debbah, {\'E}tienne Goffinet, Daniel Hesslow, Julien Launay, Quentin Malartic, et~al.
\newblock The falcon series of open language models.
\newblock \emph{arXiv preprint arXiv:2311.16867}, 2023.

\bibitem[Arora et~al.(2022)Arora, Kaffee, and Augenstein]{c-6}
Arnav Arora, Lucie{-}Aim{\'{e}}e Kaffee, and Isabelle Augenstein.
\newblock Probing pre-trained language models for cross-cultural differences in values.
\newblock \emph{CoRR}, abs/2203.13722, 2022.
\newblock \doi{10.48550/ARXIV.2203.13722}.
\newblock URL \url{https://doi.org/10.48550/arXiv.2203.13722}.

\bibitem[Artetxe et~al.(2020)Artetxe, Ruder, and Yogatama]{DBLP:conf/acl/ArtetxeRY20}
Mikel Artetxe, Sebastian Ruder, and Dani Yogatama.
\newblock On the cross-lingual transferability of monolingual representations.
\newblock In Dan Jurafsky, Joyce Chai, Natalie Schluter, and Joel~R. Tetreault (eds.), \emph{Proceedings of the 58th Annual Meeting of the Association for Computational Linguistics, {ACL} 2020, Online, July 5-10, 2020}, pp.\  4623--4637. Association for Computational Linguistics, 2020.
\newblock \doi{10.18653/V1/2020.ACL-MAIN.421}.
\newblock URL \url{https://doi.org/10.18653/v1/2020.acl-main.421}.

\bibitem[Askell et~al.(2021)Askell, Bai, Chen, Drain, Ganguli, Henighan, Jones, Joseph, Mann, DasSarma, Elhage, Hatfield{-}Dodds, Hernandez, Kernion, Ndousse, Olsson, Amodei, Brown, Clark, McCandlish, Olah, and Kaplan]{DBLP:journals/corr/abs-2112-00861}
Amanda Askell, Yuntao Bai, Anna Chen, Dawn Drain, Deep Ganguli, Tom Henighan, Andy Jones, Nicholas Joseph, Benjamin Mann, Nova DasSarma, Nelson Elhage, Zac Hatfield{-}Dodds, Danny Hernandez, Jackson Kernion, Kamal Ndousse, Catherine Olsson, Dario Amodei, Tom~B. Brown, Jack Clark, Sam McCandlish, Chris Olah, and Jared Kaplan.
\newblock A general language assistant as a laboratory for alignment.
\newblock \emph{CoRR}, abs/2112.00861, 2021.
\newblock URL \url{https://arxiv.org/abs/2112.00861}.

\bibitem[Azar et~al.(2024)Azar, Guo, Piot, Munos, Rowland, Valko, and Calandriello]{DBLP:conf/aistats/AzarGPMRVC24}
Mohammad~Gheshlaghi Azar, Zhaohan~Daniel Guo, Bilal Piot, R{\'{e}}mi Munos, Mark Rowland, Michal Valko, and Daniele Calandriello.
\newblock A general theoretical paradigm to understand learning from human preferences.
\newblock In Sanjoy Dasgupta, Stephan Mandt, and Yingzhen Li (eds.), \emph{International Conference on Artificial Intelligence and Statistics, 2-4 May 2024, Palau de Congressos, Valencia, Spain}, volume 238 of \emph{Proceedings of Machine Learning Research}, pp.\  4447--4455. {PMLR}, 2024.
\newblock URL \url{https://proceedings.mlr.press/v238/gheshlaghi-azar24a.html}.

\bibitem[Bai et~al.(2023{\natexlab{a}})Bai, Bai, Chu, Cui, Dang, Deng, Fan, Ge, Han, Huang, Hui, Ji, Li, Lin, Lin, Liu, Liu, Lu, Lu, Ma, Men, Ren, Ren, Tan, Tan, Tu, Wang, Wang, Wang, Wu, Xu, Xu, Yang, Yang, Yang, Yang, Yao, Yu, Yuan, Yuan, Zhang, Zhang, Zhang, Zhang, Zhou, Zhou, Zhou, and Zhu]{DBLP:journals/corr/abs-2309-16609}
Jinze Bai, Shuai Bai, Yunfei Chu, Zeyu Cui, Kai Dang, Xiaodong Deng, Yang Fan, Wenbin Ge, Yu~Han, Fei Huang, Binyuan Hui, Luo Ji, Mei Li, Junyang Lin, Runji Lin, Dayiheng Liu, Gao Liu, Chengqiang Lu, Keming Lu, Jianxin Ma, Rui Men, Xingzhang Ren, Xuancheng Ren, Chuanqi Tan, Sinan Tan, Jianhong Tu, Peng Wang, Shijie Wang, Wei Wang, Shengguang Wu, Benfeng Xu, Jin Xu, An~Yang, Hao Yang, Jian Yang, Shusheng Yang, Yang Yao, Bowen Yu, Hongyi Yuan, Zheng Yuan, Jianwei Zhang, Xingxuan Zhang, Yichang Zhang, Zhenru Zhang, Chang Zhou, Jingren Zhou, Xiaohuan Zhou, and Tianhang Zhu.
\newblock Qwen technical report.
\newblock \emph{CoRR}, abs/2309.16609, 2023{\natexlab{a}}.
\newblock \doi{10.48550/ARXIV.2309.16609}.
\newblock URL \url{https://doi.org/10.48550/arXiv.2309.16609}.

\bibitem[Bai et~al.(2023{\natexlab{b}})Bai, Bai, Chu, Cui, Dang, Deng, Fan, Ge, Han, Huang, et~al.]{qwen}
Jinze Bai, Shuai Bai, Yunfei Chu, Zeyu Cui, Kai Dang, Xiaodong Deng, Yang Fan, Wenbin Ge, Yu~Han, Fei Huang, et~al.
\newblock Qwen technical report.
\newblock \emph{arXiv preprint arXiv:2309.16609}, 2023{\natexlab{b}}.

\bibitem[Bai et~al.(2022)Bai, Jones, Ndousse, Askell, Chen, DasSarma, Drain, Fort, Ganguli, Henighan, Joseph, Kadavath, Kernion, Conerly, Showk, Elhage, Hatfield{-}Dodds, Hernandez, Hume, Johnston, Kravec, Lovitt, Nanda, Olsson, Amodei, Brown, Clark, McCandlish, Olah, Mann, and Kaplan]{DBLP:journals/corr/abs-2204-05862}
Yuntao Bai, Andy Jones, Kamal Ndousse, Amanda Askell, Anna Chen, Nova DasSarma, Dawn Drain, Stanislav Fort, Deep Ganguli, Tom Henighan, Nicholas Joseph, Saurav Kadavath, Jackson Kernion, Tom Conerly, Sheer~El Showk, Nelson Elhage, Zac Hatfield{-}Dodds, Danny Hernandez, Tristan Hume, Scott Johnston, Shauna Kravec, Liane Lovitt, Neel Nanda, Catherine Olsson, Dario Amodei, Tom~B. Brown, Jack Clark, Sam McCandlish, Chris Olah, Benjamin Mann, and Jared Kaplan.
\newblock Training a helpful and harmless assistant with reinforcement learning from human feedback.
\newblock \emph{CoRR}, abs/2204.05862, 2022.
\newblock \doi{10.48550/ARXIV.2204.05862}.
\newblock URL \url{https://doi.org/10.48550/arXiv.2204.05862}.

\bibitem[Bandarkar et~al.(2023)Bandarkar, Liang, Muller, Artetxe, Shukla, Husa, Goyal, Krishnan, Zettlemoyer, and Khabsa]{DBLP:journals/corr/abs-2308-16884}
Lucas Bandarkar, Davis Liang, Benjamin Muller, Mikel Artetxe, Satya~Narayan Shukla, Donald Husa, Naman Goyal, Abhinandan Krishnan, Luke Zettlemoyer, and Madian Khabsa.
\newblock The belebele benchmark: a parallel reading comprehension dataset in 122 language variants.
\newblock \emph{CoRR}, abs/2308.16884, 2023.
\newblock \doi{10.48550/ARXIV.2308.16884}.
\newblock URL \url{https://doi.org/10.48550/arXiv.2308.16884}.

\bibitem[Bang et~al.(2021)Bang, Kim, Yoo, Ha, and Choi]{cpt_7}
Jihwan Bang, Heesu Kim, Youngjoon Yoo, Jung{-}Woo Ha, and Jonghyun Choi.
\newblock Rainbow memory: Continual learning with a memory of diverse samples.
\newblock In \emph{{IEEE} Conference on Computer Vision and Pattern Recognition, {CVPR} 2021, virtual, June 19-25, 2021}, pp.\  8218--8227. Computer Vision Foundation / {IEEE}, 2021.
\newblock \doi{10.1109/CVPR46437.2021.00812}.
\newblock URL \url{https://openaccess.thecvf.com/content/CVPR2021/html/Bang\_Rainbow\_Memory\_Continual\_Learning\_With\_a\_Memory\_of\_Diverse\_Samples\_CVPR\_2021\_paper.html}.

\bibitem[Ba{\~{n}}{\'{o}}n et~al.(2020)Ba{\~{n}}{\'{o}}n, Chen, Haddow, Heafield, Hoang, Espl{\`{a}}{-}Gomis, Forcada, Kamran, Kirefu, Koehn, Ortiz{-}Rojas, Sempere, Ram{\'{\i}}rez{-}S{\'{a}}nchez, Sarr{\'{\i}}as, Strelec, Thompson, Waites, Wiggins, and Zaragoza]{ParaCrawl}
Marta Ba{\~{n}}{\'{o}}n, Pinzhen Chen, Barry Haddow, Kenneth Heafield, Hieu Hoang, Miquel Espl{\`{a}}{-}Gomis, Mikel~L. Forcada, Amir Kamran, Faheem Kirefu, Philipp Koehn, Sergio Ortiz{-}Rojas, Leopoldo~Pla Sempere, Gema Ram{\'{\i}}rez{-}S{\'{a}}nchez, Elsa Sarr{\'{\i}}as, Marek Strelec, Brian Thompson, William Waites, Dion Wiggins, and Jaume Zaragoza.
\newblock Paracrawl: Web-scale acquisition of parallel corpora.
\newblock In Dan Jurafsky, Joyce Chai, Natalie Schluter, and Joel~R. Tetreault (eds.), \emph{Proceedings of the 58th Annual Meeting of the Association for Computational Linguistics, {ACL} 2020, Online, July 5-10, 2020}, pp.\  4555--4567. Association for Computational Linguistics, 2020.
\newblock \doi{10.18653/V1/2020.ACL-MAIN.417}.
\newblock URL \url{https://doi.org/10.18653/v1/2020.acl-main.417}.

\bibitem[Bawden \& Yvon(2023)Bawden and Yvon]{DBLP:conf/eamt/BawdenY23}
Rachel Bawden and Fran{\c{c}}ois Yvon.
\newblock Investigating the translation performance of a large multilingual language model: the case of {BLOOM}.
\newblock In Mary Nurminen, Judith Brenner, Maarit Koponen, Sirkku Latomaa, Mikhail Mikhailov, Frederike Schierl, Tharindu Ranasinghe, Eva Vanmassenhove, Sergi~Alvarez Vidal, Nora Aranberri, Mara Nunziatini, Carla~Parra Escart{\'{\i}}n, Mikel~L. Forcada, Maja Popovic, Carolina Scarton, and Helena Moniz (eds.), \emph{Proceedings of the 24th Annual Conference of the European Association for Machine Translation, {EAMT} 2023, Tampere, Finland, 12-15 June 2023}, pp.\  157--170. European Association for Machine Translation, 2023.
\newblock URL \url{https://aclanthology.org/2023.eamt-1.16}.

\bibitem[Bendale et~al.(2024)Bendale, Sapienza, Ripplinger, Gibbs, Lee, and Mistry]{bendale2024sutra}
Abhijit Bendale, Michael Sapienza, Steven Ripplinger, Simon Gibbs, Jaewon Lee, and Pranav Mistry.
\newblock Sutra: Scalable multilingual language model architecture.
\newblock \emph{arXiv preprint arXiv:2405.06694}, 2024.

\bibitem[Bhattacharya \& Bojar(2023)Bhattacharya and Bojar]{DBLP:conf/blackboxnlp/BhattacharyaB23}
Sunit Bhattacharya and Ondrej Bojar.
\newblock Unveiling multilinguality in transformer models: Exploring language specificity in feed-forward networks.
\newblock In Yonatan Belinkov, Sophie Hao, Jaap Jumelet, Najoung Kim, Arya McCarthy, and Hosein Mohebbi (eds.), \emph{Proceedings of the 6th BlackboxNLP Workshop: Analyzing and Interpreting Neural Networks for NLP, BlackboxNLP@EMNLP 2023, Singapore, December 7, 2023}, pp.\  120--126. Association for Computational Linguistics, 2023.
\newblock \doi{10.18653/V1/2023.BLACKBOXNLP-1.9}.
\newblock URL \url{https://doi.org/10.18653/v1/2023.blackboxnlp-1.9}.

\bibitem[Bi et~al.(2024)Bi, Chen, Chen, Chen, Dai, Deng, Ding, Dong, Du, Fu, et~al.]{DeepseekLLM}
Xiao Bi, Deli Chen, Guanting Chen, Shanhuang Chen, Damai Dai, Chengqi Deng, Honghui Ding, Kai Dong, Qiushi Du, Zhe Fu, et~al.
\newblock Deepseek llm: Scaling open-source language models with longtermism.
\newblock \emph{arXiv preprint arXiv:2401.02954}, 2024.

\bibitem[Black et~al.(2021)Black, Gao, Wang, Leahy, and Biderman]{gpt-neo}
Sid Black, Leo Gao, Phil Wang, Connor Leahy, and Stella Biderman.
\newblock {GPT-Neo: Large Scale Autoregressive Language Modeling with Mesh-Tensorflow}, March 2021.
\newblock URL \url{https://doi.org/10.5281/zenodo.5297715}.

\bibitem[Blevins et~al.(2022)Blevins, Gonen, and Zettlemoyer]{DBLP:conf/emnlp/BlevinsGZ22}
Terra Blevins, Hila Gonen, and Luke Zettlemoyer.
\newblock Analyzing the mono- and cross-lingual pretraining dynamics of multilingual language models.
\newblock In Yoav Goldberg, Zornitsa Kozareva, and Yue Zhang (eds.), \emph{Proceedings of the 2022 Conference on Empirical Methods in Natural Language Processing, {EMNLP} 2022, Abu Dhabi, United Arab Emirates, December 7-11, 2022}, pp.\  3575--3590. Association for Computational Linguistics, 2022.
\newblock \doi{10.18653/V1/2022.EMNLP-MAIN.234}.
\newblock URL \url{https://doi.org/10.18653/v1/2022.emnlp-main.234}.

\bibitem[Blevins et~al.(2024)Blevins, Limisiewicz, Gururangan, Li, Gonen, Smith, and Zettlemoyer]{blevins2024breaking}
Terra Blevins, Tomasz Limisiewicz, Suchin Gururangan, Margaret Li, Hila Gonen, Noah~A Smith, and Luke Zettlemoyer.
\newblock Breaking the curse of multilinguality with cross-lingual expert language models.
\newblock \emph{arXiv preprint arXiv:2401.10440}, 2024.

\bibitem[Bradley \& Terry(1952)Bradley and Terry]{bradley1952rank}
Ralph~Allan Bradley and Milton~E Terry.
\newblock Rank analysis of incomplete block designs: I. the method of paired comparisons.
\newblock \emph{Biometrika}, 39\penalty0 (3/4):\penalty0 324--345, 1952.

\bibitem[Brown et~al.(2020{\natexlab{a}})Brown, Mann, Ryder, Subbiah, Kaplan, Dhariwal, Neelakantan, Shyam, Sastry, Askell, et~al.]{GPT3}
Tom Brown, Benjamin Mann, Nick Ryder, Melanie Subbiah, Jared~D Kaplan, Prafulla Dhariwal, Arvind Neelakantan, Pranav Shyam, Girish Sastry, Amanda Askell, et~al.
\newblock Language models are few-shot learners.
\newblock \emph{Advances in neural information processing systems}, 33:\penalty0 1877--1901, 2020{\natexlab{a}}.

\bibitem[Brown et~al.(2020{\natexlab{b}})Brown, Mann, Ryder, Subbiah, Kaplan, Dhariwal, Neelakantan, Shyam, Sastry, Askell, Agarwal, Herbert{-}Voss, Krueger, Henighan, Child, Ramesh, Ziegler, Wu, Winter, Hesse, Chen, Sigler, Litwin, Gray, Chess, Clark, Berner, McCandlish, Radford, Sutskever, and Amodei]{DBLP:conf/nips/BrownMRSKDNSSAA20}
Tom~B. Brown, Benjamin Mann, Nick Ryder, Melanie Subbiah, Jared Kaplan, Prafulla Dhariwal, Arvind Neelakantan, Pranav Shyam, Girish Sastry, Amanda Askell, Sandhini Agarwal, Ariel Herbert{-}Voss, Gretchen Krueger, Tom Henighan, Rewon Child, Aditya Ramesh, Daniel~M. Ziegler, Jeffrey Wu, Clemens Winter, Christopher Hesse, Mark Chen, Eric Sigler, Mateusz Litwin, Scott Gray, Benjamin Chess, Jack Clark, Christopher Berner, Sam McCandlish, Alec Radford, Ilya Sutskever, and Dario Amodei.
\newblock Language models are few-shot learners.
\newblock In Hugo Larochelle, Marc'Aurelio Ranzato, Raia Hadsell, Maria{-}Florina Balcan, and Hsuan{-}Tien Lin (eds.), \emph{Advances in Neural Information Processing Systems 33: Annual Conference on Neural Information Processing Systems 2020, NeurIPS 2020, December 6-12, 2020, virtual}, 2020{\natexlab{b}}.
\newblock URL \url{https://proceedings.neurips.cc/paper/2020/hash/1457c0d6bfcb4967418bfb8ac142f64a-Abstract.html}.

\bibitem[Buzzega et~al.(2020)Buzzega, Boschini, Porrello, Abati, and Calderara]{cpt_5}
Pietro Buzzega, Matteo Boschini, Angelo Porrello, Davide Abati, and Simone Calderara.
\newblock Dark experience for general continual learning: a strong, simple baseline.
\newblock In Hugo Larochelle, Marc'Aurelio Ranzato, Raia Hadsell, Maria{-}Florina Balcan, and Hsuan{-}Tien Lin (eds.), \emph{Advances in Neural Information Processing Systems 33: Annual Conference on Neural Information Processing Systems 2020, NeurIPS 2020, December 6-12, 2020, virtual}, 2020.
\newblock URL \url{https://proceedings.neurips.cc/paper/2020/hash/b704ea2c39778f07c617f6b7ce480e9e-Abstract.html}.

\bibitem[Cahyawijaya et~al.(2023)Cahyawijaya, Lovenia, Yu, Chung, and Fung]{trans-4}
Samuel Cahyawijaya, Holy Lovenia, Tiezheng Yu, Willy Chung, and Pascale Fung.
\newblock Instruct-align: Teaching novel languages with to llms through alignment-based cross-lingual instruction.
\newblock \emph{CoRR}, abs/2305.13627, 2023.
\newblock \doi{10.48550/ARXIV.2305.13627}.
\newblock URL \url{https://doi.org/10.48550/arXiv.2305.13627}.

\bibitem[Cai et~al.(2024)Cai, Cao, Chen, Chen, Chen, Chen, Chen, Chen, Chen, Chu, et~al.]{internlm2}
Zheng Cai, Maosong Cao, Haojiong Chen, Kai Chen, Keyu Chen, Xin Chen, Xun Chen, Zehui Chen, Zhi Chen, Pei Chu, et~al.
\newblock Internlm2 technical report.
\newblock \emph{arXiv preprint arXiv:2403.17297}, 2024.

\bibitem[Cao et~al.(2023{\natexlab{a}})Cao, Kementchedjhieva, Cui, Karamolegkou, Zhou, Dare, Donatelli, and Hershcovich]{c-32}
Yong Cao, Yova Kementchedjhieva, Ruixiang Cui, Antonia Karamolegkou, Li~Zhou, Megan Dare, Lucia Donatelli, and Daniel Hershcovich.
\newblock Cultural adaptation of recipes.
\newblock \emph{CoRR}, abs/2310.17353, 2023{\natexlab{a}}.
\newblock \doi{10.48550/ARXIV.2310.17353}.
\newblock URL \url{https://doi.org/10.48550/arXiv.2310.17353}.

\bibitem[Cao et~al.(2023{\natexlab{b}})Cao, Zhou, Lee, Cabello, Chen, and Hershcovich]{c-7}
Yong Cao, Li~Zhou, Seolhwa Lee, Laura Cabello, Min Chen, and Daniel Hershcovich.
\newblock Assessing cross-cultural alignment between chatgpt and human societies: An empirical study.
\newblock \emph{CoRR}, abs/2303.17466, 2023{\natexlab{b}}.
\newblock \doi{10.48550/ARXIV.2303.17466}.
\newblock URL \url{https://doi.org/10.48550/arXiv.2303.17466}.

\bibitem[Cao et~al.(2024)Cao, Chen, and Hershcovich]{c-33}
Yong Cao, Min Chen, and Daniel Hershcovich.
\newblock Bridging cultural nuances in dialogue agents through cultural value surveys.
\newblock In Yvette Graham and Matthew Purver (eds.), \emph{Findings of the Association for Computational Linguistics: {EACL} 2024, St. Julian's, Malta, March 17-22, 2024}, pp.\  929--945. Association for Computational Linguistics, 2024.
\newblock URL \url{https://aclanthology.org/2024.findings-eacl.63}.

\bibitem[Chai et~al.(2024{\natexlab{a}})Chai, Yang, Sun, Guo, Liu, Wang, Liang, Bai, Li, Peng, et~al.]{chai2024xcot}
Linzheng Chai, Jian Yang, Tao Sun, Hongcheng Guo, Jiaheng Liu, Bing Wang, Xiannian Liang, Jiaqi Bai, Tongliang Li, Qiyao Peng, et~al.
\newblock xcot: Cross-lingual instruction tuning for cross-lingual chain-of-thought reasoning.
\newblock \emph{arXiv preprint arXiv:2401.07037}, 2024{\natexlab{a}}.

\bibitem[Chai et~al.(2024{\natexlab{b}})Chai, Yang, Sun, Guo, Liu, Wang, Liang, Bai, Li, Peng, and Li]{trans-7}
Linzheng Chai, Jian Yang, Tao Sun, Hongcheng Guo, Jiaheng Liu, Bing Wang, Xinnian Liang, Jiaqi Bai, Tongliang Li, Qiyao Peng, and Zhoujun Li.
\newblock xcot: Cross-lingual instruction tuning for cross-lingual chain-of-thought reasoning.
\newblock \emph{CoRR}, abs/2401.07037, 2024{\natexlab{b}}.
\newblock \doi{10.48550/ARXIV.2401.07037}.
\newblock URL \url{https://doi.org/10.48550/arXiv.2401.07037}.

\bibitem[Chalkidis et~al.(2020)Chalkidis, Fergadiotis, Malakasiotis, Aletras, and Androutsopoulos]{DBLP:journals/corr/abs-2010-02559}
Ilias Chalkidis, Manos Fergadiotis, Prodromos Malakasiotis, Nikolaos Aletras, and Ion Androutsopoulos.
\newblock {LEGAL-BERT:} the muppets straight out of law school.
\newblock \emph{CoRR}, abs/2010.02559, 2020.
\newblock URL \url{https://arxiv.org/abs/2010.02559}.

\bibitem[Chang et~al.(2022)Chang, Tu, and Bergen]{DBLP:conf/emnlp/ChangTB22}
Tyler~A. Chang, Zhuowen Tu, and Benjamin~K. Bergen.
\newblock The geometry of multilingual language model representations.
\newblock In Yoav Goldberg, Zornitsa Kozareva, and Yue Zhang (eds.), \emph{Proceedings of the 2022 Conference on Empirical Methods in Natural Language Processing, {EMNLP} 2022, Abu Dhabi, United Arab Emirates, December 7-11, 2022}, pp.\  119--136. Association for Computational Linguistics, 2022.
\newblock \doi{10.18653/V1/2022.EMNLP-MAIN.9}.
\newblock URL \url{https://doi.org/10.18653/v1/2022.emnlp-main.9}.

\bibitem[Chang et~al.(2024)Chang, Wang, Wang, Wu, Yang, Zhu, Chen, Yi, Wang, Wang, et~al.]{chang2024survey}
Yupeng Chang, Xu~Wang, Jindong Wang, Yuan Wu, Linyi Yang, Kaijie Zhu, Hao Chen, Xiaoyuan Yi, Cunxiang Wang, Yidong Wang, et~al.
\newblock A survey on evaluation of large language models.
\newblock \emph{ACM Transactions on Intelligent Systems and Technology}, 15\penalty0 (3):\penalty0 1--45, 2024.

\bibitem[Chaudhry et~al.(2019)Chaudhry, Rohrbach, Elhoseiny, Ajanthan, Dokania, Torr, and Ranzato]{cpt_3}
Arslan Chaudhry, Marcus Rohrbach, Mohamed Elhoseiny, Thalaiyasingam Ajanthan, Puneet~Kumar Dokania, Philip H.~S. Torr, and Marc'Aurelio Ranzato.
\newblock Continual learning with tiny episodic memories.
\newblock \emph{CoRR}, abs/1902.10486, 2019.
\newblock URL \url{http://arxiv.org/abs/1902.10486}.

\bibitem[Chen et~al.(2024{\natexlab{a}})Chen, Lou, Chen, Bai, Xiang, Yang, Zhao, and Zhang]{chen2024dual}
Andong Chen, Lianzhang Lou, Kehai Chen, Xuefeng Bai, Yang Xiang, Muyun Yang, Tiejun Zhao, and Min Zhang.
\newblock Dual-reflect: Enhancing large language models for reflective translation through dual learning feedback mechanisms.
\newblock \emph{arXiv preprint arXiv:2406.07232}, 2024{\natexlab{a}}.

\bibitem[Chen et~al.(2021)Chen, Tworek, Jun, Yuan, de~Oliveira~Pinto, Kaplan, Edwards, Burda, Joseph, Brockman, Ray, Puri, Krueger, Petrov, Khlaaf, Sastry, Mishkin, Chan, Gray, Ryder, Pavlov, Power, Kaiser, Bavarian, Winter, Tillet, Such, Cummings, Plappert, Chantzis, Barnes, Herbert{-}Voss, Guss, Nichol, Paino, Tezak, Tang, Babuschkin, Balaji, Jain, Saunders, Hesse, Carr, Leike, Achiam, Misra, Morikawa, Radford, Knight, Brundage, Murati, Mayer, Welinder, McGrew, Amodei, McCandlish, Sutskever, and Zaremba]{DBLP:journals/corr/abs-2107-03374}
Mark Chen, Jerry Tworek, Heewoo Jun, Qiming Yuan, Henrique~Pond{\'{e}} de~Oliveira~Pinto, Jared Kaplan, Harrison Edwards, Yuri Burda, Nicholas Joseph, Greg Brockman, Alex Ray, Raul Puri, Gretchen Krueger, Michael Petrov, Heidy Khlaaf, Girish Sastry, Pamela Mishkin, Brooke Chan, Scott Gray, Nick Ryder, Mikhail Pavlov, Alethea Power, Lukasz Kaiser, Mohammad Bavarian, Clemens Winter, Philippe Tillet, Felipe~Petroski Such, Dave Cummings, Matthias Plappert, Fotios Chantzis, Elizabeth Barnes, Ariel Herbert{-}Voss, William~Hebgen Guss, Alex Nichol, Alex Paino, Nikolas Tezak, Jie Tang, Igor Babuschkin, Suchir Balaji, Shantanu Jain, William Saunders, Christopher Hesse, Andrew~N. Carr, Jan Leike, Joshua Achiam, Vedant Misra, Evan Morikawa, Alec Radford, Matthew Knight, Miles Brundage, Mira Murati, Katie Mayer, Peter Welinder, Bob McGrew, Dario Amodei, Sam McCandlish, Ilya Sutskever, and Wojciech Zaremba.
\newblock Evaluating large language models trained on code.
\newblock \emph{CoRR}, abs/2107.03374, 2021.
\newblock URL \url{https://arxiv.org/abs/2107.03374}.

\bibitem[Chen et~al.(2024{\natexlab{b}})Chen, Ji, Bogoychev, Kutuzov, Haddow, and Heafield]{tuning-11}
Pinzhen Chen, Shaoxiong Ji, Nikolay Bogoychev, Andrey Kutuzov, Barry Haddow, and Kenneth Heafield.
\newblock Monolingual or multilingual instruction tuning: Which makes a better alpaca.
\newblock In Yvette Graham and Matthew Purver (eds.), \emph{Findings of the Association for Computational Linguistics: {EACL} 2024, St. Julian's, Malta, March 17-22, 2024}, pp.\  1347--1356. Association for Computational Linguistics, 2024{\natexlab{b}}.
\newblock URL \url{https://aclanthology.org/2024.findings-eacl.90}.

\bibitem[Chen et~al.(2023{\natexlab{a}})Chen, Liu, Meng, Chen, Xu, and Zhou]{mt-8}
Yijie Chen, Yijin Liu, Fandong Meng, Yufeng Chen, Jinan Xu, and Jie Zhou.
\newblock Improving translation faithfulness of large language models via augmenting instructions.
\newblock \emph{CoRR}, abs/2308.12674, 2023{\natexlab{a}}.
\newblock \doi{10.48550/ARXIV.2308.12674}.
\newblock URL \url{https://doi.org/10.48550/arXiv.2308.12674}.

\bibitem[Chen et~al.(2023{\natexlab{b}})Chen, Wang, Xing, Zheng, Xu, Fang, Wang, Li, Wu, Liu, and Xu]{DBLP:journals/corr/abs-2310-15896}
Yirong Chen, Zhenyu Wang, Xiaofen Xing, Huimin Zheng, Zhipei Xu, Kai Fang, Junhong Wang, Sihang Li, Jieling Wu, Qi~Liu, and Xiangmin Xu.
\newblock Bianque: Balancing the questioning and suggestion ability of health llms with multi-turn health conversations polished by chatgpt.
\newblock \emph{CoRR}, abs/2310.15896, 2023{\natexlab{b}}.
\newblock \doi{10.48550/ARXIV.2310.15896}.
\newblock URL \url{https://doi.org/10.48550/arXiv.2310.15896}.

\bibitem[Chen et~al.(2023{\natexlab{c}})Chen, Xing, Lin, Zheng, Wang, Liu, and Xu]{DBLP:conf/emnlp/ChenXLZWLX23}
Yirong Chen, Xiaofen Xing, Jingkai Lin, Huimin Zheng, Zhenyu Wang, Qi~Liu, and Xiangmin Xu.
\newblock Soulchat: Improving llms' empathy, listening, and comfort abilities through fine-tuning with multi-turn empathy conversations.
\newblock In Houda Bouamor, Juan Pino, and Kalika Bali (eds.), \emph{Findings of the Association for Computational Linguistics: {EMNLP} 2023, Singapore, December 6-10, 2023}, pp.\  1170--1183. Association for Computational Linguistics, 2023{\natexlab{c}}.
\newblock \doi{10.18653/V1/2023.FINDINGS-EMNLP.83}.
\newblock URL \url{https://doi.org/10.18653/v1/2023.findings-emnlp.83}.

\bibitem[Chen et~al.(2023{\natexlab{d}})Chen, Chen, Zhang, Jiang, Chen, Yu, Wang, Liang, Zhang, Zhang, Li, Wan, Li, and Wang]{llm-zoo-2023}
Zhihong Chen, Junying Chen, Hongbo Zhang, Feng Jiang, Guiming Chen, Fei Yu, Tiannan Wang, Juhao Liang, Chen Zhang, Zhiyi Zhang, Jianquan Li, Xiang Wan, Haizhou Li, and Benyou Wang.
\newblock Llm zoo: democratizing chatgpt.
\newblock \url{https://github.com/FreedomIntelligence/LLMZoo}, 2023{\natexlab{d}}.

\bibitem[Chen et~al.(2023{\natexlab{e}})Chen, Jiang, Chen, Wang, Yu, Chen, Zhang, Liang, Zhang, Zhang, Li, Wan, Wang, and Li]{phoenix-2023}
Zhihong Chen, Feng Jiang, Junying Chen, Tiannan Wang, Fei Yu, Guiming Chen, Hongbo Zhang, Juhao Liang, Chen Zhang, Zhiyi Zhang, Jianquan Li, Xiang Wan, Benyou Wang, and Haizhou Li.
\newblock Phoenix: Democratizing chatgpt across languages.
\newblock \emph{arXiv preprint arXiv:2304.10453}, 2023{\natexlab{e}}.

\bibitem[Chen et~al.(2023{\natexlab{f}})Chen, Jiang, Chen, Wang, Yu, Chen, Zhang, Liang, Zhang, Zhang, Li, Wan, Wang, and Li]{tuning-4}
Zhihong Chen, Feng Jiang, Junying Chen, Tiannan Wang, Fei Yu, Guiming Chen, Hongbo Zhang, Juhao Liang, Chen Zhang, Zhiyi Zhang, Jianquan Li, Xiang Wan, Benyou Wang, and Haizhou Li.
\newblock Phoenix: Democratizing chatgpt across languages.
\newblock \emph{CoRR}, abs/2304.10453, 2023{\natexlab{f}}.
\newblock \doi{10.48550/ARXIV.2304.10453}.
\newblock URL \url{https://doi.org/10.48550/arXiv.2304.10453}.

\bibitem[Chern et~al.(2023)Chern, Zou, Li, Hu, Feng, Li, and Liu]{abel}
Ethan Chern, Haoyang Zou, Xuefeng Li, Jiewen Hu, Kehua Feng, Junlong Li, and Pengfei Liu.
\newblock Generative ai for math: Abel.
\newblock \url{https://github.com/GAIR-NLP/abel}, 2023.

\bibitem[Chirkova \& Nikoulina(2024)Chirkova and Nikoulina]{tuning-13}
Nadezhda Chirkova and Vassilina Nikoulina.
\newblock Zero-shot cross-lingual transfer in instruction tuning of large language model.
\newblock \emph{CoRR}, abs/2402.14778, 2024.
\newblock \doi{10.48550/ARXIV.2402.14778}.
\newblock URL \url{https://doi.org/10.48550/arXiv.2402.14778}.

\bibitem[Chiu et~al.(2024)Chiu, Jiang, Antoniak, Park, Li, Bhatia, Ravi, Tsvetkov, Shwartz, and Choi]{c-29}
Yu~Ying Chiu, Liwei Jiang, Maria Antoniak, Chan~Young Park, Shuyue~Stella Li, Mehar Bhatia, Sahithya Ravi, Yulia Tsvetkov, Vered Shwartz, and Yejin Choi.
\newblock Culturalteaming: Ai-assisted interactive red-teaming for challenging llms' (lack of) multicultural knowledge.
\newblock \emph{CoRR}, abs/2404.06664, 2024.
\newblock \doi{10.48550/ARXIV.2404.06664}.
\newblock URL \url{https://doi.org/10.48550/arXiv.2404.06664}.

\bibitem[Cho et~al.(2014)Cho, van Merrienboer, G{\"{u}}l{\c{c}}ehre, Bahdanau, Bougares, Schwenk, and Bengio]{DBLP:conf/emnlp/ChoMGBBSB14}
Kyunghyun Cho, Bart van Merrienboer, {\c{C}}aglar G{\"{u}}l{\c{c}}ehre, Dzmitry Bahdanau, Fethi Bougares, Holger Schwenk, and Yoshua Bengio.
\newblock Learning phrase representations using {RNN} encoder-decoder for statistical machine translation.
\newblock In Alessandro Moschitti, Bo~Pang, and Walter Daelemans (eds.), \emph{Proceedings of the 2014 Conference on Empirical Methods in Natural Language Processing, {EMNLP} 2014, October 25-29, 2014, Doha, Qatar, {A} meeting of SIGDAT, a Special Interest Group of the {ACL}}, pp.\  1724--1734. {ACL}, 2014.
\newblock \doi{10.3115/V1/D14-1179}.
\newblock URL \url{https://doi.org/10.3115/v1/d14-1179}.

\bibitem[Choenni et~al.(2024)Choenni, Lauscher, and Shutova]{c-36}
Rochelle Choenni, Anne Lauscher, and Ekaterina Shutova.
\newblock The echoes of multilinguality: Tracing cultural value shifts during lm fine-tuning.
\newblock \emph{CoRR}, abs/2405.12744, 2024.
\newblock \doi{10.48550/ARXIV.2405.12744}.
\newblock URL \url{https://doi.org/10.48550/arXiv.2405.12744}.

\bibitem[Chowdhery et~al.(2022)Chowdhery, Narang, Devlin, Bosma, Mishra, Roberts, Barham, Chung, Sutton, Gehrmann, Schuh, Shi, Tsvyashchenko, Maynez, Rao, Barnes, Tay, Shazeer, Prabhakaran, Reif, Du, Hutchinson, Pope, Bradbury, Austin, Isard, Gur-Ari, Yin, Duke, Levskaya, Ghemawat, Dev, Michalewski, Garcia, Misra, Robinson, Fedus, Zhou, Ippolito, Luan, Lim, Zoph, Spiridonov, Sepassi, Dohan, Agrawal, Omernick, Dai, Pillai, Pellat, Lewkowycz, Moreira, Child, Polozov, Lee, Zhou, Wang, Saeta, Diaz, Firat, Catasta, Wei, Meier-Hellstern, Eck, Dean, Petrov, and Fiedel]{chowdhery2022palmscalinglanguagemodeling}
Aakanksha Chowdhery, Sharan Narang, Jacob Devlin, Maarten Bosma, Gaurav Mishra, Adam Roberts, Paul Barham, Hyung~Won Chung, Charles Sutton, Sebastian Gehrmann, Parker Schuh, Kensen Shi, Sasha Tsvyashchenko, Joshua Maynez, Abhishek Rao, Parker Barnes, Yi~Tay, Noam Shazeer, Vinodkumar Prabhakaran, Emily Reif, Nan Du, Ben Hutchinson, Reiner Pope, James Bradbury, Jacob Austin, Michael Isard, Guy Gur-Ari, Pengcheng Yin, Toju Duke, Anselm Levskaya, Sanjay Ghemawat, Sunipa Dev, Henryk Michalewski, Xavier Garcia, Vedant Misra, Kevin Robinson, Liam Fedus, Denny Zhou, Daphne Ippolito, David Luan, Hyeontaek Lim, Barret Zoph, Alexander Spiridonov, Ryan Sepassi, David Dohan, Shivani Agrawal, Mark Omernick, Andrew~M. Dai, Thanumalayan~Sankaranarayana Pillai, Marie Pellat, Aitor Lewkowycz, Erica Moreira, Rewon Child, Oleksandr Polozov, Katherine Lee, Zongwei Zhou, Xuezhi Wang, Brennan Saeta, Mark Diaz, Orhan Firat, Michele Catasta, Jason Wei, Kathy Meier-Hellstern, Douglas Eck, Jeff Dean, Slav Petrov, and Noah Fiedel.
\newblock Palm: Scaling language modeling with pathways, 2022.
\newblock URL \url{https://arxiv.org/abs/2204.02311}.

\bibitem[Chowdhery et~al.(2023)Chowdhery, Narang, Devlin, Bosma, Mishra, Roberts, Barham, Chung, Sutton, Gehrmann, Schuh, Shi, Tsvyashchenko, Maynez, Rao, Barnes, Tay, Shazeer, Prabhakaran, Reif, Du, Hutchinson, Pope, Bradbury, Austin, Isard, Gur{-}Ari, Yin, Duke, Levskaya, Ghemawat, Dev, Michalewski, Garcia, Misra, Robinson, Fedus, Zhou, Ippolito, Luan, Lim, Zoph, Spiridonov, Sepassi, Dohan, Agrawal, Omernick, Dai, Pillai, Pellat, Lewkowycz, Moreira, Child, Polozov, Lee, Zhou, Wang, Saeta, Diaz, Firat, Catasta, Wei, Meier{-}Hellstern, Eck, Dean, Petrov, and Fiedel]{DBLP:journals/jmlr/ChowdheryNDBMRBCSGSSTMRBTSPRDHPBAI23}
Aakanksha Chowdhery, Sharan Narang, Jacob Devlin, Maarten Bosma, Gaurav Mishra, Adam Roberts, Paul Barham, Hyung~Won Chung, Charles Sutton, Sebastian Gehrmann, Parker Schuh, Kensen Shi, Sasha Tsvyashchenko, Joshua Maynez, Abhishek Rao, Parker Barnes, Yi~Tay, Noam Shazeer, Vinodkumar Prabhakaran, Emily Reif, Nan Du, Ben Hutchinson, Reiner Pope, James Bradbury, Jacob Austin, Michael Isard, Guy Gur{-}Ari, Pengcheng Yin, Toju Duke, Anselm Levskaya, Sanjay Ghemawat, Sunipa Dev, Henryk Michalewski, Xavier Garcia, Vedant Misra, Kevin Robinson, Liam Fedus, Denny Zhou, Daphne Ippolito, David Luan, Hyeontaek Lim, Barret Zoph, Alexander Spiridonov, Ryan Sepassi, David Dohan, Shivani Agrawal, Mark Omernick, Andrew~M. Dai, Thanumalayan~Sankaranarayana Pillai, Marie Pellat, Aitor Lewkowycz, Erica Moreira, Rewon Child, Oleksandr Polozov, Katherine Lee, Zongwei Zhou, Xuezhi Wang, Brennan Saeta, Mark Diaz, Orhan Firat, Michele Catasta, Jason Wei, Kathy Meier{-}Hellstern, Douglas Eck, Jeff Dean, Slav Petrov, and Noah Fiedel.
\newblock Palm: Scaling language modeling with pathways.
\newblock \emph{J. Mach. Learn. Res.}, 24:\penalty0 240:1--240:113, 2023.
\newblock URL \url{http://jmlr.org/papers/v24/22-1144.html}.

\bibitem[Clark et~al.(2020)Clark, Palomaki, Nikolaev, Choi, Garrette, Collins, and Kwiatkowski]{DBLP:journals/tacl/ClarkPNCGCK20}
Jonathan~H. Clark, Jennimaria Palomaki, Vitaly Nikolaev, Eunsol Choi, Dan Garrette, Michael Collins, and Tom Kwiatkowski.
\newblock Tydi {QA:} {A} benchmark for information-seeking question answering in typologically diverse languages.
\newblock \emph{Trans. Assoc. Comput. Linguistics}, 8:\penalty0 454--470, 2020.
\newblock \doi{10.1162/TACL\_A\_00317}.
\newblock URL \url{https://doi.org/10.1162/tacl\_a\_00317}.

\bibitem[Computer(2023)]{together2023redpajama}
Together Computer.
\newblock Redpajama: an open dataset for training large language models, October 2023.
\newblock URL \url{https://github.com/togethercomputer/RedPajama-Data}.

\bibitem[Conneau \& Lample(2019{\natexlab{a}})Conneau and Lample]{DBLP:conf/nips/ConneauL19}
Alexis Conneau and Guillaume Lample.
\newblock Cross-lingual language model pretraining.
\newblock In Hanna~M. Wallach, Hugo Larochelle, Alina Beygelzimer, Florence d'Alch{\'{e}}{-}Buc, Emily~B. Fox, and Roman Garnett (eds.), \emph{Advances in Neural Information Processing Systems 32: Annual Conference on Neural Information Processing Systems 2019, NeurIPS 2019, December 8-14, 2019, Vancouver, BC, Canada}, pp.\  7057--7067, 2019{\natexlab{a}}.
\newblock URL \url{https://proceedings.neurips.cc/paper/2019/hash/c04c19c2c2474dbf5f7ac4372c5b9af1-Abstract.html}.

\bibitem[Conneau \& Lample(2019{\natexlab{b}})Conneau and Lample]{XLM}
Alexis Conneau and Guillaume Lample.
\newblock Cross-lingual language model pretraining.
\newblock \emph{Advances in neural information processing systems}, 32, 2019{\natexlab{b}}.

\bibitem[Conneau et~al.(2018)Conneau, Rinott, Lample, Williams, Bowman, Schwenk, and Stoyanov]{DBLP:conf/emnlp/ConneauRLWBSS18}
Alexis Conneau, Ruty Rinott, Guillaume Lample, Adina Williams, Samuel~R. Bowman, Holger Schwenk, and Veselin Stoyanov.
\newblock {XNLI:} evaluating cross-lingual sentence representations.
\newblock In Ellen Riloff, David Chiang, Julia Hockenmaier, and Jun'ichi Tsujii (eds.), \emph{Proceedings of the 2018 Conference on Empirical Methods in Natural Language Processing, Brussels, Belgium, October 31 - November 4, 2018}, pp.\  2475--2485. Association for Computational Linguistics, 2018.
\newblock \doi{10.18653/V1/D18-1269}.
\newblock URL \url{https://doi.org/10.18653/v1/d18-1269}.

\bibitem[Conneau et~al.(2019)Conneau, Khandelwal, Goyal, Chaudhary, Wenzek, Guzm{\'a}n, Grave, Ott, Zettlemoyer, and Stoyanov]{XLM-R}
Alexis Conneau, Kartikay Khandelwal, Naman Goyal, Vishrav Chaudhary, Guillaume Wenzek, Francisco Guzm{\'a}n, Edouard Grave, Myle Ott, Luke Zettlemoyer, and Veselin Stoyanov.
\newblock Unsupervised cross-lingual representation learning at scale.
\newblock \emph{arXiv preprint arXiv:1911.02116}, 2019.

\bibitem[Conneau et~al.(2020{\natexlab{a}})Conneau, Khandelwal, Goyal, Chaudhary, Wenzek, Guzm{\'{a}}n, Grave, Ott, Zettlemoyer, and Stoyanov]{DBLP:conf/acl/ConneauKGCWGGOZ20}
Alexis Conneau, Kartikay Khandelwal, Naman Goyal, Vishrav Chaudhary, Guillaume Wenzek, Francisco Guzm{\'{a}}n, Edouard Grave, Myle Ott, Luke Zettlemoyer, and Veselin Stoyanov.
\newblock Unsupervised cross-lingual representation learning at scale.
\newblock In Dan Jurafsky, Joyce Chai, Natalie Schluter, and Joel~R. Tetreault (eds.), \emph{Proceedings of the 58th Annual Meeting of the Association for Computational Linguistics, {ACL} 2020, Online, July 5-10, 2020}, pp.\  8440--8451. Association for Computational Linguistics, 2020{\natexlab{a}}.
\newblock \doi{10.18653/V1/2020.ACL-MAIN.747}.
\newblock URL \url{https://doi.org/10.18653/v1/2020.acl-main.747}.

\bibitem[Conneau et~al.(2020{\natexlab{b}})Conneau, Wu, Li, Zettlemoyer, and Stoyanov]{DBLP:conf/acl/ConneauWLZS20}
Alexis Conneau, Shijie Wu, Haoran Li, Luke Zettlemoyer, and Veselin Stoyanov.
\newblock Emerging cross-lingual structure in pretrained language models.
\newblock In Dan Jurafsky, Joyce Chai, Natalie Schluter, and Joel~R. Tetreault (eds.), \emph{Proceedings of the 58th Annual Meeting of the Association for Computational Linguistics, {ACL} 2020, Online, July 5-10, 2020}, pp.\  6022--6034. Association for Computational Linguistics, 2020{\natexlab{b}}.
\newblock \doi{10.18653/V1/2020.ACL-MAIN.536}.
\newblock URL \url{https://doi.org/10.18653/v1/2020.acl-main.536}.

\bibitem[Conover et~al.(2023{\natexlab{a}})Conover, Hayes, Mathur, Meng, Xie, Wan, Shah, Ghodsi, Wendell, Zaharia, and Xin]{dolly}
Mike Conover, Matt Hayes, Ankit Mathur, Xiangrui Meng, Jianwei Xie, Jun Wan, Sam Shah, Ali Ghodsi, Patrick Wendell, Matei Zaharia, and Reynold Xin.
\newblock Free dolly: Introducing the world’s first truly open instruction-tuned llm.
\newblock 2023{\natexlab{a}}.
\newblock URL \url{https://www.databricks.com/}.

\bibitem[Conover et~al.(2023{\natexlab{b}})Conover, Hayes, Mathur, Xie, Wan, Shah, Ghodsi, Wendell, Zaharia, and Xin]{DatabricksBlog2023DollyV2}
Mike Conover, Matt Hayes, Ankit Mathur, Jianwei Xie, Jun Wan, Sam Shah, Ali Ghodsi, Patrick Wendell, Matei Zaharia, and Reynold Xin.
\newblock Free dolly: Introducing the world's first truly open instruction-tuned llm, 2023{\natexlab{b}}.
\newblock URL \url{https://www.databricks.com/blog/2023/04/12/dolly-first-open-commercially-viable-instruction-tuned-llm}.

\bibitem[Costa-juss{\`a} et~al.(2022)Costa-juss{\`a}, Cross, {\c{C}}elebi, Elbayad, Heafield, Heffernan, Kalbassi, Lam, Licht, Maillard, et~al.]{NLLB}
Marta~R Costa-juss{\`a}, James Cross, Onur {\c{C}}elebi, Maha Elbayad, Kenneth Heafield, Kevin Heffernan, Elahe Kalbassi, Janice Lam, Daniel Licht, Jean Maillard, et~al.
\newblock No language left behind: Scaling human-centered machine translation.
\newblock \emph{arXiv preprint arXiv:2207.04672}, 2022.

\bibitem[Cui et~al.(2023{\natexlab{a}})Cui, Yuan, Ding, Yao, Zhu, Ni, Xie, Liu, and Sun]{DBLP:journals/corr/abs-2310-01377}
Ganqu Cui, Lifan Yuan, Ning Ding, Guanming Yao, Wei Zhu, Yuan Ni, Guotong Xie, Zhiyuan Liu, and Maosong Sun.
\newblock Ultrafeedback: Boosting language models with high-quality feedback.
\newblock \emph{CoRR}, abs/2310.01377, 2023{\natexlab{a}}.
\newblock \doi{10.48550/ARXIV.2310.01377}.
\newblock URL \url{https://doi.org/10.48550/arXiv.2310.01377}.

\bibitem[Cui et~al.(2023{\natexlab{b}})Cui, Li, Yan, Chen, and Yuan]{DBLP:journals/corr/abs-2306-16092}
Jiaxi Cui, Zongjian Li, Yang Yan, Bohua Chen, and Li~Yuan.
\newblock Chatlaw: Open-source legal large language model with integrated external knowledge bases.
\newblock \emph{CoRR}, abs/2306.16092, 2023{\natexlab{b}}.
\newblock \doi{10.48550/ARXIV.2306.16092}.
\newblock URL \url{https://doi.org/10.48550/arXiv.2306.16092}.

\bibitem[Cui et~al.(2024)Cui, Du, Zhu, and Xiong]{cui2024efficiently}
Menglong Cui, Jiangcun Du, Shaolin Zhu, and Deyi Xiong.
\newblock Efficiently exploring large language models for document-level machine translation with in-context learning.
\newblock \emph{arXiv preprint arXiv:2406.07081}, 2024.

\bibitem[Cui \& Yao(2024)Cui and Yao]{adapt-10}
Yiming Cui and Xin Yao.
\newblock Rethinking {LLM} language adaptation: {A} case study on chinese mixtral.
\newblock \emph{CoRR}, abs/2403.01851, 2024.
\newblock \doi{10.48550/ARXIV.2403.01851}.
\newblock URL \url{https://doi.org/10.48550/arXiv.2403.01851}.

\bibitem[Cui et~al.(2023{\natexlab{c}})Cui, Yang, and Yao]{adapt-1}
Yiming Cui, Ziqing Yang, and Xin Yao.
\newblock Efficient and effective text encoding for chinese llama and alpaca.
\newblock \emph{CoRR}, abs/2304.08177, 2023{\natexlab{c}}.
\newblock \doi{10.48550/ARXIV.2304.08177}.
\newblock URL \url{https://doi.org/10.48550/arXiv.2304.08177}.

\bibitem[Dai et~al.(2024)Dai, Deng, Zhao, Xu, Gao, Chen, Li, Zeng, Yu, Wu, et~al.]{DeepSeekMoE}
Damai Dai, Chengqi Deng, Chenggang Zhao, RX~Xu, Huazuo Gao, Deli Chen, Jiashi Li, Wangding Zeng, Xingkai Yu, Y~Wu, et~al.
\newblock Deepseekmoe: Towards ultimate expert specialization in mixture-of-experts language models.
\newblock \emph{arXiv preprint arXiv:2401.06066}, 2024.

\bibitem[de~Varda \& Marelli(2024)de~Varda and Marelli]{DBLP:conf/coling/VardaM24}
Andrea~Gregor de~Varda and Marco Marelli.
\newblock The emergence of semantic units in massively multilingual models.
\newblock In Nicoletta Calzolari, Min{-}Yen Kan, V{\'{e}}ronique Hoste, Alessandro Lenci, Sakriani Sakti, and Nianwen Xue (eds.), \emph{Proceedings of the 2024 Joint International Conference on Computational Linguistics, Language Resources and Evaluation, {LREC/COLING} 2024, 20-25 May, 2024, Torino, Italy}, pp.\  15910--15921. {ELRA} and {ICCL}, 2024.
\newblock URL \url{https://aclanthology.org/2024.lrec-main.1382}.

\bibitem[de~Wynter et~al.(2024)de~Wynter, Watts, Altintoprak, Wongsangaroonsri, Zhang, Farra, Baur, Claudet, Gajdusek, G{\"{o}}ren, Gu, Kaminska, Kaminski, Kuo, Kyuba, Lee, Mathur, Merok, Milovanovic, Paananen, Paananen, Pavlenko, Vidal, Strika, Tsao, Turcato, Vakhno, Velcsov, Vickers, Visser, Widarmanto, Zaikin, and Chen]{DBLP:journals/corr/abs-2404-14397}
Adrian de~Wynter, Ishaan Watts, Nektar~Ege Altintoprak, Tua Wongsangaroonsri, Minghui Zhang, Noura Farra, Lena Baur, Samantha Claudet, Pavel Gajdusek, Can G{\"{o}}ren, Qilong Gu, Anna Kaminska, Tomasz Kaminski, Ruby Kuo, Akiko Kyuba, Jongho Lee, Kartik Mathur, Petter Merok, Ivana Milovanovic, Nani Paananen, Vesa{-}Matti Paananen, Anna Pavlenko, Bruno~Pereira Vidal, Luciano Strika, Yueh Tsao, Davide Turcato, Oleksandr Vakhno, Judit Velcsov, Anna Vickers, St{\'{e}}phanie Visser, Herdyan Widarmanto, Andrey Zaikin, and Si{-}Qing Chen.
\newblock {RTP-LX:} can llms evaluate toxicity in multilingual scenarios?
\newblock \emph{CoRR}, abs/2404.14397, 2024.
\newblock \doi{10.48550/ARXIV.2404.14397}.
\newblock URL \url{https://doi.org/10.48550/arXiv.2404.14397}.

\bibitem[DeepSeek{-}AI et~al.(2024{\natexlab{a}})DeepSeek{-}AI, Liu, Feng, Wang, Wang, Liu, Zhao, Deng, Ruan, Dai, Guo, Yang, Chen, Ji, Li, Lin, Luo, Hao, Chen, Li, Zhang, Xu, Yang, Zhang, Ding, Xin, Gao, Li, Qu, Cai, Liang, Guo, Ni, Li, Chen, Yuan, Qiu, Song, Dong, Gao, Guan, Wang, Zhang, Xu, Xia, Zhao, Zhang, Li, Wang, Zhang, Zhang, Tang, Li, Tian, Huang, Wang, Zhang, Zhu, Chen, Du, Chen, Jin, Ge, Pan, Xu, Chen, Li, Lu, Zhou, Chen, Wu, Ye, Ma, Wang, Zhou, Yu, Zhou, Zheng, Wang, Pei, Yuan, Sun, Xiao, Zeng, An, Liu, Liang, Gao, Zhang, Li, Jin, Wang, Bi, Liu, Wang, Shen, Chen, Chen, Nie, and Sun]{DeepseekV2}
DeepSeek{-}AI, Aixin Liu, Bei Feng, Bin Wang, Bingxuan Wang, Bo~Liu, Chenggang Zhao, Chengqi Deng, Chong Ruan, Damai Dai, Daya Guo, Dejian Yang, Deli Chen, Dongjie Ji, Erhang Li, Fangyun Lin, Fuli Luo, Guangbo Hao, Guanting Chen, Guowei Li, Hao Zhang, Hanwei Xu, Hao Yang, Haowei Zhang, Honghui Ding, Huajian Xin, Huazuo Gao, Hui Li, Hui Qu, J.~L. Cai, Jian Liang, Jianzhong Guo, Jiaqi Ni, Jiashi Li, Jin Chen, Jingyang Yuan, Junjie Qiu, Junxiao Song, Kai Dong, Kaige Gao, Kang Guan, Lean Wang, Lecong Zhang, Lei Xu, Leyi Xia, Liang Zhao, Liyue Zhang, Meng Li, Miaojun Wang, Mingchuan Zhang, Minghua Zhang, Minghui Tang, Mingming Li, Ning Tian, Panpan Huang, Peiyi Wang, Peng Zhang, Qihao Zhu, Qinyu Chen, Qiushi Du, R.~J. Chen, R.~L. Jin, Ruiqi Ge, Ruizhe Pan, Runxin Xu, Ruyi Chen, S.~S. Li, Shanghao Lu, Shangyan Zhou, Shanhuang Chen, Shaoqing Wu, Shengfeng Ye, Shirong Ma, Shiyu Wang, Shuang Zhou, Shuiping Yu, Shunfeng Zhou, Size Zheng, Tao Wang, Tian Pei, Tian Yuan, Tianyu Sun, W.~L. Xiao, Wangding Zeng, Wei An, Wen
  Liu, Wenfeng Liang, Wenjun Gao, Wentao Zhang, X.~Q. Li, Xiangyue Jin, Xianzu Wang, Xiao Bi, Xiaodong Liu, Xiaohan Wang, Xiaojin Shen, Xiaokang Chen, Xiaosha Chen, Xiaotao Nie, and Xiaowen Sun.
\newblock Deepseek-v2: {A} strong, economical, and efficient mixture-of-experts language model.
\newblock \emph{CoRR}, abs/2405.04434, 2024{\natexlab{a}}.
\newblock \doi{10.48550/ARXIV.2405.04434}.
\newblock URL \url{https://doi.org/10.48550/arXiv.2405.04434}.

\bibitem[DeepSeek{-}AI et~al.(2024{\natexlab{b}})DeepSeek{-}AI, Liu, Feng, Wang, Wang, Liu, Zhao, Deng, Ruan, Dai, Guo, Yang, Chen, Ji, Li, Lin, Luo, Hao, Chen, Li, Zhang, Xu, Yang, Zhang, Ding, Xin, Gao, Li, Qu, Cai, Liang, Guo, Ni, Li, Chen, Yuan, Qiu, Song, Dong, Gao, Guan, Wang, Zhang, Xu, Xia, Zhao, Zhang, Li, Wang, Zhang, Zhang, Tang, Li, Tian, Huang, Wang, Zhang, Zhu, Chen, Du, Chen, Jin, Ge, Pan, Xu, Chen, Li, Lu, Zhou, Chen, Wu, Ye, Ma, Wang, Zhou, Yu, Zhou, Zheng, Wang, Pei, Yuan, Sun, Xiao, Zeng, An, Liu, Liang, Gao, Zhang, Li, Jin, Wang, Bi, Liu, Wang, Shen, Chen, Chen, Nie, and Sun]{DBLP:journals/corr/abs-2405-04434}
DeepSeek{-}AI, Aixin Liu, Bei Feng, Bin Wang, Bingxuan Wang, Bo~Liu, Chenggang Zhao, Chengqi Deng, Chong Ruan, Damai Dai, Daya Guo, Dejian Yang, Deli Chen, Dongjie Ji, Erhang Li, Fangyun Lin, Fuli Luo, Guangbo Hao, Guanting Chen, Guowei Li, Hao Zhang, Hanwei Xu, Hao Yang, Haowei Zhang, Honghui Ding, Huajian Xin, Huazuo Gao, Hui Li, Hui Qu, J.~L. Cai, Jian Liang, Jianzhong Guo, Jiaqi Ni, Jiashi Li, Jin Chen, Jingyang Yuan, Junjie Qiu, Junxiao Song, Kai Dong, Kaige Gao, Kang Guan, Lean Wang, Lecong Zhang, Lei Xu, Leyi Xia, Liang Zhao, Liyue Zhang, Meng Li, Miaojun Wang, Mingchuan Zhang, Minghua Zhang, Minghui Tang, Mingming Li, Ning Tian, Panpan Huang, Peiyi Wang, Peng Zhang, Qihao Zhu, Qinyu Chen, Qiushi Du, R.~J. Chen, R.~L. Jin, Ruiqi Ge, Ruizhe Pan, Runxin Xu, Ruyi Chen, S.~S. Li, Shanghao Lu, Shangyan Zhou, Shanhuang Chen, Shaoqing Wu, Shengfeng Ye, Shirong Ma, Shiyu Wang, Shuang Zhou, Shuiping Yu, Shunfeng Zhou, Size Zheng, Tao Wang, Tian Pei, Tian Yuan, Tianyu Sun, W.~L. Xiao, Wangding Zeng, Wei An, Wen
  Liu, Wenfeng Liang, Wenjun Gao, Wentao Zhang, X.~Q. Li, Xiangyue Jin, Xianzu Wang, Xiao Bi, Xiaodong Liu, Xiaohan Wang, Xiaojin Shen, Xiaokang Chen, Xiaosha Chen, Xiaotao Nie, and Xiaowen Sun.
\newblock Deepseek-v2: {A} strong, economical, and efficient mixture-of-experts language model.
\newblock \emph{CoRR}, abs/2405.04434, 2024{\natexlab{b}}.
\newblock \doi{10.48550/ARXIV.2405.04434}.
\newblock URL \url{https://doi.org/10.48550/arXiv.2405.04434}.

\bibitem[Deng et~al.(2023)Deng, Zhang, Pan, and Bing]{DBLP:journals/corr/abs-2310-06474}
Yue Deng, Wenxuan Zhang, Sinno~Jialin Pan, and Lidong Bing.
\newblock Multilingual jailbreak challenges in large language models.
\newblock \emph{CoRR}, abs/2310.06474, 2023.
\newblock \doi{10.48550/ARXIV.2310.06474}.
\newblock URL \url{https://doi.org/10.48550/arXiv.2310.06474}.

\bibitem[Deshpande et~al.(2022)Deshpande, Talukdar, and Narasimhan]{DBLP:conf/naacl/DeshpandeTN22}
Ameet Deshpande, Partha Talukdar, and Karthik Narasimhan.
\newblock When is {BERT} multilingual? isolating crucial ingredients for cross-lingual transfer.
\newblock In Marine Carpuat, Marie{-}Catherine de~Marneffe, and Iv{\'{a}}n Vladimir~Meza Ru{\'{\i}}z (eds.), \emph{Proceedings of the 2022 Conference of the North American Chapter of the Association for Computational Linguistics: Human Language Technologies, {NAACL} 2022, Seattle, WA, United States, July 10-15, 2022}, pp.\  3610--3623. Association for Computational Linguistics, 2022.
\newblock \doi{10.18653/V1/2022.NAACL-MAIN.264}.
\newblock URL \url{https://doi.org/10.18653/v1/2022.naacl-main.264}.

\bibitem[Dettmers et~al.(2023)Dettmers, Pagnoni, Holtzman, and Zettlemoyer]{cpt_10}
Tim Dettmers, Artidoro Pagnoni, Ari Holtzman, and Luke Zettlemoyer.
\newblock Qlora: Efficient finetuning of quantized llms.
\newblock In Alice Oh, Tristan Naumann, Amir Globerson, Kate Saenko, Moritz Hardt, and Sergey Levine (eds.), \emph{Advances in Neural Information Processing Systems 36: Annual Conference on Neural Information Processing Systems 2023, NeurIPS 2023, New Orleans, LA, USA, December 10 - 16, 2023}, 2023.
\newblock URL \url{http://papers.nips.cc/paper\_files/paper/2023/hash/1feb87871436031bdc0f2beaa62a049b-Abstract-Conference.html}.

\bibitem[Devlin et~al.(2018)Devlin, Chang, Lee, and Toutanova]{bert}
Jacob Devlin, Ming-Wei Chang, Kenton Lee, and Kristina Toutanova.
\newblock Bert: Pre-training of deep bidirectional transformers for language understanding.
\newblock \emph{arXiv preprint arXiv:1810.04805}, 2018.

\bibitem[Devlin et~al.(2019)Devlin, Chang, Lee, and Toutanova]{DBLP:conf/naacl/DevlinCLT19}
Jacob Devlin, Ming{-}Wei Chang, Kenton Lee, and Kristina Toutanova.
\newblock {BERT:} pre-training of deep bidirectional transformers for language understanding.
\newblock In Jill Burstein, Christy Doran, and Thamar Solorio (eds.), \emph{Proceedings of the 2019 Conference of the North American Chapter of the Association for Computational Linguistics: Human Language Technologies, {NAACL-HLT} 2019, Minneapolis, MN, USA, June 2-7, 2019, Volume 1 (Long and Short Papers)}, pp.\  4171--4186. Association for Computational Linguistics, 2019.
\newblock \doi{10.18653/V1/N19-1423}.
\newblock URL \url{https://doi.org/10.18653/v1/n19-1423}.

\bibitem[Doddapaneni et~al.(2022)Doddapaneni, Aralikatte, Ramesh, Goyal, Khapra, Kunchukuttan, and Kumar]{DBLP:journals/corr/abs-2212-05409}
Sumanth Doddapaneni, Rahul Aralikatte, Gowtham Ramesh, Shreya Goyal, Mitesh~M. Khapra, Anoop Kunchukuttan, and Pratyush Kumar.
\newblock Indicxtreme: {A} multi-task benchmark for evaluating indic languages.
\newblock \emph{CoRR}, abs/2212.05409, 2022.
\newblock \doi{10.48550/ARXIV.2212.05409}.
\newblock URL \url{https://doi.org/10.48550/arXiv.2212.05409}.

\bibitem[Doddapaneni et~al.(2024)Doddapaneni, Khan, Venkatesh, Dabre, Kunchukuttan, and Khapra]{doddapaneni2024cross}
Sumanth Doddapaneni, Mohammed Safi Ur~Rahman Khan, Dilip Venkatesh, Raj Dabre, Anoop Kunchukuttan, and Mitesh~M Khapra.
\newblock Cross-lingual auto evaluation for assessing multilingual llms.
\newblock \emph{arXiv preprint arXiv:2410.13394}, 2024.

\bibitem[Eisele \& Chen(2010)Eisele and Chen]{MultiUN}
Andreas Eisele and Yu~Chen.
\newblock Multiun: {A} multilingual corpus from united nation documents.
\newblock In Nicoletta Calzolari, Khalid Choukri, Bente Maegaard, Joseph Mariani, Jan Odijk, Stelios Piperidis, Mike Rosner, and Daniel Tapias (eds.), \emph{Proceedings of the International Conference on Language Resources and Evaluation, {LREC} 2010, 17-23 May 2010, Valletta, Malta}. European Language Resources Association, 2010.
\newblock URL \url{http://www.lrec-conf.org/proceedings/lrec2010/summaries/686.html}.

\bibitem[Ethayarajh et~al.(2024)Ethayarajh, Xu, Muennighoff, Jurafsky, and Kiela]{DBLP:journals/corr/abs-2402-01306}
Kawin Ethayarajh, Winnie Xu, Niklas Muennighoff, Dan Jurafsky, and Douwe Kiela.
\newblock {KTO:} model alignment as prospect theoretic optimization.
\newblock \emph{CoRR}, abs/2402.01306, 2024.
\newblock \doi{10.48550/ARXIV.2402.01306}.
\newblock URL \url{https://doi.org/10.48550/arXiv.2402.01306}.

\bibitem[Faisal \& Anastasopoulos(2024)Faisal and Anastasopoulos]{faisal2024efficient}
Fahim Faisal and Antonios Anastasopoulos.
\newblock An efficient approach for studying cross-lingual transfer in multilingual language models.
\newblock \emph{arXiv preprint arXiv:2403.20088}, 2024.

\bibitem[Fedus et~al.(2022)Fedus, Zoph, and Shazeer]{switchtransformer}
William Fedus, Barret Zoph, and Noam Shazeer.
\newblock Switch transformers: Scaling to trillion parameter models with simple and efficient sparsity.
\newblock \emph{Journal of Machine Learning Research}, 23\penalty0 (120):\penalty0 1--39, 2022.

\bibitem[Feng et~al.(2020)Feng, Guo, Tang, Duan, Feng, Gong, Shou, Qin, Liu, Jiang, and Zhou]{DBLP:conf/emnlp/FengGTDFGS0LJZ20}
Zhangyin Feng, Daya Guo, Duyu Tang, Nan Duan, Xiaocheng Feng, Ming Gong, Linjun Shou, Bing Qin, Ting Liu, Daxin Jiang, and Ming Zhou.
\newblock Codebert: {A} pre-trained model for programming and natural languages.
\newblock In Trevor Cohn, Yulan He, and Yang Liu (eds.), \emph{Findings of the Association for Computational Linguistics: {EMNLP} 2020, Online Event, 16-20 November 2020}, volume {EMNLP} 2020 of \emph{Findings of {ACL}}, pp.\  1536--1547. Association for Computational Linguistics, 2020.
\newblock \doi{10.18653/V1/2020.FINDINGS-EMNLP.139}.
\newblock URL \url{https://doi.org/10.18653/v1/2020.findings-emnlp.139}.

\bibitem[Ferron et~al.(2023)Ferron, Shore, Mitra, and Agrawal]{DBLP:conf/emnlp/FerronSMA23}
Amila Ferron, Amber Shore, Ekata Mitra, and Ameeta Agrawal.
\newblock {MEEP:} is this engaging? prompting large language models for dialogue evaluation in multilingual settings.
\newblock In Houda Bouamor, Juan Pino, and Kalika Bali (eds.), \emph{Findings of the Association for Computational Linguistics: {EMNLP} 2023, Singapore, December 6-10, 2023}, pp.\  2078--2100. Association for Computational Linguistics, 2023.
\newblock \doi{10.18653/V1/2023.FINDINGS-EMNLP.137}.
\newblock URL \url{https://doi.org/10.18653/v1/2023.findings-emnlp.137}.

\bibitem[Fried et~al.(2023)Fried, Aghajanyan, Lin, Wang, Wallace, Shi, Zhong, Yih, Zettlemoyer, and Lewis]{DBLP:conf/iclr/FriedAL0WSZYZL23}
Daniel Fried, Armen Aghajanyan, Jessy Lin, Sida Wang, Eric Wallace, Freda Shi, Ruiqi Zhong, Scott Yih, Luke Zettlemoyer, and Mike Lewis.
\newblock Incoder: {A} generative model for code infilling and synthesis.
\newblock In \emph{The Eleventh International Conference on Learning Representations, {ICLR} 2023, Kigali, Rwanda, May 1-5, 2023}. OpenReview.net, 2023.
\newblock URL \url{https://openreview.net/pdf?id=hQwb-lbM6EL}.

\bibitem[Fu et~al.(2024)Fu, Feng, Huang, Huo, Li, Wang, Qin, and Liu]{mt-12}
Chengpeng Fu, Xiaocheng Feng, Yichong Huang, Wenshuai Huo, Baohang Li, Hui Wang, Bin Qin, and Ting Liu.
\newblock Relay decoding: Concatenating large language models for machine translation.
\newblock \emph{CoRR}, abs/2405.02933, 2024.
\newblock \doi{10.48550/ARXIV.2405.02933}.
\newblock URL \url{https://doi.org/10.48550/arXiv.2405.02933}.

\bibitem[Fu et~al.(2023)Fu, Dao, Saab, Thomas, Rudra, and Re]{fu2023hungryhungryhipposlanguage}
Daniel~Y. Fu, Tri Dao, Khaled~K. Saab, Armin~W. Thomas, Atri Rudra, and Christopher Re.
\newblock Hungry hungry hippos: Towards language modeling with state space models, 2023.
\newblock URL \url{https://arxiv.org/abs/2212.14052}.

\bibitem[Fujii et~al.(2024)Fujii, Nakamura, Loem, Iida, Ohi, Hattori, Shota, Mizuki, Yokota, and Okazaki]{adapt-3}
Kazuki Fujii, Taishi Nakamura, Mengsay Loem, Hiroki Iida, Masanari Ohi, Kakeru Hattori, Hirai Shota, Sakae Mizuki, Rio Yokota, and Naoaki Okazaki.
\newblock Continual pre-training for cross-lingual {LLM} adaptation: Enhancing japanese language capabilities.
\newblock \emph{CoRR}, abs/2404.17790, 2024.
\newblock \doi{10.48550/ARXIV.2404.17790}.
\newblock URL \url{https://doi.org/10.48550/arXiv.2404.17790}.

\bibitem[Fung et~al.(2023)Fung, Chakrabarty, Guo, Rambow, Muresan, and Ji]{c-17}
Yi~Fung, Tuhin Chakrabarty, Hao Guo, Owen Rambow, Smaranda Muresan, and Heng Ji.
\newblock {NORMSAGE:} multi-lingual multi-cultural norm discovery from conversations on-the-fly.
\newblock In Houda Bouamor, Juan Pino, and Kalika Bali (eds.), \emph{Proceedings of the 2023 Conference on Empirical Methods in Natural Language Processing, {EMNLP} 2023, Singapore, December 6-10, 2023}, pp.\  15217--15230. Association for Computational Linguistics, 2023.
\newblock \doi{10.18653/V1/2023.EMNLP-MAIN.941}.
\newblock URL \url{https://doi.org/10.18653/v1/2023.emnlp-main.941}.

\bibitem[Fung et~al.(2024)Fung, Zhao, Doo, Sun, and Ji]{c-19}
Yi~Fung, Ruining Zhao, Jae Doo, Chenkai Sun, and Heng Ji.
\newblock Massively multi-cultural knowledge acquisition {\&} {LM} benchmarking.
\newblock \emph{CoRR}, abs/2402.09369, 2024.
\newblock \doi{10.48550/ARXIV.2402.09369}.
\newblock URL \url{https://doi.org/10.48550/arXiv.2402.09369}.

\bibitem[Gala et~al.(2023)Gala, Chitale, AK, Doddapaneni, Gumma, Kumar, Nawale, Sujatha, Puduppully, Raghavan, Kumar, Khapra, Dabre, and Kunchukuttan]{DBLP:journals/corr/abs-2305-16307}
Jay~P. Gala, Pranjal~A. Chitale, Raghavan AK, Sumanth Doddapaneni, Varun Gumma, Aswanth Kumar, Janki Nawale, Anupama Sujatha, Ratish Puduppully, Vivek Raghavan, Pratyush Kumar, Mitesh~M. Khapra, Raj Dabre, and Anoop Kunchukuttan.
\newblock Indictrans2: Towards high-quality and accessible machine translation models for all 22 scheduled indian languages.
\newblock \emph{CoRR}, abs/2305.16307, 2023.
\newblock \doi{10.48550/ARXIV.2305.16307}.
\newblock URL \url{https://doi.org/10.48550/arXiv.2305.16307}.

\bibitem[Ganguli et~al.(2022)Ganguli, Lovitt, Kernion, Askell, Bai, Kadavath, Mann, Perez, Schiefer, Ndousse, Jones, Bowman, Chen, Conerly, DasSarma, Drain, Elhage, Showk, Fort, Hatfield{-}Dodds, Henighan, Hernandez, Hume, Jacobson, Johnston, Kravec, Olsson, Ringer, Tran{-}Johnson, Amodei, Brown, Joseph, McCandlish, Olah, Kaplan, and Clark]{DBLP:journals/corr/abs-2209-07858}
Deep Ganguli, Liane Lovitt, Jackson Kernion, Amanda Askell, Yuntao Bai, Saurav Kadavath, Ben Mann, Ethan Perez, Nicholas Schiefer, Kamal Ndousse, Andy Jones, Sam Bowman, Anna Chen, Tom Conerly, Nova DasSarma, Dawn Drain, Nelson Elhage, Sheer~El Showk, Stanislav Fort, Zac Hatfield{-}Dodds, Tom Henighan, Danny Hernandez, Tristan Hume, Josh Jacobson, Scott Johnston, Shauna Kravec, Catherine Olsson, Sam Ringer, Eli Tran{-}Johnson, Dario Amodei, Tom Brown, Nicholas Joseph, Sam McCandlish, Chris Olah, Jared Kaplan, and Jack Clark.
\newblock Red teaming language models to reduce harms: Methods, scaling behaviors, and lessons learned.
\newblock \emph{CoRR}, abs/2209.07858, 2022.
\newblock \doi{10.48550/ARXIV.2209.07858}.
\newblock URL \url{https://doi.org/10.48550/arXiv.2209.07858}.

\bibitem[Gao et~al.(2024)Gao, He, Wu, and Wang]{mt-10}
Pengzhi Gao, Zhongjun He, Hua Wu, and Haifeng Wang.
\newblock Towards boosting many-to-many multilingual machine translation with large language models.
\newblock \emph{CoRR}, abs/2401.05861, 2024.
\newblock \doi{10.48550/ARXIV.2401.05861}.
\newblock URL \url{https://doi.org/10.48550/arXiv.2401.05861}.

\bibitem[Garc{\'\i}a-Ferrero et~al.(2024)Garc{\'\i}a-Ferrero, Agerri, Salazar, Cabrio, de~la Iglesia, Lavelli, Magnini, Molinet, Ramirez-Romero, Rigau, et~al.]{garcia2024medical}
Iker Garc{\'\i}a-Ferrero, Rodrigo Agerri, Aitziber~Atutxa Salazar, Elena Cabrio, Iker de~la Iglesia, Alberto Lavelli, Bernardo Magnini, Benjamin Molinet, Johana Ramirez-Romero, German Rigau, et~al.
\newblock Medical mt5: an open-source multilingual text-to-text llm for the medical domain.
\newblock \emph{arXiv preprint arXiv:2404.07613}, 2024.

\bibitem[Gehman et~al.(2020)Gehman, Gururangan, Sap, Choi, and Smith]{DBLP:conf/emnlp/GehmanGSCS20}
Samuel Gehman, Suchin Gururangan, Maarten Sap, Yejin Choi, and Noah~A. Smith.
\newblock Realtoxicityprompts: Evaluating neural toxic degeneration in language models.
\newblock In Trevor Cohn, Yulan He, and Yang Liu (eds.), \emph{Findings of the Association for Computational Linguistics: {EMNLP} 2020, Online Event, 16-20 November 2020}, volume {EMNLP} 2020 of \emph{Findings of {ACL}}, pp.\  3356--3369. Association for Computational Linguistics, 2020.
\newblock \doi{10.18653/V1/2020.FINDINGS-EMNLP.301}.
\newblock URL \url{https://doi.org/10.18653/v1/2020.findings-emnlp.301}.

\bibitem[Gooding \& Mansoor(2023)Gooding and Mansoor]{DBLP:journals/corr/abs-2311-04919}
Sian Gooding and Hassan Mansoor.
\newblock The impact of preference agreement in reinforcement learning from human feedback: {A} case study in summarization.
\newblock \emph{CoRR}, abs/2311.04919, 2023.
\newblock \doi{10.48550/ARXIV.2311.04919}.
\newblock URL \url{https://doi.org/10.48550/arXiv.2311.04919}.

\bibitem[Goyal et~al.(2022)Goyal, Gao, Chaudhary, Chen, Wenzek, Ju, Krishnan, Ranzato, Guzm{\'{a}}n, and Fan]{DBLP:journals/tacl/GoyalGCCWJKRGF22}
Naman Goyal, Cynthia Gao, Vishrav Chaudhary, Peng{-}Jen Chen, Guillaume Wenzek, Da~Ju, Sanjana Krishnan, Marc'Aurelio Ranzato, Francisco Guzm{\'{a}}n, and Angela Fan.
\newblock The flores-101 evaluation benchmark for low-resource and multilingual machine translation.
\newblock \emph{Trans. Assoc. Comput. Linguistics}, 10:\penalty0 522--538, 2022.
\newblock \doi{10.1162/TACL\_A\_00474}.
\newblock URL \url{https://doi.org/10.1162/tacl\_a\_00474}.

\bibitem[Gu \& Dao(2023)Gu and Dao]{mamba}
Albert Gu and Tri Dao.
\newblock Mamba: Linear-time sequence modeling with selective state spaces.
\newblock \emph{arXiv preprint arXiv:2312.00752}, 2023.

\bibitem[Gu et~al.(2022)Gu, Goel, and Re]{gu2022efficientlymodelinglongsequences}
Albert Gu, Karan Goel, and Christopher Re.
\newblock Efficiently modeling long sequences with structured state spaces, 2022.
\newblock URL \url{https://arxiv.org/abs/2111.00396}.

\bibitem[Gurgurov et~al.(2024)Gurgurov, B{\"a}umel, and Anikina]{gurgurov2024multilingual}
Daniil Gurgurov, Tanja B{\"a}umel, and Tatiana Anikina.
\newblock Multilingual large language models and curse of multilinguality.
\newblock \emph{arXiv preprint arXiv:2406.10602}, 2024.

\bibitem[Hada et~al.(2024{\natexlab{a}})Hada, Gumma, Ahmed, Bali, and Sitaram]{DBLP:journals/corr/abs-2404-01667}
Rishav Hada, Varun Gumma, Mohamed Ahmed, Kalika Bali, and Sunayana Sitaram.
\newblock {METAL:} towards multilingual meta-evaluation.
\newblock \emph{CoRR}, abs/2404.01667, 2024{\natexlab{a}}.
\newblock \doi{10.48550/ARXIV.2404.01667}.
\newblock URL \url{https://doi.org/10.48550/arXiv.2404.01667}.

\bibitem[Hada et~al.(2024{\natexlab{b}})Hada, Gumma, de~Wynter, Diddee, Ahmed, Choudhury, Bali, and Sitaram]{DBLP:conf/eacl/HadaGWDACBS24}
Rishav Hada, Varun Gumma, Adrian de~Wynter, Harshita Diddee, Mohamed Ahmed, Monojit Choudhury, Kalika Bali, and Sunayana Sitaram.
\newblock Are large language model-based evaluators the solution to scaling up multilingual evaluation?
\newblock In Yvette Graham and Matthew Purver (eds.), \emph{Findings of the Association for Computational Linguistics: {EACL} 2024, St. Julian's, Malta, March 17-22, 2024}, pp.\  1051--1070. Association for Computational Linguistics, 2024{\natexlab{b}}.
\newblock URL \url{https://aclanthology.org/2024.findings-eacl.71}.

\bibitem[Hasan et~al.(2021)Hasan, Bhattacharjee, Islam, Mubasshir, Li, Kang, Rahman, and Shahriyar]{DBLP:conf/acl/HasanBIMLKRS21}
Tahmid Hasan, Abhik Bhattacharjee, Md.~Saiful Islam, Kazi~Samin Mubasshir, Yuan{-}Fang Li, Yong{-}Bin Kang, M.~Sohel Rahman, and Rifat Shahriyar.
\newblock Xl-sum: Large-scale multilingual abstractive summarization for 44 languages.
\newblock In Chengqing Zong, Fei Xia, Wenjie Li, and Roberto Navigli (eds.), \emph{Findings of the Association for Computational Linguistics: {ACL/IJCNLP} 2021, Online Event, August 1-6, 2021}, volume {ACL/IJCNLP} 2021 of \emph{Findings of {ACL}}, pp.\  4693--4703. Association for Computational Linguistics, 2021.
\newblock \doi{10.18653/V1/2021.FINDINGS-ACL.413}.
\newblock URL \url{https://doi.org/10.18653/v1/2021.findings-acl.413}.

\bibitem[Hershcovich et~al.(2022)Hershcovich, Frank, Lent, de~Lhoneux, Abdou, Brandl, Bugliarello, Piqueras, Chalkidis, Cui, Fierro, Margatina, Rust, and S{\o}gaard]{c-2}
Daniel Hershcovich, Stella Frank, Heather~C. Lent, Miryam de~Lhoneux, Mostafa Abdou, Stephanie Brandl, Emanuele Bugliarello, Laura~Cabello Piqueras, Ilias Chalkidis, Ruixiang Cui, Constanza Fierro, Katerina Margatina, Phillip Rust, and Anders S{\o}gaard.
\newblock Challenges and strategies in cross-cultural {NLP}.
\newblock In Smaranda Muresan, Preslav Nakov, and Aline Villavicencio (eds.), \emph{Proceedings of the 60th Annual Meeting of the Association for Computational Linguistics (Volume 1: Long Papers), {ACL} 2022, Dublin, Ireland, May 22-27, 2022}, pp.\  6997--7013. Association for Computational Linguistics, 2022.
\newblock \doi{10.18653/V1/2022.ACL-LONG.482}.
\newblock URL \url{https://doi.org/10.18653/v1/2022.acl-long.482}.

\bibitem[Holtermann et~al.(2024)Holtermann, R{\"{o}}ttger, Dill, and Lauscher]{DBLP:journals/corr/abs-2403-03814}
Carolin Holtermann, Paul R{\"{o}}ttger, Timm Dill, and Anne Lauscher.
\newblock Evaluating the elementary multilingual capabilities of large language models with multiq.
\newblock \emph{CoRR}, abs/2403.03814, 2024.
\newblock \doi{10.48550/ARXIV.2403.03814}.
\newblock URL \url{https://doi.org/10.48550/arXiv.2403.03814}.

\bibitem[Hu et~al.(2024)Hu, Tu, Han, He, Cui, Long, Zheng, Fang, Huang, Zhao, et~al.]{MiniCPM}
Shengding Hu, Yuge Tu, Xu~Han, Chaoqun He, Ganqu Cui, Xiang Long, Zhi Zheng, Yewei Fang, Yuxiang Huang, Weilin Zhao, et~al.
\newblock Minicpm: Unveiling the potential of small language models with scalable training strategies.
\newblock \emph{arXiv preprint arXiv:2404.06395}, 2024.

\bibitem[Hua et~al.(2022)Hua, Zhang, Chen, Li, and Weber]{DBLP:journals/corr/abs-2212-08204}
Wenyue Hua, Yuchen Zhang, Zhe Chen, Josie Li, and Melanie Weber.
\newblock Legalrelectra: Mixed-domain language modeling for long-range legal text comprehension.
\newblock \emph{CoRR}, abs/2212.08204, 2022.
\newblock \doi{10.48550/ARXIV.2212.08204}.
\newblock URL \url{https://doi.org/10.48550/arXiv.2212.08204}.

\bibitem[Huang \& Yang(2023)Huang and Yang]{c-30}
Jing Huang and Diyi Yang.
\newblock Culturally aware natural language inference.
\newblock In Houda Bouamor, Juan Pino, and Kalika Bali (eds.), \emph{Findings of the Association for Computational Linguistics: {EMNLP} 2023, Singapore, December 6-10, 2023}, pp.\  7591--7609. Association for Computational Linguistics, 2023.
\newblock \doi{10.18653/V1/2023.FINDINGS-EMNLP.509}.
\newblock URL \url{https://doi.org/10.18653/v1/2023.findings-emnlp.509}.

\bibitem[Huang et~al.(2023)Huang, Tao, An, Zhang, Jiang, Chen, Wu, and Feng]{DBLP:journals/corr/abs-2305-15062}
Quzhe Huang, Mingxu Tao, Zhenwei An, Chen Zhang, Cong Jiang, Zhibin Chen, Zirui Wu, and Yansong Feng.
\newblock Lawyer llama technical report.
\newblock \emph{CoRR}, abs/2305.15062, 2023.
\newblock \doi{10.48550/ARXIV.2305.15062}.
\newblock URL \url{https://doi.org/10.48550/arXiv.2305.15062}.

\bibitem[Huerta-Enochian \& Ko(2024)Huerta-Enochian and Ko]{huertaenochian2024instruction}
Mathew Huerta-Enochian and Seung~Yong Ko.
\newblock Instruction fine-tuning: Does prompt loss matter?, 2024.

\bibitem[Husain et~al.(2024)Husain, Dabre, Kumar, Puduppully, and Kunchukuttan]{adapt-7}
Jaavid~Aktar Husain, Raj Dabre, Aswanth Kumar, Ratish Puduppully, and Anoop Kunchukuttan.
\newblock Romansetu: Efficiently unlocking multilingual capabilities of large language models models via romanization.
\newblock \emph{CoRR}, abs/2401.14280, 2024.
\newblock \doi{10.48550/ARXIV.2401.14280}.
\newblock URL \url{https://doi.org/10.48550/arXiv.2401.14280}.

\bibitem[Hutchins(1999)]{hutchins-1999-retrospect}
John Hutchins.
\newblock Retrospect and prospect in computer-based translation.
\newblock In \emph{Proceedings of Machine Translation Summit VII}, pp.\  30--36, Singapore, Singapore, September 13-17 1999.
\newblock URL \url{https://aclanthology.org/1999.mtsummit-1.5}.

\bibitem[Ian~Kivlichan(2020)]{jigsaw-multilingual-toxic-comment-classification}
Julia Elliott et~al. Ian~Kivlichan, Jeffrey~Sorensen.
\newblock Jigsaw multilingual toxic comment classification, 2020.
\newblock URL \url{https://kaggle.com/competitions/jigsaw-multilingual-toxic-comment-classification}.

\bibitem[Jain et~al.(2024)Jain, Kumar, Gehman, Zhou, Hartvigsen, and Sap]{jain2024polyglotoxicityprompts}
Devansh Jain, Priyanshu Kumar, Samuel Gehman, Xuhui Zhou, Thomas Hartvigsen, and Maarten Sap.
\newblock Polyglotoxicityprompts: Multilingual evaluation of neural toxic degeneration in large language models, 2024.

\bibitem[Ji \& Chen(2024)Ji and Chen]{tuning-9}
Shaoxiong Ji and Pinzhen Chen.
\newblock Lucky 52: How many languages are needed to instruction fine-tune large language models?
\newblock \emph{CoRR}, abs/2404.04850, 2024.
\newblock \doi{10.48550/ARXIV.2404.04850}.
\newblock URL \url{https://doi.org/10.48550/arXiv.2404.04850}.

\bibitem[Ji et~al.(2021)Ji, Zhou, Liu, and Davuluri]{DBLP:journals/bioinformatics/JiZLD21}
Yanrong Ji, Zhihan Zhou, Han Liu, and Ramana~V. Davuluri.
\newblock {DNABERT:} pre-trained bidirectional encoder representations from transformers model for dna-language in genome.
\newblock \emph{Bioinform.}, 37\penalty0 (15):\penalty0 2112--2120, 2021.
\newblock \doi{10.1093/BIOINFORMATICS/BTAB083}.
\newblock URL \url{https://doi.org/10.1093/bioinformatics/btab083}.

\bibitem[Jiang et~al.(2023)Jiang, Sablayrolles, Mensch, Bamford, Chaplot, Casas, Bressand, Lengyel, Lample, Saulnier, et~al.]{mistral}
Albert~Q Jiang, Alexandre Sablayrolles, Arthur Mensch, Chris Bamford, Devendra~Singh Chaplot, Diego de~las Casas, Florian Bressand, Gianna Lengyel, Guillaume Lample, Lucile Saulnier, et~al.
\newblock Mistral 7b.
\newblock \emph{arXiv preprint arXiv:2310.06825}, 2023.

\bibitem[Jiang et~al.(2024)Jiang, Sablayrolles, Roux, Mensch, Savary, Bamford, Chaplot, Casas, Hanna, Bressand, et~al.]{Mixtral}
Albert~Q Jiang, Alexandre Sablayrolles, Antoine Roux, Arthur Mensch, Blanche Savary, Chris Bamford, Devendra~Singh Chaplot, Diego de~las Casas, Emma~Bou Hanna, Florian Bressand, et~al.
\newblock Mixtral of experts.
\newblock \emph{arXiv preprint arXiv:2401.04088}, 2024.

\bibitem[Jiang et~al.(2020)Jiang, Anastasopoulos, Araki, Ding, and Neubig]{DBLP:conf/emnlp/JiangAADN20}
Zhengbao Jiang, Antonios Anastasopoulos, Jun Araki, Haibo Ding, and Graham Neubig.
\newblock {X-FACTR:} multilingual factual knowledge retrieval from pretrained language models.
\newblock In Bonnie Webber, Trevor Cohn, Yulan He, and Yang Liu (eds.), \emph{Proceedings of the 2020 Conference on Empirical Methods in Natural Language Processing, {EMNLP} 2020, Online, November 16-20, 2020}, pp.\  5943--5959. Association for Computational Linguistics, 2020.
\newblock \doi{10.18653/V1/2020.EMNLP-MAIN.479}.
\newblock URL \url{https://doi.org/10.18653/v1/2020.emnlp-main.479}.

\bibitem[K et~al.(2020)K, Wang, Mayhew, and Roth]{DBLP:conf/iclr/KWMR20}
Karthikeyan K, Zihan Wang, Stephen Mayhew, and Dan Roth.
\newblock Cross-lingual ability of multilingual {BERT:} an empirical study.
\newblock In \emph{8th International Conference on Learning Representations, {ICLR} 2020, Addis Ababa, Ethiopia, April 26-30, 2020}. OpenReview.net, 2020.
\newblock URL \url{https://openreview.net/forum?id=HJeT3yrtDr}.

\bibitem[Kabra et~al.(2023)Kabra, Liu, Khanuja, Aji, Winata, Cahyawijaya, Anuoluwapo, Ogayo, and Neubig]{c-14}
Anubha Kabra, Emmy Liu, Simran Khanuja, Alham~Fikri Aji, Genta~Indra Winata, Samuel Cahyawijaya, Aremu Anuoluwapo, Perez Ogayo, and Graham Neubig.
\newblock Multi-lingual and multi-cultural figurative language understanding.
\newblock In Anna Rogers, Jordan~L. Boyd{-}Graber, and Naoaki Okazaki (eds.), \emph{Findings of the Association for Computational Linguistics: {ACL} 2023, Toronto, Canada, July 9-14, 2023}, pp.\  8269--8284. Association for Computational Linguistics, 2023.
\newblock \doi{10.18653/V1/2023.FINDINGS-ACL.525}.
\newblock URL \url{https://doi.org/10.18653/v1/2023.findings-acl.525}.

\bibitem[Kakwani et~al.(2020)Kakwani, Kunchukuttan, Golla, N.C., Bhattacharyya, Khapra, and Kumar]{kakwani-etal-2020-indicnlpsuite}
Divyanshu Kakwani, Anoop Kunchukuttan, Satish Golla, Gokul N.C., Avik Bhattacharyya, Mitesh~M. Khapra, and Pratyush Kumar.
\newblock {I}ndic{NLPS}uite: Monolingual corpora, evaluation benchmarks and pre-trained multilingual language models for {I}ndian languages.
\newblock In Trevor Cohn, Yulan He, and Yang Liu (eds.), \emph{Findings of the Association for Computational Linguistics: EMNLP 2020}, pp.\  4948--4961, Online, November 2020. Association for Computational Linguistics.
\newblock \doi{10.18653/v1/2020.findings-emnlp.445}.
\newblock URL \url{https://aclanthology.org/2020.findings-emnlp.445}.

\bibitem[Kanade et~al.(2020)Kanade, Maniatis, Balakrishnan, and Shi]{DBLP:conf/icml/KanadeMBS20}
Aditya Kanade, Petros Maniatis, Gogul Balakrishnan, and Kensen Shi.
\newblock Learning and evaluating contextual embedding of source code.
\newblock In \emph{Proceedings of the 37th International Conference on Machine Learning, {ICML} 2020, 13-18 July 2020, Virtual Event}, volume 119 of \emph{Proceedings of Machine Learning Research}, pp.\  5110--5121. {PMLR}, 2020.
\newblock URL \url{http://proceedings.mlr.press/v119/kanade20a.html}.

\bibitem[Kaneko et~al.(2022)Kaneko, Imankulova, Bollegala, and Okazaki]{DBLP:conf/naacl/KanekoIBO22}
Masahiro Kaneko, Aizhan Imankulova, Danushka Bollegala, and Naoaki Okazaki.
\newblock Gender bias in masked language models for multiple languages.
\newblock In Marine Carpuat, Marie{-}Catherine de~Marneffe, and Iv{\'{a}}n Vladimir~Meza Ru{\'{\i}}z (eds.), \emph{Proceedings of the 2022 Conference of the North American Chapter of the Association for Computational Linguistics: Human Language Technologies, {NAACL} 2022, Seattle, WA, United States, July 10-15, 2022}, pp.\  2740--2750. Association for Computational Linguistics, 2022.
\newblock \doi{10.18653/V1/2022.NAACL-MAIN.197}.
\newblock URL \url{https://doi.org/10.18653/v1/2022.naacl-main.197}.

\bibitem[Kassner et~al.(2021)Kassner, Dufter, and Sch{\"{u}}tze]{DBLP:conf/eacl/KassnerDS21}
Nora Kassner, Philipp Dufter, and Hinrich Sch{\"{u}}tze.
\newblock Multilingual {LAMA:} investigating knowledge in multilingual pretrained language models.
\newblock In Paola Merlo, J{\"{o}}rg Tiedemann, and Reut Tsarfaty (eds.), \emph{Proceedings of the 16th Conference of the European Chapter of the Association for Computational Linguistics: Main Volume, {EACL} 2021, Online, April 19 - 23, 2021}, pp.\  3250--3258. Association for Computational Linguistics, 2021.
\newblock \doi{10.18653/V1/2021.EACL-MAIN.284}.
\newblock URL \url{https://doi.org/10.18653/v1/2021.eacl-main.284}.

\bibitem[Keleg \& Magdy(2023)Keleg and Magdy]{c-9}
Amr Keleg and Walid Magdy.
\newblock {DLAMA:} {A} framework for curating culturally diverse facts for probing the knowledge of pretrained language models.
\newblock In Anna Rogers, Jordan~L. Boyd{-}Graber, and Naoaki Okazaki (eds.), \emph{Findings of the Association for Computational Linguistics: {ACL} 2023, Toronto, Canada, July 9-14, 2023}, pp.\  6245--6266. Association for Computational Linguistics, 2023.
\newblock \doi{10.18653/V1/2023.FINDINGS-ACL.389}.
\newblock URL \url{https://doi.org/10.18653/v1/2023.findings-acl.389}.

\bibitem[Kew et~al.(2023)Kew, Schottmann, and Sennrich]{tuning-10}
Tannon Kew, Florian Schottmann, and Rico Sennrich.
\newblock Turning english-centric llms into polyglots: How much multilinguality is needed?
\newblock \emph{CoRR}, abs/2312.12683, 2023.
\newblock \doi{10.48550/ARXIV.2312.12683}.
\newblock URL \url{https://doi.org/10.48550/arXiv.2312.12683}.

\bibitem[Khanuja et~al.(2020)Khanuja, Dandapat, Srinivasan, Sitaram, and Choudhury]{DBLP:conf/acl/KhanujaDSSC20}
Simran Khanuja, Sandipan Dandapat, Anirudh Srinivasan, Sunayana Sitaram, and Monojit Choudhury.
\newblock Gluecos: An evaluation benchmark for code-switched {NLP}.
\newblock In Dan Jurafsky, Joyce Chai, Natalie Schluter, and Joel~R. Tetreault (eds.), \emph{Proceedings of the 58th Annual Meeting of the Association for Computational Linguistics, {ACL} 2020, Online, July 5-10, 2020}, pp.\  3575--3585. Association for Computational Linguistics, 2020.
\newblock \doi{10.18653/V1/2020.ACL-MAIN.329}.
\newblock URL \url{https://doi.org/10.18653/v1/2020.acl-main.329}.

\bibitem[Kim et~al.(2024{\natexlab{a}})Kim, Choi, and Jeong]{adapt-2}
Seungduk Kim, Seungtaek Choi, and Myeongho Jeong.
\newblock Efficient and effective vocabulary expansion towards multilingual large language models.
\newblock \emph{CoRR}, abs/2402.14714, 2024{\natexlab{a}}.
\newblock \doi{10.48550/ARXIV.2402.14714}.
\newblock URL \url{https://doi.org/10.48550/arXiv.2402.14714}.

\bibitem[Kim et~al.(2024{\natexlab{b}})Kim, Choi, and Jeong]{kim2024efficient}
Seungduk Kim, Seungtaek Choi, and Myeongho Jeong.
\newblock Efficient and effective vocabulary expansion towards multilingual large language models.
\newblock \emph{arXiv preprint arXiv:2402.14714}, 2024{\natexlab{b}}.

\bibitem[Kirk et~al.(2024)Kirk, Whitefield, R{\"{o}}ttger, Bean, Margatina, Ciro, Mosquera, Bartolo, Williams, He, Vidgen, and Hale]{DBLP:journals/corr/abs-2404-16019}
Hannah~Rose Kirk, Alexander Whitefield, Paul R{\"{o}}ttger, Andrew~M. Bean, Katerina Margatina, Juan Ciro, Rafael Mosquera, Max Bartolo, Adina Williams, He~He, Bertie Vidgen, and Scott~A. Hale.
\newblock The {PRISM} alignment project: What participatory, representative and individualised human feedback reveals about the subjective and multicultural alignment of large language models.
\newblock \emph{CoRR}, abs/2404.16019, 2024.
\newblock \doi{10.48550/ARXIV.2404.16019}.
\newblock URL \url{https://doi.org/10.48550/arXiv.2404.16019}.

\bibitem[Kirk et~al.(2023)Kirk, Mediratta, Nalmpantis, Luketina, Hambro, Grefenstette, and Raileanu]{DBLP:journals/corr/abs-2310-06452}
Robert Kirk, Ishita Mediratta, Christoforos Nalmpantis, Jelena Luketina, Eric Hambro, Edward Grefenstette, and Roberta Raileanu.
\newblock Understanding the effects of {RLHF} on {LLM} generalisation and diversity.
\newblock \emph{CoRR}, abs/2310.06452, 2023.
\newblock \doi{10.48550/ARXIV.2310.06452}.
\newblock URL \url{https://doi.org/10.48550/arXiv.2310.06452}.

\bibitem[Kocetkov et~al.(2022)Kocetkov, Li, Allal, Li, Mou, Ferrandis, Jernite, Mitchell, Hughes, Wolf, Bahdanau, von Werra, and de~Vries]{DBLP:journals/corr/abs-2211-15533}
Denis Kocetkov, Raymond Li, Loubna~Ben Allal, Jia Li, Chenghao Mou, Carlos~Mu{\~{n}}oz Ferrandis, Yacine Jernite, Margaret Mitchell, Sean Hughes, Thomas Wolf, Dzmitry Bahdanau, Leandro von Werra, and Harm de~Vries.
\newblock The stack: 3 {TB} of permissively licensed source code.
\newblock \emph{CoRR}, abs/2211.15533, 2022.
\newblock \doi{10.48550/ARXIV.2211.15533}.
\newblock URL \url{https://doi.org/10.48550/arXiv.2211.15533}.

\bibitem[Kojima et~al.(2024{\natexlab{a}})Kojima, Okimura, Iwasawa, Yanaka, and Matsuo]{DBLP:journals/corr/abs-2404-02431}
Takeshi Kojima, Itsuki Okimura, Yusuke Iwasawa, Hitomi Yanaka, and Yutaka Matsuo.
\newblock On the multilingual ability of decoder-based pre-trained language models: Finding and controlling language-specific neurons.
\newblock \emph{CoRR}, abs/2404.02431, 2024{\natexlab{a}}.
\newblock \doi{10.48550/ARXIV.2404.02431}.
\newblock URL \url{https://doi.org/10.48550/arXiv.2404.02431}.

\bibitem[Kojima et~al.(2024{\natexlab{b}})Kojima, Okimura, Iwasawa, Yanaka, and Matsuo]{kojima2024multilingual}
Takeshi Kojima, Itsuki Okimura, Yusuke Iwasawa, Hitomi Yanaka, and Yutaka Matsuo.
\newblock On the multilingual ability of decoder-based pre-trained language models: Finding and controlling language-specific neurons.
\newblock \emph{arXiv preprint arXiv:2404.02431}, 2024{\natexlab{b}}.

\bibitem[K{\"{o}}pf et~al.(2023)K{\"{o}}pf, Kilcher, von R{\"{u}}tte, Anagnostidis, Tam, Stevens, Barhoum, Nguyen, Stanley, Nagyfi, ES, Suri, Glushkov, Dantuluri, Maguire, Schuhmann, Nguyen, and Mattick]{DBLP:conf/nips/KopfKRATSBNSNES23}
Andreas K{\"{o}}pf, Yannic Kilcher, Dimitri von R{\"{u}}tte, Sotiris Anagnostidis, Zhi~Rui Tam, Keith Stevens, Abdullah Barhoum, Duc Nguyen, Oliver Stanley, Rich{\'{a}}rd Nagyfi, Shahul ES, Sameer Suri, David Glushkov, Arnav Dantuluri, Andrew Maguire, Christoph Schuhmann, Huu Nguyen, and Alexander Mattick.
\newblock Openassistant conversations - democratizing large language model alignment.
\newblock In Alice Oh, Tristan Naumann, Amir Globerson, Kate Saenko, Moritz Hardt, and Sergey Levine (eds.), \emph{Advances in Neural Information Processing Systems 36: Annual Conference on Neural Information Processing Systems 2023, NeurIPS 2023, New Orleans, LA, USA, December 10 - 16, 2023}, 2023.
\newblock URL \url{http://papers.nips.cc/paper\_files/paper/2023/hash/949f0f8f32267d297c2d4e3ee10a2e7e-Abstract-Datasets\_and\_Benchmarks.html}.

\bibitem[Kovac et~al.(2023)Kovac, Sawayama, Portelas, Colas, Dominey, and Oudeyer]{c-16}
Grgur Kovac, Masataka Sawayama, R{\'{e}}my Portelas, C{\'{e}}dric Colas, Peter~Ford Dominey, and Pierre{-}Yves Oudeyer.
\newblock Large language models as superpositions of cultural perspectives.
\newblock \emph{CoRR}, abs/2307.07870, 2023.
\newblock \doi{10.48550/ARXIV.2307.07870}.
\newblock URL \url{https://doi.org/10.48550/arXiv.2307.07870}.

\bibitem[Kraljevic et~al.(2021)Kraljevic, Shek, Bean, Bendayan, Teo, and Dobson]{DBLP:journals/corr/abs-2107-03134}
Zeljko Kraljevic, Anthony Shek, Daniel Bean, Rebecca Bendayan, James~T. Teo, and Richard J.~B. Dobson.
\newblock Medgpt: Medical concept prediction from clinical narratives.
\newblock \emph{CoRR}, abs/2107.03134, 2021.
\newblock URL \url{https://arxiv.org/abs/2107.03134}.

\bibitem[Ladhak et~al.(2020)Ladhak, Durmus, Cardie, and McKeown]{DBLP:conf/emnlp/LadhakDCM20}
Faisal Ladhak, Esin Durmus, Claire Cardie, and Kathleen~R. McKeown.
\newblock Wikilingua: {A} new benchmark dataset for multilingual abstractive summarization.
\newblock In Trevor Cohn, Yulan He, and Yang Liu (eds.), \emph{Findings of the Association for Computational Linguistics: {EMNLP} 2020, Online Event, 16-20 November 2020}, volume {EMNLP} 2020 of \emph{Findings of {ACL}}, pp.\  4034--4048. Association for Computational Linguistics, 2020.
\newblock \doi{10.18653/V1/2020.FINDINGS-EMNLP.360}.
\newblock URL \url{https://doi.org/10.18653/v1/2020.findings-emnlp.360}.

\bibitem[Lai et~al.(2023{\natexlab{a}})Lai, Ngo, Veyseh, Man, Dernoncourt, Bui, and Nguyen]{DBLP:conf/emnlp/LaiNVMDBN23}
Viet~Dac Lai, Nghia~Trung Ngo, Amir Pouran~Ben Veyseh, Hieu Man, Franck Dernoncourt, Trung Bui, and Thien~Huu Nguyen.
\newblock Chatgpt beyond english: Towards a comprehensive evaluation of large language models in multilingual learning.
\newblock In Houda Bouamor, Juan Pino, and Kalika Bali (eds.), \emph{Findings of the Association for Computational Linguistics: {EMNLP} 2023, Singapore, December 6-10, 2023}, pp.\  13171--13189. Association for Computational Linguistics, 2023{\natexlab{a}}.
\newblock \doi{10.18653/V1/2023.FINDINGS-EMNLP.878}.
\newblock URL \url{https://doi.org/10.18653/v1/2023.findings-emnlp.878}.

\bibitem[Lai et~al.(2023{\natexlab{b}})Lai, Nguyen, Ngo, Nguyen, Dernoncourt, Rossi, and Nguyen]{tuning-1}
Viet~Dac Lai, Chien~Van Nguyen, Nghia~Trung Ngo, Thuat Nguyen, Franck Dernoncourt, Ryan~A. Rossi, and Thien~Huu Nguyen.
\newblock Okapi: Instruction-tuned large language models in multiple languages with reinforcement learning from human feedback.
\newblock In Yansong Feng and Els Lefever (eds.), \emph{Proceedings of the 2023 Conference on Empirical Methods in Natural Language Processing, {EMNLP} 2023 - System Demonstrations, Singapore, December 6-10, 2023}, pp.\  318--327. Association for Computational Linguistics, 2023{\natexlab{b}}.
\newblock \doi{10.18653/V1/2023.EMNLP-DEMO.28}.
\newblock URL \url{https://doi.org/10.18653/v1/2023.emnlp-demo.28}.

\bibitem[Lauren{\c{c}}on et~al.(2023)Lauren{\c{c}}on, Saulnier, Wang, Akiki, del Moral, Scao, von Werra, Mou, Ponferrada, Nguyen, Frohberg, Sasko, Lhoest, McMillan{-}Major, Dupont, Biderman, Rogers, Allal, Toni, Pistilli, Nguyen, Nikpoor, Masoud, Colombo, de~la Rosa, Villegas, Thrush, Longpre, Nagel, Weber, Mu{\~{n}}oz, Zhu, van Strien, Alyafeai, Almubarak, Vu, Gonzalez{-}Dios, Soroa, Lo, Dey, Suarez, Gokaslan, Bose, Adelani, Phan, Tran, Yu, Pai, Chim, Lepercq, Ilic, Mitchell, Luccioni, and Jernite]{ROOTS}
Hugo Lauren{\c{c}}on, Lucile Saulnier, Thomas Wang, Christopher Akiki, Albert~Villanova del Moral, Teven~Le Scao, Leandro von Werra, Chenghao Mou, Eduardo~Gonz{\'{a}}lez Ponferrada, Huu Nguyen, J{\"{o}}rg Frohberg, Mario Sasko, Quentin Lhoest, Angelina McMillan{-}Major, G{\'{e}}rard Dupont, Stella Biderman, Anna Rogers, Loubna~Ben Allal, Francesco~De Toni, Giada Pistilli, Olivier Nguyen, Somaieh Nikpoor, Maraim Masoud, Pierre Colombo, Javier de~la Rosa, Paulo Villegas, Tristan Thrush, Shayne Longpre, Sebastian Nagel, Leon Weber, Manuel Mu{\~{n}}oz, Jian Zhu, Daniel van Strien, Zaid Alyafeai, Khalid Almubarak, Minh~Chien Vu, Itziar Gonzalez{-}Dios, Aitor Soroa, Kyle Lo, Manan Dey, Pedro~Ortiz Suarez, Aaron Gokaslan, Shamik Bose, David~Ifeoluwa Adelani, Long Phan, Hieu Tran, Ian Yu, Suhas Pai, Jenny Chim, Violette Lepercq, Suzana Ilic, Margaret Mitchell, Sasha Luccioni, and Yacine Jernite.
\newblock The bigscience {ROOTS} corpus: {A} 1.6tb composite multilingual dataset.
\newblock \emph{CoRR}, abs/2303.03915, 2023.
\newblock \doi{10.48550/ARXIV.2303.03915}.
\newblock URL \url{https://doi.org/10.48550/arXiv.2303.03915}.

\bibitem[Le~Scao et~al.(2023)Le~Scao, Fan, Akiki, Pavlick, Ili{\'c}, Hesslow, Castagn{\'e}, Luccioni, Yvon, Gall{\'e}, et~al.]{BLOOM}
Teven Le~Scao, Angela Fan, Christopher Akiki, Ellie Pavlick, Suzana Ili{\'c}, Daniel Hesslow, Roman Castagn{\'e}, Alexandra~Sasha Luccioni, Fran{\c{c}}ois Yvon, Matthias Gall{\'e}, et~al.
\newblock Bloom: A 176b-parameter open-access multilingual language model.
\newblock 2023.

\bibitem[Lee et~al.(2020)Lee, Yoon, Kim, Kim, Kim, So, and Kang]{DBLP:journals/bioinformatics/LeeYKKKSK20}
Jinhyuk Lee, Wonjin Yoon, Sungdong Kim, Donghyeon Kim, Sunkyu Kim, Chan~Ho So, and Jaewoo Kang.
\newblock Biobert: a pre-trained biomedical language representation model for biomedical text mining.
\newblock \emph{Bioinform.}, 36\penalty0 (4):\penalty0 1234--1240, 2020.
\newblock \doi{10.1093/BIOINFORMATICS/BTZ682}.
\newblock URL \url{https://doi.org/10.1093/bioinformatics/btz682}.

\bibitem[Lee et~al.(2022)Lee, Ippolito, Nystrom, Zhang, Eck, Callison-Burch, and Carlini]{deduplicating-lee}
Katherine Lee, Daphne Ippolito, Andrew Nystrom, Chiyuan Zhang, Douglas Eck, Chris Callison-Burch, and Nicholas Carlini.
\newblock Deduplicating training data makes language models better.
\newblock In Smaranda Muresan, Preslav Nakov, and Aline Villavicencio (eds.), \emph{Proceedings of the 60th Annual Meeting of the Association for Computational Linguistics (Volume 1: Long Papers)}, pp.\  8424--8445, Dublin, Ireland, May 2022. Association for Computational Linguistics.
\newblock \doi{10.18653/v1/2022.acl-long.577}.
\newblock URL \url{https://aclanthology.org/2022.acl-long.577}.

\bibitem[Lepikhin et~al.(2020)Lepikhin, Lee, Xu, Chen, Firat, Huang, Krikun, Shazeer, and Chen]{GShard}
Dmitry Lepikhin, HyoukJoong Lee, Yuanzhong Xu, Dehao Chen, Orhan Firat, Yanping Huang, Maxim Krikun, Noam Shazeer, and Zhifeng Chen.
\newblock Gshard: Scaling giant models with conditional computation and automatic sharding.
\newblock \emph{arXiv preprint arXiv:2006.16668}, 2020.

\bibitem[Lester et~al.(2021)Lester, Al{-}Rfou, and Constant]{cpt_12}
Brian Lester, Rami Al{-}Rfou, and Noah Constant.
\newblock The power of scale for parameter-efficient prompt tuning.
\newblock In Marie{-}Francine Moens, Xuanjing Huang, Lucia Specia, and Scott~Wen{-}tau Yih (eds.), \emph{Proceedings of the 2021 Conference on Empirical Methods in Natural Language Processing, {EMNLP} 2021, Virtual Event / Punta Cana, Dominican Republic, 7-11 November, 2021}, pp.\  3045--3059. Association for Computational Linguistics, 2021.
\newblock \doi{10.18653/V1/2021.EMNLP-MAIN.243}.
\newblock URL \url{https://doi.org/10.18653/v1/2021.emnlp-main.243}.

\bibitem[Levy et~al.(2022)Levy, Allaway, Subbiah, Chilton, Patton, McKeown, and Wang]{DBLP:conf/emnlp/LevyASCPMW22}
Sharon Levy, Emily Allaway, Melanie Subbiah, Lydia~B. Chilton, Desmond Patton, Kathleen~R. McKeown, and William~Yang Wang.
\newblock Safetext: {A} benchmark for exploring physical safety in language models.
\newblock In Yoav Goldberg, Zornitsa Kozareva, and Yue Zhang (eds.), \emph{Proceedings of the 2022 Conference on Empirical Methods in Natural Language Processing, {EMNLP} 2022, Abu Dhabi, United Arab Emirates, December 7-11, 2022}, pp.\  2407--2421. Association for Computational Linguistics, 2022.
\newblock \doi{10.18653/V1/2022.EMNLP-MAIN.154}.
\newblock URL \url{https://doi.org/10.18653/v1/2022.emnlp-main.154}.

\bibitem[Lewis et~al.(2019)Lewis, Liu, Goyal, Ghazvininejad, Mohamed, Levy, Stoyanov, and Zettlemoyer]{BART}
Mike Lewis, Yinhan Liu, Naman Goyal, Marjan Ghazvininejad, Abdelrahman Mohamed, Omer Levy, Ves Stoyanov, and Luke Zettlemoyer.
\newblock Bart: Denoising sequence-to-sequence pre-training for natural language generation, translation, and comprehension.
\newblock \emph{arXiv preprint arXiv:1910.13461}, 2019.

\bibitem[Lewis et~al.(2020)Lewis, Oguz, Rinott, Riedel, and Schwenk]{DBLP:conf/acl/LewisORRS20}
Patrick S.~H. Lewis, Barlas Oguz, Ruty Rinott, Sebastian Riedel, and Holger Schwenk.
\newblock {MLQA:} evaluating cross-lingual extractive question answering.
\newblock In Dan Jurafsky, Joyce Chai, Natalie Schluter, and Joel~R. Tetreault (eds.), \emph{Proceedings of the 58th Annual Meeting of the Association for Computational Linguistics, {ACL} 2020, Online, July 5-10, 2020}, pp.\  7315--7330. Association for Computational Linguistics, 2020.
\newblock \doi{10.18653/V1/2020.ACL-MAIN.653}.
\newblock URL \url{https://doi.org/10.18653/v1/2020.acl-main.653}.

\bibitem[Li et~al.(2024{\natexlab{a}})Li, Chen, Wang, Sitaram, and Xie]{c-34}
Cheng Li, Mengzhou Chen, Jindong Wang, Sunayana Sitaram, and Xing Xie.
\newblock Culturellm: Incorporating cultural differences into large language models.
\newblock \emph{CoRR}, abs/2402.10946, 2024{\natexlab{a}}.
\newblock \doi{10.48550/ARXIV.2402.10946}.
\newblock URL \url{https://doi.org/10.48550/arXiv.2402.10946}.

\bibitem[Li et~al.(2024{\natexlab{b}})Li, Teney, Yang, Wen, Xie, and Wang]{c-27}
Cheng Li, Damien Teney, Linyi Yang, Qingsong Wen, Xing Xie, and Jindong Wang.
\newblock Culturepark: Boosting cross-cultural understanding in large language models.
\newblock \emph{CoRR}, abs/2405.15145, 2024{\natexlab{b}}.
\newblock \doi{10.48550/ARXIV.2405.15145}.
\newblock URL \url{https://doi.org/10.48550/arXiv.2405.15145}.

\bibitem[Li et~al.(2024{\natexlab{c}})Li, Wang, Zhang, and Zong]{li2024improving}
Chong Li, Shaonan Wang, Jiajun Zhang, and Chengqing Zong.
\newblock Improving in-context learning of multilingual generative language models with cross-lingual alignment.
\newblock In \emph{Proceedings of the 2024 Conference of the North American Chapter of the Association for Computational Linguistics: Human Language Technologies (Volume 1: Long Papers)}, pp.\  8051--8069, 2024{\natexlab{c}}.

\bibitem[Li et~al.(2023{\natexlab{a}})Li, Hammoud, Itani, Khizbullin, and Ghanem]{DBLP:conf/nips/LiHIKG23}
Guohao Li, Hasan Hammoud, Hani Itani, Dmitrii Khizbullin, and Bernard Ghanem.
\newblock {CAMEL:} communicative agents for "mind" exploration of large language model society.
\newblock In Alice Oh, Tristan Naumann, Amir Globerson, Kate Saenko, Moritz Hardt, and Sergey Levine (eds.), \emph{Advances in Neural Information Processing Systems 36: Annual Conference on Neural Information Processing Systems 2023, NeurIPS 2023, New Orleans, LA, USA, December 10 - 16, 2023}, 2023{\natexlab{a}}.
\newblock URL \url{http://papers.nips.cc/paper\_files/paper/2023/hash/a3621ee907def47c1b952ade25c67698-Abstract-Conference.html}.

\bibitem[Li et~al.(2023{\natexlab{b}})Li, Ai, Chen, Dong, Wu, Liu, Chen, and Tian]{DBLP:conf/sigir/LiACDW0CT23}
Haitao Li, Qingyao Ai, Jia Chen, Qian Dong, Yueyue Wu, Yiqun Liu, Chong Chen, and Qi~Tian.
\newblock {SAILER:} structure-aware pre-trained language model for legal case retrieval.
\newblock In Hsin{-}Hsi Chen, Wei{-}Jou~(Edward) Duh, Hen{-}Hsen Huang, Makoto~P. Kato, Josiane Mothe, and Barbara Poblete (eds.), \emph{Proceedings of the 46th International {ACM} {SIGIR} Conference on Research and Development in Information Retrieval, {SIGIR} 2023, Taipei, Taiwan, July 23-27, 2023}, pp.\  1035--1044. {ACM}, 2023{\natexlab{b}}.
\newblock \doi{10.1145/3539618.3591761}.
\newblock URL \url{https://doi.org/10.1145/3539618.3591761}.

\bibitem[Li et~al.(2023{\natexlab{c}})Li, Koto, Wu, Aji, and Baldwin]{DBLP:journals/corr/abs-2305-15011}
Haonan Li, Fajri Koto, Minghao Wu, Alham~Fikri Aji, and Timothy Baldwin.
\newblock Bactrian-x : {A} multilingual replicable instruction-following model with low-rank adaptation.
\newblock \emph{CoRR}, abs/2305.15011, 2023{\natexlab{c}}.
\newblock \doi{10.48550/ARXIV.2305.15011}.
\newblock URL \url{https://doi.org/10.48550/arXiv.2305.15011}.

\bibitem[Li et~al.(2023{\natexlab{d}})Li, Koto, Wu, Aji, and Baldwin]{tuning-2}
Haonan Li, Fajri Koto, Minghao Wu, Alham~Fikri Aji, and Timothy Baldwin.
\newblock Bactrian-x : {A} multilingual replicable instruction-following model with low-rank adaptation.
\newblock \emph{CoRR}, abs/2305.15011, 2023{\natexlab{d}}.
\newblock \doi{10.48550/ARXIV.2305.15011}.
\newblock URL \url{https://doi.org/10.48550/arXiv.2305.15011}.

\bibitem[Li et~al.(2024{\natexlab{d}})Li, Jiang, Huang, Kim, Santy, Sorensen, Lin, Dziri, Ren, and Choi]{c-24}
Huihan Li, Liwei Jiang, Jena~D. Huang, Hyunwoo Kim, Sebastin Santy, Taylor Sorensen, Bill~Yuchen Lin, Nouha Dziri, Xiang Ren, and Yejin Choi.
\newblock {CULTURE-GEN:} revealing global cultural perception in language models through natural language prompting.
\newblock \emph{CoRR}, abs/2404.10199, 2024{\natexlab{d}}.
\newblock \doi{10.48550/ARXIV.2404.10199}.
\newblock URL \url{https://doi.org/10.48550/arXiv.2404.10199}.

\bibitem[Li et~al.(2023{\natexlab{e}})Li, Zhou, Huang, Chen, and Chen]{mt-3}
Jiahuan Li, Hao Zhou, Shujian Huang, Shanbo Chen, and Jiajun Chen.
\newblock Eliciting the translation ability of large language models via multilingual finetuning with translation instructions.
\newblock \emph{CoRR}, abs/2305.15083, 2023{\natexlab{e}}.
\newblock \doi{10.48550/ARXIV.2305.15083}.
\newblock URL \url{https://doi.org/10.48550/arXiv.2305.15083}.

\bibitem[Li et~al.(2024{\natexlab{e}})Li, Liu, Liu, Shi, Ren, Zheng, Liu, and Xue]{DBLP:journals/corr/abs-2401-16765}
Jie Li, Yi~Liu, Chongyang Liu, Ling Shi, Xiaoning Ren, Yaowen Zheng, Yang Liu, and Yinxing Xue.
\newblock A cross-language investigation into jailbreak attacks in large language models.
\newblock \emph{CoRR}, abs/2401.16765, 2024{\natexlab{e}}.
\newblock \doi{10.48550/ARXIV.2401.16765}.
\newblock URL \url{https://doi.org/10.48550/arXiv.2401.16765}.

\bibitem[Li et~al.(2023{\natexlab{f}})Li, Allal, Zi, Muennighoff, Kocetkov, Mou, Marone, Akiki, Li, Chim, Liu, Zheltonozhskii, Zhuo, Wang, Dehaene, Davaadorj, Lamy{-}Poirier, Monteiro, Shliazhko, Gontier, Meade, Zebaze, Yee, Umapathi, Zhu, Lipkin, Oblokulov, Wang, V, Stillerman, Patel, Abulkhanov, Zocca, Dey, Zhang, Moustafa{-}Fahmy, Bhattacharyya, Yu, Singh, Luccioni, Villegas, Kunakov, Zhdanov, Romero, Lee, Timor, Ding, Schlesinger, Schoelkopf, Ebert, Dao, Mishra, Gu, Robinson, Anderson, Dolan{-}Gavitt, Contractor, Reddy, Fried, Bahdanau, Jernite, Ferrandis, Hughes, Wolf, Guha, von Werra, and de~Vries]{DBLP:journals/corr/abs-2305-06161}
Raymond Li, Loubna~Ben Allal, Yangtian Zi, Niklas Muennighoff, Denis Kocetkov, Chenghao Mou, Marc Marone, Christopher Akiki, Jia Li, Jenny Chim, Qian Liu, Evgenii Zheltonozhskii, Terry~Yue Zhuo, Thomas Wang, Olivier Dehaene, Mishig Davaadorj, Joel Lamy{-}Poirier, Jo{\~{a}}o Monteiro, Oleh Shliazhko, Nicolas Gontier, Nicholas Meade, Armel Zebaze, Ming{-}Ho Yee, Logesh~Kumar Umapathi, Jian Zhu, Benjamin Lipkin, Muhtasham Oblokulov, Zhiruo Wang, Rudra~Murthy V, Jason Stillerman, Siva~Sankalp Patel, Dmitry Abulkhanov, Marco Zocca, Manan Dey, Zhihan Zhang, Nour Moustafa{-}Fahmy, Urvashi Bhattacharyya, Wenhao Yu, Swayam Singh, Sasha Luccioni, Paulo Villegas, Maxim Kunakov, Fedor Zhdanov, Manuel Romero, Tony Lee, Nadav Timor, Jennifer Ding, Claire Schlesinger, Hailey Schoelkopf, Jan Ebert, Tri Dao, Mayank Mishra, Alex Gu, Jennifer Robinson, Carolyn~Jane Anderson, Brendan Dolan{-}Gavitt, Danish Contractor, Siva Reddy, Daniel Fried, Dzmitry Bahdanau, Yacine Jernite, Carlos~Mu{\~{n}}oz Ferrandis, Sean Hughes, Thomas
  Wolf, Arjun Guha, Leandro von Werra, and Harm de~Vries.
\newblock Starcoder: may the source be with you!
\newblock \emph{CoRR}, abs/2305.06161, 2023{\natexlab{f}}.
\newblock \doi{10.48550/ARXIV.2305.06161}.
\newblock URL \url{https://doi.org/10.48550/arXiv.2305.06161}.

\bibitem[Li et~al.(2023{\natexlab{g}})Li, Yu, Zhou, Schick, Zettlemoyer, Levy, Weston, and Lewis]{DBLP:journals/corr/abs-2308-06259}
Xian Li, Ping Yu, Chunting Zhou, Timo Schick, Luke Zettlemoyer, Omer Levy, Jason Weston, and Mike Lewis.
\newblock Self-alignment with instruction backtranslation.
\newblock \emph{CoRR}, abs/2308.06259, 2023{\natexlab{g}}.
\newblock \doi{10.48550/ARXIV.2308.06259}.
\newblock URL \url{https://doi.org/10.48550/arXiv.2308.06259}.

\bibitem[Li \& Liang(2021)Li and Liang]{cpt_13}
Xiang~Lisa Li and Percy Liang.
\newblock Prefix-tuning: Optimizing continuous prompts for generation.
\newblock In Chengqing Zong, Fei Xia, Wenjie Li, and Roberto Navigli (eds.), \emph{Proceedings of the 59th Annual Meeting of the Association for Computational Linguistics and the 11th International Joint Conference on Natural Language Processing, {ACL/IJCNLP} 2021, (Volume 1: Long Papers), Virtual Event, August 1-6, 2021}, pp.\  4582--4597. Association for Computational Linguistics, 2021.
\newblock \doi{10.18653/V1/2021.ACL-LONG.353}.
\newblock URL \url{https://doi.org/10.18653/v1/2021.acl-long.353}.

\bibitem[Li et~al.(2022)Li, Choi, Chung, Kushman, Schrittwieser, Leblond, Eccles, Keeling, Gimeno, Lago, Hubert, Choy, de~Masson~d'Autume, Babuschkin, Chen, Huang, Welbl, Gowal, Cherepanov, Molloy, Mankowitz, Robson, Kohli, de~Freitas, Kavukcuoglu, and Vinyals]{DBLP:journals/corr/abs-2203-07814}
Yujia Li, David~H. Choi, Junyoung Chung, Nate Kushman, Julian Schrittwieser, R{\'{e}}mi Leblond, Tom Eccles, James Keeling, Felix Gimeno, Agustin~Dal Lago, Thomas Hubert, Peter Choy, Cyprien de~Masson~d'Autume, Igor Babuschkin, Xinyun Chen, Po{-}Sen Huang, Johannes Welbl, Sven Gowal, Alexey Cherepanov, James Molloy, Daniel~J. Mankowitz, Esme~Sutherland Robson, Pushmeet Kohli, Nando de~Freitas, Koray Kavukcuoglu, and Oriol Vinyals.
\newblock Competition-level code generation with alphacode.
\newblock \emph{CoRR}, abs/2203.07814, 2022.
\newblock \doi{10.48550/ARXIV.2203.07814}.
\newblock URL \url{https://doi.org/10.48550/arXiv.2203.07814}.

\bibitem[Li et~al.(2023{\natexlab{h}})Li, Li, Zhang, Dan, and Zhang]{DBLP:journals/corr/abs-2303-14070}
Yunxiang Li, Zihan Li, Kai Zhang, Ruilong Dan, and You Zhang.
\newblock Chatdoctor: {A} medical chat model fine-tuned on llama model using medical domain knowledge.
\newblock \emph{CoRR}, abs/2303.14070, 2023{\natexlab{h}}.
\newblock \doi{10.48550/ARXIV.2303.14070}.
\newblock URL \url{https://doi.org/10.48550/arXiv.2303.14070}.

\bibitem[Li et~al.(2024{\natexlab{f}})Li, Ji, Mickus, Segonne, and Tiedemann]{li2024comparison}
Zihao Li, Shaoxiong Ji, Timothee Mickus, Vincent Segonne, and J{\"o}rg Tiedemann.
\newblock A comparison of language modeling and translation as multilingual pretraining objectives.
\newblock In \emph{Proceedings of the 2024 Conference on Empirical Methods in Natural Language Processing}, pp.\  15882--15894, 2024{\natexlab{f}}.

\bibitem[Liang et~al.(2020)Liang, Duan, Gong, Wu, Guo, Qi, Gong, Shou, Jiang, Cao, Fan, Zhang, Agrawal, Cui, Wei, Bharti, Qiao, Chen, Wu, Liu, Yang, Campos, Majumder, and Zhou]{DBLP:conf/emnlp/LiangDGWGQGSJCF20}
Yaobo Liang, Nan Duan, Yeyun Gong, Ning Wu, Fenfei Guo, Weizhen Qi, Ming Gong, Linjun Shou, Daxin Jiang, Guihong Cao, Xiaodong Fan, Ruofei Zhang, Rahul Agrawal, Edward Cui, Sining Wei, Taroon Bharti, Ying Qiao, Jiun{-}Hung Chen, Winnie Wu, Shuguang Liu, Fan Yang, Daniel Campos, Rangan Majumder, and Ming Zhou.
\newblock {XGLUE:} {A} new benchmark dataset for cross-lingual pre-training, understanding and generation.
\newblock In Bonnie Webber, Trevor Cohn, Yulan He, and Yang Liu (eds.), \emph{Proceedings of the 2020 Conference on Empirical Methods in Natural Language Processing, {EMNLP} 2020, Online, November 16-20, 2020}, pp.\  6008--6018. Association for Computational Linguistics, 2020.
\newblock \doi{10.18653/V1/2020.EMNLP-MAIN.484}.
\newblock URL \url{https://doi.org/10.18653/v1/2020.emnlp-main.484}.

\bibitem[Liao et~al.(2024)Liao, Jiang, Wang, and Wang]{DBLP:journals/corr/abs-2404-09027}
Yusheng Liao, Shuyang Jiang, Yu~Wang, and Yanfeng Wang.
\newblock {MING-MOE:} enhancing medical multi-task learning in large language models with sparse mixture of low-rank adapter experts.
\newblock \emph{CoRR}, abs/2404.09027, 2024.
\newblock \doi{10.48550/ARXIV.2404.09027}.
\newblock URL \url{https://doi.org/10.48550/arXiv.2404.09027}.

\bibitem[Lieber et~al.(2024)Lieber, Lenz, Bata, Cohen, Osin, Dalmedigos, Safahi, Meirom, Belinkov, Shalev-Shwartz, et~al.]{jamba}
Opher Lieber, Barak Lenz, Hofit Bata, Gal Cohen, Jhonathan Osin, Itay Dalmedigos, Erez Safahi, Shaked Meirom, Yonatan Belinkov, Shai Shalev-Shwartz, et~al.
\newblock Jamba: A hybrid transformer-mamba language model.
\newblock \emph{arXiv preprint arXiv:2403.19887}, 2024.

\bibitem[Lifelo et~al.(2024)Lifelo, Ning, and Dhelim]{lifelo2024adapting}
Zita Lifelo, Huansheng Ning, and Sahraoui Dhelim.
\newblock Adapting mental health prediction tasks for cross-lingual learning via meta-training and in-context learning with large language model.
\newblock \emph{arXiv preprint arXiv:2404.09045}, 2024.

\bibitem[Lin et~al.(2021)Lin, Lee, Qiao, and Ren]{DBLP:conf/acl/LinLQ020}
Bill~Yuchen Lin, Seyeon Lee, Xiaoyang Qiao, and Xiang Ren.
\newblock Common sense beyond english: Evaluating and improving multilingual language models for commonsense reasoning.
\newblock In Chengqing Zong, Fei Xia, Wenjie Li, and Roberto Navigli (eds.), \emph{Proceedings of the 59th Annual Meeting of the Association for Computational Linguistics and the 11th International Joint Conference on Natural Language Processing, {ACL/IJCNLP} 2021, (Volume 1: Long Papers), Virtual Event, August 1-6, 2021}, pp.\  1274--1287. Association for Computational Linguistics, 2021.
\newblock \doi{10.18653/V1/2021.ACL-LONG.102}.
\newblock URL \url{https://doi.org/10.18653/v1/2021.acl-long.102}.

\bibitem[Lin et~al.(2024)Lin, Ji, Tiedemann, Martins, and Sch{\"{u}}tze]{adapt-4}
Peiqin Lin, Shaoxiong Ji, J{\"{o}}rg Tiedemann, Andr{\'{e}} F.~T. Martins, and Hinrich Sch{\"{u}}tze.
\newblock Mala-500: Massive language adaptation of large language models.
\newblock \emph{CoRR}, abs/2401.13303, 2024.
\newblock \doi{10.48550/ARXIV.2401.13303}.
\newblock URL \url{https://doi.org/10.48550/arXiv.2401.13303}.

\bibitem[Lin et~al.(2022)Lin, Mihaylov, Artetxe, Wang, Chen, Simig, Ott, Goyal, Bhosale, Du, Pasunuru, Shleifer, Koura, Chaudhary, O'Horo, Wang, Zettlemoyer, Kozareva, Diab, Stoyanov, and Li]{DBLP:conf/emnlp/LinMAWCSOGBDPSK22}
Xi~Victoria Lin, Todor Mihaylov, Mikel Artetxe, Tianlu Wang, Shuohui Chen, Daniel Simig, Myle Ott, Naman Goyal, Shruti Bhosale, Jingfei Du, Ramakanth Pasunuru, Sam Shleifer, Punit~Singh Koura, Vishrav Chaudhary, Brian O'Horo, Jeff Wang, Luke Zettlemoyer, Zornitsa Kozareva, Mona~T. Diab, Veselin Stoyanov, and Xian Li.
\newblock Few-shot learning with multilingual generative language models.
\newblock In Yoav Goldberg, Zornitsa Kozareva, and Yue Zhang (eds.), \emph{Proceedings of the 2022 Conference on Empirical Methods in Natural Language Processing, {EMNLP} 2022, Abu Dhabi, United Arab Emirates, December 7-11, 2022}, pp.\  9019--9052. Association for Computational Linguistics, 2022.
\newblock \doi{10.18653/V1/2022.EMNLP-MAIN.616}.
\newblock URL \url{https://doi.org/10.18653/v1/2022.emnlp-main.616}.

\bibitem[Liu et~al.(2023)Liu, Koto, Baldwin, and Gurevych]{c-5}
Chen~Cecilia Liu, Fajri Koto, Timothy Baldwin, and Iryna Gurevych.
\newblock Are multilingual llms culturally-diverse reasoners? an investigation into multicultural proverbs and sayings.
\newblock \emph{CoRR}, abs/2309.08591, 2023.
\newblock \doi{10.48550/ARXIV.2309.08591}.
\newblock URL \url{https://doi.org/10.48550/arXiv.2309.08591}.

\bibitem[Liu et~al.(2021)Liu, Bugliarello, Ponti, Reddy, Collier, and Elliott]{DBLP:conf/emnlp/0001BPRCE21}
Fangyu Liu, Emanuele Bugliarello, Edoardo~Maria Ponti, Siva Reddy, Nigel Collier, and Desmond Elliott.
\newblock Visually grounded reasoning across languages and cultures.
\newblock In Marie{-}Francine Moens, Xuanjing Huang, Lucia Specia, and Scott~Wen{-}tau Yih (eds.), \emph{Proceedings of the 2021 Conference on Empirical Methods in Natural Language Processing, {EMNLP} 2021, Virtual Event / Punta Cana, Dominican Republic, 7-11 November, 2021}, pp.\  10467--10485. Association for Computational Linguistics, 2021.
\newblock \doi{10.18653/V1/2021.EMNLP-MAIN.818}.
\newblock URL \url{https://doi.org/10.18653/v1/2021.emnlp-main.818}.

\bibitem[Liu et~al.(2022)Liu, Tam, Muqeeth, Mohta, Huang, Bansal, and Raffel]{IA3}
Haokun Liu, Derek Tam, Mohammed Muqeeth, Jay Mohta, Tenghao Huang, Mohit Bansal, and Colin Raffel.
\newblock Few-shot parameter-efficient fine-tuning is better and cheaper than in-context learning.
\newblock In Sanmi Koyejo, S.~Mohamed, A.~Agarwal, Danielle Belgrave, K.~Cho, and A.~Oh (eds.), \emph{Advances in Neural Information Processing Systems 35: Annual Conference on Neural Information Processing Systems 2022, NeurIPS 2022, New Orleans, LA, USA, November 28 - December 9, 2022}, 2022.
\newblock URL \url{http://papers.nips.cc/paper\_files/paper/2022/hash/0cde695b83bd186c1fd456302888454c-Abstract-Conference.html}.

\bibitem[Liu et~al.(2024{\natexlab{a}})Liu, Xu, Xu, Chen, Hu, and Wu]{DBLP:journals/corr/abs-2402-16367}
Weize Liu, Yinlong Xu, Hongxia Xu, Jintai Chen, Xuming Hu, and Jian Wu.
\newblock Unraveling babel: Exploring multilingual activation patterns within large language models.
\newblock \emph{CoRR}, abs/2402.16367, 2024{\natexlab{a}}.
\newblock \doi{10.48550/ARXIV.2402.16367}.
\newblock URL \url{https://doi.org/10.48550/arXiv.2402.16367}.

\bibitem[Liu et~al.(2024{\natexlab{b}})Liu, Xu, Xu, Chen, Hu, and Wu]{liu2024unraveling}
Weize Liu, Yinlong Xu, Hongxia Xu, Jintai Chen, Xuming Hu, and Jian Wu.
\newblock Unraveling babel: Exploring multilingual activation patterns within large language models.
\newblock \emph{arXiv preprint arXiv:2402.16367}, 2024{\natexlab{b}}.

\bibitem[Liu et~al.(2020)Liu, Gu, Goyal, Li, Edunov, Ghazvininejad, Lewis, and Zettlemoyer]{DBLP:journals/tacl/LiuGGLEGLZ20}
Yinhan Liu, Jiatao Gu, Naman Goyal, Xian Li, Sergey Edunov, Marjan Ghazvininejad, Mike Lewis, and Luke Zettlemoyer.
\newblock Multilingual denoising pre-training for neural machine translation.
\newblock \emph{Trans. Assoc. Comput. Linguistics}, 8:\penalty0 726--742, 2020.
\newblock \doi{10.1162/TACL\_A\_00343}.
\newblock URL \url{https://doi.org/10.1162/tacl\_a\_00343}.

\bibitem[Longpre et~al.(2023)Longpre, Hou, Vu, Webson, Chung, Tay, Zhou, Le, Zoph, Wei, et~al.]{flan2022}
Shayne Longpre, Le~Hou, Tu~Vu, Albert Webson, Hyung~Won Chung, Yi~Tay, Denny Zhou, Quoc~V Le, Barret Zoph, Jason Wei, et~al.
\newblock The flan collection: Designing data and methods for effective instruction tuning.
\newblock \emph{arXiv preprint arXiv:2301.13688}, 2023.

\bibitem[Lozhkov et~al.(2024)Lozhkov, Li, Allal, Cassano, Lamy{-}Poirier, Tazi, Tang, Pykhtar, Liu, Wei, Liu, Tian, Kocetkov, Zucker, Belkada, Wang, Liu, Abulkhanov, Paul, Li, Li, Risdal, Li, Zhu, Zhuo, Zheltonozhskii, Dade, Yu, Krau{\ss}, Jain, Su, He, Dey, Abati, Chai, Muennighoff, Tang, Oblokulov, Akiki, Marone, Mou, Mishra, Gu, Hui, Dao, Zebaze, Dehaene, Patry, Xu, McAuley, Hu, Scholak, Paquet, Robinson, Anderson, Chapados, and et~al.]{DBLP:journals/corr/abs-2402-19173}
Anton Lozhkov, Raymond Li, Loubna~Ben Allal, Federico Cassano, Joel Lamy{-}Poirier, Nouamane Tazi, Ao~Tang, Dmytro Pykhtar, Jiawei Liu, Yuxiang Wei, Tianyang Liu, Max Tian, Denis Kocetkov, Arthur Zucker, Younes Belkada, Zijian Wang, Qian Liu, Dmitry Abulkhanov, Indraneil Paul, Zhuang Li, Wen{-}Ding Li, Megan Risdal, Jia Li, Jian Zhu, Terry~Yue Zhuo, Evgenii Zheltonozhskii, Nii Osae~Osae Dade, Wenhao Yu, Lucas Krau{\ss}, Naman Jain, Yixuan Su, Xuanli He, Manan Dey, Edoardo Abati, Yekun Chai, Niklas Muennighoff, Xiangru Tang, Muhtasham Oblokulov, Christopher Akiki, Marc Marone, Chenghao Mou, Mayank Mishra, Alex Gu, Binyuan Hui, Tri Dao, Armel Zebaze, Olivier Dehaene, Nicolas Patry, Canwen Xu, Julian~J. McAuley, Han Hu, Torsten Scholak, S{\'{e}}bastien Paquet, Jennifer Robinson, Carolyn~Jane Anderson, Nicolas Chapados, and et~al.
\newblock Starcoder 2 and the stack v2: The next generation.
\newblock \emph{CoRR}, abs/2402.19173, 2024.
\newblock \doi{10.48550/ARXIV.2402.19173}.
\newblock URL \url{https://doi.org/10.48550/arXiv.2402.19173}.

\bibitem[Luo et~al.(2023{\natexlab{a}})Luo, Sun, Xu, Zhao, Lou, Tao, Geng, Lin, Chen, and Zhang]{DBLP:journals/corr/abs-2308-09583}
Haipeng Luo, Qingfeng Sun, Can Xu, Pu~Zhao, Jianguang Lou, Chongyang Tao, Xiubo Geng, Qingwei Lin, Shifeng Chen, and Dongmei Zhang.
\newblock Wizardmath: Empowering mathematical reasoning for large language models via reinforced evol-instruct.
\newblock \emph{CoRR}, abs/2308.09583, 2023{\natexlab{a}}.
\newblock \doi{10.48550/ARXIV.2308.09583}.
\newblock URL \url{https://doi.org/10.48550/arXiv.2308.09583}.

\bibitem[Luo et~al.(2023{\natexlab{b}})Luo, Xu, Zhao, Sun, Geng, Hu, Tao, Ma, Lin, and Jiang]{DBLP:journals/corr/abs-2306-08568}
Ziyang Luo, Can Xu, Pu~Zhao, Qingfeng Sun, Xiubo Geng, Wenxiang Hu, Chongyang Tao, Jing Ma, Qingwei Lin, and Daxin Jiang.
\newblock Wizardcoder: Empowering code large language models with evol-instruct.
\newblock \emph{CoRR}, abs/2306.08568, 2023{\natexlab{b}}.
\newblock \doi{10.48550/ARXIV.2306.08568}.
\newblock URL \url{https://doi.org/10.48550/arXiv.2306.08568}.

\bibitem[Ma et~al.(2022)Ma, Datta, Wang, and Vosoughi]{c-15}
Weicheng Ma, Samiha Datta, Lili Wang, and Soroush Vosoughi.
\newblock Encbp: {A} new benchmark dataset for finer-grained cultural background prediction in english.
\newblock In Smaranda Muresan, Preslav Nakov, and Aline Villavicencio (eds.), \emph{Findings of the Association for Computational Linguistics: {ACL} 2022, Dublin, Ireland, May 22-27, 2022}, pp.\  2811--2823. Association for Computational Linguistics, 2022.
\newblock \doi{10.18653/V1/2022.FINDINGS-ACL.221}.
\newblock URL \url{https://doi.org/10.18653/v1/2022.findings-acl.221}.

\bibitem[Malmasi et~al.(2022)Malmasi, Fang, Fetahu, Kar, and Rokhlenko]{DBLP:conf/semeval/MalmasiFFKR22}
Shervin Malmasi, Anjie Fang, Besnik Fetahu, Sudipta Kar, and Oleg Rokhlenko.
\newblock Semeval-2022 task 11: Multilingual complex named entity recognition (multiconer).
\newblock In Guy Emerson, Natalie Schluter, Gabriel Stanovsky, Ritesh Kumar, Alexis Palmer, Nathan Schneider, Siddharth Singh, and Shyam Ratan (eds.), \emph{Proceedings of the 16th International Workshop on Semantic Evaluation, SemEval@NAACL 2022, Seattle, Washington, United States, July 14-15, 2022}, pp.\  1412--1437. Association for Computational Linguistics, 2022.
\newblock \doi{10.18653/V1/2022.SEMEVAL-1.196}.
\newblock URL \url{https://doi.org/10.18653/v1/2022.semeval-1.196}.

\bibitem[Mao \& Yu(2024)Mao and Yu]{mt-11}
Zhuoyuan Mao and Yen Yu.
\newblock Tuning llms with contrastive alignment instructions for machine translation in unseen, low-resource languages.
\newblock \emph{CoRR}, abs/2401.05811, 2024.
\newblock \doi{10.48550/ARXIV.2401.05811}.
\newblock URL \url{https://doi.org/10.48550/arXiv.2401.05811}.

\bibitem[Masoud et~al.(2023)Masoud, Liu, Ferianc, Treleaven, and Rodrigues]{c-8}
Reem~I. Masoud, Ziquan Liu, Martin Ferianc, Philip~C. Treleaven, and Miguel Rodrigues.
\newblock Cultural alignment in large language models: An explanatory analysis based on hofstede's cultural dimensions.
\newblock \emph{CoRR}, abs/2309.12342, 2023.
\newblock \doi{10.48550/ARXIV.2309.12342}.
\newblock URL \url{https://doi.org/10.48550/arXiv.2309.12342}.

\bibitem[Mehta et~al.(2024)Mehta, Sekhavat, Cao, Horton, Jin, Sun, Mirzadeh, Najibi, Belenko, Zatloukal, et~al.]{OpenELM}
Sachin Mehta, Mohammad~Hossein Sekhavat, Qingqing Cao, Maxwell Horton, Yanzi Jin, Chenfan Sun, Iman Mirzadeh, Mahyar Najibi, Dmitry Belenko, Peter Zatloukal, et~al.
\newblock Openelm: An efficient language model family with open-source training and inference framework.
\newblock \emph{arXiv preprint arXiv:2404.14619}, 2024.

\bibitem[Meng et~al.(2024)Meng, Xia, and Chen]{DBLP:journals/corr/abs-2405-14734}
Yu~Meng, Mengzhou Xia, and Danqi Chen.
\newblock Simpo: Simple preference optimization with a reference-free reward.
\newblock \emph{CoRR}, abs/2405.14734, 2024.
\newblock \doi{10.48550/ARXIV.2405.14734}.
\newblock URL \url{https://doi.org/10.48550/arXiv.2405.14734}.

\bibitem[Mesnard et~al.(2024)Mesnard, Hardin, Dadashi, Bhupatiraju, Pathak, Sifre, Rivi{\`{e}}re, Kale, Love, Tafti, Hussenot, Chowdhery, Roberts, Barua, Botev, Castro{-}Ros, Slone, H{\'{e}}liou, Tacchetti, Bulanova, Paterson, Tsai, Shahriari, Lan, Choquette{-}Choo, Crepy, Cer, Ippolito, Reid, Buchatskaya, Ni, Noland, Yan, Tucker, Muraru, Rozhdestvenskiy, Michalewski, Tenney, Grishchenko, Austin, Keeling, Labanowski, Lespiau, Stanway, Brennan, Chen, Ferret, Chiu, and et~al.]{DBLP:journals/corr/abs-2403-08295}
Thomas Mesnard, Cassidy Hardin, Robert Dadashi, Surya Bhupatiraju, Shreya Pathak, Laurent Sifre, Morgane Rivi{\`{e}}re, Mihir~Sanjay Kale, Juliette Love, Pouya Tafti, L{\'{e}}onard Hussenot, Aakanksha Chowdhery, Adam Roberts, Aditya Barua, Alex Botev, Alex Castro{-}Ros, Ambrose Slone, Am{\'{e}}lie H{\'{e}}liou, Andrea Tacchetti, Anna Bulanova, Antonia Paterson, Beth Tsai, Bobak Shahriari, Charline~Le Lan, Christopher~A. Choquette{-}Choo, Cl{\'{e}}ment Crepy, Daniel Cer, Daphne Ippolito, David Reid, Elena Buchatskaya, Eric Ni, Eric Noland, Geng Yan, George Tucker, George{-}Cristian Muraru, Grigory Rozhdestvenskiy, Henryk Michalewski, Ian Tenney, Ivan Grishchenko, Jacob Austin, James Keeling, Jane Labanowski, Jean{-}Baptiste Lespiau, Jeff Stanway, Jenny Brennan, Jeremy Chen, Johan Ferret, Justin Chiu, and et~al.
\newblock Gemma: Open models based on gemini research and technology.
\newblock \emph{CoRR}, abs/2403.08295, 2024.
\newblock \doi{10.48550/ARXIV.2403.08295}.
\newblock URL \url{https://doi.org/10.48550/arXiv.2403.08295}.

\bibitem[Mikhailov et~al.(2021)Mikhailov, Serikov, and Artemova]{DBLP:journals/corr/abs-2104-12847}
Vladislav Mikhailov, Oleg Serikov, and Ekaterina Artemova.
\newblock Morph call: Probing morphosyntactic content of multilingual transformers.
\newblock \emph{CoRR}, abs/2104.12847, 2021.
\newblock URL \url{https://arxiv.org/abs/2104.12847}.

\bibitem[Mitra et~al.(2024)Mitra, Khanpour, Rosset, and Awadallah]{DBLP:journals/corr/abs-2402-14830}
Arindam Mitra, Hamed Khanpour, Corby Rosset, and Ahmed Awadallah.
\newblock Orca-math: Unlocking the potential of slms in grade school math.
\newblock \emph{CoRR}, abs/2402.14830, 2024.
\newblock \doi{10.48550/ARXIV.2402.14830}.
\newblock URL \url{https://doi.org/10.48550/arXiv.2402.14830}.

\bibitem[Moradshahi et~al.(2023)Moradshahi, Shen, Bali, Choudhury, de~Chalendar, Goel, Kim, Kodali, Kumaraguru, Semmar, Semnani, Seo, Seshadri, Shrivastava, Sun, Yadavalli, You, Xiong, and Lam]{DBLP:conf/acl/MoradshahiSBCCG23}
Mehrad Moradshahi, Tianhao Shen, Kalika Bali, Monojit Choudhury, Ga{\"{e}}l de~Chalendar, Anmol Goel, Sungkyun Kim, Prashant Kodali, Ponnurangam Kumaraguru, Nasredine Semmar, Sina~J. Semnani, Jiwon Seo, Vivek Seshadri, Manish Shrivastava, Michael Sun, Aditya Yadavalli, Chaobin You, Deyi Xiong, and Monica~S. Lam.
\newblock X-risawoz: High-quality end-to-end multilingual dialogue datasets and few-shot agents.
\newblock In Anna Rogers, Jordan~L. Boyd{-}Graber, and Naoaki Okazaki (eds.), \emph{Findings of the Association for Computational Linguistics: {ACL} 2023, Toronto, Canada, July 9-14, 2023}, pp.\  2773--2794. Association for Computational Linguistics, 2023.
\newblock \doi{10.18653/V1/2023.FINDINGS-ACL.174}.
\newblock URL \url{https://doi.org/10.18653/v1/2023.findings-acl.174}.

\bibitem[Mu et~al.(2024{\natexlab{a}})Mu, Feng, Cao, Wu, Li, Wang, Xiao, Song, Liu, Zhang, et~al.]{mu2024large}
Yongyu Mu, Peinan Feng, Zhiquan Cao, Yuzhang Wu, Bei Li, Chenglong Wang, Tong Xiao, Kai Song, Tongran Liu, Chunliang Zhang, et~al.
\newblock Large language models are parallel multilingual learners.
\newblock \emph{arXiv preprint arXiv:2403.09073}, 2024{\natexlab{a}}.

\bibitem[Mu et~al.(2024{\natexlab{b}})Mu, Feng, Cao, Wu, Li, Wang, Xiao, Song, Liu, Zhang, et~al.]{mu2024revealing}
Yongyu Mu, Peinan Feng, Zhiquan Cao, Yuzhang Wu, Bei Li, Chenglong Wang, Tong Xiao, Kai Song, Tongran Liu, Chunliang Zhang, et~al.
\newblock Revealing the parallel multilingual learning within large language models.
\newblock In \emph{Proceedings of the 2024 Conference on Empirical Methods in Natural Language Processing}, pp.\  6976--6997, 2024{\natexlab{b}}.

\bibitem[Muennighoff et~al.(2023{\natexlab{a}})Muennighoff, Wang, Sutawika, Roberts, Biderman, Scao, Bari, Shen, Yong, Schoelkopf, Tang, Radev, Aji, Almubarak, Albanie, Alyafeai, Webson, Raff, and Raffel]{DBLP:conf/acl/MuennighoffWSRB23}
Niklas Muennighoff, Thomas Wang, Lintang Sutawika, Adam Roberts, Stella Biderman, Teven~Le Scao, M.~Saiful Bari, Sheng Shen, Zheng~Xin Yong, Hailey Schoelkopf, Xiangru Tang, Dragomir Radev, Alham~Fikri Aji, Khalid Almubarak, Samuel Albanie, Zaid Alyafeai, Albert Webson, Edward Raff, and Colin Raffel.
\newblock Crosslingual generalization through multitask finetuning.
\newblock In Anna Rogers, Jordan~L. Boyd{-}Graber, and Naoaki Okazaki (eds.), \emph{Proceedings of the 61st Annual Meeting of the Association for Computational Linguistics (Volume 1: Long Papers), {ACL} 2023, Toronto, Canada, July 9-14, 2023}, pp.\  15991--16111. Association for Computational Linguistics, 2023{\natexlab{a}}.
\newblock \doi{10.18653/V1/2023.ACL-LONG.891}.
\newblock URL \url{https://doi.org/10.18653/v1/2023.acl-long.891}.

\bibitem[Muennighoff et~al.(2023{\natexlab{b}})Muennighoff, Wang, Sutawika, Roberts, Biderman, Scao, Bari, Shen, Yong, Schoelkopf, Tang, Radev, Aji, Almubarak, Albanie, Alyafeai, Webson, Raff, and Raffel]{tuning-3}
Niklas Muennighoff, Thomas Wang, Lintang Sutawika, Adam Roberts, Stella Biderman, Teven~Le Scao, M.~Saiful Bari, Sheng Shen, Zheng~Xin Yong, Hailey Schoelkopf, Xiangru Tang, Dragomir Radev, Alham~Fikri Aji, Khalid Almubarak, Samuel Albanie, Zaid Alyafeai, Albert Webson, Edward Raff, and Colin Raffel.
\newblock Crosslingual generalization through multitask finetuning.
\newblock In Anna Rogers, Jordan~L. Boyd{-}Graber, and Naoaki Okazaki (eds.), \emph{Proceedings of the 61st Annual Meeting of the Association for Computational Linguistics (Volume 1: Long Papers), {ACL} 2023, Toronto, Canada, July 9-14, 2023}, pp.\  15991--16111. Association for Computational Linguistics, 2023{\natexlab{b}}.
\newblock \doi{10.18653/V1/2023.ACL-LONG.891}.
\newblock URL \url{https://doi.org/10.18653/v1/2023.acl-long.891}.

\bibitem[Mukherjee et~al.(2024)Mukherjee, Caliskan, Zhu, and Anastasopoulos]{c-35}
Anjishnu Mukherjee, Aylin Caliskan, Ziwei Zhu, and Antonios Anastasopoulos.
\newblock Global gallery: The fine art of painting culture portraits through multilingual instruction tuning.
\newblock \emph{CoRR}, 2024.

\bibitem[Mukherjee et~al.(2023)Mukherjee, Mitra, Jawahar, Agarwal, Palangi, and Awadallah]{DBLP:journals/corr/abs-2306-02707}
Subhabrata Mukherjee, Arindam Mitra, Ganesh Jawahar, Sahaj Agarwal, Hamid Palangi, and Ahmed Awadallah.
\newblock Orca: Progressive learning from complex explanation traces of {GPT-4}.
\newblock \emph{CoRR}, abs/2306.02707, 2023.
\newblock \doi{10.48550/ARXIV.2306.02707}.
\newblock URL \url{https://doi.org/10.48550/arXiv.2306.02707}.

\bibitem[Nakano et~al.(2021)Nakano, Hilton, Balaji, Wu, Ouyang, Kim, Hesse, Jain, Kosaraju, Saunders, Jiang, Cobbe, Eloundou, Krueger, Button, Knight, Chess, and Schulman]{DBLP:journals/corr/abs-2112-09332}
Reiichiro Nakano, Jacob Hilton, Suchir Balaji, Jeff Wu, Long Ouyang, Christina Kim, Christopher Hesse, Shantanu Jain, Vineet Kosaraju, William Saunders, Xu~Jiang, Karl Cobbe, Tyna Eloundou, Gretchen Krueger, Kevin Button, Matthew Knight, Benjamin Chess, and John Schulman.
\newblock Webgpt: Browser-assisted question-answering with human feedback.
\newblock \emph{CoRR}, abs/2112.09332, 2021.
\newblock URL \url{https://arxiv.org/abs/2112.09332}.

\bibitem[Naous et~al.(2023)Naous, Ryan, and Xu]{c-4}
Tarek Naous, Michael~J. Ryan, and Wei Xu.
\newblock Having beer after prayer? measuring cultural bias in large language models.
\newblock \emph{CoRR}, abs/2305.14456, 2023.
\newblock \doi{10.48550/ARXIV.2305.14456}.
\newblock URL \url{https://doi.org/10.48550/arXiv.2305.14456}.

\bibitem[Narayan et~al.(2018)Narayan, Cohen, and Lapata]{DBLP:conf/emnlp/NarayanCL18}
Shashi Narayan, Shay~B. Cohen, and Mirella Lapata.
\newblock Don't give me the details, just the summary! topic-aware convolutional neural networks for extreme summarization.
\newblock In Ellen Riloff, David Chiang, Julia Hockenmaier, and Jun'ichi Tsujii (eds.), \emph{Proceedings of the 2018 Conference on Empirical Methods in Natural Language Processing, Brussels, Belgium, October 31 - November 4, 2018}, pp.\  1797--1807. Association for Computational Linguistics, 2018.
\newblock \doi{10.18653/V1/D18-1206}.
\newblock URL \url{https://doi.org/10.18653/v1/d18-1206}.

\bibitem[Nezhad \& Agrawal(2024)Nezhad and Agrawal]{nezhad2024drives}
Sina~Bagheri Nezhad and Ameeta Agrawal.
\newblock What drives performance in multilingual language models?
\newblock \emph{arXiv preprint arXiv:2404.19159}, 2024.

\bibitem[Nguyen(2023)]{DBLP:journals/corr/abs-2302-05729}
Ha{-}Thanh Nguyen.
\newblock A brief report on lawgpt 1.0: {A} virtual legal assistant based on {GPT-3}.
\newblock \emph{CoRR}, abs/2302.05729, 2023.
\newblock \doi{10.48550/ARXIV.2302.05729}.
\newblock URL \url{https://doi.org/10.48550/arXiv.2302.05729}.

\bibitem[Nguyen et~al.(2024)Nguyen, Nguyen, Lai, Man, Ngo, Dernoncourt, Rossi, and Nguyen]{DBLP:conf/coling/NguyenNLMNDRN24}
Thuat Nguyen, Chien~Van Nguyen, Viet~Dac Lai, Hieu Man, Nghia~Trung Ngo, Franck Dernoncourt, Ryan~A. Rossi, and Thien~Huu Nguyen.
\newblock Culturax: {A} cleaned, enormous, and multilingual dataset for large language models in 167 languages.
\newblock In Nicoletta Calzolari, Min{-}Yen Kan, V{\'{e}}ronique Hoste, Alessandro Lenci, Sakriani Sakti, and Nianwen Xue (eds.), \emph{Proceedings of the 2024 Joint International Conference on Computational Linguistics, Language Resources and Evaluation, {LREC/COLING} 2024, 20-25 May, 2024, Torino, Italy}, pp.\  4226--4237. {ELRA} and {ICCL}, 2024.
\newblock URL \url{https://aclanthology.org/2024.lrec-main.377}.

\bibitem[Nguyen et~al.(2023{\natexlab{a}})Nguyen, Zhang, Li, Aljunied, Tan, Cheng, Chen, Deng, Yang, Liu, Zhang, and Bing]{adapt-5}
Xuan{-}Phi Nguyen, Wenxuan Zhang, Xin Li, Mahani Aljunied, Qingyu Tan, Liying Cheng, Guanzheng Chen, Yue Deng, Sen Yang, Chaoqun Liu, Hang Zhang, and Lidong Bing.
\newblock Seallms - large language models for southeast asia.
\newblock \emph{CoRR}, abs/2312.00738, 2023{\natexlab{a}}.
\newblock \doi{10.48550/ARXIV.2312.00738}.
\newblock URL \url{https://doi.org/10.48550/arXiv.2312.00738}.

\bibitem[Nguyen et~al.(2023{\natexlab{b}})Nguyen, Zhang, Li, Aljunied, Tan, Cheng, Chen, Deng, Yang, Liu, Zhang, and Bing]{tuning-8}
Xuan{-}Phi Nguyen, Wenxuan Zhang, Xin Li, Mahani Aljunied, Qingyu Tan, Liying Cheng, Guanzheng Chen, Yue Deng, Sen Yang, Chaoqun Liu, Hang Zhang, and Lidong Bing.
\newblock Seallms - large language models for southeast asia.
\newblock \emph{CoRR}, abs/2312.00738, 2023{\natexlab{b}}.
\newblock \doi{10.48550/ARXIV.2312.00738}.
\newblock URL \url{https://doi.org/10.48550/arXiv.2312.00738}.

\bibitem[Nijkamp et~al.(2023)Nijkamp, Pang, Hayashi, Tu, Wang, Zhou, Savarese, and Xiong]{DBLP:conf/iclr/NijkampPHTWZSX23}
Erik Nijkamp, Bo~Pang, Hiroaki Hayashi, Lifu Tu, Huan Wang, Yingbo Zhou, Silvio Savarese, and Caiming Xiong.
\newblock Codegen: An open large language model for code with multi-turn program synthesis.
\newblock In \emph{The Eleventh International Conference on Learning Representations, {ICLR} 2023, Kigali, Rwanda, May 1-5, 2023}. OpenReview.net, 2023.
\newblock URL \url{https://openreview.net/pdf?id=iaYcJKpY2B\_}.

\bibitem[Nivre et~al.(2018)Nivre, Abrams, Željko Agic, Ahrenberg, Antonsen, Aranzabe, Arutie, Asahara, Ateyah, and et~al.]{UD22}
Joakim Nivre, Mitchell Abrams, Željko Agic, Lars Ahrenberg, Lene Antonsen, Maria~Jesus Aranzabe, Gashaw Arutie, Masayuki Asahara, Luma Ateyah, and Mohammed~Attia et~al.
\newblock Universal dependencies 2.2.
\newblock 2018.

\bibitem[Ogundepo et~al.(2023)Ogundepo, Gwadabe, Rivera, Clark, Ruder, Adelani, Dossou, Diop, Sikasote, Hacheme, Buzaaba, Ezeani, Mabuya, Osei, Emezue, Kahira, Muhammad, Oladipo, Owodunni, Tonja, Shode, Asai, Ajayi, Siro, Arthur, Adeyemi, Ahia, Anuoluwapo, Awosan, Chukwuneke, Opoku, Ayodele, Otiende, Mwase, Sinkala, Rubungo, Ajisafe, Onwuegbuzia, Mbow, Niyomutabazi, Mukonde, Lawan, Ahmad, Alabi, Namukombo, Mbonu, Phiri, Putini, Mngoma, Amuok, Iro, and Adhiambo]{DBLP:journals/corr/abs-2305-06897}
Odunayo Ogundepo, Tajuddeen~R. Gwadabe, Clara~E. Rivera, Jonathan~H. Clark, Sebastian Ruder, David~Ifeoluwa Adelani, Bonaventure F.~P. Dossou, Abdou~Aziz Diop, Claytone Sikasote, Gilles Hacheme, Happy Buzaaba, Ignatius Ezeani, Rooweither Mabuya, Salomey Osei, Chris Emezue, Albert~Njoroge Kahira, Shamsuddeen~Hassan Muhammad, Akintunde Oladipo, Abraham~Toluwase Owodunni, Atnafu~Lambebo Tonja, Iyanuoluwa Shode, Akari Asai, Tunde~Oluwaseyi Ajayi, Clemencia Siro, Steven Arthur, Mofetoluwa Adeyemi, Orevaoghene Ahia, Aremu Anuoluwapo, Oyinkansola Awosan, Chiamaka Chukwuneke, Bernard Opoku, Awokoya Ayodele, Verrah Otiende, Christine Mwase, Boyd Sinkala, Andre~Niyongabo Rubungo, Daniel~A. Ajisafe, Emeka~Felix Onwuegbuzia, Habib Mbow, Emile Niyomutabazi, Eunice Mukonde, Falalu~Ibrahim Lawan, Ibrahim~Said Ahmad, Jesujoba~O. Alabi, Martin Namukombo, Chinedu~Emmanuel Mbonu, Mofya Phiri, Neo Putini, Ndumiso Mngoma, Priscilla~A. Amuok, Ruqayya~Nasir Iro, and Sonia Adhiambo.
\newblock Afriqa: Cross-lingual open-retrieval question answering for african languages.
\newblock \emph{CoRR}, abs/2305.06897, 2023.
\newblock \doi{10.48550/ARXIV.2305.06897}.
\newblock URL \url{https://doi.org/10.48550/arXiv.2305.06897}.

\bibitem[{Ortiz Su{'a}rez} et~al.(2019){Ortiz Su{'a}rez}, Sagot, and Romary]{OrtizSuarezSagotRomary2019}
Pedro~Javier {Ortiz Su{'a}rez}, Benoit Sagot, and Laurent Romary.
\newblock Asynchronous pipelines for processing huge corpora on medium to low resource infrastructures.
\newblock Proceedings of the Workshop on Challenges in the Management of Large Corpora (CMLC-7) 2019. Cardiff, 22nd July 2019, pp.\  9 -- 16, Mannheim, 2019. Leibniz-Institut f{"u}r Deutsche Sprache.
\newblock \doi{10.14618/ids-pub-9021}.
\newblock URL \url{http://nbn-resolving.de/urn:nbn:de:bsz:mh39-90215}.

\bibitem[Ortiz~Su{'a}rez et~al.(2020)Ortiz~Su{'a}rez, Romary, and Sagot]{ortiz-suarez-etal-2020-monolingual}
Pedro~Javier Ortiz~Su{'a}rez, Laurent Romary, and Benoit Sagot.
\newblock A monolingual approach to contextualized word embeddings for mid-resource languages.
\newblock In \emph{Proceedings of the 58th Annual Meeting of the Association for Computational Linguistics}, pp.\  1703--1714, Online, July 2020. Association for Computational Linguistics.
\newblock URL \url{https://www.aclweb.org/anthology/2020.acl-main.156}.

\bibitem[Ouyang et~al.(2022{\natexlab{a}})Ouyang, Wu, Jiang, Almeida, Wainwright, Mishkin, Zhang, Agarwal, Slama, Ray, et~al.]{InstructGPT}
Long Ouyang, Jeffrey Wu, Xu~Jiang, Diogo Almeida, Carroll Wainwright, Pamela Mishkin, Chong Zhang, Sandhini Agarwal, Katarina Slama, Alex Ray, et~al.
\newblock Training language models to follow instructions with human feedback.
\newblock \emph{Advances in neural information processing systems}, 35:\penalty0 27730--27744, 2022{\natexlab{a}}.

\bibitem[Ouyang et~al.(2022{\natexlab{b}})Ouyang, Wu, Jiang, Almeida, Wainwright, Mishkin, Zhang, Agarwal, Slama, Ray, Schulman, Hilton, Kelton, Miller, Simens, Askell, Welinder, Christiano, Leike, and Lowe]{DBLP:conf/nips/Ouyang0JAWMZASR22}
Long Ouyang, Jeffrey Wu, Xu~Jiang, Diogo Almeida, Carroll~L. Wainwright, Pamela Mishkin, Chong Zhang, Sandhini Agarwal, Katarina Slama, Alex Ray, John Schulman, Jacob Hilton, Fraser Kelton, Luke Miller, Maddie Simens, Amanda Askell, Peter Welinder, Paul~F. Christiano, Jan Leike, and Ryan Lowe.
\newblock Training language models to follow instructions with human feedback.
\newblock In Sanmi Koyejo, S.~Mohamed, A.~Agarwal, Danielle Belgrave, K.~Cho, and A.~Oh (eds.), \emph{Advances in Neural Information Processing Systems 35: Annual Conference on Neural Information Processing Systems 2022, NeurIPS 2022, New Orleans, LA, USA, November 28 - December 9, 2022}, 2022{\natexlab{b}}.
\newblock URL \url{http://papers.nips.cc/paper\_files/paper/2022/hash/b1efde53be364a73914f58805a001731-Abstract-Conference.html}.

\bibitem[Pagnoni et~al.(2021)Pagnoni, Balachandran, and Tsvetkov]{DBLP:conf/naacl/PagnoniBT21}
Artidoro Pagnoni, Vidhisha Balachandran, and Yulia Tsvetkov.
\newblock Understanding factuality in abstractive summarization with {FRANK:} {A} benchmark for factuality metrics.
\newblock In Kristina Toutanova, Anna Rumshisky, Luke Zettlemoyer, Dilek Hakkani{-}T{\"{u}}r, Iz~Beltagy, Steven Bethard, Ryan Cotterell, Tanmoy Chakraborty, and Yichao Zhou (eds.), \emph{Proceedings of the 2021 Conference of the North American Chapter of the Association for Computational Linguistics: Human Language Technologies, {NAACL-HLT} 2021, Online, June 6-11, 2021}, pp.\  4812--4829. Association for Computational Linguistics, 2021.
\newblock \doi{10.18653/V1/2021.NAACL-MAIN.383}.
\newblock URL \url{https://doi.org/10.18653/v1/2021.naacl-main.383}.

\bibitem[Palta \& Rudinger(2023)Palta and Rudinger]{c-10}
Shramay Palta and Rachel Rudinger.
\newblock {FORK:} {A} bite-sized test set for probing culinary cultural biases in commonsense reasoning models.
\newblock In Anna Rogers, Jordan~L. Boyd{-}Graber, and Naoaki Okazaki (eds.), \emph{Findings of the Association for Computational Linguistics: {ACL} 2023, Toronto, Canada, July 9-14, 2023}, pp.\  9952--9962. Association for Computational Linguistics, 2023.
\newblock \doi{10.18653/V1/2023.FINDINGS-ACL.631}.
\newblock URL \url{https://doi.org/10.18653/v1/2023.findings-acl.631}.

\bibitem[Pan et~al.(2017)Pan, Zhang, May, Nothman, Knight, and Ji]{DBLP:conf/acl/PanZMNKJ17}
Xiaoman Pan, Boliang Zhang, Jonathan May, Joel Nothman, Kevin Knight, and Heng Ji.
\newblock Cross-lingual name tagging and linking for 282 languages.
\newblock In Regina Barzilay and Min{-}Yen Kan (eds.), \emph{Proceedings of the 55th Annual Meeting of the Association for Computational Linguistics, {ACL} 2017, Vancouver, Canada, July 30 - August 4, Volume 1: Long Papers}, pp.\  1946--1958. Association for Computational Linguistics, 2017.
\newblock \doi{10.18653/V1/P17-1178}.
\newblock URL \url{https://doi.org/10.18653/v1/P17-1178}.

\bibitem[Park et~al.(2024)Park, Lee, Park, Jeong, Koo, Hwang, Park, and Lee]{park2024multiprageval}
Dojun Park, Jiwoo Lee, Seohyun Park, Hyeyun Jeong, Youngeun Koo, Soonha Hwang, Seonwoo Park, and Sungeun Lee.
\newblock Multiprageval: Multilingual pragmatic evaluation of large language models.
\newblock \emph{arXiv preprint arXiv:2406.07736}, 2024.

\bibitem[Peng et~al.(2023)Peng, Alcaide, Anthony, Albalak, Arcadinho, Biderman, Cao, Cheng, Chung, Grella, et~al.]{rwkv}
Bo~Peng, Eric Alcaide, Quentin Anthony, Alon Albalak, Samuel Arcadinho, Stella Biderman, Huanqi Cao, Xin Cheng, Michael Chung, Matteo Grella, et~al.
\newblock Rwkv: Reinventing rnns for the transformer era.
\newblock \emph{arXiv preprint arXiv:2305.13048}, 2023.

\bibitem[Peng \& S{\o}gaard(2024)Peng and S{\o}gaard]{peng2024concept}
Qiwei Peng and Anders S{\o}gaard.
\newblock Concept space alignment in multilingual llms.
\newblock \emph{arXiv preprint arXiv:2410.01079}, 2024.

\bibitem[Petrov et~al.(2023)Petrov, Malfa, Torr, and Bibi]{DBLP:conf/nips/PetrovMTB23}
Aleksandar Petrov, Emanuele~La Malfa, Philip H.~S. Torr, and Adel Bibi.
\newblock Language model tokenizers introduce unfairness between languages.
\newblock In Alice Oh, Tristan Naumann, Amir Globerson, Kate Saenko, Moritz Hardt, and Sergey Levine (eds.), \emph{Advances in Neural Information Processing Systems 36: Annual Conference on Neural Information Processing Systems 2023, NeurIPS 2023, New Orleans, LA, USA, December 10 - 16, 2023}, 2023.
\newblock URL \url{http://papers.nips.cc/paper\_files/paper/2023/hash/74bb24dca8334adce292883b4b651eda-Abstract-Conference.html}.

\bibitem[Pfeiffer et~al.(2020)Pfeiffer, Vulic, Gurevych, and Ruder]{MAD-X}
Jonas Pfeiffer, Ivan Vulic, Iryna Gurevych, and Sebastian Ruder.
\newblock {MAD-X:} an adapter-based framework for multi-task cross-lingual transfer.
\newblock In Bonnie Webber, Trevor Cohn, Yulan He, and Yang Liu (eds.), \emph{Proceedings of the 2020 Conference on Empirical Methods in Natural Language Processing, {EMNLP} 2020, Online, November 16-20, 2020}, pp.\  7654--7673. Association for Computational Linguistics, 2020.
\newblock \doi{10.18653/V1/2020.EMNLP-MAIN.617}.
\newblock URL \url{https://doi.org/10.18653/v1/2020.emnlp-main.617}.

\bibitem[Philippy et~al.(2023)Philippy, Guo, and Haddadan]{DBLP:conf/acl/PhilippyGH23}
Fred Philippy, Siwen Guo, and Shohreh Haddadan.
\newblock Towards a common understanding of contributing factors for cross-lingual transfer in multilingual language models: {A} review.
\newblock In Anna Rogers, Jordan~L. Boyd{-}Graber, and Naoaki Okazaki (eds.), \emph{Proceedings of the 61st Annual Meeting of the Association for Computational Linguistics (Volume 1: Long Papers), {ACL} 2023, Toronto, Canada, July 9-14, 2023}, pp.\  5877--5891. Association for Computational Linguistics, 2023.
\newblock \doi{10.18653/V1/2023.ACL-LONG.323}.
\newblock URL \url{https://doi.org/10.18653/v1/2023.acl-long.323}.

\bibitem[Pistilli et~al.(2024)Pistilli, Leidinger, Jernite, Kasirzadeh, Luccioni, and Mitchell]{c-23}
Giada Pistilli, Alina Leidinger, Yacine Jernite, Atoosa Kasirzadeh, Alexandra~Sasha Luccioni, and Margaret Mitchell.
\newblock Civics: Building a dataset for examining culturally-informed values in large language models.
\newblock \emph{CoRR}, abs/2405.13974, 2024.
\newblock \doi{10.48550/ARXIV.2405.13974}.
\newblock URL \url{https://doi.org/10.48550/arXiv.2405.13974}.

\bibitem[Ponti et~al.(2020)Ponti, Glavas, Majewska, Liu, Vulic, and Korhonen]{DBLP:conf/emnlp/PontiGMLVK20}
Edoardo~Maria Ponti, Goran Glavas, Olga Majewska, Qianchu Liu, Ivan Vulic, and Anna Korhonen.
\newblock {XCOPA:} {A} multilingual dataset for causal commonsense reasoning.
\newblock In Bonnie Webber, Trevor Cohn, Yulan He, and Yang Liu (eds.), \emph{Proceedings of the 2020 Conference on Empirical Methods in Natural Language Processing, {EMNLP} 2020, Online, November 16-20, 2020}, pp.\  2362--2376. Association for Computational Linguistics, 2020.
\newblock \doi{10.18653/V1/2020.EMNLP-MAIN.185}.
\newblock URL \url{https://doi.org/10.18653/v1/2020.emnlp-main.185}.

\bibitem[Press et~al.(2021)Press, Smith, and Lewis]{ALiBi}
Ofir Press, Noah~A Smith, and Mike Lewis.
\newblock Train short, test long: Attention with linear biases enables input length extrapolation.
\newblock \emph{arXiv preprint arXiv:2108.12409}, 2021.

\bibitem[Qi et~al.(2023)Qi, Fern{\'{a}}ndez, and Bisazza]{DBLP:conf/emnlp/QiFB23}
Jirui Qi, Raquel Fern{\'{a}}ndez, and Arianna Bisazza.
\newblock Cross-lingual consistency of factual knowledge in multilingual language models.
\newblock In Houda Bouamor, Juan Pino, and Kalika Bali (eds.), \emph{Proceedings of the 2023 Conference on Empirical Methods in Natural Language Processing, {EMNLP} 2023, Singapore, December 6-10, 2023}, pp.\  10650--10666. Association for Computational Linguistics, 2023.
\newblock \doi{10.18653/V1/2023.EMNLP-MAIN.658}.
\newblock URL \url{https://doi.org/10.18653/v1/2023.emnlp-main.658}.

\bibitem[Qin et~al.(2024)Qin, Chen, Zhou, Chen, Li, Liao, Li, Che, and Yu]{qin2024multilingual}
Libo Qin, Qiguang Chen, Yuhang Zhou, Zhi Chen, Yinghui Li, Lizi Liao, Min Li, Wanxiang Che, and Philip~S Yu.
\newblock Multilingual large language model: A survey of resources, taxonomy and frontiers.
\newblock \emph{arXiv preprint arXiv:2404.04925}, 2024.

\bibitem[Qiu et~al.(2024)Qiu, Wu, Zhang, Lin, Wang, Zhang, Wang, and Xie]{qiu2024towards}
Pengcheng Qiu, Chaoyi Wu, Xiaoman Zhang, Weixiong Lin, Haicheng Wang, Ya~Zhang, Yanfeng Wang, and Weidi Xie.
\newblock Towards building multilingual language model for medicine.
\newblock \emph{Nature Communications}, 15\penalty0 (1):\penalty0 8384, 2024.

\bibitem[Rae et~al.(2021)Rae, Borgeaud, Cai, Millican, Hoffmann, Song, Aslanides, Henderson, Ring, Young, Rutherford, Hennigan, Menick, Cassirer, Powell, van~den Driessche, Hendricks, Rauh, Huang, Glaese, Welbl, Dathathri, Huang, Uesato, Mellor, Higgins, Creswell, McAleese, Wu, Elsen, Jayakumar, Buchatskaya, Budden, Sutherland, Simonyan, Paganini, Sifre, Martens, Li, Kuncoro, Nematzadeh, Gribovskaya, Donato, Lazaridou, Mensch, Lespiau, Tsimpoukelli, Grigorev, Fritz, Sottiaux, Pajarskas, Pohlen, Gong, Toyama, de~Masson~d'Autume, Li, Terzi, Mikulik, Babuschkin, Clark, de~Las~Casas, Guy, Jones, Bradbury, Johnson, Hechtman, Weidinger, Gabriel, Isaac, Lockhart, Osindero, Rimell, Dyer, Vinyals, Ayoub, Stanway, Bennett, Hassabis, Kavukcuoglu, and Irving]{DBLP:journals/corr/abs-2112-11446}
Jack~W. Rae, Sebastian Borgeaud, Trevor Cai, Katie Millican, Jordan Hoffmann, H.~Francis Song, John Aslanides, Sarah Henderson, Roman Ring, Susannah Young, Eliza Rutherford, Tom Hennigan, Jacob Menick, Albin Cassirer, Richard Powell, George van~den Driessche, Lisa~Anne Hendricks, Maribeth Rauh, Po{-}Sen Huang, Amelia Glaese, Johannes Welbl, Sumanth Dathathri, Saffron Huang, Jonathan Uesato, John Mellor, Irina Higgins, Antonia Creswell, Nat McAleese, Amy Wu, Erich Elsen, Siddhant~M. Jayakumar, Elena Buchatskaya, David Budden, Esme Sutherland, Karen Simonyan, Michela Paganini, Laurent Sifre, Lena Martens, Xiang~Lorraine Li, Adhiguna Kuncoro, Aida Nematzadeh, Elena Gribovskaya, Domenic Donato, Angeliki Lazaridou, Arthur Mensch, Jean{-}Baptiste Lespiau, Maria Tsimpoukelli, Nikolai Grigorev, Doug Fritz, Thibault Sottiaux, Mantas Pajarskas, Toby Pohlen, Zhitao Gong, Daniel Toyama, Cyprien de~Masson~d'Autume, Yujia Li, Tayfun Terzi, Vladimir Mikulik, Igor Babuschkin, Aidan Clark, Diego de~Las~Casas, Aurelia Guy,
  Chris Jones, James Bradbury, Matthew~J. Johnson, Blake~A. Hechtman, Laura Weidinger, Iason Gabriel, William Isaac, Edward Lockhart, Simon Osindero, Laura Rimell, Chris Dyer, Oriol Vinyals, Kareem Ayoub, Jeff Stanway, Lorrayne Bennett, Demis Hassabis, Koray Kavukcuoglu, and Geoffrey Irving.
\newblock Scaling language models: Methods, analysis {\&} insights from training gopher.
\newblock \emph{CoRR}, abs/2112.11446, 2021.
\newblock URL \url{https://arxiv.org/abs/2112.11446}.

\bibitem[Rafailov et~al.(2023)Rafailov, Sharma, Mitchell, Manning, Ermon, and Finn]{DBLP:conf/nips/RafailovSMMEF23}
Rafael Rafailov, Archit Sharma, Eric Mitchell, Christopher~D. Manning, Stefano Ermon, and Chelsea Finn.
\newblock Direct preference optimization: Your language model is secretly a reward model.
\newblock In Alice Oh, Tristan Naumann, Amir Globerson, Kate Saenko, Moritz Hardt, and Sergey Levine (eds.), \emph{Advances in Neural Information Processing Systems 36: Annual Conference on Neural Information Processing Systems 2023, NeurIPS 2023, New Orleans, LA, USA, December 10 - 16, 2023}, 2023.
\newblock URL \url{http://papers.nips.cc/paper\_files/paper/2023/hash/a85b405ed65c6477a4fe8302b5e06ce7-Abstract-Conference.html}.

\bibitem[Raffel et~al.(2020{\natexlab{a}})Raffel, Shazeer, Roberts, Lee, Narang, Matena, Zhou, Li, and Liu]{DBLP:journals/jmlr/RaffelSRLNMZLL20}
Colin Raffel, Noam Shazeer, Adam Roberts, Katherine Lee, Sharan Narang, Michael Matena, Yanqi Zhou, Wei Li, and Peter~J. Liu.
\newblock Exploring the limits of transfer learning with a unified text-to-text transformer.
\newblock \emph{J. Mach. Learn. Res.}, 21:\penalty0 140:1--140:67, 2020{\natexlab{a}}.
\newblock URL \url{http://jmlr.org/papers/v21/20-074.html}.

\bibitem[Raffel et~al.(2020{\natexlab{b}})Raffel, Shazeer, Roberts, Lee, Narang, Matena, Zhou, Li, and Liu]{T5}
Colin Raffel, Noam Shazeer, Adam Roberts, Katherine Lee, Sharan Narang, Michael Matena, Yanqi Zhou, Wei Li, and Peter~J Liu.
\newblock Exploring the limits of transfer learning with a unified text-to-text transformer.
\newblock \emph{Journal of machine learning research}, 21\penalty0 (140):\penalty0 1--67, 2020{\natexlab{b}}.

\bibitem[Rajaee \& Pilehvar(2022)Rajaee and Pilehvar]{DBLP:conf/acl/RajaeeP22}
Sara Rajaee and Mohammad~Taher Pilehvar.
\newblock An isotropy analysis in the multilingual {BERT} embedding space.
\newblock In Smaranda Muresan, Preslav Nakov, and Aline Villavicencio (eds.), \emph{Findings of the Association for Computational Linguistics: {ACL} 2022, Dublin, Ireland, May 22-27, 2022}, pp.\  1309--1316. Association for Computational Linguistics, 2022.
\newblock \doi{10.18653/V1/2022.FINDINGS-ACL.103}.
\newblock URL \url{https://doi.org/10.18653/v1/2022.findings-acl.103}.

\bibitem[Ramezani \& Xu(2023)Ramezani and Xu]{c-12}
Aida Ramezani and Yang Xu.
\newblock Knowledge of cultural moral norms in large language models.
\newblock In Anna Rogers, Jordan~L. Boyd{-}Graber, and Naoaki Okazaki (eds.), \emph{Proceedings of the 61st Annual Meeting of the Association for Computational Linguistics (Volume 1: Long Papers), {ACL} 2023, Toronto, Canada, July 9-14, 2023}, pp.\  428--446. Association for Computational Linguistics, 2023.
\newblock \doi{10.18653/V1/2023.ACL-LONG.26}.
\newblock URL \url{https://doi.org/10.18653/v1/2023.acl-long.26}.

\bibitem[Ranaldi et~al.(2023)Ranaldi, Pucci, and Freitas]{trans-1}
Leonardo Ranaldi, Giulia Pucci, and Andr{\'{e}} Freitas.
\newblock Empowering cross-lingual abilities of instruction-tuned large language models by translation-following demonstrations.
\newblock \emph{CoRR}, abs/2308.14186, 2023.
\newblock \doi{10.48550/ARXIV.2308.14186}.
\newblock URL \url{https://doi.org/10.48550/arXiv.2308.14186}.

\bibitem[Rao et~al.(2023)Rao, Khandelwal, Tanmay, Agarwal, and Choudhury]{DBLP:conf/emnlp/RaoKTAC23}
Abhinav Rao, Aditi Khandelwal, Kumar Tanmay, Utkarsh Agarwal, and Monojit Choudhury.
\newblock Ethical reasoning over moral alignment: {A} case and framework for in-context ethical policies in llms.
\newblock In Houda Bouamor, Juan Pino, and Kalika Bali (eds.), \emph{Findings of the Association for Computational Linguistics: {EMNLP} 2023, Singapore, December 6-10, 2023}, pp.\  13370--13388. Association for Computational Linguistics, 2023.
\newblock \doi{10.18653/V1/2023.FINDINGS-EMNLP.892}.
\newblock URL \url{https://doi.org/10.18653/v1/2023.findings-emnlp.892}.

\bibitem[Rao et~al.(2024)Rao, Yerukola, Shah, Reinecke, and Sap]{c-26}
Abhinav Rao, Akhila Yerukola, Vishwa Shah, Katharina Reinecke, and Maarten Sap.
\newblock {NORMAD:} {A} benchmark for measuring the cultural adaptability of large language models.
\newblock \emph{CoRR}, abs/2404.12464, 2024.
\newblock \doi{10.48550/ARXIV.2404.12464}.
\newblock URL \url{https://doi.org/10.48550/arXiv.2404.12464}.

\bibitem[Razumovskaia et~al.(2024)Razumovskaia, Vuli{\'c}, and Korhonen]{razumovskaia2024analyzing}
Evgeniia Razumovskaia, Ivan Vuli{\'c}, and Anna Korhonen.
\newblock Analyzing and adapting large language models for few-shot multilingual nlu: Are we there yet?
\newblock \emph{arXiv preprint arXiv:2403.01929}, 2024.

\bibitem[Riemer et~al.(2019)Riemer, Cases, Ajemian, Liu, Rish, Tu, and Tesauro]{cpt_4}
Matthew Riemer, Ignacio Cases, Robert Ajemian, Miao Liu, Irina Rish, Yuhai Tu, and Gerald Tesauro.
\newblock Learning to learn without forgetting by maximizing transfer and minimizing interference.
\newblock In \emph{7th International Conference on Learning Representations, {ICLR} 2019, New Orleans, LA, USA, May 6-9, 2019}. OpenReview.net, 2019.
\newblock URL \url{https://openreview.net/forum?id=B1gTShAct7}.

\bibitem[Rozi{\`{e}}re et~al.(2023)Rozi{\`{e}}re, Gehring, Gloeckle, Sootla, Gat, Tan, Adi, Liu, Remez, Rapin, Kozhevnikov, Evtimov, Bitton, Bhatt, Canton{-}Ferrer, Grattafiori, Xiong, D{\'{e}}fossez, Copet, Azhar, Touvron, Martin, Usunier, Scialom, and Synnaeve]{DBLP:journals/corr/abs-2308-12950}
Baptiste Rozi{\`{e}}re, Jonas Gehring, Fabian Gloeckle, Sten Sootla, Itai Gat, Xiaoqing~Ellen Tan, Yossi Adi, Jingyu Liu, Tal Remez, J{\'{e}}r{\'{e}}my Rapin, Artyom Kozhevnikov, Ivan Evtimov, Joanna Bitton, Manish Bhatt, Cristian Canton{-}Ferrer, Aaron Grattafiori, Wenhan Xiong, Alexandre D{\'{e}}fossez, Jade Copet, Faisal Azhar, Hugo Touvron, Louis Martin, Nicolas Usunier, Thomas Scialom, and Gabriel Synnaeve.
\newblock Code llama: Open foundation models for code.
\newblock \emph{CoRR}, abs/2308.12950, 2023.
\newblock \doi{10.48550/ARXIV.2308.12950}.
\newblock URL \url{https://doi.org/10.48550/arXiv.2308.12950}.

\bibitem[Rust et~al.(2021)Rust, Pfeiffer, Vulic, Ruder, and Gurevych]{DBLP:conf/acl/RustPVRG20}
Phillip Rust, Jonas Pfeiffer, Ivan Vulic, Sebastian Ruder, and Iryna Gurevych.
\newblock How good is your tokenizer? on the monolingual performance of multilingual language models.
\newblock In Chengqing Zong, Fei Xia, Wenjie Li, and Roberto Navigli (eds.), \emph{Proceedings of the 59th Annual Meeting of the Association for Computational Linguistics and the 11th International Joint Conference on Natural Language Processing, {ACL/IJCNLP} 2021, (Volume 1: Long Papers), Virtual Event, August 1-6, 2021}, pp.\  3118--3135. Association for Computational Linguistics, 2021.
\newblock \doi{10.18653/V1/2021.ACL-LONG.243}.
\newblock URL \url{https://doi.org/10.18653/v1/2021.acl-long.243}.

\bibitem[Sanh et~al.(2022{\natexlab{a}})Sanh, Webson, Raffel, Bach, Sutawika, Alyafeai, Chaffin, Stiegler, Raja, Dey, Bari, Xu, Thakker, Sharma, Szczechla, Kim, Chhablani, Nayak, Datta, Chang, Jiang, Wang, Manica, Shen, Yong, Pandey, Bawden, Wang, Neeraj, Rozen, Sharma, Santilli, F{\'{e}}vry, Fries, Teehan, Scao, Biderman, Gao, Wolf, and Rush]{DBLP:conf/iclr/SanhWRBSACSRDBX22}
Victor Sanh, Albert Webson, Colin Raffel, Stephen~H. Bach, Lintang Sutawika, Zaid Alyafeai, Antoine Chaffin, Arnaud Stiegler, Arun Raja, Manan Dey, M~Saiful Bari, Canwen Xu, Urmish Thakker, Shanya~Sharma Sharma, Eliza Szczechla, Taewoon Kim, Gunjan Chhablani, Nihal~V. Nayak, Debajyoti Datta, Jonathan Chang, Mike~Tian{-}Jian Jiang, Han Wang, Matteo Manica, Sheng Shen, Zheng~Xin Yong, Harshit Pandey, Rachel Bawden, Thomas Wang, Trishala Neeraj, Jos Rozen, Abheesht Sharma, Andrea Santilli, Thibault F{\'{e}}vry, Jason~Alan Fries, Ryan Teehan, Teven~Le Scao, Stella Biderman, Leo Gao, Thomas Wolf, and Alexander~M. Rush.
\newblock Multitask prompted training enables zero-shot task generalization.
\newblock In \emph{The Tenth International Conference on Learning Representations, {ICLR} 2022, Virtual Event, April 25-29, 2022}. OpenReview.net, 2022{\natexlab{a}}.
\newblock URL \url{https://openreview.net/forum?id=9Vrb9D0WI4}.

\bibitem[Sanh et~al.(2022{\natexlab{b}})Sanh, Webson, Raffel, Bach, Sutawika, Alyafeai, Chaffin, Stiegler, Raja, Dey, Bari, Xu, Thakker, Sharma, Szczechla, Kim, Chhablani, Nayak, Datta, Chang, Jiang, Wang, Manica, Shen, Yong, Pandey, Bawden, Wang, Neeraj, Rozen, Sharma, Santilli, F{\'{e}}vry, Fries, Teehan, Scao, Biderman, Gao, Wolf, and Rush]{p3}
Victor Sanh, Albert Webson, Colin Raffel, Stephen~H. Bach, Lintang Sutawika, Zaid Alyafeai, Antoine Chaffin, Arnaud Stiegler, Arun Raja, Manan Dey, M~Saiful Bari, Canwen Xu, Urmish Thakker, Shanya~Sharma Sharma, Eliza Szczechla, Taewoon Kim, Gunjan Chhablani, Nihal~V. Nayak, Debajyoti Datta, Jonathan Chang, Mike~Tian{-}Jian Jiang, Han Wang, Matteo Manica, Sheng Shen, Zheng~Xin Yong, Harshit Pandey, Rachel Bawden, Thomas Wang, Trishala Neeraj, Jos Rozen, Abheesht Sharma, Andrea Santilli, Thibault F{\'{e}}vry, Jason~Alan Fries, Ryan Teehan, Teven~Le Scao, Stella Biderman, Leo Gao, Thomas Wolf, and Alexander~M. Rush.
\newblock Multitask prompted training enables zero-shot task generalization.
\newblock In \emph{The Tenth International Conference on Learning Representations, {ICLR} 2022, Virtual Event, April 25-29, 2022}. OpenReview.net, 2022{\natexlab{b}}.
\newblock URL \url{https://openreview.net/forum?id=9Vrb9D0WI4}.

\bibitem[Sarfraz et~al.(2023)Sarfraz, Arani, and Zonooz]{cpt_6}
Fahad Sarfraz, Elahe Arani, and Bahram Zonooz.
\newblock Error sensitivity modulation based experience replay: Mitigating abrupt representation drift in continual learning.
\newblock In \emph{The Eleventh International Conference on Learning Representations, {ICLR} 2023, Kigali, Rwanda, May 1-5, 2023}. OpenReview.net, 2023.
\newblock URL \url{https://openreview.net/pdf?id=zlbci7019Z3}.

\bibitem[Scao et~al.(2022)Scao, Fan, Akiki, Pavlick, Ilic, Hesslow, Castagn{\'{e}}, Luccioni, Yvon, Gall{\'{e}}, Tow, Rush, Biderman, Webson, Ammanamanchi, Wang, Sagot, Muennighoff, del Moral, Ruwase, Bawden, Bekman, McMillan{-}Major, Beltagy, Nguyen, Saulnier, Tan, Suarez, Sanh, Lauren{\c{c}}on, Jernite, Launay, Mitchell, Raffel, Gokaslan, Simhi, Soroa, Aji, Alfassy, Rogers, Nitzav, Xu, Mou, Emezue, Klamm, Leong, van Strien, Adelani, and et~al.]{DBLP:journals/corr/abs-2211-05100}
Teven~Le Scao, Angela Fan, Christopher Akiki, Ellie Pavlick, Suzana Ilic, Daniel Hesslow, Roman Castagn{\'{e}}, Alexandra~Sasha Luccioni, Fran{\c{c}}ois Yvon, Matthias Gall{\'{e}}, Jonathan Tow, Alexander~M. Rush, Stella Biderman, Albert Webson, Pawan~Sasanka Ammanamanchi, Thomas Wang, Beno{\^{\i}}t Sagot, Niklas Muennighoff, Albert~Villanova del Moral, Olatunji Ruwase, Rachel Bawden, Stas Bekman, Angelina McMillan{-}Major, Iz~Beltagy, Huu Nguyen, Lucile Saulnier, Samson Tan, Pedro~Ortiz Suarez, Victor Sanh, Hugo Lauren{\c{c}}on, Yacine Jernite, Julien Launay, Margaret Mitchell, Colin Raffel, Aaron Gokaslan, Adi Simhi, Aitor Soroa, Alham~Fikri Aji, Amit Alfassy, Anna Rogers, Ariel~Kreisberg Nitzav, Canwen Xu, Chenghao Mou, Chris Emezue, Christopher Klamm, Colin Leong, Daniel van Strien, David~Ifeoluwa Adelani, and et~al.
\newblock {BLOOM:} {A} 176b-parameter open-access multilingual language model.
\newblock \emph{CoRR}, abs/2211.05100, 2022.
\newblock \doi{10.48550/ARXIV.2211.05100}.
\newblock URL \url{https://doi.org/10.48550/arXiv.2211.05100}.

\bibitem[Scarlatos \& Lan(2023)Scarlatos and Lan]{scarlatos-lan-2023-tree}
Alexander Scarlatos and Andrew Lan.
\newblock Tree-based representation and generation of natural and mathematical language.
\newblock In \emph{Proceedings of the 61st Annual Meeting of the Association for Computational Linguistics (Volume 1: Long Papers)}, pp.\  3714--3730, Toronto, Canada, July 2023. Association for Computational Linguistics.
\newblock URL \url{https://aclanthology.org/2023.acl-long.205}.

\bibitem[Schulman et~al.(2017)Schulman, Wolski, Dhariwal, Radford, and Klimov]{DBLP:journals/corr/SchulmanWDRK17}
John Schulman, Filip Wolski, Prafulla Dhariwal, Alec Radford, and Oleg Klimov.
\newblock Proximal policy optimization algorithms.
\newblock \emph{CoRR}, abs/1707.06347, 2017.
\newblock URL \url{http://arxiv.org/abs/1707.06347}.

\bibitem[Seganti et~al.(2021)Seganti, Firlag, Skowronska, Satlawa, and Andruszkiewicz]{DBLP:conf/eacl/SegantiFSSA21}
Alessandro Seganti, Klaudia Firlag, Helena Skowronska, Michal Satlawa, and Piotr Andruszkiewicz.
\newblock Multilingual entity and relation extraction dataset and model.
\newblock In Paola Merlo, J{\"{o}}rg Tiedemann, and Reut Tsarfaty (eds.), \emph{Proceedings of the 16th Conference of the European Chapter of the Association for Computational Linguistics: Main Volume, {EACL} 2021, Online, April 19 - 23, 2021}, pp.\  1946--1955. Association for Computational Linguistics, 2021.
\newblock \doi{10.18653/V1/2021.EACL-MAIN.166}.
\newblock URL \url{https://doi.org/10.18653/v1/2021.eacl-main.166}.

\bibitem[Shaham et~al.(2024{\natexlab{a}})Shaham, Herzig, Aharoni, Szpektor, Tsarfaty, and Eyal]{shaham2024multilingual}
Uri Shaham, Jonathan Herzig, Roee Aharoni, Idan Szpektor, Reut Tsarfaty, and Matan Eyal.
\newblock Multilingual instruction tuning with just a pinch of multilinguality.
\newblock \emph{arXiv preprint arXiv:2401.01854}, 2024{\natexlab{a}}.

\bibitem[Shaham et~al.(2024{\natexlab{b}})Shaham, Herzig, Aharoni, Szpektor, Tsarfaty, and Eyal]{tuning-5}
Uri Shaham, Jonathan Herzig, Roee Aharoni, Idan Szpektor, Reut Tsarfaty, and Matan Eyal.
\newblock Multilingual instruction tuning with just a pinch of multilinguality.
\newblock \emph{CoRR}, abs/2401.01854, 2024{\natexlab{b}}.
\newblock \doi{10.48550/ARXIV.2401.01854}.
\newblock URL \url{https://doi.org/10.48550/arXiv.2401.01854}.

\bibitem[Shao et~al.(2024)Shao, Wang, Zhu, Xu, Song, Zhang, Li, Wu, and Guo]{DBLP:journals/corr/abs-2402-03300}
Zhihong Shao, Peiyi Wang, Qihao Zhu, Runxin Xu, Junxiao Song, Mingchuan Zhang, Y.~K. Li, Y.~Wu, and Daya Guo.
\newblock Deepseekmath: Pushing the limits of mathematical reasoning in open language models.
\newblock \emph{CoRR}, abs/2402.03300, 2024.
\newblock \doi{10.48550/ARXIV.2402.03300}.
\newblock URL \url{https://doi.org/10.48550/arXiv.2402.03300}.

\bibitem[Shazeer et~al.(2017)Shazeer, Mirhoseini, Maziarz, Davis, Le, Hinton, and Dean]{Outrageously}
Noam Shazeer, Azalia Mirhoseini, Krzysztof Maziarz, Andy Davis, Quoc Le, Geoffrey Hinton, and Jeff Dean.
\newblock Outrageously large neural networks: The sparsely-gated mixture-of-experts layer.
\newblock \emph{arXiv preprint arXiv:1701.06538}, 2017.

\bibitem[She et~al.(2024)She, Huang, Zou, Zhu, Liu, Geng, and Chen]{trans-10}
Shuaijie She, Shujian Huang, Wei Zou, Wenhao Zhu, Xiang Liu, Xiang Geng, and Jiajun Chen.
\newblock {MAPO:} advancing multilingual reasoning through multilingual alignment-as-preference optimization.
\newblock \emph{CoRR}, abs/2401.06838, 2024.
\newblock \doi{10.48550/ARXIV.2401.06838}.
\newblock URL \url{https://doi.org/10.48550/arXiv.2401.06838}.

\bibitem[Shen et~al.(2024{\natexlab{a}})Shen, Tan, Chen, Chen, Zhang, Xu, Zheng, Koehn, and Khashabi]{shen2024language}
Lingfeng Shen, Weiting Tan, Sihao Chen, Yunmo Chen, Jingyu Zhang, Haoran Xu, Boyuan Zheng, Philipp Koehn, and Daniel Khashabi.
\newblock The language barrier: Dissecting safety challenges of llms in multilingual contexts.
\newblock \emph{arXiv preprint arXiv:2401.13136}, 2024{\natexlab{a}}.

\bibitem[Shen et~al.(2024{\natexlab{b}})Shen, Tan, Chen, Chen, Zhang, Xu, Zheng, Koehn, and Khashabi]{tuning-12}
Lingfeng Shen, Weiting Tan, Sihao Chen, Yunmo Chen, Jingyu Zhang, Haoran Xu, Boyuan Zheng, Philipp Koehn, and Daniel Khashabi.
\newblock The language barrier: Dissecting safety challenges of llms in multilingual contexts.
\newblock \emph{CoRR}, abs/2401.13136, 2024{\natexlab{b}}.
\newblock \doi{10.48550/ARXIV.2401.13136}.
\newblock URL \url{https://doi.org/10.48550/arXiv.2401.13136}.

\bibitem[Shen et~al.(2024{\natexlab{c}})Shen, Logeswaran, Lee, Lee, Poria, and Mihalcea]{c-20}
Siqi Shen, Lajanugen Logeswaran, Moontae Lee, Honglak Lee, Soujanya Poria, and Rada Mihalcea.
\newblock Understanding the capabilities and limitations of large language models for cultural commonsense.
\newblock \emph{CoRR}, abs/2405.04655, 2024{\natexlab{c}}.
\newblock \doi{10.48550/ARXIV.2405.04655}.
\newblock URL \url{https://doi.org/10.48550/arXiv.2405.04655}.

\bibitem[Shen et~al.(2023)Shen, Tao, Ma, Neiswanger, Liu, Wang, Tan, Hestness, Vassilieva, Soboleva, et~al.]{cerebras2023slimpajama}
Zhiqiang Shen, Tianhua Tao, Liqun Ma, Willie Neiswanger, Zhengzhong Liu, Hongyi Wang, Bowen Tan, Joel Hestness, Natalia Vassilieva, Daria Soboleva, et~al.
\newblock Slimpajama-dc: Understanding data combinations for llm training.
\newblock \emph{arXiv preprint arXiv:2309.10818}, 2023.

\bibitem[Shi et~al.(2024{\natexlab{a}})Shi, Xu, Wang, Qin, Wang, Wang, and Wang]{cpt_1}
Haizhou Shi, Zihao Xu, Hengyi Wang, Weiyi Qin, Wenyuan Wang, Yibin Wang, and Hao Wang.
\newblock Continual learning of large language models: {A} comprehensive survey.
\newblock \emph{CoRR}, abs/2404.16789, 2024{\natexlab{a}}.
\newblock \doi{10.48550/ARXIV.2404.16789}.
\newblock URL \url{https://doi.org/10.48550/arXiv.2404.16789}.

\bibitem[Shi et~al.(2024{\natexlab{b}})Shi, Xu, Wang, Qin, Wang, Wang, and Wang]{cpt_2}
Haizhou Shi, Zihao Xu, Hengyi Wang, Weiyi Qin, Wenyuan Wang, Yibin Wang, and Hao Wang.
\newblock Continual learning of large language models: {A} comprehensive survey.
\newblock \emph{CoRR}, abs/2404.16789, 2024{\natexlab{b}}.
\newblock \doi{10.48550/ARXIV.2404.16789}.
\newblock URL \url{https://doi.org/10.48550/arXiv.2404.16789}.

\bibitem[Shi et~al.(2024{\natexlab{c}})Shi, Li, Zhang, Ziems, yu, Horesh, de~Paula, and Yang]{c-22}
Weiyan Shi, Ryan Li, Yutong Zhang, Caleb Ziems, Chunhua yu, Raya Horesh, Rog{\'{e}}rio~Abreu de~Paula, and Diyi Yang.
\newblock Culturebank: An online community-driven knowledge base towards culturally aware language technologies.
\newblock \emph{CoRR}, abs/2404.15238, 2024{\natexlab{c}}.
\newblock \doi{10.48550/ARXIV.2404.15238}.
\newblock URL \url{https://doi.org/10.48550/arXiv.2404.15238}.

\bibitem[Shi et~al.(2024{\natexlab{d}})Shi, Yang, Wu, Aitchison, Yilmaz, and Lipani]{shi2024instruction}
Zhengyan Shi, Adam~X. Yang, Bin Wu, Laurence Aitchison, Emine Yilmaz, and Aldo Lipani.
\newblock Instruction tuning with loss over instructions, 2024{\natexlab{d}}.

\bibitem[Shwartz(2022)]{c-11}
Vered Shwartz.
\newblock Good night at 4 pm?! time expressions in different cultures.
\newblock In Smaranda Muresan, Preslav Nakov, and Aline Villavicencio (eds.), \emph{Findings of the Association for Computational Linguistics: {ACL} 2022, Dublin, Ireland, May 22-27, 2022}, pp.\  2842--2853. Association for Computational Linguistics, 2022.
\newblock \doi{10.18653/V1/2022.FINDINGS-ACL.224}.
\newblock URL \url{https://doi.org/10.18653/v1/2022.findings-acl.224}.

\bibitem[Si et~al.(2024)Si, Zhang, and Zhang]{si2024mpn}
Nianwen Si, Hao Zhang, and Weiqiang Zhang.
\newblock Mpn: Leveraging multilingual patch neuron for cross-lingual model editing.
\newblock \emph{arXiv preprint arXiv:2401.03190}, 2024.

\bibitem[Singh et~al.(2024)Singh, Vargus, D'souza, Karlsson, Mahendiran, Ko, Shandilya, Patel, Mataciunas, O'Mahony, Zhang, Hettiarachchi, Wilson, Machado, Moura, Krzeminski, Fadaei, Erg{\"{u}}n, Okoh, Alaagib, Mudannayake, Alyafeai, Vu, Ruder, Guthikonda, Alghamdi, Gehrmann, Muennighoff, Bartolo, Kreutzer, {\"{U}}st{\"{u}}n, Fadaee, and Hooker]{DBLP:journals/corr/abs-2402-06619}
Shivalika Singh, Freddie Vargus, Daniel D'souza, B{\"{o}}rje~F. Karlsson, Abinaya Mahendiran, Wei{-}Yin Ko, Herumb Shandilya, Jay Patel, Deividas Mataciunas, Laura O'Mahony, Mike Zhang, Ramith Hettiarachchi, Joseph Wilson, Marina Machado, Luisa~Souza Moura, Dominik Krzeminski, Hakimeh Fadaei, Irem Erg{\"{u}}n, Ifeoma Okoh, Aisha Alaagib, Oshan Mudannayake, Zaid Alyafeai, Minh~Chien Vu, Sebastian Ruder, Surya Guthikonda, Emad~A. Alghamdi, Sebastian Gehrmann, Niklas Muennighoff, Max Bartolo, Julia Kreutzer, Ahmet {\"{U}}st{\"{u}}n, Marzieh Fadaee, and Sara Hooker.
\newblock Aya dataset: An open-access collection for multilingual instruction tuning.
\newblock \emph{CoRR}, abs/2402.06619, 2024.
\newblock \doi{10.48550/ARXIV.2402.06619}.
\newblock URL \url{https://doi.org/10.48550/arXiv.2402.06619}.

\bibitem[Singhal et~al.(2023)Singhal, Tu, Gottweis, Sayres, Wulczyn, Hou, Clark, Pfohl, Cole{-}Lewis, Neal, Schaekermann, Wang, Amin, Lachgar, Mansfield, Prakash, Green, Dominowska, y~Arcas, Tomasev, Liu, Wong, Semturs, Mahdavi, Barral, Webster, Corrado, Matias, Azizi, Karthikesalingam, and Natarajan]{DBLP:journals/corr/abs-2305-09617}
Karan Singhal, Tao Tu, Juraj Gottweis, Rory Sayres, Ellery Wulczyn, Le~Hou, Kevin Clark, Stephen Pfohl, Heather Cole{-}Lewis, Darlene Neal, Mike Schaekermann, Amy Wang, Mohamed Amin, Sami Lachgar, Philip~Andrew Mansfield, Sushant Prakash, Bradley Green, Ewa Dominowska, Blaise~Ag{\"{u}}era y~Arcas, Nenad Tomasev, Yun Liu, Renee Wong, Christopher Semturs, S.~Sara Mahdavi, Joelle~K. Barral, Dale~R. Webster, Gregory~S. Corrado, Yossi Matias, Shekoofeh Azizi, Alan Karthikesalingam, and Vivek Natarajan.
\newblock Towards expert-level medical question answering with large language models.
\newblock \emph{CoRR}, abs/2305.09617, 2023.
\newblock \doi{10.48550/ARXIV.2305.09617}.
\newblock URL \url{https://doi.org/10.48550/arXiv.2305.09617}.

\bibitem[Song et~al.(2024)Song, Yu, Li, Yu, Huang, Li, and Wang]{DBLP:conf/aaai/00010LYHLW24}
Feifan Song, Bowen Yu, Minghao Li, Haiyang Yu, Fei Huang, Yongbin Li, and Houfeng Wang.
\newblock Preference ranking optimization for human alignment.
\newblock In Michael~J. Wooldridge, Jennifer~G. Dy, and Sriraam Natarajan (eds.), \emph{Thirty-Eighth {AAAI} Conference on Artificial Intelligence, {AAAI} 2024, Thirty-Sixth Conference on Innovative Applications of Artificial Intelligence, {IAAI} 2024, Fourteenth Symposium on Educational Advances in Artificial Intelligence, {EAAI} 2014, February 20-27, 2024, Vancouver, Canada}, pp.\  18990--18998. {AAAI} Press, 2024.
\newblock \doi{10.1609/AAAI.V38I17.29865}.
\newblock URL \url{https://doi.org/10.1609/aaai.v38i17.29865}.

\bibitem[Sorensen et~al.(2024)Sorensen, Moore, Fisher, Gordon, Mireshghallah, Rytting, Ye, Jiang, Lu, Dziri, Althoff, and Choi]{c-1}
Taylor Sorensen, Jared Moore, Jillian Fisher, Mitchell~L. Gordon, Niloofar Mireshghallah, Christopher~Michael Rytting, Andre Ye, Liwei Jiang, Ximing Lu, Nouha Dziri, Tim Althoff, and Yejin Choi.
\newblock A roadmap to pluralistic alignment.
\newblock \emph{CoRR}, abs/2402.05070, 2024.
\newblock \doi{10.48550/ARXIV.2402.05070}.
\newblock URL \url{https://doi.org/10.48550/arXiv.2402.05070}.

\bibitem[Stanczak et~al.(2022)Stanczak, Ponti, Hennigen, Cotterell, and Augenstein]{DBLP:conf/naacl/StanczakPHCA22}
Karolina Stanczak, Edoardo~M. Ponti, Lucas~Torroba Hennigen, Ryan Cotterell, and Isabelle Augenstein.
\newblock Same neurons, different languages: Probing morphosyntax in multilingual pre-trained models.
\newblock In Marine Carpuat, Marie{-}Catherine de~Marneffe, and Iv{\'{a}}n Vladimir~Meza Ru{\'{\i}}z (eds.), \emph{Proceedings of the 2022 Conference of the North American Chapter of the Association for Computational Linguistics: Human Language Technologies, {NAACL} 2022, Seattle, WA, United States, July 10-15, 2022}, pp.\  1589--1598. Association for Computational Linguistics, 2022.
\newblock \doi{10.18653/V1/2022.NAACL-MAIN.114}.
\newblock URL \url{https://doi.org/10.18653/v1/2022.naacl-main.114}.

\bibitem[Stanovsky et~al.(2019)Stanovsky, Smith, and Zettlemoyer]{DBLP:conf/acl/StanovskySZ19}
Gabriel Stanovsky, Noah~A. Smith, and Luke Zettlemoyer.
\newblock Evaluating gender bias in machine translation.
\newblock In Anna Korhonen, David~R. Traum, and Llu{\'{\i}}s M{\`{a}}rquez (eds.), \emph{Proceedings of the 57th Conference of the Association for Computational Linguistics, {ACL} 2019, Florence, Italy, July 28- August 2, 2019, Volume 1: Long Papers}, pp.\  1679--1684. Association for Computational Linguistics, 2019.
\newblock \doi{10.18653/V1/P19-1164}.
\newblock URL \url{https://doi.org/10.18653/v1/p19-1164}.

\bibitem[Starace et~al.(2023)Starace, Papakostas, Choenni, Panagiotopoulos, Rosati, Leidinger, and Shutova]{DBLP:conf/emnlp/StaracePCPRLS23}
Giulio Starace, Konstantinos Papakostas, Rochelle Choenni, Apostolos Panagiotopoulos, Matteo Rosati, Alina Leidinger, and Ekaterina Shutova.
\newblock Probing llms for joint encoding of linguistic categories.
\newblock In Houda Bouamor, Juan Pino, and Kalika Bali (eds.), \emph{Findings of the Association for Computational Linguistics: {EMNLP} 2023, Singapore, December 6-10, 2023}, pp.\  7158--7179. Association for Computational Linguistics, 2023.
\newblock \doi{10.18653/V1/2023.FINDINGS-EMNLP.476}.
\newblock URL \url{https://doi.org/10.18653/v1/2023.findings-emnlp.476}.

\bibitem[Stiennon et~al.(2020)Stiennon, Ouyang, Wu, Ziegler, Lowe, Voss, Radford, Amodei, and Christiano]{DBLP:conf/nips/StiennonO0ZLVRA20}
Nisan Stiennon, Long Ouyang, Jeffrey Wu, Daniel~M. Ziegler, Ryan Lowe, Chelsea Voss, Alec Radford, Dario Amodei, and Paul~F. Christiano.
\newblock Learning to summarize with human feedback.
\newblock In Hugo Larochelle, Marc'Aurelio Ranzato, Raia Hadsell, Maria{-}Florina Balcan, and Hsuan{-}Tien Lin (eds.), \emph{Advances in Neural Information Processing Systems 33: Annual Conference on Neural Information Processing Systems 2020, NeurIPS 2020, December 6-12, 2020, virtual}, 2020.
\newblock URL \url{https://proceedings.neurips.cc/paper/2020/hash/1f89885d556929e98d3ef9b86448f951-Abstract.html}.

\bibitem[Su et~al.(2024)Su, Ahmed, Lu, Pan, Bo, and Liu]{rope}
Jianlin Su, Murtadha Ahmed, Yu~Lu, Shengfeng Pan, Wen Bo, and Yunfeng Liu.
\newblock Roformer: Enhanced transformer with rotary position embedding.
\newblock \emph{Neurocomputing}, 568:\penalty0 127063, 2024.

\bibitem[Sun et~al.(2023)Sun, Zhang, Deng, Cheng, and Huang]{DBLP:journals/corr/abs-2304-10436}
Hao Sun, Zhexin Zhang, Jiawen Deng, Jiale Cheng, and Minlie Huang.
\newblock Safety assessment of chinese large language models.
\newblock \emph{CoRR}, abs/2304.10436, 2023.
\newblock \doi{10.48550/ARXIV.2304.10436}.
\newblock URL \url{https://doi.org/10.48550/arXiv.2304.10436}.

\bibitem[Sun et~al.(2024)Sun, Jin, Xu, Pan, Cui, Dui, Lei, Yang, Shi, Xiao, et~al.]{fuxitranyu}
Haoran Sun, Renren Jin, Shaoyang Xu, Leiyu Pan, Menglong Cui, Jiangcun Dui, Yikun Lei, Lei Yang, Ling Shi, Juesi Xiao, et~al.
\newblock Fuxitranyu: A multilingual large language model trained with balanced data.
\newblock \emph{arXiv preprint arXiv:2408.06273}, 2024.

\bibitem[Sun et~al.(2021)Sun, Wang, Feng, Ding, Pang, Shang, Liu, Chen, Zhao, Lu, et~al.]{erine}
Yu~Sun, Shuohuan Wang, Shikun Feng, Siyu Ding, Chao Pang, Junyuan Shang, Jiaxiang Liu, Xuyi Chen, Yanbin Zhao, Yuxiang Lu, et~al.
\newblock Ernie 3.0: Large-scale knowledge enhanced pre-training for language understanding and generation.
\newblock \emph{arXiv preprint arXiv:2107.02137}, 2021.

\bibitem[Survey(2022)]{wsv}
World~Values Survey.
\newblock World values survey.
\newblock 2022.
\newblock URL \url{https://www.worldvaluessurvey.org/wvs.jsp}.

\bibitem[Talmor et~al.(2019)Talmor, Herzig, Lourie, and Berant]{DBLP:conf/naacl/TalmorHLB19}
Alon Talmor, Jonathan Herzig, Nicholas Lourie, and Jonathan Berant.
\newblock Commonsenseqa: {A} question answering challenge targeting commonsense knowledge.
\newblock In Jill Burstein, Christy Doran, and Thamar Solorio (eds.), \emph{Proceedings of the 2019 Conference of the North American Chapter of the Association for Computational Linguistics: Human Language Technologies, {NAACL-HLT} 2019, Minneapolis, MN, USA, June 2-7, 2019, Volume 1 (Long and Short Papers)}, pp.\  4149--4158. Association for Computational Linguistics, 2019.
\newblock \doi{10.18653/V1/N19-1421}.
\newblock URL \url{https://doi.org/10.18653/v1/n19-1421}.

\bibitem[Tang et~al.(2024)Tang, Luo, Huang, Zhang, Wang, Zhao, Wei, and Wen]{DBLP:journals/corr/abs-2402-16438}
Tianyi Tang, Wenyang Luo, Haoyang Huang, Dongdong Zhang, Xiaolei Wang, Xin Zhao, Furu Wei, and Ji{-}Rong Wen.
\newblock Language-specific neurons: The key to multilingual capabilities in large language models.
\newblock \emph{CoRR}, abs/2402.16438, 2024.
\newblock \doi{10.48550/ARXIV.2402.16438}.
\newblock URL \url{https://doi.org/10.48550/arXiv.2402.16438}.

\bibitem[Taori et~al.(2023)Taori, Gulrajani, Zhang, Dubois, Li, Guestrin, Liang, and Hashimoto]{alpaca}
Rohan Taori, Ishaan Gulrajani, Tianyi Zhang, Yann Dubois, Xuechen Li, Carlos Guestrin, Percy Liang, and Tatsunori~B. Hashimoto.
\newblock Stanford alpaca: An instruction-following llama model.
\newblock 2023.
\newblock URL \url{https://github.com/tatsu-lab/stanford_alpaca}.

\bibitem[Tars et~al.(2022)Tars, T{\"{a}}ttar, and Fishel]{DBLP:journals/bjmc/TarsTF22}
Maali Tars, Andre T{\"{a}}ttar, and Mark Fishel.
\newblock Cross-lingual transfer from large multilingual translation models to unseen under-resourced languages.
\newblock \emph{Balt. J. Mod. Comput.}, 10\penalty0 (3), 2022.
\newblock \doi{10.22364/BJMC.2022.10.3.16}.
\newblock URL \url{https://doi.org/10.22364/bjmc.2022.10.3.16}.

\bibitem[Tay et~al.(2022)Tay, Dehghani, Tran, Garcia, Wei, Wang, Chung, Shakeri, Bahri, Schuster, et~al.]{UL2}
Yi~Tay, Mostafa Dehghani, Vinh~Q Tran, Xavier Garcia, Jason Wei, Xuezhi Wang, Hyung~Won Chung, Siamak Shakeri, Dara Bahri, Tal Schuster, et~al.
\newblock Ul2: Unifying language learning paradigms.
\newblock \emph{arXiv preprint arXiv:2205.05131}, 2022.

\bibitem[Team et~al.(2024)Team, Mesnard, Hardin, Dadashi, Bhupatiraju, Pathak, Sifre, Rivi{\`e}re, Kale, Love, et~al.]{Gemma}
Gemma Team, Thomas Mesnard, Cassidy Hardin, Robert Dadashi, Surya Bhupatiraju, Shreya Pathak, Laurent Sifre, Morgane Rivi{\`e}re, Mihir~Sanjay Kale, Juliette Love, et~al.
\newblock Gemma: Open models based on gemini research and technology.
\newblock \emph{arXiv preprint arXiv:2403.08295}, 2024.

\bibitem[Team(2023)]{internlm}
InternLM Team.
\newblock Internlm: A multilingual language model with progressively enhanced capabilities, 2023.

\bibitem[Team(2024{\natexlab{a}})]{qwen_moe}
Qwen Team.
\newblock Qwen1.5-moe: Matching 7b model performance with 1/3 activated parameters", February 2024{\natexlab{a}}.
\newblock URL \url{https://qwenlm.github.io/blog/qwen-moe/}.

\bibitem[Team(2024{\natexlab{b}})]{DBRX}
The Mosaic~Research Team.
\newblock Dbrx: A new state-of-the-art open llm, March 2024{\natexlab{b}}.
\newblock URL \url{https://www.databricks.com/blog/introducing-dbrx-new-state-art-open-llm}.

\bibitem[Thapliyal et~al.(2022)Thapliyal, Pont{-}Tuset, Chen, and Soricut]{DBLP:conf/emnlp/ThapliyalPCS22}
Ashish~V. Thapliyal, Jordi Pont{-}Tuset, Xi~Chen, and Radu Soricut.
\newblock Crossmodal-3600: {A} massively multilingual multimodal evaluation dataset.
\newblock In Yoav Goldberg, Zornitsa Kozareva, and Yue Zhang (eds.), \emph{Proceedings of the 2022 Conference on Empirical Methods in Natural Language Processing, {EMNLP} 2022, Abu Dhabi, United Arab Emirates, December 7-11, 2022}, pp.\  715--729. Association for Computational Linguistics, 2022.
\newblock \doi{10.18653/V1/2022.EMNLP-MAIN.45}.
\newblock URL \url{https://doi.org/10.18653/v1/2022.emnlp-main.45}.

\bibitem[Tiyajamorn et~al.(2021)Tiyajamorn, Kajiwara, Arase, and Onizuka]{DBLP:conf/emnlp/TiyajamornKAO21}
Nattapong Tiyajamorn, Tomoyuki Kajiwara, Yuki Arase, and Makoto Onizuka.
\newblock Language-agnostic representation from multilingual sentence encoders for cross-lingual similarity estimation.
\newblock In Marie{-}Francine Moens, Xuanjing Huang, Lucia Specia, and Scott~Wen{-}tau Yih (eds.), \emph{Proceedings of the 2021 Conference on Empirical Methods in Natural Language Processing, {EMNLP} 2021, Virtual Event / Punta Cana, Dominican Republic, 7-11 November, 2021}, pp.\  7764--7774. Association for Computational Linguistics, 2021.
\newblock \doi{10.18653/V1/2021.EMNLP-MAIN.612}.
\newblock URL \url{https://doi.org/10.18653/v1/2021.emnlp-main.612}.

\bibitem[Tokpanov et~al.(2024)Tokpanov, Millidge, Glorioso, Pilault, Ibrahim, Whittington, and Anthony]{tokpanov2024zyda}
Yury Tokpanov, Beren Millidge, Paolo Glorioso, Jonathan Pilault, Adam Ibrahim, James Whittington, and Quentin Anthony.
\newblock Zyda: A 1.3t dataset for open language modeling, 2024.

\bibitem[Touvron et~al.(2023{\natexlab{a}})Touvron, Lavril, Izacard, Martinet, Lachaux, Lacroix, Rozi{\`{e}}re, Goyal, Hambro, Azhar, Rodriguez, Joulin, Grave, and Lample]{DBLP:journals/corr/abs-2302-13971}
Hugo Touvron, Thibaut Lavril, Gautier Izacard, Xavier Martinet, Marie{-}Anne Lachaux, Timoth{\'{e}}e Lacroix, Baptiste Rozi{\`{e}}re, Naman Goyal, Eric Hambro, Faisal Azhar, Aur{\'{e}}lien Rodriguez, Armand Joulin, Edouard Grave, and Guillaume Lample.
\newblock Llama: Open and efficient foundation language models.
\newblock \emph{CoRR}, abs/2302.13971, 2023{\natexlab{a}}.
\newblock \doi{10.48550/ARXIV.2302.13971}.
\newblock URL \url{https://doi.org/10.48550/arXiv.2302.13971}.

\bibitem[Touvron et~al.(2023{\natexlab{b}})Touvron, Lavril, Izacard, Martinet, Lachaux, Lacroix, Rozi{\`e}re, Goyal, Hambro, Azhar, et~al.]{llama-1}
Hugo Touvron, Thibaut Lavril, Gautier Izacard, Xavier Martinet, Marie-Anne Lachaux, Timoth{\'e}e Lacroix, Baptiste Rozi{\`e}re, Naman Goyal, Eric Hambro, Faisal Azhar, et~al.
\newblock Llama: Open and efficient foundation language models.
\newblock \emph{arXiv preprint arXiv:2302.13971}, 2023{\natexlab{b}}.

\bibitem[Touvron et~al.(2023{\natexlab{c}})Touvron, Martin, Stone, Albert, Almahairi, Babaei, Bashlykov, Batra, Bhargava, Bhosale, Bikel, Blecher, Canton{-}Ferrer, Chen, Cucurull, Esiobu, Fernandes, Fu, Fu, Fuller, Gao, Goswami, Goyal, Hartshorn, Hosseini, Hou, Inan, Kardas, Kerkez, Khabsa, Kloumann, Korenev, Koura, Lachaux, Lavril, Lee, Liskovich, Lu, Mao, Martinet, Mihaylov, Mishra, Molybog, Nie, Poulton, Reizenstein, Rungta, Saladi, Schelten, Silva, Smith, Subramanian, Tan, Tang, Taylor, Williams, Kuan, Xu, Yan, Zarov, Zhang, Fan, Kambadur, Narang, Rodriguez, Stojnic, Edunov, and Scialom]{DBLP:journals/corr/abs-2307-09288}
Hugo Touvron, Louis Martin, Kevin Stone, Peter Albert, Amjad Almahairi, Yasmine Babaei, Nikolay Bashlykov, Soumya Batra, Prajjwal Bhargava, Shruti Bhosale, Dan Bikel, Lukas Blecher, Cristian Canton{-}Ferrer, Moya Chen, Guillem Cucurull, David Esiobu, Jude Fernandes, Jeremy Fu, Wenyin Fu, Brian Fuller, Cynthia Gao, Vedanuj Goswami, Naman Goyal, Anthony Hartshorn, Saghar Hosseini, Rui Hou, Hakan Inan, Marcin Kardas, Viktor Kerkez, Madian Khabsa, Isabel Kloumann, Artem Korenev, Punit~Singh Koura, Marie{-}Anne Lachaux, Thibaut Lavril, Jenya Lee, Diana Liskovich, Yinghai Lu, Yuning Mao, Xavier Martinet, Todor Mihaylov, Pushkar Mishra, Igor Molybog, Yixin Nie, Andrew Poulton, Jeremy Reizenstein, Rashi Rungta, Kalyan Saladi, Alan Schelten, Ruan Silva, Eric~Michael Smith, Ranjan Subramanian, Xiaoqing~Ellen Tan, Binh Tang, Ross Taylor, Adina Williams, Jian~Xiang Kuan, Puxin Xu, Zheng Yan, Iliyan Zarov, Yuchen Zhang, Angela Fan, Melanie Kambadur, Sharan Narang, Aur{\'{e}}lien Rodriguez, Robert Stojnic, Sergey Edunov,
  and Thomas Scialom.
\newblock Llama 2: Open foundation and fine-tuned chat models.
\newblock \emph{CoRR}, abs/2307.09288, 2023{\natexlab{c}}.
\newblock \doi{10.48550/ARXIV.2307.09288}.
\newblock URL \url{https://doi.org/10.48550/arXiv.2307.09288}.

\bibitem[Touvron et~al.(2023{\natexlab{d}})Touvron, Martin, Stone, Albert, Almahairi, Babaei, Bashlykov, Batra, Bhargava, Bhosale, et~al.]{llama-2}
Hugo Touvron, Louis Martin, Kevin Stone, Peter Albert, Amjad Almahairi, Yasmine Babaei, Nikolay Bashlykov, Soumya Batra, Prajjwal Bhargava, Shruti Bhosale, et~al.
\newblock Llama 2: Open foundation and fine-tuned chat models.
\newblock \emph{arXiv preprint arXiv:2307.09288}, 2023{\natexlab{d}}.

\bibitem[Upadhayay \& Behzadan(2023)Upadhayay and Behzadan]{trans-9}
Bibek Upadhayay and Vahid Behzadan.
\newblock Taco: Enhancing cross-lingual transfer for low-resource languages in llms through translation-assisted chain-of-thought processes.
\newblock \emph{CoRR}, abs/2311.10797, 2023.
\newblock \doi{10.48550/ARXIV.2311.10797}.
\newblock URL \url{https://doi.org/10.48550/arXiv.2311.10797}.

\bibitem[{\"{U}}st{\"{u}}n et~al.(2024){\"{U}}st{\"{u}}n, Aryabumi, Yong, Ko, D'souza, Onilude, Bhandari, Singh, Ooi, Kayid, Vargus, Blunsom, Longpre, Muennighoff, Fadaee, Kreutzer, and Hooker]{tuning-14}
Ahmet {\"{U}}st{\"{u}}n, Viraat Aryabumi, Zheng~Xin Yong, Wei{-}Yin Ko, Daniel D'souza, Gbemileke Onilude, Neel Bhandari, Shivalika Singh, Hui{-}Lee Ooi, Amr Kayid, Freddie Vargus, Phil Blunsom, Shayne Longpre, Niklas Muennighoff, Marzieh Fadaee, Julia Kreutzer, and Sara Hooker.
\newblock Aya model: An instruction finetuned open-access multilingual language model.
\newblock \emph{CoRR}, abs/2402.07827, 2024.
\newblock \doi{10.48550/ARXIV.2402.07827}.
\newblock URL \url{https://doi.org/10.48550/arXiv.2402.07827}.

\bibitem[Vashishtha et~al.(2023)Vashishtha, Ahuja, and Sitaram]{DBLP:conf/acl/VashishthaAS23}
Aniket Vashishtha, Kabir Ahuja, and Sunayana Sitaram.
\newblock On evaluating and mitigating gender biases in multilingual settings.
\newblock In Anna Rogers, Jordan~L. Boyd{-}Graber, and Naoaki Okazaki (eds.), \emph{Findings of the Association for Computational Linguistics: {ACL} 2023, Toronto, Canada, July 9-14, 2023}, pp.\  307--318. Association for Computational Linguistics, 2023.
\newblock \doi{10.18653/V1/2023.FINDINGS-ACL.21}.
\newblock URL \url{https://doi.org/10.18653/v1/2023.findings-acl.21}.

\bibitem[Vaswani et~al.(2017)Vaswani, Shazeer, Parmar, Uszkoreit, Jones, Gomez, Kaiser, and Polosukhin]{vaswani2017attention}
Ashish Vaswani, Noam Shazeer, Niki Parmar, Jakob Uszkoreit, Llion Jones, Aidan~N Gomez, {\L}ukasz Kaiser, and Illia Polosukhin.
\newblock Attention is all you need.
\newblock \emph{Advances in neural information processing systems}, 30, 2017.

\bibitem[Vilares et~al.(2016)Vilares, Alonso, and G{\'{o}}mez{-}Rodr{\'{\i}}guez]{DBLP:conf/lrec/VilaresAG16}
David Vilares, Miguel~A. Alonso, and Carlos G{\'{o}}mez{-}Rodr{\'{\i}}guez.
\newblock {EN-ES-CS:} an english-spanish code-switching twitter corpus for multilingual sentiment analysis.
\newblock In Nicoletta Calzolari, Khalid Choukri, Thierry Declerck, Sara Goggi, Marko Grobelnik, Bente Maegaard, Joseph Mariani, H{\'{e}}l{\`{e}}ne Mazo, Asunci{\'{o}}n Moreno, Jan Odijk, and Stelios Piperidis (eds.), \emph{Proceedings of the Tenth International Conference on Language Resources and Evaluation {LREC} 2016, Portoro{\v{z}}, Slovenia, May 23-28, 2016}. European Language Resources Association {(ELRA)}, 2016.
\newblock URL \url{http://www.lrec-conf.org/proceedings/lrec2016/summaries/43.html}.

\bibitem[Vulic et~al.(2023)Vulic, Glavas, Liu, Collier, Ponti, and Korhonen]{DBLP:conf/eacl/VulicGLCPK23}
Ivan Vulic, Goran Glavas, Fangyu Liu, Nigel Collier, Edoardo~Maria Ponti, and Anna Korhonen.
\newblock Probing cross-lingual lexical knowledge from multilingual sentence encoders.
\newblock In Andreas Vlachos and Isabelle Augenstein (eds.), \emph{Proceedings of the 17th Conference of the European Chapter of the Association for Computational Linguistics, {EACL} 2023, Dubrovnik, Croatia, May 2-6, 2023}, pp.\  2081--2097. Association for Computational Linguistics, 2023.
\newblock \doi{10.18653/V1/2023.EACL-MAIN.153}.
\newblock URL \url{https://doi.org/10.18653/v1/2023.eacl-main.153}.

\bibitem[Wang et~al.(2024{\natexlab{a}})Wang, Lin, Liu, Wei, and Chen]{c-37}
Bin Wang, Geyu Lin, Zhengyuan Liu, Chengwei Wei, and Nancy~F. Chen.
\newblock Craft: Extracting and tuning cultural instructions from the wild.
\newblock \emph{CoRR}, abs/2405.03138, 2024{\natexlab{a}}.
\newblock \doi{10.48550/ARXIV.2405.03138}.
\newblock URL \url{https://doi.org/10.48550/arXiv.2405.03138}.

\bibitem[Wang et~al.(2023{\natexlab{a}})Wang, Yang, Du, Fan, and Li]{DBLP:journals/corr/abs-2306-09968}
Guangyu Wang, Guoxing Yang, Zongxin Du, Longjun Fan, and Xiaohu Li.
\newblock Clinicalgpt: Large language models finetuned with diverse medical data and comprehensive evaluation.
\newblock \emph{CoRR}, abs/2306.09968, 2023{\natexlab{a}}.
\newblock \doi{10.48550/ARXIV.2306.09968}.
\newblock URL \url{https://doi.org/10.48550/arXiv.2306.09968}.

\bibitem[Wang et~al.(2024{\natexlab{b}})Wang, Minervini, and Ponti]{wang2024probing}
Hetong Wang, Pasquale Minervini, and Edoardo~M Ponti.
\newblock Probing the emergence of cross-lingual alignment during llm training.
\newblock \emph{arXiv preprint arXiv:2406.13229}, 2024{\natexlab{b}}.

\bibitem[Wang et~al.(2023{\natexlab{b}})Wang, Duan, Lam, Chen, Xu, Chen, Liu, Pang, and Tan]{DBLP:conf/cicba/WangDLCXCLPT23}
Rongsheng Wang, Yaofei Duan, Chan{-}Tong Lam, Jiexi Chen, Jiangsheng Xu, Haoming Chen, Xiaohong Liu, Patrick~Cheong{-}Iao Pang, and Tao Tan.
\newblock Ivygpt: Interactive chinese pathway language model in medical domain.
\newblock In Lu~Fang, Jian Pei, Guangtao Zhai, and Ruiping Wang (eds.), \emph{Artificial Intelligence - Third {CAAI} International Conference, {CICAI} 2023, Fuzhou, China, July 22-23, 2023, Revised Selected Papers, Part {II}}, volume 14474 of \emph{Lecture Notes in Computer Science}, pp.\  378--382. Springer, 2023{\natexlab{b}}.
\newblock \doi{10.1007/978-981-99-9119-8\_34}.
\newblock URL \url{https://doi.org/10.1007/978-981-99-9119-8\_34}.

\bibitem[Wang et~al.(2023{\natexlab{c}})Wang, Jiao, Huang, Dai, Huang, Tu, and Lyu]{c-13}
Wenxuan Wang, Wenxiang Jiao, Jingyuan Huang, Ruyi Dai, Jen{-}tse Huang, Zhaopeng Tu, and Michael~R. Lyu.
\newblock Not all countries celebrate thanksgiving: On the cultural dominance in large language models.
\newblock \emph{CoRR}, abs/2310.12481, 2023{\natexlab{c}}.
\newblock \doi{10.48550/ARXIV.2310.12481}.
\newblock URL \url{https://doi.org/10.48550/arXiv.2310.12481}.

\bibitem[Wang et~al.(2023{\natexlab{d}})Wang, Tu, Chen, Yuan, Huang, Jiao, and Lyu]{DBLP:journals/corr/abs-2310-00905}
Wenxuan Wang, Zhaopeng Tu, Chang Chen, Youliang Yuan, Jen{-}tse Huang, Wenxiang Jiao, and Michael~R. Lyu.
\newblock All languages matter: On the multilingual safety of large language models.
\newblock \emph{CoRR}, abs/2310.00905, 2023{\natexlab{d}}.
\newblock \doi{10.48550/ARXIV.2310.00905}.
\newblock URL \url{https://doi.org/10.48550/arXiv.2310.00905}.

\bibitem[Wang et~al.(2023{\natexlab{e}})Wang, Chen, Ge, Xia, Bao, Zheng, Zhang, Gui, and Huang]{cpt_16}
Xiao Wang, Tianze Chen, Qiming Ge, Han Xia, Rong Bao, Rui Zheng, Qi~Zhang, Tao Gui, and Xuanjing Huang.
\newblock Orthogonal subspace learning for language model continual learning.
\newblock In Houda Bouamor, Juan Pino, and Kalika Bali (eds.), \emph{Findings of the Association for Computational Linguistics: {EMNLP} 2023, Singapore, December 6-10, 2023}, pp.\  10658--10671. Association for Computational Linguistics, 2023{\natexlab{e}}.
\newblock \doi{10.18653/V1/2023.FINDINGS-EMNLP.715}.
\newblock URL \url{https://doi.org/10.18653/v1/2023.findings-emnlp.715}.

\bibitem[Wang et~al.(2022)Wang, Mishra, Alipoormolabashi, Kordi, Mirzaei, Naik, Ashok, Dhanasekaran, Arunkumar, Stap, Pathak, Karamanolakis, Lai, Purohit, Mondal, Anderson, Kuznia, Doshi, Pal, Patel, Moradshahi, Parmar, Purohit, Varshney, Kaza, Verma, Puri, Karia, Doshi, Sampat, Mishra, A, Patro, Dixit, and Shen]{DBLP:conf/emnlp/WangMAKMNADASPK22}
Yizhong Wang, Swaroop Mishra, Pegah Alipoormolabashi, Yeganeh Kordi, Amirreza Mirzaei, Atharva Naik, Arjun Ashok, Arut~Selvan Dhanasekaran, Anjana Arunkumar, David Stap, Eshaan Pathak, Giannis Karamanolakis, Haizhi~Gary Lai, Ishan Purohit, Ishani Mondal, Jacob Anderson, Kirby Kuznia, Krima Doshi, Kuntal~Kumar Pal, Maitreya Patel, Mehrad Moradshahi, Mihir Parmar, Mirali Purohit, Neeraj Varshney, Phani~Rohitha Kaza, Pulkit Verma, Ravsehaj~Singh Puri, Rushang Karia, Savan Doshi, Shailaja~Keyur Sampat, Siddhartha Mishra, Sujan~Reddy A, Sumanta Patro, Tanay Dixit, and Xudong Shen.
\newblock Super-naturalinstructions: Generalization via declarative instructions on 1600+ {NLP} tasks.
\newblock In Yoav Goldberg, Zornitsa Kozareva, and Yue Zhang (eds.), \emph{Proceedings of the 2022 Conference on Empirical Methods in Natural Language Processing, {EMNLP} 2022, Abu Dhabi, United Arab Emirates, December 7-11, 2022}, pp.\  5085--5109. Association for Computational Linguistics, 2022.
\newblock \doi{10.18653/V1/2022.EMNLP-MAIN.340}.
\newblock URL \url{https://doi.org/10.18653/v1/2022.emnlp-main.340}.

\bibitem[Wang et~al.(2023{\natexlab{f}})Wang, Kordi, Mishra, Liu, Smith, Khashabi, and Hajishirzi]{DBLP:conf/acl/WangKMLSKH23}
Yizhong Wang, Yeganeh Kordi, Swaroop Mishra, Alisa Liu, Noah~A. Smith, Daniel Khashabi, and Hannaneh Hajishirzi.
\newblock Self-instruct: Aligning language models with self-generated instructions.
\newblock In Anna Rogers, Jordan~L. Boyd{-}Graber, and Naoaki Okazaki (eds.), \emph{Proceedings of the 61st Annual Meeting of the Association for Computational Linguistics (Volume 1: Long Papers), {ACL} 2023, Toronto, Canada, July 9-14, 2023}, pp.\  13484--13508. Association for Computational Linguistics, 2023{\natexlab{f}}.
\newblock \doi{10.18653/V1/2023.ACL-LONG.754}.
\newblock URL \url{https://doi.org/10.18653/v1/2023.acl-long.754}.

\bibitem[Wang et~al.(2021)Wang, Wang, Joty, and Hoi]{DBLP:conf/emnlp/0034WJH21}
Yue Wang, Weishi Wang, Shafiq~R. Joty, and Steven C.~H. Hoi.
\newblock Codet5: Identifier-aware unified pre-trained encoder-decoder models for code understanding and generation.
\newblock In Marie{-}Francine Moens, Xuanjing Huang, Lucia Specia, and Scott~Wen{-}tau Yih (eds.), \emph{Proceedings of the 2021 Conference on Empirical Methods in Natural Language Processing, {EMNLP} 2021, Virtual Event / Punta Cana, Dominican Republic, 7-11 November, 2021}, pp.\  8696--8708. Association for Computational Linguistics, 2021.
\newblock \doi{10.18653/V1/2021.EMNLP-MAIN.685}.
\newblock URL \url{https://doi.org/10.18653/v1/2021.emnlp-main.685}.

\bibitem[Wang et~al.(2023{\natexlab{g}})Wang, Le, Gotmare, Bui, Li, and Hoi]{DBLP:conf/emnlp/WangLGB0H23}
Yue Wang, Hung Le, Akhilesh Gotmare, Nghi D.~Q. Bui, Junnan Li, and Steven C.~H. Hoi.
\newblock Codet5+: Open code large language models for code understanding and generation.
\newblock In Houda Bouamor, Juan Pino, and Kalika Bali (eds.), \emph{Proceedings of the 2023 Conference on Empirical Methods in Natural Language Processing, {EMNLP} 2023, Singapore, December 6-10, 2023}, pp.\  1069--1088. Association for Computational Linguistics, 2023{\natexlab{g}}.
\newblock \doi{10.18653/V1/2023.EMNLP-MAIN.68}.
\newblock URL \url{https://doi.org/10.18653/v1/2023.emnlp-main.68}.

\bibitem[Weber et~al.(2024)Weber, Thellmann, Ebert, Flores{-}Herr, Lehmann, Fromm, and Ali]{tuning-15}
Alexander~Arno Weber, Klaudia Thellmann, Jan Ebert, Nicolas Flores{-}Herr, Jens Lehmann, Michael Fromm, and Mehdi Ali.
\newblock Investigating multilingual instruction-tuning: Do polyglot models demand for multilingual instructions?
\newblock \emph{CoRR}, abs/2402.13703, 2024.
\newblock \doi{10.48550/ARXIV.2402.13703}.
\newblock URL \url{https://doi.org/10.48550/arXiv.2402.13703}.

\bibitem[Webster et~al.(2020)Webster, Wang, Tenney, Beutel, Pitler, Pavlick, Chen, and Petrov]{DBLP:journals/corr/abs-2010-06032}
Kellie Webster, Xuezhi Wang, Ian Tenney, Alex Beutel, Emily Pitler, Ellie Pavlick, Jilin Chen, and Slav Petrov.
\newblock Measuring and reducing gendered correlations in pre-trained models.
\newblock \emph{CoRR}, abs/2010.06032, 2020.
\newblock URL \url{https://arxiv.org/abs/2010.06032}.

\bibitem[Wei et~al.()Wei, Bosma, Zhao, Guu, Yu, Lester, Du, Dai, and Le]{flan2021}
Jason Wei, Maarten Bosma, Vincent Zhao, Kelvin Guu, Adams~Wei Yu, Brian Lester, Nan Du, Andrew~M Dai, and Quoc~V Le.
\newblock Finetuned language models are zero-shot learners.
\newblock In \emph{International Conference on Learning Representations}.

\bibitem[Wei et~al.(2022)Wei, Bosma, Zhao, Guu, Yu, Lester, Du, Dai, and Le]{DBLP:conf/iclr/WeiBZGYLDDL22}
Jason Wei, Maarten Bosma, Vincent~Y. Zhao, Kelvin Guu, Adams~Wei Yu, Brian Lester, Nan Du, Andrew~M. Dai, and Quoc~V. Le.
\newblock Finetuned language models are zero-shot learners.
\newblock In \emph{The Tenth International Conference on Learning Representations, {ICLR} 2022, Virtual Event, April 25-29, 2022}. OpenReview.net, 2022.
\newblock URL \url{https://openreview.net/forum?id=gEZrGCozdqR}.

\bibitem[Wei et~al.(2023)Wei, Wei, Lin, Li, Zhang, Ren, Li, Wan, Cao, Xie, Hu, Li, Hui, Yu, Liu, Yang, Huang, and Xie]{tuning-7}
Xiangpeng Wei, Haoran Wei, Huan Lin, Tianhao Li, Pei Zhang, Xingzhang Ren, Mei Li, Yu~Wan, Zhiwei Cao, Binbin Xie, Tianxiang Hu, Shangjie Li, Binyuan Hui, Bowen Yu, Dayiheng Liu, Baosong Yang, Fei Huang, and Jun Xie.
\newblock Polylm: An open source polyglot large language model.
\newblock \emph{CoRR}, abs/2307.06018, 2023.
\newblock \doi{10.48550/ARXIV.2307.06018}.
\newblock URL \url{https://doi.org/10.48550/arXiv.2307.06018}.

\bibitem[Wendler et~al.(2024)Wendler, Veselovsky, Monea, and West]{DBLP:journals/corr/abs-2402-10588}
Chris Wendler, Veniamin Veselovsky, Giovanni Monea, and Robert West.
\newblock Do llamas work in english? on the latent language of multilingual transformers.
\newblock \emph{CoRR}, abs/2402.10588, 2024.
\newblock \doi{10.48550/ARXIV.2402.10588}.
\newblock URL \url{https://doi.org/10.48550/arXiv.2402.10588}.

\bibitem[Wenzek et~al.(2020)Wenzek, Lachaux, Conneau, Chaudhary, Guzm{\'{a}}n, Joulin, and Grave]{DBLP:conf/lrec/WenzekLCCGJG20}
Guillaume Wenzek, Marie{-}Anne Lachaux, Alexis Conneau, Vishrav Chaudhary, Francisco Guzm{\'{a}}n, Armand Joulin, and Edouard Grave.
\newblock Ccnet: Extracting high quality monolingual datasets from web crawl data.
\newblock In Nicoletta Calzolari, Fr{\'{e}}d{\'{e}}ric B{\'{e}}chet, Philippe Blache, Khalid Choukri, Christopher Cieri, Thierry Declerck, Sara Goggi, Hitoshi Isahara, Bente Maegaard, Joseph Mariani, H{\'{e}}l{\`{e}}ne Mazo, Asunci{\'{o}}n Moreno, Jan Odijk, and Stelios Piperidis (eds.), \emph{Proceedings of The 12th Language Resources and Evaluation Conference, {LREC} 2020, Marseille, France, May 11-16, 2020}, pp.\  4003--4012. European Language Resources Association, 2020.
\newblock URL \url{https://aclanthology.org/2020.lrec-1.494/}.

\bibitem[Wu et~al.(2024{\natexlab{a}})Wu, Gan, Ge, Lu, Wang, Feng, Luo, and Shan]{cpt_15}
Chengyue Wu, Yukang Gan, Yixiao Ge, Zeyu Lu, Jiahao Wang, Ye~Feng, Ping Luo, and Ying Shan.
\newblock Llama pro: Progressive llama with block expansion.
\newblock \emph{CoRR}, abs/2401.02415, 2024{\natexlab{a}}.
\newblock \doi{10.48550/ARXIV.2401.02415}.
\newblock URL \url{https://doi.org/10.48550/arXiv.2401.02415}.

\bibitem[Wu et~al.(2024{\natexlab{b}})Wu, Xie, Yang, Wu, Gao, Ding, Wang, and He]{wu2024betadpodirectpreferenceoptimization}
Junkang Wu, Yuexiang Xie, Zhengyi Yang, Jiancan Wu, Jinyang Gao, Bolin Ding, Xiang Wang, and Xiangnan He.
\newblock $\beta$-dpo: Direct preference optimization with dynamic $\beta$, 2024{\natexlab{b}}.
\newblock URL \url{https://arxiv.org/abs/2407.08639}.

\bibitem[Wu et~al.(2024{\natexlab{c}})Wu, Nagata, Miao, and Tsuruoka]{mt-6}
Qiyu Wu, Masaaki Nagata, Zhongtao Miao, and Yoshimasa Tsuruoka.
\newblock Word alignment as preference for machine translation.
\newblock \emph{CoRR}, abs/2405.09223, 2024{\natexlab{c}}.
\newblock \doi{10.48550/ARXIV.2405.09223}.
\newblock URL \url{https://doi.org/10.48550/arXiv.2405.09223}.

\bibitem[Wu et~al.(2024{\natexlab{d}})Wu, Balashankar, Kim, Eisenstein, and Beirami]{tuning-6}
Zhaofeng Wu, Ananth Balashankar, Yoon Kim, Jacob Eisenstein, and Ahmad Beirami.
\newblock Reuse your rewards: Reward model transfer for zero-shot cross-lingual alignment.
\newblock \emph{CoRR}, abs/2404.12318, 2024{\natexlab{d}}.
\newblock \doi{10.48550/ARXIV.2404.12318}.
\newblock URL \url{https://doi.org/10.48550/arXiv.2404.12318}.

\bibitem[Xiao et~al.(2021)Xiao, Hu, Liu, Tu, and Sun]{DBLP:journals/aiopen/XiaoHLTS21}
Chaojun Xiao, Xueyu Hu, Zhiyuan Liu, Cunchao Tu, and Maosong Sun.
\newblock Lawformer: {A} pre-trained language model for chinese legal long documents.
\newblock \emph{{AI} Open}, 2:\penalty0 79--84, 2021.
\newblock \doi{10.1016/J.AIOPEN.2021.06.003}.
\newblock URL \url{https://doi.org/10.1016/j.aiopen.2021.06.003}.

\bibitem[Xie et~al.(2024)Xie, Zeng, Yu, Gao, Zhang, and Ye]{DBLP:journals/corr/abs-2403-15747}
Rui Xie, Zhengran Zeng, Zhuohao Yu, Chang Gao, Shikun Zhang, and Wei Ye.
\newblock Codeshell technical report.
\newblock \emph{CoRR}, abs/2403.15747, 2024.
\newblock \doi{10.48550/ARXIV.2403.15747}.
\newblock URL \url{https://doi.org/10.48550/arXiv.2403.15747}.

\bibitem[Xie et~al.(2022)Xie, Zhao, Yu, and Li]{DBLP:conf/emnlp/XieZ0L22}
Zhihui Xie, Handong Zhao, Tong Yu, and Shuai Li.
\newblock Discovering low-rank subspaces for language-agnostic multilingual representations.
\newblock In Yoav Goldberg, Zornitsa Kozareva, and Yue Zhang (eds.), \emph{Proceedings of the 2022 Conference on Empirical Methods in Natural Language Processing, {EMNLP} 2022, Abu Dhabi, United Arab Emirates, December 7-11, 2022}, pp.\  5617--5633. Association for Computational Linguistics, 2022.
\newblock \doi{10.18653/V1/2022.EMNLP-MAIN.379}.
\newblock URL \url{https://doi.org/10.18653/v1/2022.emnlp-main.379}.

\bibitem[Xiong et~al.(2023)Xiong, Wang, Zhu, Zhao, Liu, Huang, Wang, and Shen]{DBLP:journals/corr/abs-2304-01097}
Honglin Xiong, Sheng Wang, Yitao Zhu, Zihao Zhao, Yuxiao Liu, Linlin Huang, Qian Wang, and Dinggang Shen.
\newblock Doctorglm: Fine-tuning your chinese doctor is not a herculean task.
\newblock \emph{CoRR}, abs/2304.01097, 2023.
\newblock \doi{10.48550/ARXIV.2304.01097}.
\newblock URL \url{https://doi.org/10.48550/arXiv.2304.01097}.

\bibitem[Xiong et~al.(2024)Xiong, Dong, Ye, Wang, Zhong, Ji, Jiang, and Zhang]{xiong2024iterativepreferencelearninghuman}
Wei Xiong, Hanze Dong, Chenlu Ye, Ziqi Wang, Han Zhong, Heng Ji, Nan Jiang, and Tong Zhang.
\newblock Iterative preference learning from human feedback: Bridging theory and practice for rlhf under kl-constraint, 2024.
\newblock URL \url{https://arxiv.org/abs/2312.11456}.

\bibitem[Xu et~al.(2023{\natexlab{a}})Xu, Sun, Zheng, Geng, Zhao, Feng, Tao, and Jiang]{DBLP:journals/corr/abs-2304-12244}
Can Xu, Qingfeng Sun, Kai Zheng, Xiubo Geng, Pu~Zhao, Jiazhan Feng, Chongyang Tao, and Daxin Jiang.
\newblock Wizardlm: Empowering large language models to follow complex instructions.
\newblock \emph{CoRR}, abs/2304.12244, 2023{\natexlab{a}}.
\newblock \doi{10.48550/ARXIV.2304.12244}.
\newblock URL \url{https://doi.org/10.48550/arXiv.2304.12244}.

\bibitem[Xu et~al.(2022)Xu, Alon, Neubig, and Hellendoorn]{DBLP:conf/pldi/Xu0NH22}
Frank~F. Xu, Uri Alon, Graham Neubig, and Vincent~Josua Hellendoorn.
\newblock A systematic evaluation of large language models of code.
\newblock In Swarat Chaudhuri and Charles Sutton (eds.), \emph{MAPS@PLDI 2022: 6th {ACM} {SIGPLAN} International Symposium on Machine Programming, San Diego, CA, USA, 13 June 2022}, pp.\  1--10. {ACM}, 2022.
\newblock \doi{10.1145/3520312.3534862}.
\newblock URL \url{https://doi.org/10.1145/3520312.3534862}.

\bibitem[Xu et~al.(2023{\natexlab{b}})Xu, Kim, Sharaf, and Awadalla]{mt-2}
Haoran Xu, Young~Jin Kim, Amr Sharaf, and Hany~Hassan Awadalla.
\newblock A paradigm shift in machine translation: Boosting translation performance of large language models.
\newblock \emph{CoRR}, abs/2309.11674, 2023{\natexlab{b}}.
\newblock \doi{10.48550/ARXIV.2309.11674}.
\newblock URL \url{https://doi.org/10.48550/arXiv.2309.11674}.

\bibitem[Xu et~al.(2024{\natexlab{a}})Xu, Sharaf, Chen, Tan, Shen, Durme, Murray, and Kim]{mt-5}
Haoran Xu, Amr Sharaf, Yunmo Chen, Weiting Tan, Lingfeng Shen, Benjamin~Van Durme, Kenton Murray, and Young~Jin Kim.
\newblock Contrastive preference optimization: Pushing the boundaries of {LLM} performance in machine translation.
\newblock \emph{CoRR}, abs/2401.08417, 2024{\natexlab{a}}.
\newblock \doi{10.48550/ARXIV.2401.08417}.
\newblock URL \url{https://doi.org/10.48550/arXiv.2401.08417}.

\bibitem[Xu et~al.(2023{\natexlab{c}})Xu, Zhang, Ye, Zhang, and Huang]{DBLP:conf/emnlp/XuZYZH23}
Ningyu Xu, Qi~Zhang, Jingting Ye, Menghan Zhang, and Xuanjing Huang.
\newblock Are structural concepts universal in transformer language models? towards interpretable cross-lingual generalization.
\newblock In Houda Bouamor, Juan Pino, and Kalika Bali (eds.), \emph{Findings of the Association for Computational Linguistics: {EMNLP} 2023, Singapore, December 6-10, 2023}, pp.\  13951--13976. Association for Computational Linguistics, 2023{\natexlab{c}}.
\newblock \doi{10.18653/V1/2023.FINDINGS-EMNLP.931}.
\newblock URL \url{https://doi.org/10.18653/v1/2023.findings-emnlp.931}.

\bibitem[Xu et~al.(2024{\natexlab{b}})Xu, Zhao, Zu, Li, Chen, Zhang, Zheng, Dou, Qin, Gui, Zhang, and Huang]{mt-9}
Nuo Xu, Jun Zhao, Can Zu, Sixian Li, Lu~Chen, Zhihao Zhang, Rui Zheng, Shihan Dou, Wenjuan Qin, Tao Gui, Qi~Zhang, and Xuanjing Huang.
\newblock Advancing translation preference modeling with {RLHF:} {A} step towards cost-effective solution.
\newblock \emph{CoRR}, abs/2402.11525, 2024{\natexlab{b}}.
\newblock \doi{10.48550/ARXIV.2402.11525}.
\newblock URL \url{https://doi.org/10.48550/arXiv.2402.11525}.

\bibitem[Xu et~al.(2023{\natexlab{d}})Xu, Li, and Xiong]{DBLP:conf/emnlp/XuLX23}
Shaoyang Xu, Junzhuo Li, and Deyi Xiong.
\newblock Language representation projection: Can we transfer factual knowledge across languages in multilingual language models?
\newblock In Houda Bouamor, Juan Pino, and Kalika Bali (eds.), \emph{Proceedings of the 2023 Conference on Empirical Methods in Natural Language Processing, {EMNLP} 2023, Singapore, December 6-10, 2023}, pp.\  3692--3702. Association for Computational Linguistics, 2023{\natexlab{d}}.
\newblock \doi{10.18653/V1/2023.EMNLP-MAIN.226}.
\newblock URL \url{https://doi.org/10.18653/v1/2023.emnlp-main.226}.

\bibitem[Xu et~al.(2023{\natexlab{e}})Xu, Li, and Xiong]{xu2023language}
Shaoyang Xu, Junzhuo Li, and Deyi Xiong.
\newblock Language representation projection: Can we transfer factual knowledge across languages in multilingual language models?
\newblock \emph{arXiv preprint arXiv:2311.03788}, 2023{\natexlab{e}}.

\bibitem[Xu et~al.(2024{\natexlab{c}})Xu, Leng, Yu, and Xiong]{c-34-1}
Shaoyang Xu, Yongqi Leng, Linhao Yu, and Deyi Xiong.
\newblock Self-pluralising culture alignment for large language models.
\newblock \emph{CoRR}, abs/2410.12971, 2024{\natexlab{c}}.
\newblock URL \url{https://arxiv.org/abs/2410.12971}.

\bibitem[Xu et~al.(2024{\natexlab{d}})Xu, Liu, Liu, Hou, Li, Zhang, Wang, Zeng, Du, Zhao, Tang, and Dong]{DBLP:journals/corr/abs-2404-02893}
Yifan Xu, Xiao Liu, Xinghan Liu, Zhenyu Hou, Yueyan Li, Xiaohan Zhang, Zihan Wang, Aohan Zeng, Zhengxiao Du, Wenyi Zhao, Jie Tang, and Yuxiao Dong.
\newblock Chatglm-math: Improving math problem-solving in large language models with a self-critique pipeline.
\newblock \emph{CoRR}, abs/2404.02893, 2024{\natexlab{d}}.
\newblock \doi{10.48550/ARXIV.2404.02893}.
\newblock URL \url{https://doi.org/10.48550/arXiv.2404.02893}.

\bibitem[Xu et~al.(2024{\natexlab{e}})Xu, Hu, Zhao, Qiu, Ye, and Gu]{xu2024survey}
Yuemei Xu, Ling Hu, Jiayi Zhao, Zihan Qiu, Yuqi Ye, and Hanwen Gu.
\newblock A survey on multilingual large language models: Corpora, alignment, and bias.
\newblock \emph{arXiv preprint arXiv:2404.00929}, 2024{\natexlab{e}}.

\bibitem[Xu et~al.(2023{\natexlab{f}})Xu, Xie, Gu, Chen, Chang, Zhang, Chen, Zhang, and Tian]{cpt_11}
Yuhui Xu, Lingxi Xie, Xiaotao Gu, Xin Chen, Heng Chang, Hengheng Zhang, Zhengsu Chen, Xiaopeng Zhang, and Qi~Tian.
\newblock Qa-lora: Quantization-aware low-rank adaptation of large language models.
\newblock \emph{CoRR}, abs/2309.14717, 2023{\natexlab{f}}.
\newblock \doi{10.48550/ARXIV.2309.14717}.
\newblock URL \url{https://doi.org/10.48550/arXiv.2309.14717}.

\bibitem[Xue et~al.(2024)Xue, Zheng, Fu, Ni, Zheng, Zhou, and You]{OpenMoE}
Fuzhao Xue, Zian Zheng, Yao Fu, Jinjie Ni, Zangwei Zheng, Wangchunshu Zhou, and Yang You.
\newblock Openmoe: An early effort on open mixture-of-experts language models.
\newblock \emph{arXiv preprint arXiv:2402.01739}, 2024.

\bibitem[Xue et~al.(2020)Xue, Constant, Roberts, Kale, Al-Rfou, Siddhant, Barua, and Raffel]{mT5}
Linting Xue, Noah Constant, Adam Roberts, Mihir Kale, Rami Al-Rfou, Aditya Siddhant, Aditya Barua, and Colin Raffel.
\newblock mt5: A massively multilingual pre-trained text-to-text transformer.
\newblock \emph{arXiv preprint arXiv:2010.11934}, 2020.

\bibitem[Xue et~al.(2021)Xue, Constant, Roberts, Kale, Al{-}Rfou, Siddhant, Barua, and Raffel]{DBLP:conf/naacl/XueCRKASBR21}
Linting Xue, Noah Constant, Adam Roberts, Mihir Kale, Rami Al{-}Rfou, Aditya Siddhant, Aditya Barua, and Colin Raffel.
\newblock mt5: {A} massively multilingual pre-trained text-to-text transformer.
\newblock In Kristina Toutanova, Anna Rumshisky, Luke Zettlemoyer, Dilek Hakkani{-}T{\"{u}}r, Iz~Beltagy, Steven Bethard, Ryan Cotterell, Tanmoy Chakraborty, and Yichao Zhou (eds.), \emph{Proceedings of the 2021 Conference of the North American Chapter of the Association for Computational Linguistics: Human Language Technologies, {NAACL-HLT} 2021, Online, June 6-11, 2021}, pp.\  483--498. Association for Computational Linguistics, 2021.
\newblock \doi{10.18653/V1/2021.NAACL-MAIN.41}.
\newblock URL \url{https://doi.org/10.18653/v1/2021.naacl-main.41}.

\bibitem[Xue et~al.(2022)Xue, Barua, Constant, Al-Rfou, Narang, Kale, Roberts, and Raffel]{xue-etal-2022-byt5}
Linting Xue, Aditya Barua, Noah Constant, Rami Al-Rfou, Sharan Narang, Mihir Kale, Adam Roberts, and Colin Raffel.
\newblock {B}y{T}5: Towards a token-free future with pre-trained byte-to-byte models.
\newblock \emph{Transactions of the Association for Computational Linguistics}, 10:\penalty0 291--306, 2022.
\newblock \doi{10.1162/tacl_a_00461}.
\newblock URL \url{https://aclanthology.org/2022.tacl-1.17}.

\bibitem[Yang et~al.(2023{\natexlab{a}})Yang, Xiao, Wang, Zhang, Bian, Yin, Lv, Pan, Wang, Yan, Yang, Deng, Wang, Liu, Ai, Dong, Zhao, Xu, Sun, Zhang, Liu, Ji, Xie, Dai, Fang, Su, Song, Liu, Ru, Ma, Wang, Liu, Lin, Nie, Guo, Sun, Zhang, Li, Li, Cheng, Chen, Zeng, Wang, Chen, Men, Yu, Pan, Shen, Wang, Li, Jiang, Gao, Zhang, Zhou, and Wu]{DBLP:journals/corr/abs-2309-10305}
Aiyuan Yang, Bin Xiao, Bingning Wang, Borong Zhang, Ce~Bian, Chao Yin, Chenxu Lv, Da~Pan, Dian Wang, Dong Yan, Fan Yang, Fei Deng, Feng Wang, Feng Liu, Guangwei Ai, Guosheng Dong, Haizhou Zhao, Hang Xu, Haoze Sun, Hongda Zhang, Hui Liu, Jiaming Ji, Jian Xie, Juntao Dai, Kun Fang, Lei Su, Liang Song, Lifeng Liu, Liyun Ru, Luyao Ma, Mang Wang, Mickel Liu, MingAn Lin, Nuolan Nie, Peidong Guo, Ruiyang Sun, Tao Zhang, Tianpeng Li, Tianyu Li, Wei Cheng, Weipeng Chen, Xiangrong Zeng, Xiaochuan Wang, Xiaoxi Chen, Xin Men, Xin Yu, Xuehai Pan, Yanjun Shen, Yiding Wang, Yiyu Li, Youxin Jiang, Yuchen Gao, Yupeng Zhang, Zenan Zhou, and Zhiying Wu.
\newblock Baichuan 2: Open large-scale language models.
\newblock \emph{CoRR}, abs/2309.10305, 2023{\natexlab{a}}.
\newblock \doi{10.48550/ARXIV.2309.10305}.
\newblock URL \url{https://doi.org/10.48550/arXiv.2309.10305}.

\bibitem[Yang et~al.(2023{\natexlab{b}})Yang, Xiao, Wang, Zhang, Bian, Yin, Lv, Pan, Wang, Yan, et~al.]{baichuan}
Aiyuan Yang, Bin Xiao, Bingning Wang, Borong Zhang, Ce~Bian, Chao Yin, Chenxu Lv, Da~Pan, Dian Wang, Dong Yan, et~al.
\newblock Baichuan 2: Open large-scale language models.
\newblock \emph{arXiv preprint arXiv:2309.10305}, 2023{\natexlab{b}}.

\bibitem[Yang et~al.(2024{\natexlab{a}})Yang, Guo, Yin, Bai, Wang, Liu, Liang, Chai, Yang, and Li]{mt-13}
Jian Yang, Hongcheng Guo, Yuwei Yin, Jiaqi Bai, Bing Wang, Jiaheng Liu, Xinnian Liang, Linzheng Chai, Liqun Yang, and Zhoujun Li.
\newblock m3p: Towards multimodal multilingual translation with multimodal prompt.
\newblock In Nicoletta Calzolari, Min{-}Yen Kan, V{\'{e}}ronique Hoste, Alessandro Lenci, Sakriani Sakti, and Nianwen Xue (eds.), \emph{Proceedings of the 2024 Joint International Conference on Computational Linguistics, Language Resources and Evaluation, {LREC/COLING} 2024, 20-25 May, 2024, Torino, Italy}, pp.\  10858--10871. {ELRA} and {ICCL}, 2024{\natexlab{a}}.
\newblock URL \url{https://aclanthology.org/2024.lrec-main.948}.

\bibitem[Yang et~al.(2024{\natexlab{b}})Yang, Ali, Wang, Hu, and Wang]{cpt_14}
Shu Yang, Muhammad~Asif Ali, Cheng{-}Long Wang, Lijie Hu, and Di~Wang.
\newblock Moral: Moe augmented lora for llms' lifelong learning.
\newblock \emph{CoRR}, abs/2402.11260, 2024{\natexlab{b}}.
\newblock \doi{10.48550/ARXIV.2402.11260}.
\newblock URL \url{https://doi.org/10.48550/arXiv.2402.11260}.

\bibitem[Yang et~al.(2023{\natexlab{c}})Yang, Li, Zhang, and Zong]{mt-1}
Wen Yang, Chong Li, Jiajun Zhang, and Chengqing Zong.
\newblock Bigtrans: Augmenting large language models with multilingual translation capability over 100 languages.
\newblock \emph{CoRR}, abs/2305.18098, 2023{\natexlab{c}}.
\newblock \doi{10.48550/ARXIV.2305.18098}.
\newblock URL \url{https://doi.org/10.48550/arXiv.2305.18098}.

\bibitem[Yang et~al.(2019)Yang, Zhang, Tar, and Baldridge]{DBLP:conf/emnlp/YangZTB19}
Yinfei Yang, Yuan Zhang, Chris Tar, and Jason Baldridge.
\newblock {PAWS-X:} {A} cross-lingual adversarial dataset for paraphrase identification.
\newblock In Kentaro Inui, Jing Jiang, Vincent Ng, and Xiaojun Wan (eds.), \emph{Proceedings of the 2019 Conference on Empirical Methods in Natural Language Processing and the 9th International Joint Conference on Natural Language Processing, {EMNLP-IJCNLP} 2019, Hong Kong, China, November 3-7, 2019}, pp.\  3685--3690. Association for Computational Linguistics, 2019.
\newblock \doi{10.18653/V1/D19-1382}.
\newblock URL \url{https://doi.org/10.18653/v1/D19-1382}.

\bibitem[Yao et~al.(2024)Yao, Jiang, Yang, and Hu]{c-25}
Binwei Yao, Ming Jiang, Diyi Yang, and Junjie Hu.
\newblock Benchmarking llm-based machine translation on cultural awareness.
\newblock \emph{CoRR}, abs/2305.14328, 2024.
\newblock \doi{10.48550/ARXIV.2305.14328}.
\newblock URL \url{https://doi.org/10.48550/arXiv.2305.14328}.

\bibitem[Yin et~al.(2022)Yin, Bansal, Monajatipoor, Li, and Chang]{c-21}
Da~Yin, Hritik Bansal, Masoud Monajatipoor, Liunian~Harold Li, and Kai{-}Wei Chang.
\newblock Geomlama: Geo-diverse commonsense probing on multilingual pre-trained language models.
\newblock In Yoav Goldberg, Zornitsa Kozareva, and Yue Zhang (eds.), \emph{Proceedings of the 2022 Conference on Empirical Methods in Natural Language Processing, {EMNLP} 2022, Abu Dhabi, United Arab Emirates, December 7-11, 2022}, pp.\  2039--2055. Association for Computational Linguistics, 2022.
\newblock \doi{10.18653/V1/2022.EMNLP-MAIN.132}.
\newblock URL \url{https://doi.org/10.18653/v1/2022.emnlp-main.132}.

\bibitem[Yong et~al.(2023{\natexlab{a}})Yong, Menghini, and Bach]{DBLP:journals/corr/abs-2310-02446}
Zheng~Xin Yong, Cristina Menghini, and Stephen~H. Bach.
\newblock Low-resource languages jailbreak {GPT-4}.
\newblock \emph{CoRR}, abs/2310.02446, 2023{\natexlab{a}}.
\newblock \doi{10.48550/ARXIV.2310.02446}.
\newblock URL \url{https://doi.org/10.48550/arXiv.2310.02446}.

\bibitem[Yong et~al.(2023{\natexlab{b}})Yong, Schoelkopf, Muennighoff, Aji, Adelani, Almubarak, Bari, Sutawika, Kasai, Baruwa, Winata, Biderman, Raff, Radev, and Nikoulina]{adapt-8}
Zheng~Xin Yong, Hailey Schoelkopf, Niklas Muennighoff, Alham~Fikri Aji, David~Ifeoluwa Adelani, Khalid Almubarak, M.~Saiful Bari, Lintang Sutawika, Jungo Kasai, Ahmed Baruwa, Genta~Indra Winata, Stella Biderman, Edward Raff, Dragomir Radev, and Vassilina Nikoulina.
\newblock {BLOOM+1:} adding language support to {BLOOM} for zero-shot prompting.
\newblock In Anna Rogers, Jordan~L. Boyd{-}Graber, and Naoaki Okazaki (eds.), \emph{Proceedings of the 61st Annual Meeting of the Association for Computational Linguistics (Volume 1: Long Papers), {ACL} 2023, Toronto, Canada, July 9-14, 2023}, pp.\  11682--11703. Association for Computational Linguistics, 2023{\natexlab{b}}.
\newblock \doi{10.18653/V1/2023.ACL-LONG.653}.
\newblock URL \url{https://doi.org/10.18653/v1/2023.acl-long.653}.

\bibitem[Yoon et~al.(2024)Yoon, Jang, Kim, Kim, Shafayat, and Seo]{adapt-6}
Dongkeun Yoon, Joel Jang, Sungdong Kim, Seungone Kim, Sheikh Shafayat, and Minjoon Seo.
\newblock Langbridge: Multilingual reasoning without multilingual supervision.
\newblock \emph{CoRR}, abs/2401.10695, 2024.
\newblock \doi{10.48550/ARXIV.2401.10695}.
\newblock URL \url{https://doi.org/10.48550/arXiv.2401.10695}.

\bibitem[Young et~al.(2024)Young, Chen, Li, Huang, Zhang, Zhang, Li, Zhu, Chen, Chang, et~al.]{Yi}
Alex Young, Bei Chen, Chao Li, Chengen Huang, Ge~Zhang, Guanwei Zhang, Heng Li, Jiangcheng Zhu, Jianqun Chen, Jing Chang, et~al.
\newblock Yi: Open foundation models by 01. ai.
\newblock \emph{arXiv preprint arXiv:2403.04652}, 2024.

\bibitem[Yu et~al.(2023)Yu, Jiang, Shi, Yu, Liu, Zhang, Kwok, Li, Weller, and Liu]{DBLP:journals/corr/abs-2309-12284}
Longhui Yu, Weisen Jiang, Han Shi, Jincheng Yu, Zhengying Liu, Yu~Zhang, James~T. Kwok, Zhenguo Li, Adrian Weller, and Weiyang Liu.
\newblock Metamath: Bootstrap your own mathematical questions for large language models.
\newblock \emph{CoRR}, abs/2309.12284, 2023.
\newblock \doi{10.48550/ARXIV.2309.12284}.
\newblock URL \url{https://doi.org/10.48550/arXiv.2309.12284}.

\bibitem[Yuan et~al.(2023{\natexlab{a}})Yuan, Yuan, Wu, and Li]{DBLP:journals/corr/abs-2311-09071}
Fei Yuan, Shuai Yuan, Zhiyong Wu, and Lei Li.
\newblock How multilingual is multilingual llm?
\newblock \emph{CoRR}, abs/2311.09071, 2023{\natexlab{a}}.
\newblock \doi{10.48550/ARXIV.2311.09071}.
\newblock URL \url{https://doi.org/10.48550/arXiv.2311.09071}.

\bibitem[Yuan et~al.(2023{\natexlab{b}})Yuan, Yuan, Tan, Wang, Huang, and Huang]{DBLP:journals/corr/abs-2304-05302}
Zheng Yuan, Hongyi Yuan, Chuanqi Tan, Wei Wang, Songfang Huang, and Fei Huang.
\newblock {RRHF:} rank responses to align language models with human feedback without tears.
\newblock \emph{CoRR}, abs/2304.05302, 2023{\natexlab{b}}.
\newblock \doi{10.48550/ARXIV.2304.05302}.
\newblock URL \url{https://doi.org/10.48550/arXiv.2304.05302}.

\bibitem[Yue et~al.(2023{\natexlab{a}})Yue, Chen, Wang, Li, Shen, Liu, Zhou, Xiao, Yun, Huang, and Wei]{DBLP:journals/corr/abs-2309-11325}
Shengbin Yue, Wei Chen, Siyuan Wang, Bingxuan Li, Chenchen Shen, Shujun Liu, Yuxuan Zhou, Yao Xiao, Song Yun, Xuanjing Huang, and Zhongyu Wei.
\newblock Disc-lawllm: Fine-tuning large language models for intelligent legal services.
\newblock \emph{CoRR}, abs/2309.11325, 2023{\natexlab{a}}.
\newblock \doi{10.48550/ARXIV.2309.11325}.
\newblock URL \url{https://doi.org/10.48550/arXiv.2309.11325}.

\bibitem[Yue et~al.(2023{\natexlab{b}})Yue, Qu, Zhang, Fu, Huang, Sun, Su, and Chen]{DBLP:journals/corr/abs-2309-05653}
Xiang Yue, Xingwei Qu, Ge~Zhang, Yao Fu, Wenhao Huang, Huan Sun, Yu~Su, and Wenhu Chen.
\newblock Mammoth: Building math generalist models through hybrid instruction tuning.
\newblock \emph{CoRR}, abs/2309.05653, 2023{\natexlab{b}}.
\newblock \doi{10.48550/ARXIV.2309.05653}.
\newblock URL \url{https://doi.org/10.48550/arXiv.2309.05653}.

\bibitem[Zeng et~al.(2024)Zeng, Meng, Yin, and Zhou]{mt-7}
Jiali Zeng, Fandong Meng, Yongjing Yin, and Jie Zhou.
\newblock Teaching large language models to translate with comparison.
\newblock In Michael~J. Wooldridge, Jennifer~G. Dy, and Sriraam Natarajan (eds.), \emph{Thirty-Eighth {AAAI} Conference on Artificial Intelligence, {AAAI} 2024, Thirty-Sixth Conference on Innovative Applications of Artificial Intelligence, {IAAI} 2024, Fourteenth Symposium on Educational Advances in Artificial Intelligence, {EAAI} 2014, February 20-27, 2024, Vancouver, Canada}, pp.\  19488--19496. {AAAI} Press, 2024.
\newblock \doi{10.1609/AAAI.V38I17.29920}.
\newblock URL \url{https://doi.org/10.1609/aaai.v38i17.29920}.

\bibitem[Zeng et~al.(2021)Zeng, Ren, Su, Wang, Liao, Wang, Jiang, Yang, Wang, Zhang, Li, Gong, Yao, Huang, Wang, Yu, Guo, Yu, Zhang, Wang, Tao, Yan, Yi, Peng, Jiang, Zhang, Deng, Zhang, Lin, Zhang, Zhang, Guo, Gu, Fan, Wang, Jin, Liu, and Tian]{DBLP:journals/corr/abs-2104-12369}
Wei Zeng, Xiaozhe Ren, Teng Su, Hui Wang, Yi~Liao, Zhiwei Wang, Xin Jiang, ZhenZhang Yang, Kaisheng Wang, Xiaoda Zhang, Chen Li, Ziyan Gong, Yifan Yao, Xinjing Huang, Jun Wang, Jianfeng Yu, Qi~Guo, Yue Yu, Yan Zhang, Jin Wang, Hengtao Tao, Dasen Yan, Zexuan Yi, Fang Peng, Fangqing Jiang, Han Zhang, Lingfeng Deng, Yehong Zhang, Zhe Lin, Chao Zhang, Shaojie Zhang, Mingyue Guo, Shanzhi Gu, Gaojun Fan, Yaowei Wang, Xuefeng Jin, Qun Liu, and Yonghong Tian.
\newblock Pangu-{\(\alpha\)}: Large-scale autoregressive pretrained chinese language models with auto-parallel computation.
\newblock \emph{CoRR}, abs/2104.12369, 2021.
\newblock URL \url{https://arxiv.org/abs/2104.12369}.

\bibitem[Zhang \& Sennrich(2019)Zhang and Sennrich]{rmsnorm}
Biao Zhang and Rico Sennrich.
\newblock Root mean square layer normalization.
\newblock \emph{Advances in Neural Information Processing Systems}, 32, 2019.

\bibitem[Zhang et~al.(2023{\natexlab{a}})Zhang, D'Haro, Tang, Shi, Tang, and Li]{DBLP:conf/emnlp/ZhangDTST023}
Chen Zhang, Luis~F. D'Haro, Chengguang Tang, Ke~Shi, Guohua Tang, and Haizhou Li.
\newblock xdial-eval: {A} multilingual open-domain dialogue evaluation benchmark.
\newblock In Houda Bouamor, Juan Pino, and Kalika Bali (eds.), \emph{Findings of the Association for Computational Linguistics: {EMNLP} 2023, Singapore, December 6-10, 2023}, pp.\  5579--5601. Association for Computational Linguistics, 2023{\natexlab{a}}.
\newblock \doi{10.18653/V1/2023.FINDINGS-EMNLP.371}.
\newblock URL \url{https://doi.org/10.18653/v1/2023.findings-emnlp.371}.

\bibitem[Zhang et~al.(2024{\natexlab{a}})Zhang, Chen, Bai, Xiang, and Zhang]{zhang2024paying}
Hongbin Zhang, Kehai Chen, Xuefeng Bai, Yang Xiang, and Min Zhang.
\newblock Paying more attention to source context: Mitigating unfaithful translations from large language model.
\newblock \emph{arXiv preprint arXiv:2406.07036}, 2024{\natexlab{a}}.

\bibitem[Zhang et~al.(2023{\natexlab{b}})Zhang, Chen, Jiang, Yu, Chen, Chen, Li, Wu, Zhang, Xiao, Wan, Wang, and Li]{DBLP:conf/emnlp/ZhangCJYCCLWZXW23}
Hongbo Zhang, Junying Chen, Feng Jiang, Fei Yu, Zhihong Chen, Guiming Chen, Jianquan Li, Xiangbo Wu, Zhiyi Zhang, Qingying Xiao, Xiang Wan, Benyou Wang, and Haizhou Li.
\newblock Huatuogpt, towards taming language model to be a doctor.
\newblock In Houda Bouamor, Juan Pino, and Kalika Bali (eds.), \emph{Findings of the Association for Computational Linguistics: {EMNLP} 2023, Singapore, December 6-10, 2023}, pp.\  10859--10885. Association for Computational Linguistics, 2023{\natexlab{b}}.
\newblock \doi{10.18653/V1/2023.FINDINGS-EMNLP.725}.
\newblock URL \url{https://doi.org/10.18653/v1/2023.findings-emnlp.725}.

\bibitem[Zhang et~al.(2023{\natexlab{c}})Zhang, Fang, Zhang, Ma, Zhou, Huang, Bu, Gui, Chen, Chen, and Feng]{trans-3}
Shaolei Zhang, Qingkai Fang, Zhuocheng Zhang, Zhengrui Ma, Yan Zhou, Langlin Huang, Mengyu Bu, Shangtong Gui, Yunji Chen, Xilin Chen, and Yang Feng.
\newblock Bayling: Bridging cross-lingual alignment and instruction following through interactive translation for large language models.
\newblock \emph{CoRR}, abs/2306.10968, 2023{\natexlab{c}}.
\newblock \doi{10.48550/ARXIV.2306.10968}.
\newblock URL \url{https://doi.org/10.48550/arXiv.2306.10968}.

\bibitem[Zhang et~al.(2023{\natexlab{d}})Zhang, Dong, Li, Zhang, Sun, Wang, Li, Hu, Zhang, Wu, and Wang]{DBLP:journals/corr/abs-2308-10792}
Shengyu Zhang, Linfeng Dong, Xiaoya Li, Sen Zhang, Xiaofei Sun, Shuhe Wang, Jiwei Li, Runyi Hu, Tianwei Zhang, Fei Wu, and Guoyin Wang.
\newblock Instruction tuning for large language models: {A} survey.
\newblock \emph{CoRR}, abs/2308.10792, 2023{\natexlab{d}}.
\newblock \doi{10.48550/ARXIV.2308.10792}.
\newblock URL \url{https://doi.org/10.48550/arXiv.2308.10792}.

\bibitem[Zhang et~al.(2024{\natexlab{b}})Zhang, Gao, Zhu, Chen, Huang, Han, Feng, Deng, and Huang]{zhang2024getting}
Shimao Zhang, Changjiang Gao, Wenhao Zhu, Jiajun Chen, Xin Huang, Xue Han, Junlan Feng, Chao Deng, and Shujian Huang.
\newblock Getting more from less: Large language models are good spontaneous multilingual learners.
\newblock In \emph{Proceedings of the 2024 Conference on Empirical Methods in Natural Language Processing}, pp.\  8037--8051, 2024{\natexlab{b}}.

\bibitem[Zhang et~al.(2022)Zhang, Roller, Goyal, Artetxe, Chen, Chen, Dewan, Diab, Li, Lin, Mihaylov, Ott, Shleifer, Shuster, Simig, Koura, Sridhar, Wang, and Zettlemoyer]{DBLP:journals/corr/abs-2205-01068}
Susan Zhang, Stephen Roller, Naman Goyal, Mikel Artetxe, Moya Chen, Shuohui Chen, Christopher Dewan, Mona~T. Diab, Xian Li, Xi~Victoria Lin, Todor Mihaylov, Myle Ott, Sam Shleifer, Kurt Shuster, Daniel Simig, Punit~Singh Koura, Anjali Sridhar, Tianlu Wang, and Luke Zettlemoyer.
\newblock {OPT:} open pre-trained transformer language models.
\newblock \emph{CoRR}, abs/2205.01068, 2022.
\newblock \doi{10.48550/ARXIV.2205.01068}.
\newblock URL \url{https://doi.org/10.48550/arXiv.2205.01068}.

\bibitem[Zhang et~al.(2023{\natexlab{e}})Zhang, Aljunied, Gao, Chia, and Bing]{DBLP:conf/nips/ZhangAGCB23}
Wenxuan Zhang, Mahani Aljunied, Chang Gao, Yew~Ken Chia, and Lidong Bing.
\newblock M3exam: {A} multilingual, multimodal, multilevel benchmark for examining large language models.
\newblock In Alice Oh, Tristan Naumann, Amir Globerson, Kate Saenko, Moritz Hardt, and Sergey Levine (eds.), \emph{Advances in Neural Information Processing Systems 36: Annual Conference on Neural Information Processing Systems 2023, NeurIPS 2023, New Orleans, LA, USA, December 10 - 16, 2023}, 2023{\natexlab{e}}.
\newblock URL \url{http://papers.nips.cc/paper\_files/paper/2023/hash/117c5c8622b0d539f74f6d1fb082a2e9-Abstract-Datasets\_and\_Benchmarks.html}.

\bibitem[Zhang et~al.(2024{\natexlab{c}})Zhang, Wang, Liu, Wang, Wang, Li, Sun, and Liu]{DBLP:journals/corr/abs-2402-12204}
Yuanchi Zhang, Yile Wang, Zijun Liu, Shuo Wang, Xiaolong Wang, Peng Li, Maosong Sun, and Yang Liu.
\newblock Enhancing multilingual capabilities of large language models through self-distillation from resource-rich languages.
\newblock \emph{CoRR}, abs/2402.12204, 2024{\natexlab{c}}.
\newblock \doi{10.48550/ARXIV.2402.12204}.
\newblock URL \url{https://doi.org/10.48550/arXiv.2402.12204}.

\bibitem[Zhang et~al.(2024{\natexlab{d}})Zhang, Wang, Liu, Wang, Wang, Li, Sun, and Liu]{zhang2024enhancing}
Yuanchi Zhang, Yile Wang, Zijun Liu, Shuo Wang, Xiaolong Wang, Peng Li, Maosong Sun, and Yang Liu.
\newblock Enhancing multilingual capabilities of large language models through self-distillation from resource-rich languages.
\newblock \emph{arXiv preprint arXiv:2402.12204}, 2024{\natexlab{d}}.

\bibitem[Zhang et~al.(2023{\natexlab{f}})Zhang, Lee, Fang, Yu, Jia, Jiang, and Barbieri]{trans-8}
Zhihan Zhang, Dong{-}Ho Lee, Yuwei Fang, Wenhao Yu, Mengzhao Jia, Meng Jiang, and Francesco Barbieri.
\newblock {PLUG:} leveraging pivot language in cross-lingual instruction tuning.
\newblock \emph{CoRR}, abs/2311.08711, 2023{\natexlab{f}}.
\newblock \doi{10.48550/ARXIV.2311.08711}.
\newblock URL \url{https://doi.org/10.48550/arXiv.2311.08711}.

\bibitem[Zhang et~al.(2024{\natexlab{e}})Zhang, Zhao, Zhang, Gui, and Huang]{DBLP:journals/corr/abs-2402-14700}
Zhihao Zhang, Jun Zhao, Qi~Zhang, Tao Gui, and Xuanjing Huang.
\newblock Unveiling linguistic regions in large language models.
\newblock \emph{CoRR}, abs/2402.14700, 2024{\natexlab{e}}.
\newblock \doi{10.48550/ARXIV.2402.14700}.
\newblock URL \url{https://doi.org/10.48550/arXiv.2402.14700}.

\bibitem[Zhao et~al.(2024{\natexlab{a}})Zhao, Zhang, Gao, Zhang, Gui, and Huang]{adapt-9}
Jun Zhao, Zhihao Zhang, Luhui Gao, Qi~Zhang, Tao Gui, and Xuanjing Huang.
\newblock Llama beyond english: An empirical study on language capability transfer.
\newblock \emph{CoRR}, abs/2401.01055, 2024{\natexlab{a}}.
\newblock \doi{10.48550/ARXIV.2401.01055}.
\newblock URL \url{https://doi.org/10.48550/arXiv.2401.01055}.

\bibitem[Zhao et~al.(2024{\natexlab{b}})Zhao, Mondal, Tandon, Dillion, Gray, and Gu]{c-28}
Wenlong Zhao, Debanjan Mondal, Niket Tandon, Danica Dillion, Kurt Gray, and Yuling Gu.
\newblock Worldvaluesbench: {A} large-scale benchmark dataset for multi-cultural value awareness of language models.
\newblock In Nicoletta Calzolari, Min{-}Yen Kan, V{\'{e}}ronique Hoste, Alessandro Lenci, Sakriani Sakti, and Nianwen Xue (eds.), \emph{Proceedings of the 2024 Joint International Conference on Computational Linguistics, Language Resources and Evaluation, {LREC/COLING} 2024, 20-25 May, 2024, Torino, Italy}, pp.\  17696--17706. {ELRA} and {ICCL}, 2024{\natexlab{b}}.
\newblock URL \url{https://aclanthology.org/2024.lrec-main.1539}.

\bibitem[Zhao et~al.(2023)Zhao, Joshi, Liu, Khalman, Saleh, and Liu]{DBLP:journals/corr/abs-2305-10425}
Yao Zhao, Rishabh Joshi, Tianqi Liu, Misha Khalman, Mohammad Saleh, and Peter~J. Liu.
\newblock Slic-hf: Sequence likelihood calibration with human feedback.
\newblock \emph{CoRR}, abs/2305.10425, 2023.
\newblock \doi{10.48550/ARXIV.2305.10425}.
\newblock URL \url{https://doi.org/10.48550/arXiv.2305.10425}.

\bibitem[Zhao et~al.(2024{\natexlab{c}})Zhao, Zhang, Chen, Kawaguchi, and Bing]{DBLP:journals/corr/abs-2402-18815}
Yiran Zhao, Wenxuan Zhang, Guizhen Chen, Kenji Kawaguchi, and Lidong Bing.
\newblock How do large language models handle multilingualism?
\newblock \emph{CoRR}, abs/2402.18815, 2024{\natexlab{c}}.
\newblock \doi{10.48550/ARXIV.2402.18815}.
\newblock URL \url{https://doi.org/10.48550/arXiv.2402.18815}.

\bibitem[Zhao et~al.(2024{\natexlab{d}})Zhao, Zhang, Chen, Kawaguchi, and Bing]{zhao2024large}
Yiran Zhao, Wenxuan Zhang, Guizhen Chen, Kenji Kawaguchi, and Lidong Bing.
\newblock How do large language models handle multilingualism?
\newblock \emph{arXiv preprint arXiv:2402.18815}, 2024{\natexlab{d}}.

\bibitem[Zhao et~al.(2024{\natexlab{e}})Zhao, Zhang, Wang, Kawaguchi, and Bing]{zhao2024adamergex}
Yiran Zhao, Wenxuan Zhang, Huiming Wang, Kenji Kawaguchi, and Lidong Bing.
\newblock Adamergex: Cross-lingual transfer with large language models via adaptive adapter merging.
\newblock \emph{arXiv preprint arXiv:2402.18913}, 2024{\natexlab{e}}.

\bibitem[Zheng et~al.(2023{\natexlab{a}})Zheng, Chiang, Sheng, Li, Zhuang, Wu, Zhuang, Li, Lin, Xing, Gonzalez, Stoica, and Zhang]{DBLP:journals/corr/abs-2309-11998}
Lianmin Zheng, Wei{-}Lin Chiang, Ying Sheng, Tianle Li, Siyuan Zhuang, Zhanghao Wu, Yonghao Zhuang, Zhuohan Li, Zi~Lin, Eric~P. Xing, Joseph~E. Gonzalez, Ion Stoica, and Hao Zhang.
\newblock Lmsys-chat-1m: {A} large-scale real-world {LLM} conversation dataset.
\newblock \emph{CoRR}, abs/2309.11998, 2023{\natexlab{a}}.
\newblock \doi{10.48550/ARXIV.2309.11998}.
\newblock URL \url{https://doi.org/10.48550/arXiv.2309.11998}.

\bibitem[Zheng et~al.(2023{\natexlab{b}})Zheng, Chiang, Sheng, Zhuang, Wu, Zhuang, Lin, Li, Li, Xing, Zhang, Gonzalez, and Stoica]{zheng2023judging}
Lianmin Zheng, Wei-Lin Chiang, Ying Sheng, Siyuan Zhuang, Zhanghao Wu, Yonghao Zhuang, Zi~Lin, Zhuohan Li, Dacheng Li, Eric.~P Xing, Hao Zhang, Joseph~E. Gonzalez, and Ion Stoica.
\newblock Judging llm-as-a-judge with mt-bench and chatbot arena, 2023{\natexlab{b}}.

\bibitem[Zheng et~al.(2023{\natexlab{c}})Zheng, Xia, Zou, Dong, Wang, Xue, Shen, Wang, Wang, Li, Su, Yang, and Tang]{DBLP:conf/kdd/ZhengXZDWXSW0LS23}
Qinkai Zheng, Xiao Xia, Xu~Zou, Yuxiao Dong, Shan Wang, Yufei Xue, Lei Shen, Zihan Wang, Andi Wang, Yang Li, Teng Su, Zhilin Yang, and Jie Tang.
\newblock Codegeex: {A} pre-trained model for code generation with multilingual benchmarking on humaneval-x.
\newblock In Ambuj~K. Singh, Yizhou Sun, Leman Akoglu, Dimitrios Gunopulos, Xifeng Yan, Ravi Kumar, Fatma Ozcan, and Jieping Ye (eds.), \emph{Proceedings of the 29th {ACM} {SIGKDD} Conference on Knowledge Discovery and Data Mining, {KDD} 2023, Long Beach, CA, USA, August 6-10, 2023}, pp.\  5673--5684. {ACM}, 2023{\natexlab{c}}.
\newblock \doi{10.1145/3580305.3599790}.
\newblock URL \url{https://doi.org/10.1145/3580305.3599790}.

\bibitem[Zheng et~al.(2023{\natexlab{d}})Zheng, Xia, Zou, Dong, Wang, Xue, Wang, Shen, Wang, Li, Su, Yang, and Tang]{DBLP:journals/corr/abs-2303-17568}
Qinkai Zheng, Xiao Xia, Xu~Zou, Yuxiao Dong, Shan Wang, Yufei Xue, Zihan Wang, Lei Shen, Andi Wang, Yang Li, Teng Su, Zhilin Yang, and Jie Tang.
\newblock Codegeex: {A} pre-trained model for code generation with multilingual evaluations on humaneval-x.
\newblock \emph{CoRR}, abs/2303.17568, 2023{\natexlab{d}}.
\newblock \doi{10.48550/ARXIV.2303.17568}.
\newblock URL \url{https://doi.org/10.48550/arXiv.2303.17568}.

\bibitem[Zhou et~al.(2023{\natexlab{a}})Zhou, Liu, Xu, Iyer, Sun, Mao, Ma, Efrat, Yu, Yu, Zhang, Ghosh, Lewis, Zettlemoyer, and Levy]{lima}
Chunting Zhou, Pengfei Liu, Puxin Xu, Srinivasan Iyer, Jiao Sun, Yuning Mao, Xuezhe Ma, Avia Efrat, Ping Yu, Lili Yu, Susan Zhang, Gargi Ghosh, Mike Lewis, Luke Zettlemoyer, and Omer Levy.
\newblock {LIMA:} less is more for alignment.
\newblock In Alice Oh, Tristan Naumann, Amir Globerson, Kate Saenko, Moritz Hardt, and Sergey Levine (eds.), \emph{Advances in Neural Information Processing Systems 36: Annual Conference on Neural Information Processing Systems 2023, NeurIPS 2023, New Orleans, LA, USA, December 10 - 16, 2023}, 2023{\natexlab{a}}.
\newblock URL \url{http://papers.nips.cc/paper\_files/paper/2023/hash/ac662d74829e4407ce1d126477f4a03a-Abstract-Conference.html}.

\bibitem[Zhou \& Cao(2021)Zhou and Cao]{cpt_8}
Fan Zhou and Chengtai Cao.
\newblock Overcoming catastrophic forgetting in graph neural networks with experience replay.
\newblock In \emph{Thirty-Fifth {AAAI} Conference on Artificial Intelligence, {AAAI} 2021, Thirty-Third Conference on Innovative Applications of Artificial Intelligence, {IAAI} 2021, The Eleventh Symposium on Educational Advances in Artificial Intelligence, {EAAI} 2021, Virtual Event, February 2-9, 2021}, pp.\  4714--4722. {AAAI} Press, 2021.
\newblock \doi{10.1609/AAAI.V35I5.16602}.
\newblock URL \url{https://doi.org/10.1609/aaai.v35i5.16602}.

\bibitem[Zhou et~al.(2024)Zhou, Karidi, Garneau, Cao, Liu, Chen, and Hershcovich]{c-31}
Li~Zhou, Taelin Karidi, Nicolas Garneau, Yong Cao, Wanlong Liu, Wenyu Chen, and Daniel Hershcovich.
\newblock Does mapo tofu contain coffee? probing llms for food-related cultural knowledge.
\newblock \emph{CoRR}, abs/2404.06833, 2024.
\newblock \doi{10.48550/ARXIV.2404.06833}.
\newblock URL \url{https://doi.org/10.48550/arXiv.2404.06833}.

\bibitem[Zhou et~al.(2023{\natexlab{b}})Zhou, Ji, Li, Dutta, Davuluri, and Liu]{DBLP:journals/corr/abs-2306-15006}
Zhihan Zhou, Yanrong Ji, Weijian Li, Pratik Dutta, Ramana~V. Davuluri, and Han Liu.
\newblock {DNABERT-2:} efficient foundation model and benchmark for multi-species genome.
\newblock \emph{CoRR}, abs/2306.15006, 2023{\natexlab{b}}.
\newblock \doi{10.48550/ARXIV.2306.15006}.
\newblock URL \url{https://doi.org/10.48550/arXiv.2306.15006}.

\bibitem[Zhu et~al.(2024{\natexlab{a}})Zhu, Chen, Zhang, Haddow, Shen, and Klakow]{mt-4}
Dawei Zhu, Pinzhen Chen, Miaoran Zhang, Barry Haddow, Xiaoyu Shen, and Dietrich Klakow.
\newblock Fine-tuning large language models to translate: Will a touch of noisy data in misaligned languages suffice?
\newblock \emph{CoRR}, abs/2404.14122, 2024{\natexlab{a}}.
\newblock \doi{10.48550/ARXIV.2404.14122}.
\newblock URL \url{https://doi.org/10.48550/arXiv.2404.14122}.

\bibitem[Zhu et~al.(2024{\natexlab{b}})Zhu, Cui, and Xiong]{zhu2024towards}
Shaolin Zhu, Menglong Cui, and Deyi Xiong.
\newblock Towards robust in-context learning for machine translation with large language models.
\newblock In \emph{Proceedings of the 2024 Joint International Conference on Computational Linguistics, Language Resources and Evaluation (LREC-COLING 2024)}, pp.\  16619--16629, 2024{\natexlab{b}}.

\bibitem[Zhu et~al.(2024{\natexlab{c}})Zhu, Pan, Li, and Xiong]{zhu2024landermt}
Shaolin Zhu, Leiyu Pan, Bo~Li, and Deyi Xiong.
\newblock Landermt: Dectecting and routing language-aware neurons for selectively finetuning llms to machine translation.
\newblock In \emph{Proceedings of the 62nd Annual Meeting of the Association for Computational Linguistics (Volume 1: Long Papers)}, pp.\  12135--12148, 2024{\natexlab{c}}.

\bibitem[Zhu et~al.(2024{\natexlab{d}})Zhu, Pan, and Xiong]{zhu2024feds}
Shaolin Zhu, Leiyu Pan, and Deyi Xiong.
\newblock Feds-icl: Enhancing translation ability and efficiency of large language model by optimizing demonstration selection.
\newblock \emph{Information Processing \& Management}, 61\penalty0 (5):\penalty0 103825, 2024{\natexlab{d}}.

\bibitem[Zhu et~al.(2023{\natexlab{a}})Zhu, Liu, Dong, Xu, Kong, Chen, Li, and Huang]{DBLP:journals/corr/abs-2304-04675}
Wenhao Zhu, Hongyi Liu, Qingxiu Dong, Jingjing Xu, Lingpeng Kong, Jiajun Chen, Lei Li, and Shujian Huang.
\newblock Multilingual machine translation with large language models: Empirical results and analysis.
\newblock \emph{CoRR}, abs/2304.04675, 2023{\natexlab{a}}.
\newblock \doi{10.48550/ARXIV.2304.04675}.
\newblock URL \url{https://doi.org/10.48550/arXiv.2304.04675}.

\bibitem[Zhu et~al.(2023{\natexlab{b}})Zhu, Lv, Dong, Yuan, Xu, Huang, Kong, Chen, and Li]{trans-2}
Wenhao Zhu, Yunzhe Lv, Qingxiu Dong, Fei Yuan, Jingjing Xu, Shujian Huang, Lingpeng Kong, Jiajun Chen, and Lei Li.
\newblock Extrapolating large language models to non-english by aligning languages.
\newblock \emph{CoRR}, abs/2308.04948, 2023{\natexlab{b}}.
\newblock \doi{10.48550/ARXIV.2308.04948}.
\newblock URL \url{https://doi.org/10.48550/arXiv.2308.04948}.

\bibitem[Zhu et~al.(2024{\natexlab{e}})Zhu, Huang, Yuan, Cheng, Chen, and Birch]{trans-6}
Wenhao Zhu, Shujian Huang, Fei Yuan, Chen Cheng, Jiajun Chen, and Alexandra Birch.
\newblock The power of question translation training in multilingual reasoning: Broadened scope and deepened insights.
\newblock \emph{CoRR}, abs/2405.01345, 2024{\natexlab{e}}.
\newblock URL \url{https://doi.org/10.48550/arXiv.2405.01345}.

\bibitem[Zhu et~al.(2024{\natexlab{f}})Zhu, Huang, Yuan, She, Chen, and Birch]{trans-5}
Wenhao Zhu, Shujian Huang, Fei Yuan, Shuaijie She, Jiajun Chen, and Alexandra Birch.
\newblock Question translation training for better multilingual reasoning.
\newblock \emph{CoRR}, abs/2401.07817, 2024{\natexlab{f}}.
\newblock \doi{10.48550/ARXIV.2401.07817}.
\newblock URL \url{https://doi.org/10.48550/arXiv.2401.07817}.

\bibitem[Ziegler et~al.(2019)Ziegler, Stiennon, Wu, Brown, Radford, Amodei, Christiano, and Irving]{DBLP:journals/corr/abs-1909-08593}
Daniel~M. Ziegler, Nisan Stiennon, Jeffrey Wu, Tom~B. Brown, Alec Radford, Dario Amodei, Paul~F. Christiano, and Geoffrey Irving.
\newblock Fine-tuning language models from human preferences.
\newblock \emph{CoRR}, abs/1909.08593, 2019.
\newblock URL \url{http://arxiv.org/abs/1909.08593}.

\bibitem[Ziemski et~al.(2016)Ziemski, Junczys{-}Dowmunt, and Pouliquen]{UNCorpus}
Michal Ziemski, Marcin Junczys{-}Dowmunt, and Bruno Pouliquen.
\newblock The united nations parallel corpus v1.0.
\newblock In Nicoletta Calzolari, Khalid Choukri, Thierry Declerck, Sara Goggi, Marko Grobelnik, Bente Maegaard, Joseph Mariani, H{\'{e}}l{\`{e}}ne Mazo, Asunci{\'{o}}n Moreno, Jan Odijk, and Stelios Piperidis (eds.), \emph{Proceedings of the Tenth International Conference on Language Resources and Evaluation {LREC} 2016, Portoro{\v{z}}, Slovenia, May 23-28, 2016}. European Language Resources Association {(ELRA)}, 2016.
\newblock URL \url{http://www.lrec-conf.org/proceedings/lrec2016/summaries/1195.html}.

\bibitem[Zou et~al.(2023)Zou, Wang, Kolter, and Fredrikson]{DBLP:journals/corr/abs-2307-15043}
Andy Zou, Zifan Wang, J.~Zico Kolter, and Matt Fredrikson.
\newblock Universal and transferable adversarial attacks on aligned language models.
\newblock \emph{CoRR}, abs/2307.15043, 2023.
\newblock \doi{10.48550/ARXIV.2307.15043}.
\newblock URL \url{https://doi.org/10.48550/arXiv.2307.15043}.

\end{thebibliography}
\bibliographystyle{tmlr}

\end{document}